\documentclass[12pt,letterpaper]{article}
\usepackage[a4paper, total={7in, 10in}]{geometry}

\usepackage{graphicx}
\usepackage{helvet}
\usepackage{authblk}
\usepackage{hyperref}
\usepackage{amsmath} 
\usepackage{amssymb} 
\usepackage{orcidlink} 
\usepackage[super,comma,sort&compress]  
   {natbib}

\usepackage{float}
\usepackage{appendix}

\usepackage{fancyhdr}
\usepackage[acronym]{glossaries}
\makeglossaries

\newacronym{cbos}{CBOS}{Continuous Bag of Skip-Gram}

\newacronym{cbow}{CBOW}{Continuous Bag of Words}

\newacronym{cnn}{CNN}{Convolutional Neural Network}

\newacronym{crf}{CRF}{Conditional Random Field}

\newacronym{dl}{DL}{Deep Learning}

\newacronym{dnn}{DNN}{Deep Neural Network}

\newacronym{dsm}{DSM}{Distributional Semantic Model}

\newacronym{gec}{GEC}{Grammatical Error Correction}

\newacronym{glu}{GLU}{General Language Understanding}

\newacronym{gnn}{GNN}{Graph Neural Network}

\newacronym{gpu}{GPU}{Graphics Processing Unit}

\newacronym{gru}{GRU}{Gated Recurrent Unit}

\newacronym{gsl}{GSL}{Greek Sign Language}

\newacronym{ie}{IE}{Information Extraction}

\newacronym{ietf}{IETF}{Internet Engineering Task Force}

\newacronym{bcp}{BCP}{Best Current Practice}

\newacronym{ir}{IR}{Information Retrieval}

\newacronym{lfg/xle}{LFG/XLE}{Lexical-Functional Grammar / Xerox Linguistic Environment}

\newacronym{lgbtq}{LGBTQ}{Lesbian, Gay, Bisexual, Transgender, Queer}

\newacronym{lm}{LM}{Language Model}

\newacronym{llm}{LLM}{Large Language Model}

\newacronym{lr}{LR}{Language Resource}

\newacronym{bilstm}{Bi-LSTM}{Bidirectional Long Short-Term Memory}

\newacronym{lstm}{LSTM}{Long Short-Term Memory}

\newacronym{ml}{ML}{Machine Learning}

\newacronym{mt}{MT}{Machine Translation}

\newacronym{mbert}{mBERT}{multilingual BERT}

\newacronym{mwe}{MWE}{Multi-word expression}

\newacronym{ner}{NER}{Named Entity Recognition}

\newacronym{nli}{NLI}{Natural Language Inference}

\newacronym{nlg}{NLG}{Natural Language Generation}

\newacronym{nlp}{NLP}{Natural Language Processing}

\newacronym{nlu}{NLU}{Natural Language Understanding}

\newacronym{nmt}{NMT}{Neural Machine Translation}

\newacronym{nn}{NN}{Neural Network}

\newacronym{oie}{Open IE}{Open Information Extraction}

\newacronym{plm}{PLM}{Pre-trained Language Model}

\newacronym{pos}{POS}{Part of Speech}

\newacronym{qa}{QA}{Question Answering}
\newacronym{bow}{BoW}{Bag of Words}
\newacronym{scfg}{SCFG}{Synchronous Context-free Grammar}
\newacronym{rl}{RL}{Reinforcement Learning}

\newacronym{rdf}{RDF}{Resource Description Framework}

\newacronym{rf}{RF}{Random Forest}

\newacronym{rnn}{RNN}{Recurrent Neural Network}

\newacronym{sa}{SA}{Sentiment Analysis}

\newacronym{smt}{SMT}{Statistical Machine Translation}

\newacronym{smo}{SMO}{Sequential Minimal Optimization}

\newacronym{svm}{SVM}{Support Vector Machine}

\newacronym{tf-idf}{TF-IDF}{Term Frequency - Inverse Document Frequency}

\newacronym{ud}{UD}{Universal Dependencies}

\newacronym{xlmr}{XLM-R}{XLM-RoBERTa}

\usepackage{booktabs}
\usepackage{multirow}
\usepackage[greek,english]{babel}
\usepackage[T1]{fontenc}
\makeatletter
\renewcommand{\maketitle}{\bgroup\setlength{\parindent}{0pt}
\begin{flushleft}
  \textbf{\@title}
  
  \@author
\end{flushleft}\egroup}
\makeatother

\title{A Systematic Survey of Natural Language Processing for the Greek Language}
\date{}

\author[1]{Juli Bakagianni}
\author[2]{Kanella Pouli}
\author[2]{Maria Gavriilidou}
\author[1,3,4,5,*]{John Pavlopoulos}

\affil[1]{Department of Informatics, Athens University of Economics and Business, Athens GR10434, Greece}
\affil[2]{Institute for Language and Speech Processing, Athena Research Center, Athens GR15125, Greece}
\affil[3]{Archimedes, Athena Research Center, Athens GR15125, Greece}
\affil[4]{Department of Computer and Systems Sciences, Stockholm University, Kista 16455, Sweden}
\affil[5]{Lead contact}

\affil[*]{Correspondence: \texttt{annis@aueb.gr}}

\begin{document}

\maketitle

\begin{abstract}

Comprehensive monolingual Natural Language Processing (NLP) surveys are essential for assessing language-specific challenges, resource availability, and research gaps. However, existing surveys often lack standardized methodologies, leading to selection bias and fragmented coverage of NLP tasks and resources. This study introduces a generalizable framework for systematic monolingual NLP surveys. Our approach integrates a structured search protocol to minimize bias, an NLP task taxonomy for classification, and language resource taxonomies to identify potential benchmarks and highlight opportunities for improving resource availability. We apply this framework to Greek NLP (2012-2023), providing an in-depth analysis of its current state, task-specific progress, and resource gaps. The survey results are publicly available (\href{https://doi.org/10.5281/zenodo.15314882}{https://doi.org/10.5281/zenodo.15314882}) and are regularly updated to provide an evergreen resource. This systematic survey of Greek NLP serves as a case study, demonstrating the effectiveness of our framework and its potential for broader application to other not so well-resourced languages as regards  NLP.
\end{abstract}

\section*{Keywords}


Monolingual NLP survey, Greek NLP, language resources, task taxonomy, search protocol

\section{Introduction}\label{sec:intro}
\gls{nlp} focuses on the computational processing of human languages, enabling machines to understand and generate natural language. Recently, several \gls{nlp} tasks have advanced significantly with the help of \gls{dl}\cite{vaswani2017attention} and more recently with \glspl{llm}\cite{brown2020language}. Multilingual \gls{nlp} has benefited from these advances;\cite{pires2019multilingual,qin2025survey} however, by focusing on progress per language, we observe that well-supported languages benefit considerably more compared to the rest.\cite{blasi-etal-2022-systematic} As a result, \gls{nlp} for the myriad of languages worldwide relies heavily on research conducted for well-supported languages, often inheriting their assumptions, biases, and other characteristics that may not align with their unique linguistic features,\cite{bender2021dangers} thereby limiting equitable technological access.

Monolingual \gls{nlp} surveys offer a pathway to address these disparities by synthesizing language-specific challenges (e.g., scarce annotated data, morphological complexity), auditing resources and methodological adaptations, and identifying research gaps that hinder equitable progress. However, their utility depends on systematic rigor: reproducible search protocols and transparent filtering criteria minimize selection bias and ensure replicable results, while organizing surveyed material into coherent \gls{nlp} thematic tracks, such as Syntax and \gls{ie}, enables structured analysis of task-specific challenges, gaps, and trends. This structured presentation also supports cross-task comparisons, revealing overarching insights, such as state-of-the-art models across tasks. Furthermore, systematically documenting \glspl{lr} — including their availability, annotation status (e.g., raw, human-annotated), and annotation type (e.g., automatically labeled) — identifies potential benchmarks that can be used for pre-training, fine-tuning, and assessing \gls{nlp} models, without inheriting the assumptions and biases of well-supported languages. This process also highlights critical shortages, such as annotated datasets for understudied tasks. Although monolingual \gls{nlp} surveys exist,\cite{oflazer2014turkish,shoufan2015natural,al2018deep,Marie2019,papantoniou2020nlp,alam2021review,hamalainen2021current,desai2021taxonomic,guellil2021arabic,darwish2021panoramic,rajendran2022tamil,gonzalez2022natural} and their contributions are valuable, they do not share the surveying methods they followed, such as the search protocol, risking selection bias, and fragmented coverage of tasks, and resources. To our knowledge, no generalized framework exists to standardize monolingual survey design, hindering actionable progress for less-supported languages.

In this work, we bridge this gap by (1) proposing a generalizable methodology for systematic monolingual \gls{nlp} surveys, and (2) applying it to Greek, a language that is characterized as a low-resource language for several \gls{nlp} tasks.\cite{athanasiou2017,Papadopoulos2021penelopie,mountantonakis2022,papaioannou-etal-2022-cross} We demonstrate how our framework — tested through a comprehensive review of Greek \gls{nlp} — enables researchers to identify language-specific challenges, evaluate resource availability, and prioritize future work efficiently. Our survey of Greek \gls{nlp} research is focused on studies published between 2012 and 2023. This timeframe marks transformative advancements in \gls{nlp} (e.g., the shift from \gls{ml} to \gls{dl} and \glspl{llm}) and societal shifts driven by GenA’s digital-native upbringing. Our analysis captured how Greek \gls{nlp} evolved alongside these technological and generational trends. Using our systematic search protocol, we retrieved over a thousand research studies on Greek \gls{nlp}, of which 142 met the specific criteria outlined in our search protocol. This survey offers both task-specific insights and an overview of overarching trends in Greek \gls{nlp}. 

Our findings show that:
\begin{itemize}
    \item \textbf{Greek is moderately supported in \gls{nlp}}. We identified nine publicly available, human-annotated datasets related to nine distinct \gls{nlp} tasks, including Summarization, \gls{ner}, Intent Classification, Topic Classification, \gls{gec}, Toxicity Detection, Syntactical and Morphological Analysis, \gls{mt}, and Text Classification. These resources hold significant potential as benchmarks for advancing Greek \gls{nlp} research. This observation positions Greek as a moderately-supported language in \gls{nlp}, and is also aligned with a language support classification system we developed, that classifies languages based on their coverage in ACL publications, which also classifies Greek as a moderately-supported language.
    \item \textbf{Resource gaps exist despite cross-lingual innovations}. Despite progress, benchmarks for certain \gls{nlp} tasks, such as \gls{sa}, are missing. However, our systematic cataloguing identified 17 datasets that — with added licenses or improved maintenance — could serve as benchmarks. Cross-lingual techniques, such as translation strategies outperforming multilingual encoders,\cite{papaioannou-etal-2022-cross} offer practical pathways to mitigate data scarcity, and therefore we summarize and highlight these efforts.
    \item \textbf{Methodological shifts reveal lingering gaps.} The research landscape in Greek \gls{nlp} has shifted from traditional \gls{ml} methods, which dominated until 2018, to the increasing adoption of \gls{dl} approaches since 2019. Despite this shift, \gls{ml} methods continue to dominate certain tasks, such as Authorship Analysis, \gls{qa}, and Semantics, indicating that these areas require further \gls{dl} innovation. Conversely, newer trends for Greek, such as \gls{ie}, Ethics and \gls{nlp}, and Summarization are increasingly dominated by \gls{dl} approaches, with Greek included also in shared tasks for the last two fields.
    \item \textbf{Monolingual \glspl{lm} are preferred over multilingual ones}. Despite the global emphasis on multilingual systems, such as \gls{xlmr} and \gls{mbert}, few studies in Greek \gls{nlp} are found to use them.\cite{evdaimon2023greekbart,Ranasinghe2021,koutsikakis2020,ahn2020nlpdove,zampieri2023offenseval} Greek \gls{nlp} favors monolingual \glspl{lm}, such as GreekBERT,\cite{koutsikakis2020} which achieves state-of-the-art results in several studies addressing different tasks.
    \item \textbf{Task-specific trends differ notably from global trends.} While Greek research aligns with global \gls{nlp} trends in tasks such as \gls{sa}, where \gls{nlp} research declines,\cite{rohatgi2023acl} this is not true in areas such as Syntax, where Greek \gls{nlp} retains interest despite a global decline in syntax-related research.
\end{itemize}

In what follows, we first provide the background of the present work (\S\ref{sec:background}). In this section, we discuss the support level of human languages within the \gls{nlp} community and the characteristics of the Greek language (\S\ref{sec:language}). Also, we discuss the examined time frame along with an exploration of the methodological shifts occurring during this time period (\S\ref{sec:time}), and we present the related work (\S\ref{sec:rel-work}). Then, we present our approach (\S\ref{sec:our-method}), consisting of the search protocol (\S\ref{sec:search}) and the taxonomies adopted for tasks, \glspl{lr} availability, and annotation type (\S\ref{sec:taxonomy}). Subsequently, we present the main outcomes of our study, organized by \gls{nlp} thematic areas: Machine Learning for \gls{nlp} (\S\ref{sec:ml-nlp}), Syntax and Grammar (\S\ref{sec:syntax}), Semantics (\S\ref{sec:semantics}), \gls{ie} (\S\ref{sec:ie}), \gls{sa} (\S\ref{sec:sentiment}), Authorship Analysis (\S\ref{sec:authorship}),  Ethics and NLP (\S\ref{sec:toxicity}),  Summarization (\S\ref{sec:summarization}), \gls{qa} (\S\ref{sec:qa}), \gls{mt} (\S\ref{sec:mt}), and \gls{nlp} Applications that are not classifiable in any of the previous tracks (\S\ref{sec:nlp-apps}). Lastly, we discuss the outcomes of this study with remarks on the limitations, and our final observations (\S\ref{sec:discussion}), followed by our conclusions.  
Each of the sections presenting the main outcomes of this survey (\S\ref{sec:ml-nlp}-\S\ref{sec:nlp-apps}) is structured as follows: first, we describe the track within its global context; then, we discuss the methods identified by our study and the \glspl{lr} produced; finally, each section concludes with a summary of the track and relevant observations.

\begin{section}{Background}\label{sec:background}
\subsection{The Language}
\label{sec:language}
\subsubsection{Human Languages}
\label{ssec:languages}

Human languages encompass a rich tapestry, totaling 7,916, as cataloged by ISO 639-3,  an international standard that assigns unique codes to represent languages, including living, extinct, ancient, historic and constructed ones. Despite this linguistic diversity, \gls{nlp} research exhibits significant imbalances, with English dominating the field. To assess the level of support for different languages in the \gls{nlp} field, we conducted an analysis of the ACL Anthology, an authoritative hub of computational linguistics and \gls{nlp} research. Specifically, we counted papers published between January 2012 and January 2024 that reference each language listed in the \gls{ietf} \gls{bcp} 47 standard in their titles or abstracts. Languages were classified into three tiers based on the number of publications: well-supported, moderately-supported, and low-supported.

As shown in Figure~\ref{fig:acl_languages}, English is the most-studied language, with 6,915 publications. This figure likely underestimates the true volume, as it is common practice in the \gls{nlp} community not to explicitly mention English when it is the language of study.\cite{bender2019rule} Chinese, German, French, Arabic, and Spanish are also well-supported, each with thousands of publications. Moderately-supported languages, including Greek, constitute the second tier, with publication counts ranging from 100 to 1,000 per language. In contrast, 574 languages fall into the third tier, with one to 100 publications, while 7,312 languages are entirely unsupported.
\begin{figure}[H]
\centering
\includegraphics[width=\columnwidth]{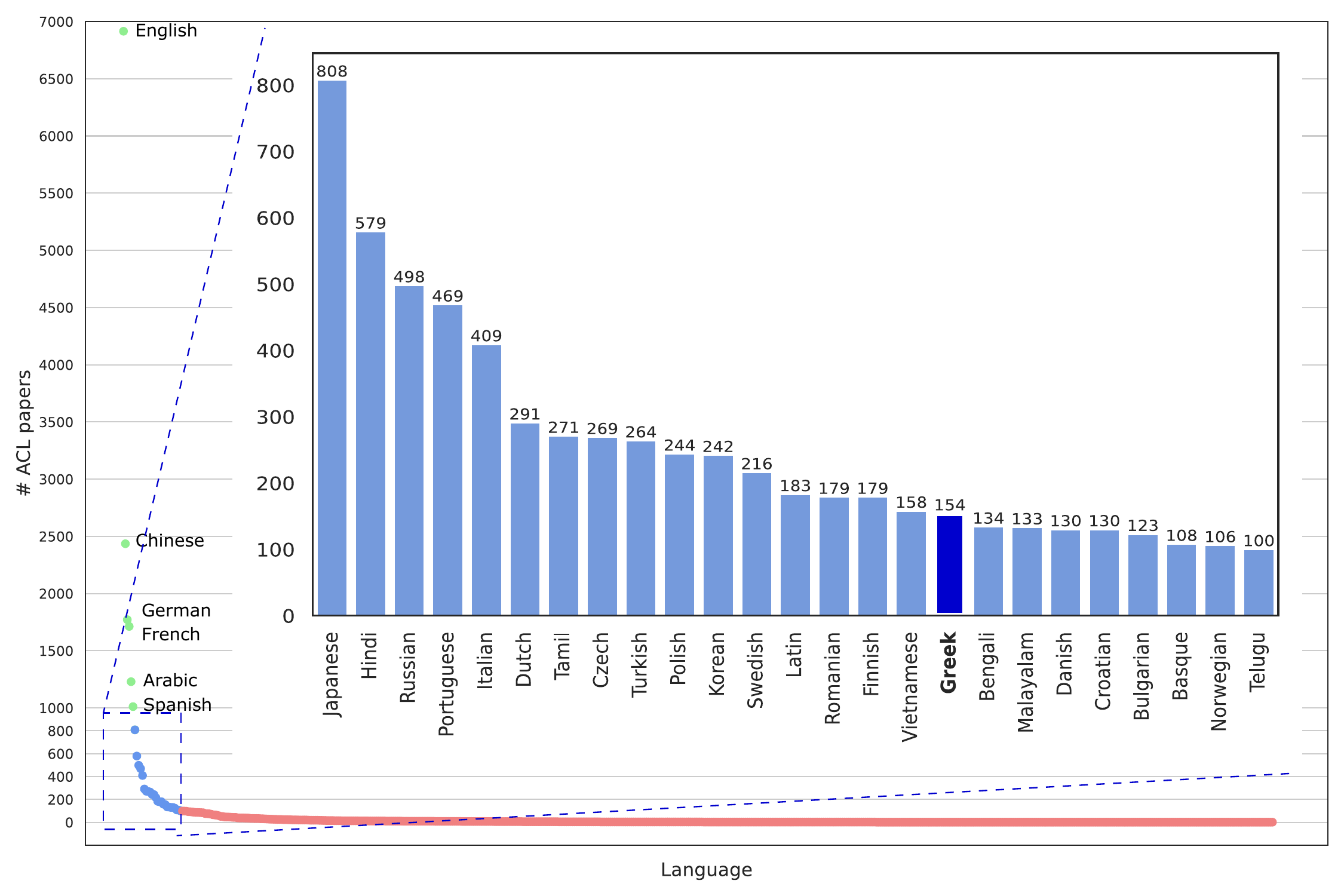}
\caption{
Number of publications in the ACL Anthology per language (shown vertically), with languages referenced in the title or abstract (horizontally). We use the collection of languages outlined in the \gls{ietf} \gls{bcp} 47 language tag (RFC 5646).\cite{rfc5646} The vast majority of languages appear in none (7,312 languages) or fewer than 100 publications (574 languages), depicted with a long red tail on the lowermost part of the figure. We refer to this group of languages as the third tier, which consists of less-supported languages. The second tier, shown in blue in the same distribution,is presented with additional detail in the upper-right part of the figure. This tier comprises moderately-supported languages, which appear in between 100 and 1,000 publications, with Greek specifically represented in 154 publications. The first tier comprises well-supported languages, each referenced in more than 1,000 publications: English (referenced in 6,915 papers), Chinese (in almost 2,500), German and French (around 1,750), Arabic (1,229), and Spanish (1,011).
}
\label{fig:acl_languages}
\end{figure}

This study focuses on Greek, a second-tier language among 25 others with moderate \gls{nlp} research interest (100–1,000 references). Within this group, Greek ranks 17th by publication count (154 papers). However, when adjusting for total speaker populations (including native and second-language speakers), Greek rises to the 10th place. Speaker population data were sourced from \citet{ethnologue} and support per speaker population was calculated by dividing publication counts by speaker populations. Latin was excluded as an extinct language. This adjustment provides a more nuanced perspective by incorporating both research output and the size of the speaker base.

\subsubsection{The Greek Language}\label{ssec:MG}
Understanding the linguistic characteristics of a language can help \gls{nlp} researchers understand the specific challenges and opportunities for developing and applying \gls{nlp} technologies in this language context. Greek, or Modern Greek, to differentiate it from earlier historical stages, is the official language of Greece and one of the two official languages of Cyprus. It is the mother tongue of approximately 95\% of the 10.5 million inhabitants of Greece and of the approximately 500,000 Greek Cypriots. It is also used by approximately five million people of Greek origin worldwide as heritage language.\cite{gavriilidou2023language} 

The Greek alphabet has been the main script for writing Greek for most of the language's recorded history.\cite{Davies2015}
The use of the standard variety in education and mass media has led to the prevalence of Standard Modern Greek over various dialects. Henceforth, the term ``Greek'' is used to refer to Standard Modern Greek, which is a highly inflected language. It has four cases for the nominal system, two numbers, and three genders. The verb conjugation system is even more complex, with multiple tenses, moods, voices, different suffixes per person, and many irregularities. Word length is an additional factor differentiating Greek from other languages, most notably English. The majority of the Greek words, typically, have two or three syllables, but words with more syllables (e.g., eight or nine) are also not rare.\cite{gavrilidou2012metanet} Moreover, Greek, unlike English, exhibits significant flexibility in word order. Its system of rich nominal inflection allows syntactic relations among clausal elements to be identified without requiring fixed positions. For instance, a simple declarative clause containing a verb, its nominal subject, and object can be constructed in all six logically possible combinations.\cite{tzanidaki1995greek} 

\subsection{The Time Period of the Survey}
\label{sec:time}
The selected time span for our survey (2012 to 2023) aimed to capture the evolution of research methodologies in \gls{nlp} in response to global technological advancements and shifts in the field. The period under investigation witnessed a transition from traditional \gls{ml} to \gls{dl}. As \citet{manning-2015-last} stated: ``\gls{dl} waves have lapped at the shores of computational linguistics for several years now, but 2015 seems like the year when the full force of the tsunami hit the major \gls{nlp} conferences''. We aim to explore how this methodological shift influenced research on Greek. In the following sections we present a brief historical overview of the scientific field itself (i.e., by disregarding the target language) and its evolution over the years under study. The methods applied to the Greek language are outlined in \S\ref{sec:ml-nlp}.

\subsubsection{The ML Era}\label{sssec:ml}

The predominant approach to \gls{nlp} research in 2012, marking the beginning of our study period, primarily relied on traditional \gls{ml} algorithms. Traditional \gls{ml} focuses on developing algorithms and models that learn statistical patterns from data to make predictions or decisions. Unlike \gls{dl}, which automates the feature extraction process through layered neural architectures, traditional \gls{ml} is highly dependent on manual feature engineering.  In traditional \gls{ml}, relevant features are extracted, selected, or created from raw data to improve model performance. Commonly employed features include character or word tokens (unigrams or n-grams) and their frequency, often using methods such as frequency counts or the \gls{tf-idf} weighting scheme. Lexicon-based features, such as lists of words with specific meanings (e.g., sentiment lexicons), are also common. 

\subsubsection{The DL Era}\label{sssec:dl}

The surge of \gls{dl} in \gls{nlp} can be attributed to its ability to automatically learn hierarchical representations of data, eliminating the need for extensive feature engineering.\cite{qiu2020pre} Coupled with the availability of vast amounts of data and increased computational power, \gls{dl} has enabled more effective handling of complex linguistic structures. As a result, \gls{dl} has demonstrated superior performance across various \gls{nlp} tasks.\cite{bommasani2021opportunities} These advancements have led to the development of \glspl{plm}, which are neural network-based statistical \glspl{lm}.\cite{minaee2024large}

\glspl{plm} are task-agnostic and follow a pre-training and fine-tuning paradigm, where \glspl{lm} are pre-trained on Web-scale unlabeled
text corpora for general tasks such as word prediction, and then fine-tuned to specific tasks using small amounts of (labeled) task-specific data.\cite{minaee2024large} Initially, models such as \glspl{rnn}\cite{pascanu2013difficulty} were used for these purposes. \glspl{rnn}, proposed in the 1980s for modeling time-series,\cite{rumelhart1986learning,elman1990finding,werbos1988generalization} are designed to explore temporal correlations between distant elements in the text. 

The introduction of the Transformer architecture was a major milestone in \gls{nlp}. Transformers\cite{vaswani2017attention} use self-attention mechanisms to compute attention scores for each word in a sentence, allowing for greater parallelization compared to \gls{rnn}.\cite{minaee2024large} Transformer-based \glspl{plm} are categorized into three main types based on their neural architectures: encoder-only, decoder-only, and encoder-decoder models. Encoder-only models, such as BERT\cite{devlin-etal-2019-bert} and its variants (RoBERTa,\cite{liu2019roberta} ALBERT,\cite{lan2020albertlitebertselfsupervised} DeBERTa,\cite{he2021debertadecodingenhancedbertdisentangled} XLM,\cite{lample2019crosslinguallanguagemodelpretraining} \gls{xlmr},\cite{conneau2019unsupervised} and XLNet)\cite{yang2020xlnetgeneralizedautoregressivepretraining} are primarily used for language understanding tasks such as text classification. A detailed discussion on the distinction between \gls{nlu} and \gls{nlg} can be found in Appendix~\ref{sec:nlu-nlg}. The fascination with the inner workings of these Transformer-based models has led to the emergence of a trend known as BERTology.\cite{bertology_hf} Decoder-only models, including GPT-1\cite{radford2018improving} and GPT-2\cite{radford2019language} from OpenAI, focus on language generation tasks. Encoder-decoder models, such as T5,\cite{raffel2020exploring} mT5,\cite{xue2021mt5} and BART,\cite{lewis2019bart} are versatile and can perform both understanding and generation tasks by framing them as sequence-to-sequence problems.

Finally, \glspl{llm} refer to transformer-based \glspl{plm} with tens to hundreds of billions of parameters. These models are not only larger in size but also exhibit stronger language understanding and generation capabilities compared to smaller models mentioned earlier.\cite{minaee2024large} Notable \gls{llm} families include OpenAI's GPT, Meta's open-source Llama, and Google's PaLM and Gemini. Other representative \glspl{llm} include FLAN,\cite{wei2021finetuned} Gopher,\cite{rae2021scaling} T0,\cite{sanh2021multitask} and GLaM\cite{du2022glam} among others.

\subsection{NLP surveys}\label{sec:rel-work}

\subsubsection{Greek NLP surveys}

Through our search protocol (\S\ref{sec:search}), we identified other Greek \gls{nlp} surveys -- both comprehensive and domain-specific -- which we discuss here. 
First, \citet{papantoniou2020nlp} provided a brief survey of \gls{nlp} for the Greek language covering Ancient Greek, Modern Greek, and various dialects. This survey included the work of 99 papers published from 1990 to 2020. The authors addressed text, video, and image modalities. For text modality, they presented papers on tasks such as Phonology, Syntax, Semantics, \gls{ie}, \gls{sa}, Argument Mining, \gls{qa}, \gls{mt}, and \gls{nlp} Applications. For image modality they outlined Optical Character Recognition (OCR), and for video modality, they discussed Lip Reading and Keyword Spotting. Regarding \glspl{lr}, they presented a limited number of \glspl{lr}, specifically three online lexica, five online corpora, two downloadable datasets, five tools, and one service. \citet{giarelis2023review} provided an overview of state-of-the-art research in Greek \gls{nlp} and chatbot applications published since 2018, establishing the search protocol they used. They reported on three \gls{dl} \glspl{lm}, two embedding-based techniques, and nine \gls{dl} \gls{nlp} applications, detailing the relevant datasets. For chatbot applications, they identified and reviewed five papers. Additionally, they offered insights into \gls{nlp} models and chatbot implementation methodologies.

The remaining three surveys are purely domain-specific. 
\citet{Nikiforos2021} provided an extensive review of 49 papers published from 2012 to 2020 related to the Social Web in Modern Greek, Greek dialects, and Greeklish script. The \gls{nlp} tasks covered include Argument Mining, Authorship Attribution, Gender Identification, Offensive Language Detection, and \gls{sa}. The authors systematically addressed the scientific contributions and unresolved issues of the reviewed papers. They also presented two tools and 21 datasets extracted from the surveyed papers, providing detailed information and links where available. 
\citet{Alexandridis2021Survey} reviewed 14 papers published from 2014 to 2020 that focus specifically on \gls{sa} and opinion mining in Greek social media. The authors discussed the methods, tools, datasets, lexical resources, and models used for \gls{sa} and opinion mining in Greek texts. 
Finally, \citet{Krasadakis2021} surveyed 43 papers related to Legal \gls{nlp} published from 2012 to 2021. The survey covered tasks such as \gls{ner}, Entity Linking, Text Segmentation, Summarization, \gls{mt}, Rationale Extraction, Judgment Prediction, and \gls{qa}.

\subsubsection{Monolingual NLP surveys in Other Languages}\label{sec:other-lang-surveys}
 
Beyond Greek, we found that comprehensive monolingual \gls{nlp} surveys are relatively rare. We searched the literature for surveys or overviews that cover a broad range of \gls{nlp} tasks -- similar in scope to our research -- for well- and moderately-supported languages, as classified in our tier system (\S\ref{ssec:languages}). Our search process involved querying Google Scholar for publications published between January 2012 and September 2023, using a specific query pattern. We searched for the name of the language of interest along with the keyword ``Natural Language Processing'', and either ``survey'' or ``overview''.

Notably, we found that only 19\% of well- and moderately-supported languages have peer-reviewed comprehensive monolingual \gls{nlp} surveys. Among the six well-referenced languages, only Arabic, a macro-language that encompasses various individual varieties, has dedicated \gls{nlp} surveys.\cite{guellil2021arabic,darwish2021panoramic,Marie2019,al2018deep,shoufan2015natural} Of the 25 moderately-referenced languages, five have peer-reviewed surveys, i.e., Tamil,\cite{rajendran2022tamil} Turkish,\cite{oflazer2014turkish} Finnish,\cite{hamalainen2021current} Greek,\cite{papantoniou2020nlp} and Basque.\cite{gonzalez2022natural} Additionally, two languages, Hindi\cite{desai2021taxonomic} and Bengali,\cite{alam2021review} have preprints available. 

\subsubsection{Limitations in Existing NLP Surveys}
The surveys mentioned above provide valuable insights into the languages they study; however, none disclose their search protocol, except for the domain-specific work of \citeauthor{giarelis2023review}.\cite{giarelis2023review} This lack of transparency makes it difficult to assess the reproducibility of the surveys and understand the criteria and rationale behind the inclusion of specific papers. Additionally, it is unclear whether the \gls{nlp} tasks presented fully encompass the research conducted in the language or if the papers were manually selected to fit the chosen tasks. Similarly, while some surveys provide information about the \glspl{lr} available for the examined language, it is often unclear why certain \glspl{lr} were selected, and whether they are accessible and properly licensed.
\end{section}

\begin{section}{Monolingual NLP Survey Methodology}\label{sec:our-method}
This section outlines the methodology proposed for constructing monolingual \gls{nlp} surveys. It includes the search protocol (\S\ref{sec:search}) applied to Greek \gls{nlp} research, as well as the taxonomies of tasks and \glspl{lr} (\S\ref{sec:taxonomy}).
\subsection{Search Protocol}\label{sec:search}

We developed a comprehensive search protocol to identify peer-reviewed research papers related to \gls{nlp} in the Greek language. Our goal was to create a process that is adaptable to any language and any publication time period. The protocol includes a search strategy for automatically locating relevant papers (\S\ref{sec:search-strategy}) and a filtering process based on well-defined criteria (\S\ref{filtering}). It uses both bibliographic metadata and additional metadata collected to support the surveying process (\S\ref{sec:metadata}).

\subsubsection{Search Strategy}\label{sec:search-strategy}

\paragraph{Scientific Databases} We used three reputable scientific databases to identify research papers related to \gls{nlp} for Greek, published between January 2012 and December 2023. The selected databases are: ACL Anthology,\cite{acl_anthology} a hub for computational linguistics and \gls{nlp} research; Semantic Scholar,\cite{semantic_scholar} an AI-powered search engine prioritizing computer science and related fields; and Scopus,\cite{scopus} a globally recognized database. These databases were chosen not only for their reputability but also for their automated publication retrieval capabilities: Semantic Scholar and Scopus offer APIs, while ACL Anthology provides publication metadata in XML format through its GitHub repository.\cite{acl_anthology_repo}

\paragraph{Querying Process} The search was conducted using tailored query terms across ACL Anthology, Scopus, and Semantic Scholar, adapting to the search capabilities of each database. Scopus allows searching in the title, abstract, and full text (including references); Semantic Scholar searches across the entire paper content; and ACL Anthology limits the search to the title and abstract. Therefore, we focused our search on the language name, i.e., ``Greek'' or ``Modern Greek'', in the title or abstract of the papers and the term ``Natural Language Processing'' in the entire paper (where feasible). This approach was chosen because papers focused on a specific language are likely to mention the language name in these sections, thereby reducing the retrieval of false positive papers (see \S\ref{ssec:challenges}). Specifically, Scopus employs Lucene queries, allowing us to search for the language name in titles and abstracts, and the term ``Natural Language Processing'' across the entire paper. Semantic Scholar does not offer specific search area options, so we used combined keywords with the + operator (AND), initially searching broadly and subsequently filtering results where the language name appeared in the title or abstract. For the ACL Anthology, which is dedicated to \gls{nlp}, we limited our search to the language name in the title or abstract.

\paragraph{Core Search Rounds} The search process comprised four rounds, with the first three being core rounds, as detailed in Table~\ref{tab:searcQueries}. The first two core rounds focused on papers published between 2012 and 2022 and differed in the language query terms used. In the first core round, we searched using ``Modern Greek'', but due to its limited usage, we shifted to ``Greek'' in the second core round to capture a wider range of relevant papers. The language-specific filtering was then applied during the filtering process stage. The third core round focused on papers published in 2023 to incorporate more recent relevant work. Unlike the earlier rounds —which were exploratory and iterative, helping to shape the survey design — this round was conducted several months later, after the finalization of our survey methodology. As such, it served as a test case for our methodology, assessing the time and the effort needed to integrate new papers into the survey. Incorporating papers from this round was one-third faster, highlighting how a well-defined monolingual survey methodology, such as the one we propose, can significantly improve efficiency and scalability for future surveys.

\begin{table}[H]
\centering
\caption{Core rounds of the search process, including the databases searched in each round, the queries used, the publication date ranges, and the dates the searches took place.}
\label{tab:searcQueries}
\resizebox{\textwidth}{!}{
\begin{tabular}{|l | l | p{7cm} | l| l |} 
 \hline
 \textbf{Round} & \textbf{Database} & \textbf{Query} & \textbf{Publication date} & \textbf{Search date} \\  
 \hline
 1st & ACL Anthology & “Modern Greek” in title or abstract & 2012-2022 & 1/11/2022 \\  
    & Scopus & TITLE-ABS(\{Modern Greek\}) AND ALL(\{Natural Language Processing\}) & 2012-2022  & 31/10/2022 \\  
    & Semantic Scholar & Modern + Greek + Natural + Language + Processing and then ``Modern Greek'' in title or abstract & 2012-2022  & 1/11/2022 \\  
 \hline
 2nd & ACL Anthology & “Greek” in title or abstract & 2012-2022  & 24/10/2023 \\  
 & Scopus & TITLE-ABS(\{Greek\}) AND ALL(\{Natural Language Processing\}) & 2012-2022  & 24/10/2023 \\  
 & Semantic Scholar & Greek + Natural + Language + Processing and then ``Greek'' in title or abstract & 2012-2022  & 24/10/2023 \\  
 \hline
 3rd & ACL Anthology & “Greek” in title or abstract & 2023  & 15/7/2024 \\  
 & Scopus & TITLE-ABS(\{Greek\}) AND ALL(\{Natural Language Processing\}) & 2023  & 15/7/2024 \\  
 & Semantic Scholar & Greek + Natural + Language + Processing and then ``Greek'' in title or abstract & 2023  & 15/7/2024 \\  
 \hline
\end{tabular}
}
\end{table}

\paragraph{Quality Assurance Round} The fourth round served as a supplementary phase for quality assurance of our search strategy and to validate the comprehensiveness of the selected query terms during the previous core search rounds. The objectives were two-fold: first, to verify that the selected queries terms retrieved all relevant publications related to \gls{nlp} research in the Greek language; and second, to address any potential gaps from excluding Google Scholar\cite{google_scholar} in the core rounds. Despite its widespread usage, Google Scholar was not included in the core rounds due to its lack of an API for automated publication retrieval. In this phase, we cherry-picked specific \gls{nlp} downstream tasks, such as Toxicity Detection, Authorship Analysis, \gls{sa}, \gls{mt}, \gls{qa}, Summarization, Syntax, and Semantics, and integrated them as additional query terms  alongside the language name and the overarching term ``Natural Language Processing'' in Google Scholar. This effort identified only five additional papers, suggesting that the original search protocol effectively captured Greek NLP publications. Therefore, we consider our approach comprehensive. Further details about this quality assurance step can be found in the Appendix~\ref{app:quality_assurance_round}.

\subsubsection{Filtering Strategy} \label{filtering}
We retrieved a total of 1,717 bibliographic records, which were reduced to 1,135 after removing duplicates. 
Each record included metadata such as the title, author names, abstract, publication date, and citations.  
Publication types were manually added when missing (e.g., conference papers, journal articles, etc.). Papers not relevant to our study were discarded based on the following qualitative and quantitative exclusion criteria: 

\begin{itemize}
    \item \textbf{Publication language}; all major \gls{nlp} conferences and journals publish in English, hence studies written in other languages (including Greek) were disregarded; 
    \item \textbf{Language of study}; with Modern Greek being the language of interest, both papers dedicated to monolingual (Greek specific) and multilingual (Greek inter alia) research were accepted; studies referring to older stages of the language (i.e., katharevousa), geographical dialects, or \gls{gsl} were not considered; 
    \item \textbf{Subject area}; papers irrelevant to \gls{nlp} were excluded;
    \item \textbf{Modality}; papers not studying textual data were not considered;
    \item \textbf{Publication venue}; only conference papers and journal articles were included, leaving out book chapters, theses, and preprints,
    \item \textbf{Number of citations}; we applied an arithmetic progression based on both the number of citations and the year of publication, beginning with zero for papers published in 2023 and increasing with step one for each preceding year. In this sense, the demand for citations was higher for older publications than for more recent ones. Consequently, any paper falling below the defined citation threshold was excluded from our selection. We used Google Scholar to manually extract citation counts, due to its high coverage. This criterion ensures the inclusion of impactful and relevant papers by balancing the recency and significance of contributions, thereby streamlining the selection process.
\end{itemize}

This process resulted in a final selection of 142 papers, all published within the selected time frame. We have identified 23 additional papers that are submissions to task-specific events, such as shared tasks or workshops. Only the top-ranked submissions for each task are cited in our survey, so not all retrieved submissions are featured in the survey and are consequently excluded from the statistics.

\subsubsection{Metadata Extraction}\label{sec:metadata}
In addition to the metadata retrieved from the databases, we gathered supplementary information to facilitate the surveying process. To ensure traceability of the retrieved papers, we recorded details about the search process, including the search date, the queried database, and the search query used. Furthermore, to aid in the filtering process, we collected information about the publication venue, as well as Google Scholar citations. After filtering and selecting the papers for review, we documented the tasks and tracks addressed by the authors, any keywords used, and the languages covered by each paper. For \glspl{lr} created for each paper, we gathered information on their availability, including the URL, license, and format (for datasets). Specifically for datasets, we recorded details about their annotation type, size, linguality type (monolingual or multilingual), translation process (if applicable), domain, and time coverage.
\subsection{The Taxonomies}\label{sec:taxonomy}
\subsubsection{The Task Taxonomy}\label{sec:task-taxonomy}

Our survey adopts a paper-driven approach to structuring the taxonomy of \gls{nlp} tasks and research themes, which we propose as a systematic framework for conducting monolingual \gls{nlp} surveys to comprehensively capture the \gls{nlp} research landscape for a specific language. This approach ensures that the selection of \gls{nlp} tasks and their presentation are guided directly by the surveyed papers, allowing for a taxonomy that reflects the actual scope of research. Instead of starting with a predefined set of tasks, we adopt a bottom-up methodology, assigning surveyed papers to the specific \gls{nlp} tasks they addressed. These tasks are then grouped into broader research themes using the comprehensive taxonomy proposed by \citeauthor{bommasani2023holistic},\cite{bommasani2023holistic} which maps \gls{nlp} tasks to thematic tracks presented at ACL 2023 edition.\cite{acl2023_call} This framework ensures that the survey aligns with contemporary research trends while systematically organizing the surveyed papers.

\begin{table}[H]
     \centering
     \caption{Taxonomy of \gls{nlp} tasks for the Greek language, organized according to the tracks of ACL 2023. The numbers in parentheses represent the count of surveyed papers that contribute to each task.}
     \label{tab:tasks-taxonomy}
     \resizebox{\textwidth}{!}{
 \begin{tabular}{|l|p{0.6\linewidth}|}\toprule
     \bf Track &\bf                                               Task \\\midrule
     Authorship Analysis & Authorship Verification (3), Author Profiling (3), Authorship Attribution (2), Author Identification (2), Author Clustering (1)\\\hline
     Ethics and \gls{nlp} & Hate Speech Detection (6), Offensive Language Detection (5), User Content Moderation (2), Bullying Detection (1), Verbal Aggression Detection (1) \\\hline
     \gls{ie} & \gls{ner} (7), Event Extraction (3), Entity Linking (3), Term Extraction (2), Open Information Extraction (1), Web Content Extraction (1) \\\hline
     Interpretability and Analysis of Models for \gls{nlp} & Grammatical Structure Bias (1), Word-Level Translation Analysis in Multilingual \glspl{lm}, Polysemy Knowledge in \glspl{plm} (1), Bias Detection in \glspl{plm} (1)\\\hline
     \gls{ml} for \gls{nlp} & Language Modeling (2)\\\hline
     \gls{mt} & \gls{mt} Evaluation (6), \gls{smt} (2), Rule-Based \gls{mt} (1) \\\hline
     Multilingualism and Cross-Lingual \gls{nlp} & Multilingual Language Learning (1), Term Translations Detection (1), Language Distance Detection (1), Language Identification (1), Cross-Lingual Data Augmentation (1), Cross-Lingual Knowledge Transfer (1)\\\hline
     \gls{nlp} Applications & Legal \gls{nlp} (3), Business \gls{nlp} (2), Clinical \gls{nlp} (2), Educational \gls{nlp} (1), Media \gls{nlp} (1) \\\hline
     \gls{qa} & \gls{qa} (4) \\\hline
     Semantics & Distributional Semantic Modeling (4), Natural Language Inference (2), Frame Semantics (2), Distributional Semantic Models Evaluation (1), Lexical Ambiguity (1), Semantic Annotation (1), Semantic Shift Detection (1), Word Sense Induction (1), Metaphor Detection (1), Paraphrase Detection (1), Contextual Interpretation (1)  \\\hline
     \gls{sa} and Argument Mining & Document-Level \gls{sa} (14), Sentence-Level \gls{sa} (13), Aspect-Based \gls{sa} (3), Argument Mining (2), Stance Detection (1), Paragraph-Level \gls{sa} (1)\\\hline
     Summarization & Summarization (5), Summarization Evaluation (1)\\\hline
     Syntax and \gls{gec} & \gls{gec} (3), Dependency Parsing (3), POS Tagging (3), Sentence Boundary Detection (2), MWE Parsing (2), Tokenization (1), Lemmatization (1) \\
     \bottomrule
 \end{tabular}
     }
 \end{table}

Canonical \gls{nlp} tasks were determined based on their established tradition in \gls{nlp} research, such as \gls{ner}. Although we acknowledge the subjectivity in defining ``canonical'', we determined which tasks could be considered canonical, drawing from our expertise in the field, thereby enabling consistent organization of tasks into manageable categories. Studies addressing non-canonical tasks were categorized based on their specific focus. Subsequently, each identified task was mapped to its corresponding thematic area, as outlined by ACL 2023, enabling systematic alignment of the surveyed papers with broader \gls{nlp} research themes. Table~\ref{tab:tasks-taxonomy} illustrates the resulting taxonomy of \gls{nlp} tasks for the Greek language.
 
In some cases, our taxonomy diverged from the ACL classification. Specifically, we present Authorship Analysis separately from \gls{sa} and Argument Mining, although there is a single ACL track for ``Sentiment Analysis, Stylistic Analysis, and Argument Mining''. This decision was dictated by the fact that Authorship Analysis has attracted increased attention in the \gls{nlp} community for Greek. Additionally, studies addressing tasks outside the scope of canonical \gls{nlp} domains, such as the  consolidation of historical revisions, were classified under the \gls{nlp} Applications track. By combining a flexible categorization strategy with a structured taxonomy, this survey comprehensively captures Greek \gls{nlp} research while offering a replicable methodology for other monolingual \gls{nlp} surveys.

\subsubsection{The Language Resource Taxonomies}\label{sec:lr-taxonomy}
One of our survey objectives was to compile a comprehensive list of the \glspl{lr} developed in the reviewed studies, including detailed metadata. This metadata includes the availability of each \gls{lr} ensuring it aligns with the FAIR Data Principles — findable, accessible, interoperable, and reusable.\cite{wilkinson2016fair} Our search focused on the availability of URLs for each resource rather than identifying whether they were assigned persistent identifiers, such as DOIs, which may limit full compliance with the ``findable'' criterion. Additionally, we addressed the annotation types used for the datasets. These types, which refer to the methods employed in annotating resources, significantly affect data quality, task suitability, reproducibility, and research transparency.

\paragraph{Availability taxonomy}\label{sec:lr-avail-taxonomy}
\begin{table}[H]
    \centering
    	\caption{
    \glspl{lr} Availability Categories: Each category corresponds to specific criteria applied to the resource's URL, license and data format.}
    \label{tab:lr-availability}
    \resizebox{\textwidth}{!}{
    \begin{tabular}{|l | l | l | l| l |}	 
	\hline
	\textbf{Availability} & \textbf{Description} & \textbf{Provided URL} & \textbf{License}  &\textbf{Data format}\\  
	\hline
	Yes & publicly available& valid & yes (open license) & machine-actionable\\  
	\hline
	Lmt & limited public availability & valid & no license or available upon request or pay & machine-actionable\textsuperscript{a}\\  
	\hline
	Err & publicly unavailable& invalid & n/a & n/a\\  
	\hline
	No & no information provided & no URL & n/a & n/a\\  
	\hline
    \end{tabular}
    }
\textsuperscript{a} For Lmt, when the \gls{lr} is available upon request, the data format is unknown unless specified in the paper.
\end{table}

The \glspl{lr} availability classification scheme is based on three parameters: the presence of a functional URL, valid license information, and a machine-actionable format. We identified the resources' URLs from the papers in which they were created, without extending our search to other web sources. The scheme presented in Table~\ref{tab:lr-availability} classifies \glspl{lr} availability into four distinct categories. The value ``Yes'' signifies resources with a valid, functional URL and a defined license, such as Creative Commons. We do not evaluate license restrictions, as even restrictive licenses provide more legal clarity and alignment with FAIR principles than the absence of a license, which creates significant legal uncertainty. These datasets and lexica are also in a machine-actionable format (e.g., txt, csv, pkl). 
The designation ``Lmt'' is used for \glspl{lr} with limited availability, referring to resources with valid URLs but no license terms, resources provided upon request, or accessible for a fee (e.g., tweets). Their data format is machine-actionable, except for those available upon request, for which their format readiness could not be verified. The value ``Err'' signifies resources for which the authors provided URLs which were found to be inaccessible due to broken links or other HTTP errors. Lastly, the value ``No'' is assigned to resources for which the creators did not provide URLs.

\paragraph{Annotation Type Taxonomy}\label{sec:lr-ann-status-taxonomy}

\begin{table}[H]
    \centering
    \caption{
    \glspl{lr} Annotation types reflecting varying levels of curation and automation.}
    \label{tab:lr-ann-status}
    \begin{tabular}{|l| l |} 
    \hline
    \textbf{Annotation type} & \textbf{Description}\\  
    \hline
    manual & human annotation \\  
	\hline
    automatic & automatic annotation\\  
	\hline
    hybrid & manual and automatic annotation\\  
	\hline
    user-generated & annotation from user edits, not curated\\  
	\hline
    curated  & metadata provided by distributor \\  
	\hline
    no annotation & no annotation \\  
	\hline
    \end{tabular}
\end{table}

The classification scheme for annotation types includes six categories as outlined in Table~\ref{tab:lr-ann-status}. Manual annotations are performed by human annotators, offering high accuracy and often serving as the gold standard. In contrast, automatic annotations are generated using algorithms or predefined rules, ensuring consistency and scalability. Hybrid annotations combine both manual and automatic methods, such as performing automatic annotation followed by manual correction and validation. User-generated annotations come from real-world interactions, like hotel review ratings from users. Curated datasets feature metadata sourced from distributors, enriching datasets with structured information like topics from news articles or author details from publishers. Finally, ``No Annotation'' refers to datasets that contain unprocessed text with no annotations.

\end{section}

\section{Track: Machine Learning for NLP}\label{sec:ml-nlp}
This section marks the beginning of the discussion on track-specific research in NLP. It focuses on Machine Learning for \gls{nlp} and the Interpretability and Analysis of Models for \gls{nlp}. \textbf{\gls{ml} for \gls{nlp}} track explores how \gls{ml} techniques are integrated to improve the ability of computers to understand, interpret, and generate human language. \textbf{Interpretability and Analysis of Models for \gls{nlp}} is rooted in the rise of \gls{dl}, which has changed radically \gls{nlp}. The use of \glspl{nn} became the dominant approach. However, their opaque nature poses challenges in understanding their inner workings, prompting a surge in research on analyzing and interpreting \gls{nn} models in \gls{nlp}.\cite{belinkov2020interpretability}

\subsection{Machine Learning for NLP in Greek: Language Models and Methods}\label{sec:ml-nlp-lms}
\subsubsection{ML vs DL approaches} 
\paragraph{\gls{ml} approaches} 
The predominant approach to Greek \gls{nlp} research in 2012 relied primarily on \gls{ml} algorithms. Given the morphological richness of the Greek language, feature engineering was a key step in traditional \gls{ml}.  Typically, a structured pipeline was followed for extracting additional features, such as \gls{pos} tags, lemmas, or word stems. Additionally, features such as named entities, dependency trees, and, more recently, word embeddings were often extracted. Most of the surveyed studies using a \gls{ml} approach derived features from frequency-based methods, such as n-grams and lexicons (used in 41 studies), or extracted information such as \gls{pos} tags, lemmas, stems, named entities, or dependency trees (used in 28 studies). Furthermore, most methods that employed word embeddings also used additional features (11 out of 15). 

Regarding word embeddings, \citet{prokopidis2020neural} trained fastText\cite{bojanowski2017enriching} on newspaper articles and the Greek part of the w2c corpus (see \S\ref{sec:res-and-eval}). Similarly, \citet{tsakalidis2018building} trained Word2Vec\cite{mikolov2013distributed} on political Greek tweets (see \S\ref{sec:sentiment}). Both sets of trained word embeddings are publicly available for research use. For the other features used in \gls{ml} approaches, the corresponding tools developed by the surveyed papers are presented in various \gls{nlp} track sections, according to the \gls{nlp} task they address. For example, tools related to syntax are presented in \S\ref{sec:syntax}, and tools for \gls{ie}, such as \gls{ner}, are discussed in \S\ref{sec:ie}.

\paragraph{Early \gls{dl} Approaches}
The adoption of \gls{rnn}-based methods in Greek \gls{nlp} began in 2017
 with the introduction of \gls{rnn}-based methods\cite{athanasiou2017,Pavlopoulos2017deep,Pavlopoulos2017improved} and \gls{cnn}-based methods.\cite{Medrouk2017,Pavlopoulos2017deep} \gls{rnn}-based methods became prevalent in Greek \gls{nlp}, and when \gls{ml}-based approaches were compared to \gls{rnn}-based ones, the latter consistently outperformed the former.\cite{Pitenis2020, Calderón2022}

\begin{table}[ht]
    \centering
    \caption{Monolingual Greek \glspl{plm}, including their availability (Yes: publicly available, Lmt: limited availability; see Table~\ref{tab:lr-availability} for details; the citations point to URLs) and the backbone model they are based on.}
    \label{tab:pfms}
    \begin{tabular}{|l | l | l |}	 
	\hline
	\textbf{Authors} & \textbf{Availability} & \textbf{Backbone} \\ 
	\hline	
    \citet{giarelis2024greekt5} & Yes\cite{imis_greekt5_mt5_small}& mT5 \\ 
    & Yes\cite{imis_greekt5_umt5_small}& umT5 \\ 
    & Yes\cite{imis_greekt5_umt5_base}& umT5 \\ 
	\hline
    \citet{evdaimon2023greekbart} & Yes\cite{GreekBART}& BART \\ 
	\hline
    \citet{koutsikakis2020} & Yes\cite{GreekBERT}& BERT \\ 
	\hline
    \citet{zaikis2023pima} & Lmt\cite{GreekMediaBERT}& BERT \\ 
	\hline
    \citet{Alexandridis2021Survey} & Lmt\cite{GreekSocialBERT2023}& BERT \\ 
	\hline
    \citet{Alexandridis2021Survey} & Lmt\cite{PaloBERT2023}& RoBERTa \\ 
	\hline
    \citet{Perifanos2021} & Lmt\cite{BERTaTweetGR2023}& RoBERTa \\ 
	\hline
    \end{tabular}
\end{table}

\paragraph{\glspl{plm}}
\glspl{plm} following the Transformer architecture have been pivotal in recent advancements in Greek \gls{nlp}. Table~\ref{tab:pfms} lists the publicly available Greek \glspl{plm} developed for the studies surveyed. These models address tasks in both \gls{nlu} and \gls{nlg} (see Appendix~\ref{sec:nlu-nlg}). Among the monolingual \glspl{plm} designed for \gls{nlu} tasks such as \gls{sa}, GreekBERT\cite{koutsikakis2020} has emerged as a standard in Greek \gls{nlp} research. It is recognized as state-of-the-art in several studies.\cite{koutsikakis2020, Perifanos2021, Alexandridis2021, Rizou2022, Bilianos2022, Kapoteli2022,evdaimon2023greekbart} GreekBERT uses the BERT-BASE-UNCASED architecture\cite{devlin-etal-2019-bert} and was pre-trained on 29 GB of Greek text from the Greek Wikipedia,\cite{greek_wikipedia_dumps} the Greek part of the European Parliament Proceedings Parallel Corpus (Europarl),\cite{koehn2005europarl} and the Greek part of OSCAR,\cite{suarez2019asynchronous} a clean version of Common Crawl.\cite{common_crawl} There are two fine-tuned variants of GreekBERT: Greek Media BERT,\cite{zaikis2023pima} which is fine-tuned on media domain data, and GreekSocialBERT,\cite{Alexandridis2021Survey} which is fine-tuned on Greek social media data. Additionally, PaloBERT,\cite{Alexandridis2021Survey} trained on social media data, and BERTaTweetGR,\cite{Perifanos2021} trained on tweets, are two monolingual models based on the RoBERTa architecture and they also address \gls{nlu} tasks. On the other hand, there are two monolingual \glspl{plm} based on the encoder-decoder architecture (see \S\ref{sec:time}), which are capable of performing all \gls{nlu} and \gls{nlg} tasks. GreekBART,\cite{evdaimon2023greekbart} based on the BART architecture,\cite{lewis2020bart} was pre-trained on the same datasets as GreekBERT plus the Greek Web Corpus,\cite{outsios2018word} incorporating diverse Greek text types, as well as formal and informal text, to enhance robustness. The GreekT5 series of models\cite{giarelis2024greekt5} was fine-tuned on the GreekSUM training dataset,\cite{evdaimon2023greekbart} using the multilingual T5 \glspl{lm}, which comprise (google/mt5-small,\cite{xue2021mt5} google/umt5-small,\cite{chungunimax} and google/umt5-base).\cite{chungunimax}

\subsubsection{Interpretability and Analysis of Models for NLP}
Research concerning interpretability and analysis of \gls{nn} models for Greek \gls{nlp} spans various languages and is quite diverse. 
\citet{papadimitriou2023multilingual} investigated grammatical structure bias in multilingual \glspl{lm}, examining how higher-resource languages influence lower-resource ones. They compared Greek and Spanish monolingual BERT models with \gls{mbert},\cite{devlin-etal-2019-bert} which is trained predominantly on English. The study found that \gls{mbert} tends to adopt English-like sentence structures in Spanish and Greek. They tested this phenomenon on the subject-verb order in Greek, which exhibits free word order (see \S\ref{ssec:MG}).
\citet{ahn2021mitigating} examined ethnic bias in BERT models across eight languages, including Greek, examining how these models reflect historical and social contexts. They proposed mitigation methods and highlighted the language-specific nature of ethnic bias. \citet{gari2021let} proposed a method to assess whether \glspl{plm} for multiple languages (including Greek) have knowledge of lexical polysemy, demonstrating their capabilities through empirical evaluation. The source code is available.\cite{monopoly_repo}
\citet{Gonen2020} revealed the inherent understanding of \gls{mbert} for word-level translations and its capacity of cross-lingual knowledge transfer, despite the fact that it is not explicitly trained on parallel data. The source code is available.\cite{mbert_repo}

\begin{figure}[H]
\centering
\includegraphics[width=\columnwidth]{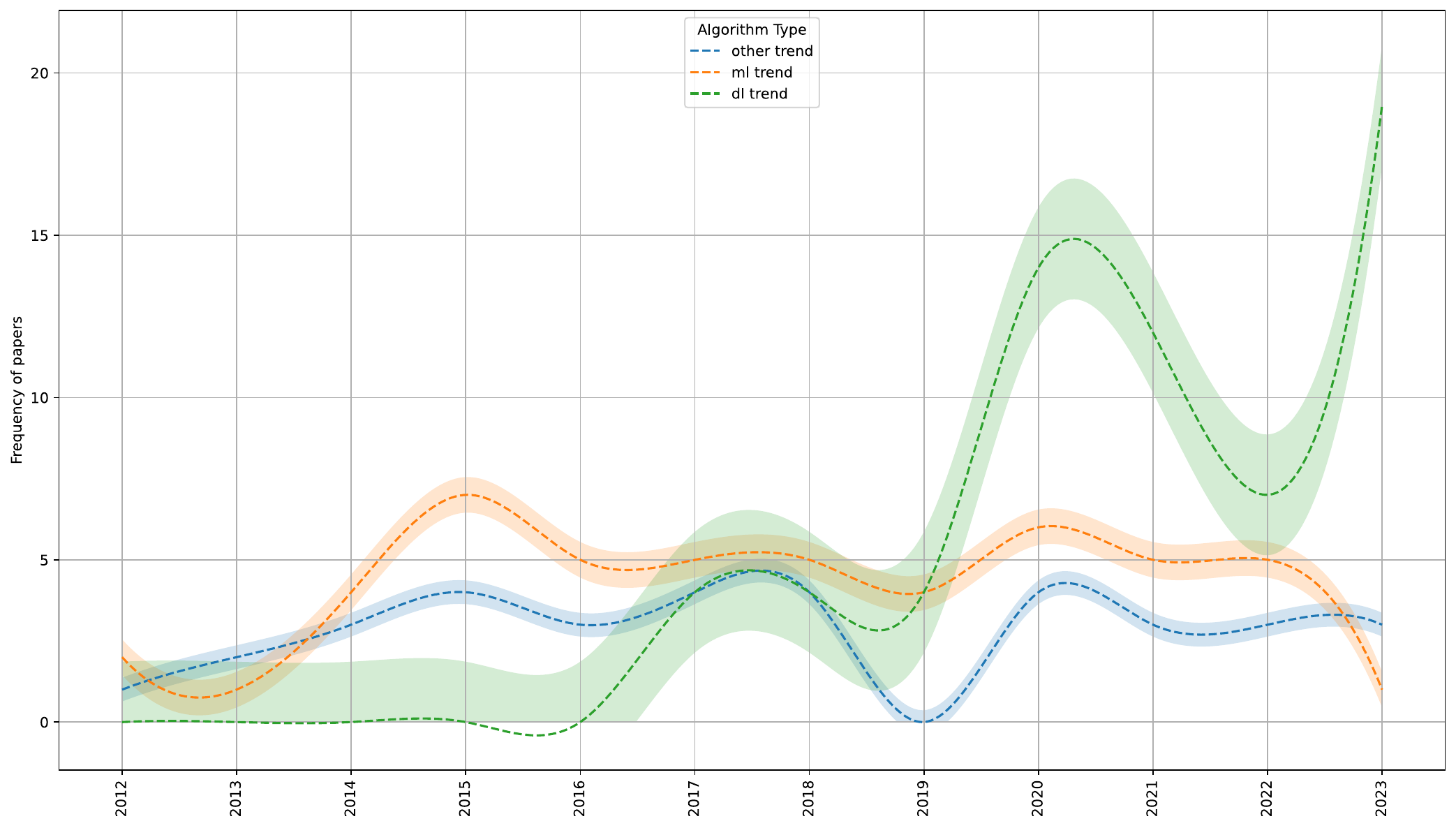}
\caption{Frequency of \gls{nlp} approaches, shown as the number of papers using each approach over the years. The approaches include \gls{dl}, traditional \gls{ml}, and other methods such as rule-based systems.}
\label{fig:ai-approach-perc}
\end{figure}

\subsection{Summary of Machine Learning for NLP in Greek}
Recently, \gls{nlp} research has increasingly been based on \glspl{llm}, with some of the most popular ones being either fully or partially closed-source.\cite{balloccu-etal-2024-leak} Notable examples for Greek include OpenAI's GPT-3.5 and GPT-4.0,\cite{achiam2023gpt} which are trained on multilingual data and can therefore process and generate texts in multiple languages, including Greek. Additionally, there are other multilingual \glspl{plm} available in open-source environments, such as \gls{xlmr}\cite{conneau2019unsupervised} used by \citeauthor{evdaimon2023greekbart},\cite{evdaimon2023greekbart} \citeauthor{Ranasinghe2021},\cite{Ranasinghe2021} \citeauthor{koutsikakis2020};\cite{koutsikakis2020} \gls{mbert}\cite{devlin-etal-2019-bert} used by \citeauthor{ahn2020nlpdove},\cite{ahn2020nlpdove} \citeauthor{koutsikakis2020};\cite{koutsikakis2020} Flan-T5-large\cite{chung2024scaling} used by \citeauthor{zampieri2023offenseval},\cite{zampieri2023offenseval} and the recent GR-NLP-Toolkit.\cite{loukas-etal-2025-gr} New \glspl{plm} emerge regularly in multilingual and monolingual settings, such as GreekBART,\cite{evdaimon2023greekbart} the GreekT5 series of models,\cite{giarelis2024greekt5} the Mistral-based Meltemi-7B,\cite{voukoutis2024meltemi} and Llama-Krikri.\cite{LlamaKrikri2024} Although covering all \glspl{plm} for Greek is beyond the scope of our study, we highlight the significance of GreekBERT, which has significantly impacted Greek \gls{nlp} research since its introduction in 2020, leading to a shift from traditional \gls{ml} to \gls{dl} approaches.

\paragraph{Historical evolution} 
Figure~\ref{fig:ai-approach-perc} shows the trends of Greek \gls{nlp} approaches, categorized into traditional \gls{ml} methods, \gls{dl} methods, and other non-\gls{ml} methods, such as rule-based systems. Traditional \gls{ml} methods remained the dominant approach until 2019, with the exception of 2013 when other methods were favored. From 2017 onwards, researchers began to use and compare both \gls{ml} and \gls{dl} approaches. As mentioned in \S\ref{sec:ml-nlp-lms}, in 2017, the first publications employing \gls{dl} techniques emerged, primarily focusing on \gls{rnn}-based and \gls{cnn}-based models, which accounted for approximately 30\% of the total papers published that year. Since the release of GreekBERT,\cite{koutsikakis2020} \gls{dl} methodologies have surpassed traditional \gls{ml} approaches in usage. While \gls{ml} methods still find applications, a significant portion of the studies employing \gls{ml} techniques, integrate both \gls{ml} and \gls{dl} techniques in their research experiments.

\begin{figure}[H]
\centering
\includegraphics[width=\columnwidth]{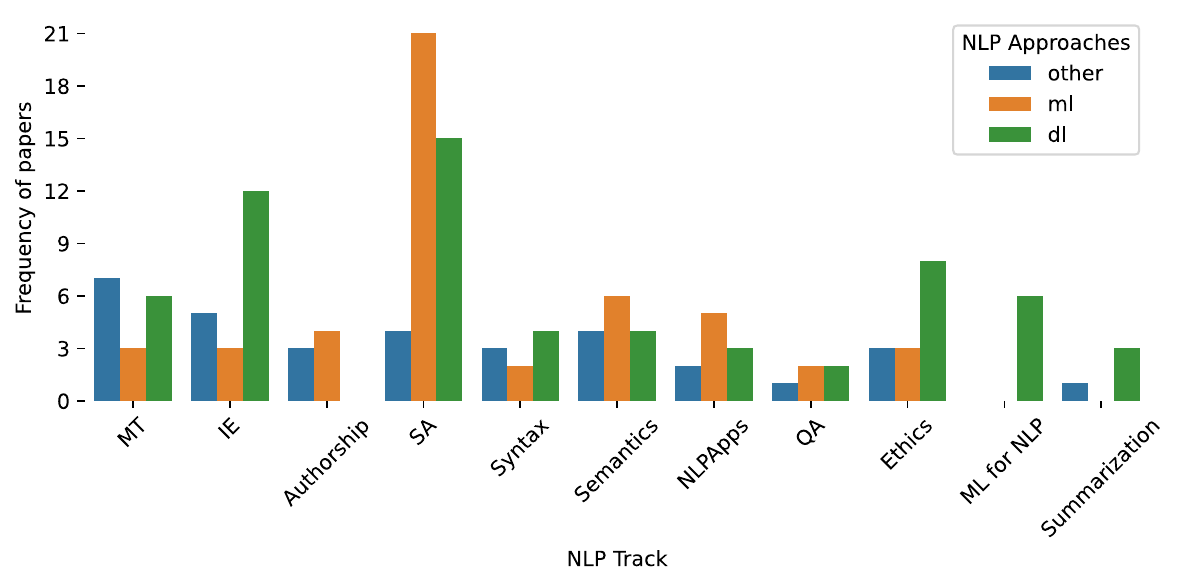}
\caption{Number of papers per \gls{nlp} track per approach (\gls{ml}, \gls{dl}, other) since 2017, the year when a study could follow multiple approaches (e.g., both \gls{ml} and \gls{dl}).}
\label{fig:ai-approach-per-track}
\end{figure}

What we also observe in Figure~\ref{fig:ai-approach-perc} is a decline in research output between 2020 and 2022, particularly in studies adopting \gls{dl} approaches, followed by an increase thereafter. Several factors might explain this temporary drop. First, the COVID-19 pandemic led to disruptions in research. Labs, conferences, and collaborative projects slowed down or paused during 2020–2021. Also, many researchers pivoted to pandemic-related applications of AI or public health instead of language-specific \gls{nlp}. At the same period, the explosion of large-scale pretraining (BERT, GPT, T5) heavily favored English and multilingual benchmarks like XGLUE\cite{liang2020xglue} or XTREME,\cite{hu2020xtreme} which often provide only shallow Greek coverage. Therefore, researchers might have preferred to contribute to multilingual efforts instead of monolingual Greek projects, effectively lowering visibility of Greek-focused work. Collectively, these elements may explain the observed short-term dip, without necessarily implying long-term stagnation.

\paragraph{NLP approaches per track} Figure~\ref{fig:ai-approach-per-track} illustrates the number of the surveyed papers (published from 2017 onward) across \gls{nlp} tracks, categorized by their \gls{nlp} approach. The starting point of 2017 reflects the emergence of \gls{dl} approaches in Greek \gls{nlp}, allowing for a clearer view of their integration across different tracks. We observe that Ethics and \gls{nlp}, \gls{ie}, Syntax, and Summarization are predominantly addressed using \gls{dl} techniques. On the other hand, \gls{qa}, \gls{sa}, \gls{mt}, Semantics, and \gls{nlp} Applications incorporate both traditional \gls{ml} and \gls{dl} approaches, either within the same study or across different studies focusing on the same task. Notably, Authorship Analysis is the only track where \gls{dl} techniques are not employed. Additionally, \gls{ml} for \gls{nlp} is a recently introduced track, consisting solely of papers that adopt \gls{dl} approaches.

\section{Track: Syntax and Grammar}\label{sec:syntax}

\textbf{Syntactic processing} encompasses various subtasks in \gls{nlp} focused on phrase and sentence structure, as well as the relation of words and constituents to each other within a phrase or sentence.\cite{woolf2010building} It involves recognition of sentence constituents, identification of their syntactic roles, and potentially establishment of the underlying semantic structure. These features are valuable for \gls{nlu},\cite{cambria2017sentiment} a topic further discussed in Appendix~\ref{sec:nlu-nlg}. Additionally, syntactic processing serves as a pre-processing step for more complex \gls{nlp} tasks, such as \gls{sa} and error correction among others.\cite{zhang2023survey} \noindent\textbf{\gls{gec}} is a user-oriented task that aims for automatically correcting diverse types of errors present in a given text, encompassing violations of rules pertaining to morphology, lexicon, syntax, and semantics.\cite{wang2021comprehensive} \gls{gec} can be used to enhance fluency, render sentences in a more natural manner, and align with the speech patterns of native speakers.\cite{wang2021comprehensive}

\subsection{Syntax and Grammar in Greek: Language Models and Methods}

\paragraph{Syntactic Processing in Greek} This task is related to sentence splitting, tokenization, and  morphosyntactic processing, including \gls{pos} tagging, lemmatization, and dependency parsing. 
\citet{prokopidis2020neural} addressed several syntax tasks, using the pre-trained Punkt model\cite{kiss2006unsupervised} for sentence splitting and a \gls{bilstm} tagger using the StanfordNLP library\cite{qi2019universal} for \gls{pos} tagging. Lemmatization involved a lexicon-based approach with a \gls{bilstm} lemmatization model as a fallback for out-of-lexicon words. For dependency parsing, the authors trained a neural attention-based parser\cite{dozat2016deep} on the Greek \gls{ud} treebank.\cite{prokopidis2017universal} On the same dataset, \citet{koutsikakis2020} performed \gls{pos} tagging using Transformer-based models, namely their GreekBERT, \gls{xlmr}, and two variants of \gls{mbert}, concluding that all four have comparable performance in terms of Accuracy.
\citet{Partalidou2019} conducted \gls{pos} tagging and \gls{ner} tasks, with the details of their \gls{ner} system summarized in \S\ref{sec:ie}. For \gls{pos} tagging they used spaCy,\cite{honnibal2017spacy} adhering to the \gls{ud} annotation schema. Additionally, they assessed the model's tolerance towards Out-of-Vocabulary (OOV) words and found that it lacked flexibility in handling such instances. Widely used \gls{nlp} pipelines in the surveyed papers are:  
an ILSP suite of \gls{nlp} tools,\cite{prokopidis2011suite} the Natural Language Toolkit (NLTK),\cite{bird2006nltk} polyglot,\cite{Polyglot2015} 
spaCy for Greek,\cite{spacy_greek_models,gsoc2018_spacy_greek}   
Stanza,\cite{Qi2020} 
and UDPipe.\cite{straka2017tokenizing} Additional research in the field of Syntax explored hybrid embeddings  proposed by \citet{zuhra2023hybrid} to enhance dependency parsing for morphologically rich, free word order languages, including Greek, using \gls{ud} treebanks. These hybrid embeddings were based on \gls{pos} tags and morphological features, significantly improving parsing accuracy. \citet{wong2014isentenizer} developed a multilingual sentence boundary detection method based on an incremental decision tree learning algorithm. Furthermore, while \citet{fotopoulou2015mwes} and \citet{samaridi2014} dealt with verbal \glspl{mwe}, the former study aimed at defining formal criteria for classifying verbal \glspl{mwe} as either idiomatic expressions or Support Verb Constructions (consisting of a support verb and a predicative noun). In contrast, the latter focused on parsing \glspl{mwe} using the \glspl{lfg/xle} framework, extending their analysis beyond traditional syntactic boundaries by incorporating lexical knowledge from lexicons.

\paragraph{\gls{gec} in Greek} \citet{korre2021elerrant} focused on the correction of grammatical errors that vary from grammatical mistakes to  punctuation, spelling, and morphology of word. The authors listed 18 main categories of grammatical errors that systems can correct, also developing a rule-based annotation tool for Greek. The tool takes an original erroneous sentence along with its correction as input. Then, it automatically produces an annotation that mainly consists of the error location and type, as well as its correction.
\citet{gakis:2017} created a rule-based grammar checker tool,\cite{neurolingo_ggc} which analyzes and corrects syntactic, grammatical, and stylistic (i.e., the formal, informal, or oral style of language used) errors in sentences, providing users with error notifications and correction hints. \citet{kavros2022soundexgr} focused on spelling errors, addressing the issue of misspelled and mispronounced words in Greek. They employed phonetic algorithms to assign the same code to different word variations based on phonetic rules. For example, they successfully grouped \textgreek{μήνυμα} (correct spelling) with \textgreek{μύνημα} (both sounding as /mínima/). They reported better results compared to stemming and edit-distance approaches. The source code is available.\cite{soundexgr_repo}

\subsection{Syntax and Grammar in Greek: Language Resources}
Table~\ref{tab:syntax-datasets} displays the pertinent monolingual \glspl{lr} for this track. It shows three publicly available resources for \gls{gec} and two resources for Syntax, of which one is publicly available.
For \gls{gec}, \citet{kavros2022soundexgr} created word lists, containing words and their misspellings. These misspellings were generated through the addition, deletion, or substitution of a letter, as well as by incorporating words with similar sounds. 
\citet{korre2021elerrant} developed two datasets, namely the Greek Native Corpus (GNC) and the Greek Wiki Edits (GWE). GNC is comprised of essays written by students who are native speakers of Greek, totaling 227 sentences. Each sentence within this dataset may contain zero, one, or multiple grammatical errors, all annotated with the corresponding grammatical error types as defined in the provided annotation schema. On the other hand, GWE consists of sentences extracted from WikiConv.\cite{hua-2018-wikiconv} Each sentence in this dataset includes the original sentence, the edited sentence, the original string that underwent editing, and the specific grammatical error type. 

Regarding Syntax, \citet{prokopidis2017universal} provided the Greek \gls{ud} treebank as part of the \gls{ud} project,\cite{nivre2016universal} a project that offers standardized treebanks with consistent annotations across languages. The dataset includes syntactic dependencies, \gls{pos} tags, morphological features, and lemmas. Derived from the Greek Dependency Treebank,\cite{prokopidis2005theoretical} it contains 2,521 sentences split into training (1,622), development (403), and test (456) sets, and was manually validated and corrected. \citet{gakis2015analysis} collected a corpus consisting of 2.05M tokens derived from student essays, literary works, and newspaper articles. They extracted morphosyntactic information automatically for this corpus with the help of a lexicon.\cite{gakis2012}

\begin{table}[H]
    \centering
    \caption{\glspl{lr} related to \gls{gec} and Syntax, with information on availability (Yes: publicly available, No: no information provided; see Table~\ref{tab:lr-availability} for details; the citations point to URLs), annotation type (see Table~\ref{tab:lr-ann-status} for details), size, size unit, and text type.}
\label{tab:syntax-datasets}
    \resizebox{\textwidth}{!}{
    \begin{tabular}{|l | l | l | l| l | p{4cm} |}	 
    \hline
    \textbf{Authors} & \textbf{Availability} & \textbf{Ann. type} & \textbf{Size}  &\textbf{Size unit} & \textbf{Text type} \\  
    \hline
    \citet{kavros2022soundexgr} & Yes\cite{soundexgr_repo} &  automatic & 1,086&  word&  word list\\ 
    \hline
    \citet{korre2021elerrant} & Yes\cite{elerrant} &  manual &227&  sentence &  student essay\\ 
    & Yes\cite{elerrant}&  user-generated & 100&  sentence&  Wikipedia Talk Page\\ 
    \hline
    \citet{prokopidis2017universal} & Yes\cite{ud-greek-gdt} &  hybrid & 2,521 &  sentence &  Wikinews, european parliament sessions\\
    \hline
    \citet{gakis2015analysis} & No &  automatic &2.05M& token&  essay, literature, news\\ 
    \hline
    \end{tabular}
    }
\end{table}

\subsection{Summary of Syntax and Grammar in Greek}

Traditionally, syntactic processing served as a pre-processing step for higher-level \gls{nlp} tasks (\S\ref{sec:ml-nlp}). However, in the era of \gls{dl}-based \gls{nlp}, syntactic processing is often neglected. Instead, \glspl{nn} are leveraged to implicitly capture syntactic information, surpassing the performance of symbolic methods that rely on manually hand-crafted features. This is also reflected by the number of ACL submissions related to Syntax  (i.e., Tagging, Chunking and Parsing), which is significantly shrinking.\cite{rohatgi2023acl} Our study partially reflects this trend, showing a slight decline in focus on syntactic tasks since 2020, though they remain active. Notably, the Syntax and Grammar track, alongside the \gls{ie} track (see \S\ref{sec:ie}), has the highest number of publicly available \glspl{lr} for Greek and the largest proportion of publicly available \glspl{lr} among all task-related \glspl{lr}.

\section{Track: Semantics}\label{sec:semantics}

The meaning in language is the focus of Semantics. In the context of \gls{nlp}, semantic analysis aims to extract, represent, and interpret meaning from textual data, bridging the gap between natural language and machine understanding.\cite{goddard2011semantic}
Semantic analysis can operate at three different levels, each focusing on different units of examination: lexical semantics, sentence-level semantics, and discourse analysis. \textbf{Lexical Semantics} pertains to the understanding of word meanings, including their various senses, relationships with other words, and roles in different linguistic contexts.\cite{cruse1986lexical} \textbf{Sentence-level Semantics} considers the meaning of individual sentences or phrases in terms of their internal structure and relationships. \textbf{Discourse Analysis}, on the other hand, deals with understanding the meaning in a broader textual context, beyond individual sentences.\cite{prasad2008penn} It involves analyzing how sentences connect and influence each other within the context of a text or a conversation.

At the core of Lexical Semantics lies the task of Distributional Semantics, which is the leading approach to lexical meaning representation in \gls{nlp}.\cite{lenci2022comparative} Founded upon the distributional hypothesis,\cite{harris1954distributional,sahlgren2008distributional} which suggests that words sharing similar linguistic contexts also share similar meanings, Distributional Semantics employs real-valued vectors, commonly known as embeddings, to encode the linguistic distribution of lexical items within textual corpora. As \citet{lenci2022comparative} explain, this field has progressed through three key generations of models: (i) count-based \glspl{dsm}, which form distributional vectors based on co-occurrence frequencies that adhere to the \gls{bow} assumption; (ii) prediction-based \glspl{dsm}, employing shallow neural networks to learn vectors by predicting adjacent words, yielding dense, static word embeddings (or simply word embeddings); and (iii) contextual \glspl{dsm}, harnessing deep neural language models to generate inherently contextualized vectors for each word token (e.g., word embeddings extracted from BERT-based models). The evolution from earlier static \glspl{dsm}, which learn a single vector per word type, to contextual \glspl{dsm} is further examined in \S\ref{sec:time}.

\subsection{Semantics in Greek: Language Models and Methods}

\paragraph{Lexical Semantics}
Studies focusing on Lexical Semantics in Greek address the following tasks: building \glspl{dsm}, \glspl{dsm} evaluation, diachronic semantic shifts of words, word sense induction, lexical ambiguity, metaphor detection, and semantic annotation.

\citet{zervanou2014word} used \gls{bow} representations to study the impact of morphology on unstructured count-based \glspl{dsm}. They proposed a selective stemming process, by using a metric to determine which words to stem, demonstrating improved performance in morphologically rich languages such as Greek. \citet{Palogiannidi2015,palogiannidi2016} used semantic similarity and \gls{bow} representations of seed words to estimate the ratings of unknown words, applying their method on affective lexica of five different languages, including Greek. 
\citet{iosif-etal-2016-cognitively} proposed word embeddings inspired by cognitive processes in human memory,\cite{kahneman2013thinking} showing that they outperform \gls{bow} representations. \citet{Lioudakis2019} introduced the \gls{cbos} method for generating word representations, combining Continuous Skip-gram with \gls{cbow}, and assessing its performance across various tasks (word analogies, word similarity, etc.). The source code is available.\cite{mikeliou_greek_word_embeddings}

\citet{outsios2020-evaluation} performed \textit{evaluation} of various word embeddings trained on diverse data sources. The evaluation framework considered tasks involving word analogies and similarity. \citet{dritsa2022greek} and \citet{Barzokas2020} investigated the \textit{diachronic semantic shifts of words} with the use of Distributional Semantics. \citet{dritsa2022greek} constructed a dataset from Greek Parliament proceedings (further discussed in \S\ref{sec:res-and-eval}). They also applied four state-of-the-art semantic shift detection algorithms, namely Orthogonal Procrustes,\cite{hamilton2016diachronic} Compass,\cite{di2019training} NN,\cite{gonen-etal-2020-simple} and Second-Order Similarity,\cite{hamilton2016cultural} to identify word usage change across time and among political parties. \citet{Barzokas2020} compiled a corpus of e-books (presented in \S\ref{sec:res-and-eval}), trained word embeddings, and used k-nearest neighbors along with cosine distances to trace semantic shifts aiming to capture both linguistic and cultural evolution.

\citet{gari2021let} introduced an approach to analyze lexical polysemy knowledge in \glspl{plm} across various languages, including Greek. They found that contextual \gls{lm} representations, like BERT, encode information about lexical polysemy, and they performed \textit{word sense induction} by enabling interpretable clustering of polysemous words based on their senses. On the other hand, \citet{gakis2015analysis} analyzed \textit{lexical ambiguity} using morphosyntactic features from a lexicon.\cite{gakis2012} They categorized ambiguous words based on their spelling and etymology. \citet{Florou2018} focused on \textit{metaphor detection} using the discriminative model of \citeauthor{steen2007finding},\cite{steen2007finding} identifying the literal and metaphorical functions of phrases through the optimal separation of hyperplanes in vector representations of word combinations. 

\citet{chowdhury2014cross} addressed the challenge of transferring \textit{semantic annotations} from a source language corpus (Italian) to a target language (Greek) using crowd-sourcing. They introduced a methodology to evaluate the quality of crowd-annotated corpora by considering inter-annotator agreement for evaluation of annotations within the target language, whereas cross-language transfer quality is evaluated by comparison against source language annotations.  

\paragraph{Sentence-Level Semantics}

We identified two tasks in Greek \gls{nlp} that fall under Sentence-Level Semantics: Semantic Parsing and \gls{nli}. Semantic Parsing involves converting natural language utterances into logical forms that can be executed on a knowledge base.\cite{kamath2018survey} \citet{li2015improving} tackled this task using \glspl{scfg}, which model language relationships by deriving coherent logical forms. They enhanced the \gls{scfg} framework by extending the translation rules with informative symbols, achieving state-of-the-art performance in English, Greek, and German on a benchmark dataset. In contrast, \gls{nli} focuses on assessing the logical relationship between sentence pairs, determining if one sentence entails, contradicts, or is neutral with respect to another. \citet{koutsikakis2020} evaluated this task using the Greek part of the XNLI corpus,\cite{conneau2018xnli} comparing their model GreekBERT with \gls{xlmr}, two variants of \gls{mbert}, and the Decomposable Attention Model (DAM).\cite{parikh2016decomposable} They found that GreekBERT outperformed the other models. Three years later, \citet{evdaimon2023greekbart} fine-tuned their model, GreekBART, on the XNLI training split and compared it with GreekBERT and \gls{xlmr} on the test split, concluding that GreekBART achieved results comparable to GreekBERT.

\paragraph{Discourse Analysis} The only study identified that performs Discourse Analysis is by \citeauthor{giachos2023systemic}.\cite{giachos2023systemic} This study focused on how the robot processes and understands sentences in context, teaching the robot to handle incomplete information and enabling a word learning procedure, beginning with 200 Greek words as a seed dictionary.

\subsection{Semantics in Greek: Language Resources}

Table~\ref{tab:semantics-datasets} presents the  \glspl{lr} for semantics-related tasks, along with their availability (classified according to Table~\ref{tab:lr-availability}), annotation type (classified as per  Table~\ref{tab:lr-ann-status}), linguality type, size, and size unit. By contrast to Syntax and Grammar (\S\ref{sec:syntax}), only one \gls{lr} regarding Semantics is publicly available. The rest six are either of limited availability (Lmt), could not be accessed (Err), or were not publicly available (No). 

\begin{table}[H]
    \centering
    \caption {\glspl{lr} related to Semantics, with information on availability (Yes: publicly available, Lmt: limited  availability, Err: unavailable, No: no information provided; see Table~\ref{tab:lr-availability} for details; the citations point to URLs), annotation type (see Table~\ref{tab:lr-ann-status} for details),  linguality type, size, and size unit (with size denoting the portion in Greek for multilingual datasets).
    }
    \label{tab:semantics-datasets}
    \resizebox{\textwidth}{!}{
    \begin{tabular}{|l | l | l | l| l | l | l |}	 
	\hline
	\textbf{Authors} & \textbf{Availability} & \textbf{Ann. type} & \textbf{Size}  & \textbf{Size unit} & \textbf{Linguality} \\  
	\hline
        \citet{ganitkevitch2014multilingual} & Yes\cite{paraphrase} & hybrid & 22.3M & paraphrase & multilingual\\  
	\hline
        \citet{gari2021let} & Lmt\cite{monopoly_repo} &  automatic & 418 & word & multilingual\\  
	\hline
        \citet{outsios2020-evaluation}& Err\cite{aueb_outsios2020ev_353tr} & manual & 353 & word-pair & monolingual\\  
        & Err\cite{aueb_outsios2020ev_Test} & automatic & 39,174 & word analogy question & monolingual\\  
	\hline
        \citet{pilitsidou2021frame}& No & manual & 73,069 & token & bilingual\\  
	\hline
        \citet{giouli2020greek} & No & manual & 3,012 & token & monolingual\\  
	\hline
        \citet{Florou2018} & No & manual & 914 & sentence & monolingual\\  
	\hline
    \end{tabular}
    }
\end{table}

The only publicly available resource is that of \citeauthor{ganitkevitch2014multilingual},\cite{ganitkevitch2014multilingual} who expanded the Paraphrase Database (PPDB)\cite{ganitkevitch2013ppdb} with paraphrases in 23 languages, including Greek. The original database contains human-annotated paraphrases in English. For the additional languages, \citet{ganitkevitch2014multilingual} extracted paraphrases using parallel corpora. This study was not mentioned earlier in this section because there was no other contributions except for the introduction of this \gls{lr}.  
\citet{gari2021let} offered a multilingual dataset comprising words, their corresponding senses, and sentences featuring the word in its specific sense. In the case of the Greek part of the corpus, sentences were extracted from the Eurosense corpus,\cite{bovi2017eurosense} which contains texts from Europarl, automatically annotated with BabelNet word senses.\cite{navigli2012babelnet} 
\citet{outsios2020-evaluation} translated to Greek the benchmark dataset WordSim353,\cite{finkelstein2001placing} which contains word pairs along with human-assigned similarity judgments. Additionally, they assembled 39,174 analogy questions to conduct word analogy tests, measuring word similarity in a low-dimensional embedding space.\cite{mikolov2013efficient} 
\glspl{lr} of the three studies that were not publicly available were about the Greek counterpart of the Global FrameNet project,\cite{giouli2020greek} a bilingual frame-semantic lexicon for the financial domain,\cite{pilitsidou2021frame} and a corpus that consists of sentences using the same transitive verbs in both metaphorical and literal contexts.\cite{Florou2018}

\subsection{Summary of Semantics in Greek}

Studies pertaining to Semantics can be found throughout the period under investigation (2012-2023), but most were published in early years (2013-2016; see Figure~\ref{fig:track-year-distr}). Most of the studies focus on Lexical Semantics. While various semantics-related tasks are addressed, typically only one study per task is observed. An exception regards \gls{dsm}, where significant attention has been directed towards prediction-based methods, with notable studies being those of \citet{iosif-etal-2016-cognitively} and \citeauthor{Lioudakis2019},\cite{Lioudakis2019} who proposed new approaches to generate word embeddings, 
and of \citet{outsios2020-evaluation} who undertook a word embedding benchmark.
We also acknowledge that contextual embeddings, which comprise rich information,\cite{nikolaev2023universe,pado2023investigating} are heavily understudied in Greek. That is despite the existence of publicly available models.\cite{koutsikakis2020,evdaimon2023greekbart} An exception is the work of \citeauthor{gari2021let},\cite{gari2021let} who investigated the potential of contextual embeddings to capture lexical polysemy. 

\section{Track: Information Extraction}\label{sec:ie}
\gls{ie} concerns the automated identification and extraction of structured data, including entities, relationships, events, or other factual information from unstructured text. The primary objective of \gls{ie} is to make the information machine-readable, facilitating analysis, search, and practical use of textual information.\cite{yang2022survey} 

\subsection{Information Extraction in Greek: Language Models and Methods}
Below, we present studies addressing \gls{ie} (in descending order of recency) and \gls{ner}.

\paragraph{\gls{ie}}  \citet{mouratidis2023comparative} conducted a study on extracting maritime terms from legal texts in the Official Government Gazette of the Hellenic Republic. They identified these terms by counting token lengths, setting a threshold, and using lexicon-based stemmed tokens from maritime dictionaries introduced in their previous study.\cite{Mouratidis2022} Additionally, they derived word embeddings and used them to train \gls{rnn}-based models, incorporating the maritime term extraction features into the training process. In a separate study, \citet{Papadopoulos2021penelopie} tackled \textbf{\gls{oie}}, a process that involves converting unstructured text into \textsc{<subject; relation; object>} tuples. Addressing the challenge of \gls{oie} in languages that are resource-lean for this task, such as Greek, \citet{Papadopoulos2021penelopie} used \gls{nmt} between English and Greek to generate English translations of Greek text (\S\ref{sec:mt}). These were then processed through a \gls{nlp} pipeline,\cite{penelopie} enabling coreference resolution, summarization, and triple extraction using existing English \glspl{lm} and tools, and then back-translated the extracted triplets to Greek. \citet{barbaresi-lejeune-2020-box} evaluated \textbf{web content extraction} tools on HTML 4 standard pages in five different languages (Greek, Chinese, English, Polish, Russian), concluding that the three best tools for Greek perform comparably to the three top tools for English; for the rest of the languages the results are much lower than in English and Greek. Finally, \citet{lejeune2015multilingual} developed a multilingual (Chinese, English, Greek, Polish, and Russian) rule- and character-based \textbf{event extraction} system, where an event is defined minimally as a pair consisting of a disease and its corresponding location. This system was also referenced in prior studies of the authors.\cite{brixtel2013any,lejeune2012daniel}

\paragraph{\gls{ner}} This task involves the identification and categorization of specific entities, such as names of people, organizations, locations, dates, and more, within unstructured text. \citet{papantoniou2023automating} conducted \gls{ner} and \textbf{entity linking} on a dataset derived from Greek Wikipedia event pages. They assessed five established methods for \gls{ner} and four methods for entity linking, including three designed for English, which required translating Greek text into English. 
\citet{rizou2023efficient} carried out \gls{ner} and \textbf{intent classification} tasks on queries from a University help desk dataset with Greek and English submissions. They employed joint-task methods using Transformer-based models. In their earlier work, \citet{Rizou2022} applied the same tasks with the same methods to the widely used English benchmark dataset, the Airline Travel Information System (ATIS),\cite{hemphill-etal-1990-atis} which they also translated into Greek.
\citet{Bartziokas2020} curated \gls{ner} datasets and evaluated five \gls{dnn} models on them, selected for their high performance on the English CoNLL-2003\cite{sang2003introduction} and OntoNotes 5\cite{pradhan2007ontonotes} datasets, showing comparable performance to English. \citet{koutsikakis2020} performed \gls{ner} using their model, GreekBERT, as well as \gls{xlmr} and two variants of \gls{mbert}, finding that GreekBERT outperformed the other three \glspl{lm} in terms of micro-F1.
\citet{Partalidou2019} employed spaCy\cite{honnibal2017spacy} for \gls{pos} tagging and \gls{ner}, discovering limited impact of \gls{pos} tags on \gls{ner}. \citet{Angelidis2018} performed \gls{ner} and entity linking in legal texts. For \gls{ner} they used \gls{lstm} models; for entity linking, Levenshtein and substring distance were evaluated; for entity representation and linking, a \gls{rdf} specification was chosen. In entity linking, \citet{papantoniou2021} performed \gls{ner} on the text, generating candidates for the extracted entities from several wiki-based knowledge bases, then conducting disambiguation.

\subsection{Information Extraction in Greek: Language Resources}
Table~\ref{tab:ie-datasets} presents the \glspl{lr} developed for \gls{ie} tasks in Greek. Four out of the nine \glspl{lr} were publicly available, three of which were about news.

\begin{table}[H]
    \centering
    \caption{Datasets related to \gls{ie} with information on availability (Yes: publicly available, Err: unavailable, No: no information provided; see Table~\ref{tab:lr-availability} for details; the citations point to URLs), annotation type (see Table~\ref{tab:lr-ann-status} for details), size, size unit, and domain.
    }
    \label{tab:ie-datasets}
    \resizebox{\textwidth}{!}{
    \begin{tabular}{|l | l | l | l| l| l|}	 
	\hline
    \textbf{Authors} & \textbf{Availability} & \textbf{Ann. type} & \textbf{Size} & \textbf{Size unit} & \textbf{Domain} \\ 
	\hline
    \citet{papantoniou2023automating} & Yes\cite{ner_nel_greek_wikipedia} &  automatic & 474,361 & token & news \\  \hline
     \citet{rizou2023efficient} & Yes\cite{uniway_dataset} &  manual & 4,302 & sentence & university \\  \hline
     \citet{Bartziokas2020} & Yes\cite{elner} &  hybrid & 623,700& token & news \\
      & Yes\cite{elner} &  hybrid & 623,700 & token &  news \\  \hline
     \citet{Rizou2022} &
     Err\cite{msensis_downloads}&  manual & 5,473 & sentence& airline travel \\  \hline
     \citet{Lioudakis2019}& Err\cite{aueb-resources_Lioudakis2019} &  hybrid & n/a & n/a & n/a\\  \hline
     \citet{Angelidis2018} & Err\cite{uoa_legislation} &  manual & 254 & piece & legal \\  \hline
     \citet{lejeune2012daniel} & Err\cite{daniel_lejeune}&  manual &390& document & epidemics \\  \hline
     \citet{Mouratidis2022} & No &  manual & 80,000 & word & maritime law\\
     \bottomrule
    \end{tabular}
}
\end{table} 
By focusing on \glspl{lr} related to named entities, \citet{papantoniou2023automating} created a dataset from the Greek Wikipedia Events pages by automatically annotating eight entity tags. The annotation was performed by identifying terms that appeared in Wikidata, which also facilitated entity linking. \citet{rizou2023efficient} created a dataset of graduate student questions to two Greek universities, requesting the students to provide their questions in both Greek and English. The dataset is manually annotated with three entity tags and six intents. \citet{Bartziokas2020} provided two annotated datasets, one with four label tags akin to the CONLL-2003 dataset,\cite{sang2003introduction} and the other incorporating 18 tags for entities, as in the OntoNotes 5 English dataset.\cite{pradhan2007ontonotes} These datasets were developed during the GSOC2018 project (discussed in \S\ref{sec:syntax}), where the initial automatic annotation was followed by manual curation.
\citet{Lioudakis2019} converted the GSOC2018 named entity annotated dataset to the CONLL-2003 format. The source dataset was annotated using Prodigy,\cite{prodigy} where the initial annotations were done manually; subsequently, model predictions were used to accelerate the annotation process.
\citet{Rizou2022} undertook the task of translating to Greek the Airline Travel Information System corpus (ATIS) dataset\cite{hemphill-etal-1990-atis} eliminating duplicate entries. The dataset consists of audio recordings and manual transcripts of inquiries related to flight information in automated airline travel systems. It is complemented by annotations for named entities within the airline travel domain  and intent categories.
\citet{Angelidis2018} curated a dataset containing 254 daily issues of the Greek Government Gazette spanning the period 2000-2017, manually annotated for six entity types. \citet{lejeune2012daniel} offered 1,681 documents in five languages, annotating them regarding diseases and locations - where applicable. \citet{Mouratidis2022} conducted stemming on legal texts related to maritime topics from the Official Government Gazette of the Hellenic Republic, annotating tokens as either maritime terms or not.

\subsection{Summary of Information Extraction in Greek}

\gls{ie} studies in Greek primarily focus on \gls{ner}, often accompanied by datasets. Of the nine reported \glspl{lr}, four are publicly available, while another four could become available in the future, as their links are provided but currently result in HTTP errors.
Figure~\ref{fig:track-year-distr} shows that there was relative interest in \gls{ie} early on (2012-2014), which was discontinued (up to 2017), and then kept an upward trend. This can explain the tendency towards \gls{dl} approaches in this track, as highlighted in Figure~\ref{fig:ai-approach-per-track}, which is probably related to efforts to create benchmark datasets.\cite{Bartziokas2020, Rizou2022, Angelidis2018} Such benchmark datasets create the resources needed to train and assess \gls{dl} models. Another notable study in light of the data scarcity in certain \gls{ie} tasks is the work of \citet{Papadopoulos2021penelopie} who leveraged cross-lingual transfer learning techniques.

\section{Track: Sentiment Analysis and Argument Mining}\label{sec:sentiment}
The \gls{sa} task concerns the detection of opinions expressed in opinionated texts, while Argument Mining concerns the detection of the reasons why people hold their opinions.\cite{lawrence2020argument} 

As its name suggests, \gls{sa} involves the analysis of human sentiments toward specific entities. In addition to the analysis of sentiment, the task also concerns opinions, appraisals, attitudes, or emotions,\cite{liu2020sentiment} while the entities discussed can be products, services, organisations, individuals, events, issues, topics, etc. Particularly active in domains such as finance, tourism, health, and social media, \gls{sa} involves applications in  recommendation-based systems,\cite{Ghuribi2021} business intelligence,\cite{rokade2019business} and predictive or trend analyses.\cite{mudinas2019market, chauhan2021emergence}
The field of \gls{sa} has evolved significantly since it was popularized by the pioneering work of \citet{Turney2002thumbs} and \citeauthor{Pang2002thumbs},\cite{Pang2002thumbs} who classified texts as positive or negative. Subsequent studies have expanded and enriched the field, moving beyond binary classification and introducing slightly different tasks and alternative terms such as opinion mining, opinion analysis, opinion extraction, sentiment mining, subjectivity analysis, affect analysis, emotion analysis, and review mining, all of which now fall under the umbrella of \gls{sa}.\cite{liu2020sentiment} Further information on this track and background knowledge are included in Appendix~\ref{apx:sentiment}.

\subsection{Sentiment Analysis and Argument Mining in Greek: Language Models and Methods} 

\paragraph{Document-Level \gls{sa}}
\citet{evdaimon2023greekbart} evaluated their GreekBART model, along with GreekBERT\cite{koutsikakis2020} and \gls{xlmr}\cite{conneau2019unsupervised} \glspl{lm}, on a user-annotated movie reviews dataset — the Athinorama\_movies\_dataset\cite{greek_movies_kaggle} — for binary \gls{sa}, and found that GreekBERT outperformed the other models. Additionally, \citet{Bilianos2022} used GreekBERT\cite{koutsikakis2020} to classify the polarity of product reviews as positive or negative, while \citet{Braoudaki2020} conducted binary polarity classification of hotel reviews, experimenting with \gls{lstm} architectures and lexicon-based input features. 
\citet{Medrouk2018, Medrouk2017} applied binary polarity classification as well, but experimented with monolingual and multilingual input.
Multilingual \gls{sa} was addressed also by \citeauthor{Manias2020},\cite{Manias2020} who investigated the impact of \gls{nmt} on \gls{sa}. 
The authors translated part of the English IMDb reviews dataset\cite{imdb_review_kaggle} to Greek and German and trained the same \gls{nn} architecture on \gls{sa} using either the source or the target language as input. Translation was also used by \citet{athanasiou2017} for data augmentation purposes.

Early document-level \gls{sa} approaches were mainly based on \gls{ml} and feature engineering, employing information such as term frequency,\cite{markopoulos2015} \gls{pos},\cite{Spatiotis2016, spatiotis2017examining, Spatiotis2019, Beleveslis2019, spatiotis2020} and sentiment lexicons (see Table~\ref{tab:sentiment-lexicons}). Features crafted from sentiment lexicons have been found beneficial compared to dense word embeddings because the latter do not carry sentiment information,\cite{GIATSOGLOU2017} while \citet{markopoulos2015} noted that the \gls{tf-idf} representation outperformed lexicon-based features. 
Feature engineering-based \gls{sa} is not optimal compared to \gls{dl} counterparts.\cite{Bilianos2022, Kapoteli2022, Alexandridis2021Survey, spatiotis2020,Spatiotis2019} \citet{Spatiotis2019, spatiotis2020} applied feature engineering for \gls{sa} in hybrid educational systems, using features such as school level, region, and gender.  

\paragraph{Sentence-Level \gls{sa}}
\citet{zaikis2023pima} created a unified media analysis framework that classifies sentiment, emotion, irony, and hate speech in sentence- and paragraph-level texts by using a joint learning approach. This method leveraged the similarities between these tasks to enhance overall performance.
\citet{patsiouras2023greekpolitics} classified political tweets across four dimensions: sentiment polarity (three-class), figurativeness (ironic, sarcastic, figurative, or literal), aggressiveness (offensive, abusive, racist, or neutral language), and bias (strongly opinionated or not). They employed a \gls{cnn} and a Transformer-based architecture for classification, using data augmentation techniques to handle imbalanced categories. Both \citet{katika2023mining} and \citet{Kapoteli2022} fine-tuned GreekBERT for binary sentiment classification of COVID-19-related tweets, with the former focusing on Long-COVID effects and the latter on COVID-19 vaccination.
\citet{Alexandridis2021Survey} performed a benchmark of \gls{sa} methods either by including the neutral class or not. 
In the binary setting, the GPT2-Greek\cite{gpt2_greek} \gls{lm} outperformed \gls{ml} methods that used GreekBERT and FastText word embeddings. In the three-class setting, only \gls{dl} methods were used. The authors created and shared two \glspl{plm}, PaloBERT\cite{PaloBERT2023} and GreekSocialBERT,\cite{GreekSocialBERT2023} with the latter outperforming the former, which in turn outperformed GreekBERT. In a subsequent study, \citet{Alexandridis2021} compared their \glspl{lm} in emotion detection and concluded that GreekBERT consistently exhibited better performance than PaloBERT. 
\citet{Drakopoulos2020} used \glspl{gnn} on tweets, which were found to provide more accurate estimations of intentions by aggregating information about the twitter account.

In earlier work on sentence-level \gls{sa}, \citet{tsakalidis2018building} highlighted the importance of considering the domain in \gls{sa}, noting that n-gram representations are more effective for intra-domain \gls{sa}, while word embeddings and lexicon-based methods are more suitable for cross-domain \gls{sa}. \citet{Charalampakis2016, Charalampakis2015} ranked the features they used in descending order of significance, based on information gain. \citet{Solakidis2014} conducted semi-supervised \gls{sa} and emotion detection using lexicon-based n-gram features of emoticons and keywords, and found that emoticons can intensify and indicate the presence and polarity of specific sentiments within a document. \citet{Chatzakou2017} categorized social media input into 12 emotions using lexicon-based features of sentiment words and emoticons, where they translated from Greek the words of the input texts to English for the usage of English sentiment lexicons. 
Besides \gls{ml}-based \gls{sa}, there are also studies exploring sentiment in real-world situations, such as COVID-19\cite{Kydros2021} and pre-election events.\cite{Beleveslis2019}

\paragraph{Aspect-Based \gls{sa}}
\citet{antonakaki2017, Antonakaki2016} analyzed political discourse on Twitter by conducting entity-based \gls{sa} and sarcasm detection. They manually identified entities and performed lexicon-based \gls{sa} at the entity level. For sarcasm detection, they trained an \gls{svm} algorithm using lexicon-based sentiment features and topics extracted through topic modeling, based on the hypothesis that certain topics are more closely associated with sarcasm. The source code of is available.\cite{antonakaki_elections_study_2017} \citet{petasis2014sentiment} performed entity-based \gls{sa} to support a real-world reputation management application, monitoring whether entities are perceived positively or negatively on the Web.

\paragraph{Stance Detection} \citet{Tsakalidis2018} aimed to nowcast on a daily basis the voting stance of Twitter users during the pre-electoral period of the 2015 Greek bailout referendum. They performed semi-supervised, time-sensitive classification of tweets, leveraging text and network information.

\paragraph{Argument Mining} \citet{sliwa-etal-2018-multi} tackled argument mining for non-English languages using parallel data. They used parallel data pairs with English as the source language and either Arabic or a Balkan language (including Greek) as the target language. They automatically annotated English sentences for argumentation using eight classifiers and extended the labels to the target languages using majority voting.
\citet{sardianos2015argument} identified segments representing argument elements (i.e., claims and premises) in online texts (e.g., news), using \glspl{crf}\cite{lafferty2001conditional} and features based on \gls{pos} tags, cue word lists, and word embeddings.

\subsection{Sentiment Analysis and Argument Mining in Greek: Language Resources} 
Table \ref{tab:sentiment-datasets} presents the datasets related to \gls{sa} and Argument Mining, along with information on their availability (see Table~\ref{tab:lr-availability}), annotation type (see Table~\ref{tab:lr-ann-status}), size, size unit and classes of annotation. Besides datasets, \glspl{lr} in Greek comprise sentiment lexicons, which are summarized in Table~\ref{tab:sentiment-lexicons} and have been used to extract features for \gls{ml} algorithms, or could have been used.\cite{chen2014building}

\paragraph{Document-Level \gls{sa}} Document-Level datasets in Greek mainly regard product reviews. 
\citet{Bilianos2022} presented 240 negative and 240 positive electronic product reviews. These reviews consist of user-generated content with ratings adjusted by the researchers to generate binary polarity. The remaining studies focusing on document-level \gls{sa} created non-publicly available datasets annotated either for emotion\cite{Kapoteli2022} or sentiment.\cite{Braoudaki2020, spatiotis2020, Spatiotis2019, Spatiotis2016, Manias2020, Medrouk2018, Medrouk2017, athanasiou2017, spatiotis2017examining, markopoulos2015,petasis2014sentiment}

\paragraph{Sentence-Level \gls{sa}} Datasets annotated for sentiment at the sentence-level in Greek primarily consist of tweets. 
\citet{patsiouras2023greekpolitics} created, and provide upon request, a dataset of 2,578 unique tweets manually annotated across four different dimensions: sentiment polarity (three-class), figurativeness (ironic, sarcastic, figurative, or literal), aggressiveness (offensive, abusive,
\begin{table}[H]
    \centering
    \caption{
    Datasets for \gls{sa} and argument mining, indicating their availability status (Lmt: limited availability, Err: unavailable, No: no information provided; see Table~\ref{tab:lr-availability} for details; the citations point to URLs), annotation type (see Table~\ref{tab:lr-ann-status} for details), size, size unit, and the sentiment annotation classes.}
    \label{tab:sentiment-datasets}
    \resizebox{\textwidth}{!}{
    \begin{tabular}{|l|l|l|l|p{2cm}|p{5cm}|}	 
	\hline
	\textbf{Authors} & \textbf{Availability} & \textbf{Ann. type} & \textbf{Size}  &\textbf{Size unit} & \textbf{Class}\\  
	\hline
	\citet{patsiouras2023greekpolitics} & Lmt\cite{auth_greekpolitics} &  manual & 2,578& tweet&  (positive, negative, neutral), (figurative, normal), (aggressive, normal), (partizan, neutral)\\ 
	\hline
	\citet{Bilianos2022} & Lmt\cite{greek_sentiment_analysis}&  user-generated & 480 & review &  positive, negative\\ 
	\hline
	\citet{Kydros2021} &  Lmt &  automatic & 44,639& tweet&  positive, negative, anxiety\\ 
	\hline
	\citet{sliwa-etal-2018-multi} & Lmt &  automatic & 166,430 & sentence & argumentative, non-argumentative\\ 
	\hline
	\citet{tsakalidis2018building} &  Lmt&  manual & 1,640& tweet &  positive, negative, neutral\\ 
	&  Lmt &  manual & 2,506&  tweet & sarcastic, non-sarcastic\\ 
	\hline
	\citet{Chatzakou2017} & Lmt\cite{chatzakou2016dataset} &  manual & 2,246& tweet &  Ekman’s six basic emotions \& enthusiasm, rejection, shame, anxiety, calm, interest\\ 
	\hline
	\citet{Antonakaki2016,antonakaki2017} & Lmt\cite{social_media_turbulence} &  automatic & 301,000 &  tweet &  -5 to -1 (negative), 1 to 5 (positive)\\ 
	& Lmt\cite{social_media_turbulence} &  automatic & 182,000&  tweet&  -5 to -1 (negative), 1 to 5 (positive)\\ 
	& Lmt\cite{social_media_turbulence} &  manual & 4,644&  tweet & sarcastic, non-sarcastic\\ 
	\hline
	\citet{Makrynioti2015} &  Lmt &  manual & 8,888& tweet &  positive, negative, neutral\\ 
	\hline
	\citet{sardianos2015argument} &  Lmt &  manual & 300& document&  argument\\ 
	\hline
	\citet{Charalampakis2016} & Err\cite{websent} &  hybrid & 44,438& tweet &  ironic, non-ironic\\ 
	\hline
	\citet{Charalampakis2015} & Err\cite{websent}&  hybrid &61,427& tweet &  ironic, non-ironic\\ 
	\hline
	\citet{katika2023mining} & No &  hybrid & 937 & tweet&  positive, negative, neutral\\ 
	\hline
	\citet{zaikis2023pima}  & No &  manual & 14,579 & sentence, paragraph&  (positive,negative, neutral), (ironic, not ironic), (hate, not hate), (Happiness, Contempt, Anger, Disgust, Surprise, Sadness, None)\\ 
	\hline
	\citet{Alexandridis2021}& No &  manual & 3,875& tweet &  Ekman's six basic emotions, anticipation, trust \& none \\ 
	& No &  manual & 54,916&  document &  positive, negative, neutral\\ 
	\hline
	\citet{Alexandridis2021Survey} & No &  manual & 59,810 & social media text &  positive, negative, neutral\\ 
	\hline
	\citet{Kapoteli2022} & No&  manual & 1,424& tweet &  positive, negative, neutral\\ 
	\hline
	\citet{Braoudaki2020} & No &  user-generated & 156,700& review &  positive, negative\\ 
	\hline
	\citet{Drakopoulos2020} & No &  automatic & 17.465M& tweet &  positive, negative\\ 
	\hline
	\citet{spatiotis2020,Spatiotis2019,Spatiotis2016} & No &  manual & 11,156& review&  very positive, positive, neutral, negative, very negative\\ 
	\hline
	\citet{Manias2020}& No &  user-generated & 4,251&  review & positive, negative, unsupported\\ 
	\hline
	\citet{Beleveslis2019}& No &  automatic & 46,705&  tweet& positive, negative, neutral\\ 
	\hline
	\citet{Medrouk2018} & No &  user-generated & 91,816 (EL, EN, FR)& review&  positive, negative\\ 
	\hline
	\citet{Tsakalidis2018} & No &  hybrid & 1.64M& tweet &  favor, against\\ 
	\hline
	\citet{Medrouk2017} & No &  user-generated & 2,600& review &  positive, negative\\ 
	& No &  user-generated & 7,200 (EL, EN, FR)&  review & positive, negative\\ 
	\hline
	\citet{athanasiou2017} & No &  manual & 740& comment &  positive, negative \\ 
	\hline
	\citet{GIATSOGLOU2017} & No &  manual & 2,800& sentence &  positive, negative \\ 
	\hline
	\citet{spatiotis2017examining} & No &  manual & 11,156& review &  very positive, positive, neutral, negative, very negative\\ 
	\hline
	\citet{markopoulos2015} & No&  manual & 1,800 & review &  positive, negative\\ 
	\hline
	\citet{petasis2014sentiment} & No & manual & 2,300& text&  positive, negative\\ 
	\hline
	\citet{Solakidis2014}& No &  hybrid & 25,700&  tweet & positive, negative, neutral, joy, love, anger, sadness\\ 
        \hline
    \end{tabular}
    }
\end{table}
 racist, or neutral language), and bias (strongly opinionated or not). The tweets span the period from March 2014 to March 2021.
\citet{Chatzakou2017} presented a dataset of randomly selected tweets annotated for 12 emotions through crowdsourcing and majority voting. \citet{Kydros2021} created, and provide upon request, a dataset of tweets related to the Covid-19 pandemic. These tweets were automatically annotated using lexicons to determine their sentiment as either positive or negative, with an additional annotation for anxiety.

\citeauthor{Antonakaki2016}\cite{Antonakaki2016,antonakaki2017} presented three datasets of tweets focused on politics. Two datasets feature automatic sentiment annotation on a scale from -5 to 5, while the third is manually annotated for sarcasm using crowdsourcing, with a binary classification. Although all datasets include the full tweet texts and are released under the Apache 2.0 license,\cite{apache_license} this conflicts with X's terms of service and copyright law.\cite{twitter_tos} As a result, they are classified as having limited availability according to the availability schema (see \S\ref{sec:lr-avail-taxonomy}). \citet{tsakalidis2018building} offered a tweet dataset related to the January 2015 General Elections in Greece annotated for positive or negative sentiment, as well as a second dataset election-related tweets annotated for sarcasm. Both datasets were filtered to include only instances where annotators agreed. \citet{Makrynioti2015} randomly sampled 8,888 tweets from August 2012 to January 2015 and annotated them as positive, negative or neutral. \citet{Charalampakis2016, Charalampakis2015} shared two tweet datasets annotated for irony. Each dataset includes 162 manually annotated tweets, with the rest of the tweets having been automatically annotated. These tweets were collected in the weeks before and after the May 2012 parliamentary elections in Greece and are characterized by the political parties and their leaders. 
The remaining studies focusing on sentence-level \gls{sa} are not publicly available and primarily use tweets\cite{katika2023mining,Kapoteli2022,Drakopoulos2020, Beleveslis2019,Tsakalidis2018,Solakidis2014} or social media content from other sources.\cite{Alexandridis2021Survey} Exceptions include \citeauthor{GIATSOGLOU2017},\cite{GIATSOGLOU2017} who segmented user reviews on mobile phones into sentences for annotation, and \citeauthor{zaikis2023pima},\cite{zaikis2023pima} who collected data from the internet, social media and press, annotating it at both the sentence and paragraph levels. Additionally, all datasets are annotated for sentiment, except for \citeauthor{Solakidis2014},\cite{Solakidis2014} who annotated both sentiment and four emotions, and \citet{zaikis2023pima} who annotated for sentiment, irony, hate speech, and emotions.

\paragraph{Argument Mining} \citet{sliwa-etal-2018-multi} provided a collection of bilingual datasets containing sentences labeled as argumentative or non-argumentative, available upon request. The sentences were automatically annotated by eight different argument mining models, and the final label was determined based on the majority vote of these models. These datasets were derived from parallel corpora where the source language is English and the target language is either a Balkan language or Arabic. Additionally, \citet{sardianos2015argument} made a dataset available upon request. This dataset consists of 300 news articles from the Greek newspaper Avgi,\cite{avgi} annotated by two human annotators (150 articles each) for argument components, i.e., premises and claims.

\begin{table*}[ht]
    \centering
    \caption{Sentiment lexica with information on availability (Lmt: limited availability, Err: unavailable, No: no information provided; see Table~\ref{tab:lr-availability} for details; the citations point to URLs), size, size unit, and the sentiment annotation classes.}
    \label{tab:sentiment-lexicons}
    \resizebox{\textwidth}{!}{
    \begin{tabular}{|l | l | l | l| l |}	 
	\hline
	\textbf{Authors} & \textbf{Availability} & \textbf{Size}  &\textbf{Size unit} & \textbf{Class}\\  
	\hline
    \citet{tsakalidis2018building} & Lmt\cite{tsakalidis2017building}&  2,260& word&  subjectivity, polarity \& Ekman’s six basic emotions\\ 
   & Lmt\cite{tsakalidis2017building}&  190,667 & ngram &  positive, negative\\ 
   & Lmt\cite{tsakalidis2017building}&  32,980 & ngram &  Ekman’s six basic emotions\\ 
    \hline
    \citet{GIATSOGLOU2017} & Err\cite{msensis_demon}& 4,658&  word &  subjectivity, polarity \& Ekman’s six basic emotions\\ 
    \hline
    \citet{palogiannidi2016}& Err\cite{greek_affective_lexicon} &  407,000& word &  valence, arousal, dominance\\  
    \hline
    \citet{antonakaki2017}& No &  ~4,915& word &  positive, negative\\ 
    \hline
    \citet{markopoulos2015}& No &  68,748& token &  positive, negative\\ 
    \hline
    \end{tabular}
 }
\end{table*}

\paragraph{\gls{sa} Lexicons} \citet{tsakalidis2018building} developed three lexicons with data collected between August 1st, 2015, and November 1st, 2015. The first lexicon, the Greek Affect and Sentiment lexicon (GrAFS), was derived from the digital version of the  Dictionary of Standard Modern Greek,\cite{triantafyllides_dictionary} which was web-crawled to gather words used in an ironic, derogatory, abusive, mocking, or vulgar manner. This process yielded 2,324 words (later reduced to 2,260 after editing) along with their definitions. These words were manually annotated as objective, strongly subjective, or weakly subjective. Subjective words were further annotated as positive, negative, or both, and each annotation was rated on a scale from one (least) to five (most) based on Ekman's six basic emotions. The annotations were then automatically extended to all inflected forms, resulting in 32,884 unique entries. To capture informalities prevalent in Twitter content, the authors also developed two Twitter-specific lexicons collecting tweets using seed words from the first lexicon: the Keyword-based lexicon (KBL) with 190,667 n-grams and the Emoticon-based lexicon (EBL) with 32,980 n-grams. \citet{GIATSOGLOU2017} proposed an expansion of the lexicon of \citeauthor{tsakalidis2014ensemble},\cite{tsakalidis2014ensemble} which included 2,315 words annotated for subjectivity, polarity, and Ekman's six emotions. This expanded lexicon incorporated synonyms grouped around each term and assigned a vector containing the average emotion over all dimensions and terms, resulting in a total of 4,658 Greek terms. \citet{palogiannidi2016} introduced an affective lexicon of 1,034 words with human ratings for valence, arousal, and dominance, originating from \citeauthor{bradley1999affective}.\cite{bradley1999affective} The terms were translated, manually annotated by multiple annotators, and then automatically expanded using a semantic model to estimate the semantic similarity between two words, resulting in a final lexicon of 407,000 words. The following lexicons are not publicly available.  \citet{antonakaki2017} presented a lexicon consisting of 4,915 words manually annotated for polarity. This lexicon is a compilation of three independent lexicons: two general-purpose lexicons and one from the political domain. \citet{markopoulos2015} developed a sentiment lexicon of Greek words from a corpus they constructed, covering terms with positive or negative meanings. The lexicon includes all inflected forms of the words, resulting in 68,748 unique entries.

\subsection{Summary of Sentiment Analysis and Argument Mining in Greek}
\gls{sa} for Greek is applied to hotel and product reviews, political posts, educational questionnaires, COVID19-related posts, and trending topics discussed on Twitter, listed here in descending order of frequency. We have also observed studies that deal with a range of distinct emotions, or a specific emotion, e.g. anxiety or irony which is a sentiment that is extremely challenging to capture in \gls{nlp}.\cite{wankhade2022survey} 
The \gls{sa} task has attracted significant attention in the field of \gls{nlp} for Greek (Figure~\ref{fig:track-year-distr}), constituting approximately one-fourth of the studies (23.4\%). This attention reached its peak in 2017 before experiencing a slight decrease. A similar trend is observed across ACL Anthology tracks, albeit slightly earlier, i.e., between 2013 and 2016.\cite{rohatgi2023acl} Despite the abundance of studies, however, we observe that a publicly available Greek dataset that can serve as a \gls{sa} benchmark does not exist. Even among the published datasets, limitations exist, such as missing licenses or paywalls (the datasets marked as ``Lmt'' in the availability type column), or unavailability due to HTTP errors (the datasets marked as ``Err'' in the availability type column). Furthermore, studies exploring \gls{sa} through lexicons have generated new ones, yet none of these are publicly accessible.

\section{Track: Authorship Analysis}\label{sec:authorship}
Authorship analysis attempts to infer information about the authorship of a piece of work.\cite{el2014authorship} It encompasses three primary tasks: \textbf{author profiling}, detecting sociolinguistic attributes of authors from their text; \textbf{authorship verification}, determining whether a text belongs to a specific author; and \textbf{authorship attribution}, pinpointing the right author of a particular text from a predefined set of potential authors.\cite{el2014authorship} Both authorship verification and authorship attribution are variations of the broader \textbf{author identification} problem, which seeks to determine the author of a text.\cite{kestemont2018overview} Another pertinent task within authorship analysis is \textbf{author clustering}, which entails grouping documents  authored by the same individual into clusters, with each cluster representing a distinct author.\cite{rosso-2016-overview} Although other tasks may relate to authorship analysis, research on Greek authorship analysis has predominantly focused on these five tasks; therefore, we concentrate on them.

\subsection{Authorship Analysis: Language Models and Methods}
In Greek, authorship analysis has been supported by a workshop series addressing various tasks within this field. Furthermore, additional studies concentrating on the fundamental tasks of authorship analysis have been identified.

\paragraph{PAN} A workshop series and a networking initiative, called PAN,\cite{pan_webis} is dedicated to the fields of digital text forensics and stylometry since 2007. Its objective is to foster collaboration among researchers and practitioners, exploring text analysis in terms of authorship, originality, trustworthiness, and ethics among others. PAN has organized shared tasks focusing on computational challenges related to authorship analysis, computational ethics, and plagiarism detection, amassing a total of 64 shared tasks with 55 datasets provided by the organizing committees and an additional nine contributed by the community.\cite{bevendorff2023overview} Among these shared tasks, four specifically addressed Greek, with three dealing with author identification,\cite{juola2013overview, stamatatos2014overview, stamatatos2015b} and one with author clustering.\cite{stamatatos:2016}

\paragraph{Authorship Attribution} 
\citet{juola2019comparative} benchmarked an attribution framework, JGAAP,\cite{jgaap} on a corpus of Greek texts that were authored by students who also translated their texts to English. 
Authorship attribution was substantially more accurate in English than in Greek. They provided three possible reasons suitable for future investigation. First, the framework or the features tested excel in English (selection bias). Second, authorship pool bias may exist due to the authors' non-native English proficiency, potentially affecting the error rate. Third, linguistically, Greek may possess inherent complexities that hinder individual feature extraction. 

\paragraph{Authorship Verification} This task has been addressed in Greek in a multilingual setting. 
\citet{kocher2017simple} suggested an unsupervised baseline, by concatenating the candidate author's texts and comparing the 200 most frequent occurring terms (words and punctuation symbols) extracted from these texts with those extracted from the disputed text. 
\citet{hurlimann2015glad} trained a binary linear classifier on top of engineered features (e.g., character n-grams, text similarity, visual text attributes); the source code is available.\cite{huerlimann15_glad} \citet{halvani2016authorship} approached the task with a single-class classification, demonstrating strong performance in the PAN-2020 competition.\cite{halvani2020taveer,kestemont2020overview}

\paragraph{Author Profiling} \citet{mikros2013systematic} and \citet{perifanos2015gender} performed gender identification in Greek tweets. They deployed \gls{ml} algorithms using stylometric features at the character and word levels, most frequent words in the text, as well as gender-related keywords lists, extracted from the texts. Specifically, \citet{mikros2013systematic} focused on stylometric features, including lexical and sub-lexical units, to analyze their distribution in texts written by male and female authors. Their study concluded that men and women use most stylometric features differently. In a prior study, \citet{mikros2012authorship} conducted author gender identification and authorship attribution using stylometric features, employing the \gls{smo} algorithm.\cite{Platt1999} Their findings indicated that author gender is conveyed through distinctive syntactical and morphological patterns, whereas authorship is linked to the over- or under-representation of certain high-frequency words.

\subsection{Authorship Analysis in Greek: Language Resources}

Table~\ref{tab:authorship} displays the \glspl{lr} corresponding to Authorship Analysis tasks in Greek. Some datasets have limited availability due to the lack of a provided license, while others are not available at all. Most of the limited availability \glspl{lr} originate from the PAN workshop series. Although our method returned the task overview papers from the PAN 2013-2014-2016 workshops, we include all Greek datasets from the PAN workshop series for comprehensive coverage.
\citet{juola2013overview} delivered a corpus for the PAN-2013 author identification task, which contains documents in Greek, English, and Spanish. The Greek part of the corpus comprises newspaper articles published in the Greek weekly newspaper To Vima,\cite{tovima} from 1996 to 2012. The authors organized the Greek corpus into 50 verification task groups, where each group comprises documents with known authors and one document with an unknown author. The grouping criteria included genre, theme, writing date, and stylistic relationships.
\begin{table*}[ht]
    \centering
    \caption{Datasets related to authorship analysis  with information on availability (Lmt: limited availability, No: no information provided; tags are linkable except for No; see Table~\ref{tab:lr-availability} for details), annotation type (see Table~\ref{tab:lr-ann-status} for details), size, size unit, and domain.}
    \label{tab:authorship}
    \resizebox{\textwidth}{!}{
    \begin{tabular}{|l | l | l | l| l| l|} 
    \hline
    \textbf{Authors} & \textbf{Availability} & \textbf{Ann. type} & \textbf{Size} & \textbf{Size unit} & \textbf{Domain} \\  
    \hline
    \citet{stamatatos:2016} & Lmt\cite{pan16_clustering} & manual & 330 & \ document & \ articles, reviews\\  
    \hline
    \citet{stamatatos2015b} & Lmt\cite{pan15_verification} & manual & 393 & \ document & \ news\\  
    \hline
    \citet{stamatatos2014overview} & Lmt\cite{stamatatos2014pan14} & manual & 385 & \ document & \ news\\  
    \hline
    \citet{juola2013overview} & Lmt\cite{juola2013pan13} & manual & 120 & \ article & \ news \\  
    \hline
    \citet{halvani2016authorship} & Lmt\cite{bitly_1OjFRhJ} & curated & 190 & \ recipe, article & \ cooking, various \\  
    \hline
    \citet{juola2019comparative} & No & manual & 200 & \ essay & \ personal \& academic topics \\  
    \hline
    \citet{mikros2013systematic} & No & curated & 479,439 & \ word & \ science, society, economy, art \\  
    \hline
    \citet{mikros2012authorship} & No & curated & 406,460 & \ word & \ blog \\  
    \hline
    \end{tabular}
    }
\end{table*}
\citet{stamatatos2014overview} provided a larger corpus for the PAN-2014 author identification task, consisting of 200 verification task groups of documents and including Dutch in addition to the previously mentioned languages. Unlike PAN-2013, no stylistic analysis was conducted on the texts to identify authors with very similar styles or texts by the same author with notable differences. The PAN-2015 corpus for the author identification task\cite{stamatatos2015b} is the same size as PAN-2014 and includes the same languages, but it allows for cross-topic and cross-genre author verification without assuming that all documents share the same genre or topic.
\citet{stamatatos:2016} introduced a dataset for the PAN-2016 author clustering task, including documents in Greek, English, and Dutch, spanning two genres (articles and reviews) and covering various topics. The Greek part of the corpus comprises six document groups specifically created for author clustering and authorship-link ranking. 

Additionally, \citet{halvani2016authorship} provided the test part of their dataset, which includes 120 recipes and 70 news articles, but it is also not accompanied by a license. The other three \glspl{lr} are not publicly available. \citet{juola2019comparative} performed authorship attribution on Greek and English student essays, which were written by one hundred students following specific instructions. In contrast, \citeauthor{mikros2013systematic}\cite{mikros2013systematic, mikros2012authorship} focused on identifying gender in news texts and blogs, creating datasets that are equally balanced by gender and topic, with this information provided by the distributors.

\subsection{Summary of Authorship Analysis in Greek}

Authorship Analysis studies in Greek address all three primary tasks: authorship attribution, verification, and profiling. A significant contribution to this field is the inclusion of Greek benchmark datasets in the PAN workshop series (from PAN-2013 to PAN-2016 workshops), which focus on digital text forensics and stylometry. The availability of Greek datasets through this workshop series has been pivotal for advancing authorship analysis research in Greek, with many early studies using these datasets. However, this initial boost was not sustained after  2016, rendering this track less popular than others (Figure~\ref{fig:track-year-distr}). The most recent study retrieved dates back to 2019, with the majority of studies concentrated between 2012 and 2017. Recent global advancements in authorship analysis tasks, such as \gls{dl} and transfer learning,\cite{savoy2020machine} have not yet been adopted for Greek. Additionally, the evolution in the field has introduced new tasks, such as bots profiling, author obfuscation, and style change detection, areas where no relevant studies have been conducted in the Greek context.\cite{potthast2019decade}

\section{Track: Ethics and NLP}\label{sec:toxicity}

Within the thematic domain of Ethics and \gls{nlp}, a wide range of topics is addressed globally, encompassing issues such as overgeneralization, dual use of \gls{nlp} technologies, privacy protection, bias in \gls{nlp} models, underrepresentation, fairness, and toxicity detection, among others.\cite{hovy2017proceedings} Studies pertaining to Greek within this domain primarily focus on \textbf{Toxicity Detection}, i.e, the automated identification of abusive, offensive, or otherwise harmful user-generated content (more details about definitions are in Appendix~\ref{apx:tox}). This task is a necessity for online content moderation, including social media moderation, content filtering, and prevention of online harassment, all aimed at fostering a safer and more respectful online environment.\cite{schmidt2017survey} A notable exception among the many toxicity detection studies in Greek is the work of \citeauthor{ahn2021mitigating},\cite{ahn2021mitigating} which  explores ethnic bias detection in \glspl{plm} (discussed in \S\ref{sec:ml-nlp}).

Toxicity detection has sparked heated debates. Criticism primarily focuses on datasets and the way researchers' approaches to the problem, often oversimplifying the issue and disregarding the variety of use cases.\cite{diaz2022accounting} Toxicity can be interpreted differently depending on factors like social group membership, social status, and privilege, leading to unequal impacts on marginalized communities.\cite{welbl2021challenges} A second point of criticism highlights the fact that commonly used datasets, on which the models are trained and evaluated, often lack sufficient context to allow reliable judgments.\cite{pavlopoulos2020toxicity,hovy2021importance} Lastly, the definition of toxicity is subjective and dependent on the annotator's perspective.\cite{sap2019risk,gordon2022jury} Recent studies indicate that annotators may not simply disagree but can be polarized in their assessments regarding the toxicity of the same text. For instance, a text can be found toxic by all women annotators but considered benign by all men.\cite{pavlopoulos2024polarized} However, such factors are rarely incorporated into research tasks, which adversely affects how the task is framed and how the automated systems are developed. Furthermore, differences in the interpretation of toxicity are evident in the datasets and models that are created.\cite{sap2019risk,gordon2022jury}

\subsection{Ethics and NLP in Greek: Language Models and Methods}

\paragraph{Toxicity Detection - Multilingual Shared Task}
In 2020, a subtask of offensive language identification for Greek was introduced as part of the SemEval-2020 Task 12 on Multilingual Offensive Language Identification in Social Media (OffensEval-2020).\cite{Zampieri2020} The task focused on four other languages besides Greek: Danish, English, Turkish, and Arabic. Overall, 145 teams submitted official runs on the test data, 37 of which made an official submission on Greek, while the submissions for English were approximately double, with 81 submissions. The three top systems\cite{ahn2020nlpdove,wang2020galileo,socha2020ks} primarily relied on BERT-based \glspl{lm}, with the first two using multilingual models and the third a monolingual one.

\paragraph{Toxicity Detection - Greek Shared Task Subtask} The Greek dataset used for the OffensEval-2020 subtask is an extended version of the Offensive Greek Tweet Dataset (OGTD), which was developed by \citeauthor{Pitenis2020}.\cite{Pitenis2020} In their original article, the authors trained \gls{ml} classifiers on extracted features, including \gls{tf-idf} unigrams, bigrams, \gls{pos} and dependency relation tags and obtained their best results from a \gls{dl} network, by feeding word embeddings\cite{outsios2018word} into \gls{lstm} and \gls{gru} cells equipped with self-attention mechanisms.\cite{plum2019rgcl} Three years after the release of the OffensEval-2020 subtask, \citet{zampieri2023offenseval} evaluated \glspl{llm} on OffensEval-2020 datasets and from the previous OffensEval-2019,\cite{zampieri2019semeval} which focused exclusively on English. They used the top three systems of each language track as baselines. While eight \glspl{llm} were evaluated on the English datasets, only the Flan-T5-large \gls{llm},\cite{chung2024scaling} which is fine-tuned in other languages besides English, was tested on the other languages, including Greek. In all non-English languages, the macro F1 score of the \gls{llm} demonstrated a significant improvement from the third-place system. Additionally, they reviewed recent popular benchmark competitions on the topic, none of which, apart from OffensEval, included Greek. Both \citet{zaikis2023pima} and \citet{Ranasinghe2021} used the extended version of OGTD to evaluate their systems. The former developed a unified media analysis framework designed to classify sentiment, emotion, irony, and hate speech in texts through a joint learning approach (see \S\ref{sec:sentiment}). They evaluated their system on this dataset and also tested their media domain fine-tuned version of GreekBERT, the Greek Media BERT.\cite{GreekMediaBERT} The latter employed cross-lingual contextual word embeddings and transfer learning techniques to adapt the \gls{xlmr}\cite{conneau2019unsupervised} model, initially trained on English offensive language data (OLID),\cite{zampieri2019predicting} for detecting offensive language in Greek and six other low-resource languages.

\paragraph{Toxicity Detection by Companies}
Toxicity detection in online content has also been explored in the Greek context in industry-led or industry-supported efforts. For instance, considering both fully and semi-automatic user content moderation in the comments section of the Gazzetta sports portal,\cite{gazzetta} \citet{Pavlopoulos2017deep} equipped a \gls{gru}-based \gls{rnn} with a self-attention mechanism for toxicity detection. Their results suggested that an ablated version, using only self-attention and disregarding the \gls{rnn}, performs considerably well given its simplicity. In a different study on the same data, \citet{Pavlopoulos2017improved} showed that user embeddings lead to improved performance.

\paragraph{Hate Speech}
Research focusing on the phenomenon of hate speech targeting specific groups has garnered significant interest in Greek. Greece, serving as a primary entry point for a large number of immigrants arriving in Europe, has witnessed the emergence of xenophobic and racist opinions directed towards immigrants and refugees.\cite{Arvanitidis_2021} \citet{Calderón2022} tackled racist and/or xenophobic hate speech detection on tweets for Greek, Spanish and Italian using BERT-based \glspl{lm}. The same task was addressed by \citeauthor{Perifanos2021},\cite{Perifanos2021} also for Greek but with a multi-modal approach, to account for hateful content that does not necessarily carry textual streams, i.e., images. For the text modality, they used GreekBERT and they also created the BERTaTweetGR \gls{lm} (described in \S\ref{sec:ml-nlp}), while,  for joint representations of text and tweet images, they used a single model that combines the representations of BERT and ResNet. 
\citet{Pontiki2020} performed verbal aggression analysis on twitter data. They identified different aspects of verbal aggression related to predefined targets of interest. \citet{patsiouras2023greekpolitics} classified political tweets into four different dimensions, one of which was aggressive language (further details provided in \S\ref{sec:sentiment}). \citet{kotsakis2023web} developed and evaluated a web framework\cite{pharmproject_scripts} which was designed for automated data collection, hate speech detection and content management of multilingual content from social media targeting refugees and migrants. The evaluation was conducted by human experts who assessed hate speech detection using both a lexicon-based approach and an \gls{rnn}-based approach. 
\citet{lekea2018detecting} detected hate speech within terrorist manifestos, classifying these manifestos into three categories: no hate speech, moderate hate speech, and evident hate speech. Finally, \citet{nikiforos2020} detected bullying within Virtual Learning Communities - online communities created for educational purposes - and applied linguistic analysis to recognize behavior patterns.

\subsection{Ethics and NLP in Greek: Language Resources}

Table \ref{tab:toxicity-datasets} presents the relevant  \glspl{lr} in Greek, along with their availability (as shown in Table~\ref{tab:lr-availability}), annotation type (detailed in Table~\ref{tab:lr-ann-status}), size, and the type of content targeted for detection. Only one dataset is publicly available, licensed, and accessible at the time of this study. Half of the datasets are publicly available but have licensing issues, another one was inaccessible, and two others are not publicly available (both with very few data). 

\begin{table}[H]
    \centering
    \caption{Datasets designed for toxicity detection related tasks, with information on availability (Yes: publicly available, Lmt: limited  availability, Err: unavailable, No: no information provided; see Table~\ref{tab:lr-availability} for details; the citations point to URLs), annotation type (see Table~\ref{tab:lr-ann-status} for details), size, size unit, and detection class.}
    \label{tab:toxicity-datasets}
    \resizebox{\textwidth}{!}{
    \begin{tabular}{|l | l | l | l| l | l |}	 
	\hline
	\textbf{Authors} & \textbf{Availability} & \textbf{Ann. type} & \textbf{Size}  &\textbf{Size unit} & \textbf{Class}\\
	\hline
    \citet{Zampieri2020} & Yes\cite{strombergnlp_offenseval_2020} &  manual & 10,287 &  tweet & offense\\  
    \hline
    \citet{Perifanos2021} & Lmt\cite{multimodal_hate_speech_detection} &  manual & 4,004 &  tweet &toxicity\\  
    \hline
    \citet{Pontiki2020} & Lmt\cite{clarin_xenophobia}&   automatic & 4.490M &  tweet  &xenophobia\\  
    \hline
    \citet{Pavlopoulos2017deep} & Lmt\cite{gazzetta_comments_dataset}&  manual & 1.450M &  comment &toxicity\\  
    \hline
    \citet{Calderón2022}& Err\cite{pharmproject}&  manual &15,761&  tweet & racism, xenophobia\\  
    \hline 
    \citet{nikiforos2020}& No&  manual &583& sentence &bullying\\  
    \hline
    \citet{lekea2018detecting}& No &  automatic\textsuperscript{a}& 81& manifesto &terrorism\\  
    \hline
    \end{tabular}
    }
    \textsuperscript{a} Unclear annotation process.
\end{table}

\citet{Pitenis2020} introduced a dataset comprising 1,401 offensive and 3,378 non-offensive tweets, sourced from popular and trending hashtags in Greece, keyword queries containing sensitive or obscene language, and tweets featuring /eisai/ (``you are'') as a keyword. \citet{Zampieri2020} extended this dataset for OffensEval-2020, increasing the total to 10,287 tweets. The extension involved manual annotation by three annotators, with the final label determined by majority voting among the annotators.

\citet{Perifanos2021} curated 1,040 racist and/or xenophobic hate-speech tweets and 2,964 non-hate-speech tweets, providing both their tweet IDs and code for retrieving them. Their dataset reflects an overlap of neo-Nazi, far-right, and alt-right social media accounts systematically targeting refugees, LGBTQ activists, feminists, and human rights advocates. The dataset was annotated by three human rights activists. The final label for each tweet was determined by majority voting among the annotators.  \citet{Pontiki2020} gathered a dataset of 4,490,572 tweets exhibiting verbal attacks against ten predefined target groups during the financial crisis in Greece. The dataset has limited availability as it does not include the tweets or tweet IDs.
\citet{Pavlopoulos2017deep} offered 1.6 million manually moderated user-generated encrypted comments from Gazzetta,\cite{gazzetta} a Greek sports news portal. \citet{Calderón2022} provided a multilingual dataset in Greek, Italian, and Spanish, containing records of racist/xenophobic hate speech from various sources such as news articles, Twitter, YouTube, and Facebook.
\citet{nikiforos2020} developed a manually annotated corpus on bullying within Virtual Learning Communities using conversations from Wikispaces. The corpus was annotated by two annotators, but the dataset was not publicly released. Finally, \citet{lekea2018detecting} annotated manifestos authored by members of the terrorist organization 17 November with hate speech tags (moderate/apparent/no hate speech). However, they did not publicly release these annotated documents, and details of the annotation process were not clearly provided in their publication.

\subsection{Summary of Ethics and NLP in Greek} 
Studies on Ethics and \gls{nlp} in the context of Greek language are predominantly focused on Toxicity Detection rather than addressing issues related to Responsible AI, Data Ethics, or Privacy Preservation, which were covered in the second ACL Workshop on Ethics in \gls{nlp}.\cite{ws-2018-acl} 
Toxicity detection in Greek has been facilitated within the setting of the OffensEval multilingual shared task. The Greek subtask attracted approximately half the submissions compared to its English counterpart. Additional studies on Toxicity detection include collaborations with private companies and efforts specifically targeting hate speech against certain groups. An analysis of the toxicity detection methods and \glspl{lr} available for Greek  reveals that published studies are more concerned with social issues pertinent to Greek society\cite{Calderón2022,Perifanos2021,lekea2018detecting,Nikiforos2021} rather than improving existing methods or sharing new \glspl{lr}. With a few exceptions focusing on hate speech detection from a social perspective,\cite{nikiforos2020, lekea2018detecting} the predominant approach in this domain relies heavily on \gls{dl} techniques. Notably, shared \glspl{lr} for Greek often lack conversational context, aligning with a global trend observed in toxicity detection research.\cite{pavlopoulos2020toxicity} Furthermore, studies predominantly target social groups responsible for generating hate speech, overlooking hate speech originating from marginalized groups, who may use it as a form of self-defense or as a way to assert their rights.

\section{Track: Summarization}\label{sec:summarization}

Summarization is the task of automatically generating concise and coherent summaries from longer texts while preserving key information and overall meaning.\cite{el2021automatic} It aims to distill the most important points, ideas, or arguments from a document or multiple documents into a shorter version, typically a paragraph or a few sentences.\cite{maybury1995generating} Summarization in \gls{nlp} research began back to 1958 with Luhn’s work,\cite{luhn1958automatic} which automatically excerpted abstracts of magazine articles and
technical papers. There are generally two main types of summarization: \textbf{Extractive Summarization}, i.e. extracting the most important sentences or phrases directly from the original text and then combining them to form a summary,\cite{giarelis2023abstractive} and \textbf{Abstractive Summarization}, i.e. generating new sentences that convey the essential information from the original text in a more concise manner. This method may involve paraphrasing and rephrasing content using \gls{nlg} techniques, such as \gls{rl} approaches and sequence-to-sequence Transformer architectures.\cite{alomari2022deep} In Greek, the most common language generation approaches use sequence-to-sequence Transformer architectures.

\subsection{Summarization in Greek: Language Models and Methods}
For the Greek language, we identified extractive summarization techniques implemented in shared task series and a workshop, as well as two \glspl{plm} used to create abstractive summaries. Additionally, another study also investigated both types of summaries in the legal domain.

\paragraph{Extractive Summarization}

The \textbf{Financial Narrative Summarization Shared Task (FNS)} has been part of the Financial Narrative Processing (FNP) workshop series since 2020. It involves generating either abstractive or extractive summaries from financial annual reports. Initially focusing on English, it expanded to include Greek and Spanish in 2022.\cite{el-haj-etal-2022-financial} In FNS-2023,\cite{zavitsanos2023financial} six systems from three participating teams competed, with extractive summarization being predominant among the top three systems.\cite{shukla2023generative, vanetik2023summarizing} The winning system\cite{shukla2023generative} employed an algorithm\cite{shukla2022dimsum} that allocates words across narrative sections based on their weights, combining and summarizing the content. The second and third best systems\cite{vanetik2023summarizing} performed summarization by first filtering out noisy content, either by retaining the first 10\% of the text or using the BertSum summarizer\cite{liu2019fine} to select 3,000-word segments, and then applying a Positional \gls{lm},\cite{lv2009positional} which incorporates positional information to understand the order of words in a sequence. In FNS-2022,\cite{el-haj-etal-2022-financial} 14 systems from seven teams competed, with most addressing Greek and Spanish.\cite{el-haj-etal-2022-financial} The top performing systems for Greek were all from one team,\cite{shukla2022dimsum} including the system that later won in FNS-2023, and all performed extractive summarization. These systems combined narrative sections either by ascending page order or descending weight order to generate summaries.
\citet{giannakopoulos2013multi} provided an overview of the \textbf{MultiLing 2013 Workshop} at ACL 2013, which introduced a multilingual, multi-document summarization challenge. This challenge assessed summarization systems across ten languages, including Greek. It featured two tasks: generating summaries that describe document sets as event sequences, and creating systems to evaluate these summaries. Among the seven participants, three submitted results for Greek.

\paragraph{Abstractive Summarization}
For Greek, there are two monolingual \glspl{plm} based on sequence-to-sequence Transformer architectures. \citet{evdaimon2023greekbart} developed GreekBART, a BART-based model\cite{lewis2020bart} pre-trained on  a corpus that included the same datasets as GreekBERT (see \S\ref{sec:ml-nlp}) and the Greek Web Corpus dataset.\cite{outsios2018word} \citet{giarelis2024greekt5} fine-tuned the multilingual T5 architecture \glspl{lm} (google/mt5-small,\cite{xue2021mt5} google/umt5-small,\cite{chungunimax} and google/umt5-base),\cite{chungunimax} which follow a sequence-to-sequence approach, on the GreekSUM train split dataset,\cite{evdaimon2023greekbart} creating the GreekT5 series of models. They evaluated both GreekT5 and GreekBART on the GreekSUM Abstract test split dataset,\cite{evdaimon2023greekbart} reporting that GreekT5 outperformed GreekBART in ROUGE metrics, while GreekBART excelled in the BERTScore metric. The evaluation code is available.\cite{nc0der_greekt5}

\paragraph{Extractive and Abstractive Summarization}
\cite{koniaris2023evaluation} addressed the challenge of summarizing Greek legal documents through both abstractive and extractive methods, using both automatic and human metrics for evaluation. They employed LexRank\cite{erkan2004lexrank} and Biased LexRank\cite{otterbacher2009biased} for extractive summarization, while for abstractive summarization, they used a sequence-to-sequence approach with a BERT-based encoder-decoder architecture, initializing both components with GreekBERT weights.

\subsection{Summarization in Greek: Language Resources}
Table \ref{tab:summ-datasets} presents the pertinent \glspl{lr} for summarization. We observe that one is publicly available, while the rest lack a license.

\citet{koniaris2023evaluation} created a legal corpus of 8,395 court decisions from Areios Pagos,\cite{areiospagos} the Supreme Civil and Criminal Court of Greece. This corpus includes the decisions, their summaries, and related metadata, all sourced from the Areios Pagos website. The dataset is divided into training, validation, and testing sets. \citet{evdaimon2023greekbart} collected articles from a news website on various topics to create two summarization datasets. In the first dataset, the titles serve as the summaries, while in the second, the abstracts are used. Both datasets are also divided into training, validation, and testing sets. However, these datasets are currently not accompanied by a license.
\begin{table}[H]
    \centering
    \caption{\glspl{lr} for summarization, with information on availability (Yes: publicly available, Lmt: limited availability; see Table~\ref{tab:lr-availability} for details; the citations point to URLs), annotation type (detailed in Table~\ref{tab:lr-ann-status}), the size and size unit of the summarized documents, and text domain.}
    \label{tab:summ-datasets}
    \resizebox{\textwidth}{!}{
    \begin{tabular}{|l | l | l | l| l | l |}
    \hline
    \textbf{Authors} & \textbf{Availability} & \textbf{Ann. type} & \textbf{Size}  &\textbf{Size unit} & \textbf{Domain}\\  
    \hline
    \citet{koniaris2023evaluation} & Yes\cite{greeklegalsum}&  curated &8,395& document & legal \\  
    \hline
    \citet{evdaimon2023greekbart} & Lmt\cite{greeksum}&  curated &151,000& document & news \\  
    \hline
    \citet{zavitsanos2023financial} & Lmt\cite{fns2023}&  manual &312& document & finance \\  
    \hline
    \citet{el-haj-etal-2022-financial} & Lmt\cite{fnp2022}&  manual &262& document & finance \\  
    \hline
    \citet{li2013multi} & Lmt\cite{multiling_datasets}&  manual & 1,350\textsuperscript{a} & document & news \\  
    \hline
    \end{tabular}
    }
\textsuperscript{a} This is the size of the multilingual dataset, not just the Greek portion.
\end{table}

\citet{zavitsanos2023financial} and \citet{el-haj-etal-2022-financial} supplied the Greek datasets for the FNS-2023 and FNS-202 shared subtasks, respectively. The first dataset included 312 financial reports, while the second included 262 financial reports. In both datasets, each report ranged from 100 to 300 pages and was accompanied by at least two human-generated gold-standard summaries.

\citet{li2013multi} provided the corpora for the MultiLing 2013 Workshop at ACL 2013,\cite{giannakopoulos2013multi} based on a subset of English content from WikiNews.\cite{wikinews} This English content was manually translated into nine languages, including Greek, resulting in nine parallel corpora. The English-Greek parallel corpus included ten topics, each with a human-generated summary serving as the gold standard.

\subsection{Summary of Summarization in Greek} 

Summarization in Greek has gained notable attention recently, mainly due to its inclusion in shared tasks (FNS 2022-2023) and the introduction of two monolingual encoder-decoder \glspl{plm} that perform \gls{nlg} tasks. While the shared tasks were fostering both extractive and abstractive summarization, the top systems performed extractive summarization. In contrast, sequence-to-sequence Transformer architectures, specifically BART\cite{lewis2020bart} and T5,\cite{raffel2020exploring} were implemented to support abstractive summarization for Greek in a monolingual setting. Recent research, building on these shared tasks and \glspl{plm}, has significantly advanced the field by introducing four new datasets specifically designed for Summarization, compared to only one dataset available before 2022.

\section{Track: Question Answering}\label{sec:qa}

\gls{qa} aims to automatically answer user questions in natural language.\cite{Zhu2021} 
There are several real-world applications linked to \gls{qa}, ranging from decision and customer support systems,\cite{goodwin2016medical,li2018question} to chatbots,\cite{adamopoulou2020chatbots} and personal assistants.\cite{DEBARCELOSSILVA2020} 
Given a question such as ``Which is the birthplace of Plato?'', a \gls{qa} system is expected to predict the correct answer: ``Athens''. 
However, \gls{qa} can be very broad in terms of the types of the questions that can be asked.\cite{Gardner2019} For example, a statement such as ``My name is John'', which is not a question, could receive an answer of the form ``That is a nice name'' or ``Nice to meet you, John''. 

\paragraph{The \gls{qa} Format} The \gls{qa} format can take multiple variations. 
For instance, it can extract the answer from a given knowledge base, whether structured (e.g., documents, database) or not, or it can generate the text of the answer without any such support, as is the case with ChatGPT.\cite{chatgpt} An example of \gls{qa} through prompting ChatGPT is shown in Table \ref{tab:chatgpt}.

\begin{table*}[ht]
    \centering
    \resizebox{\textwidth}{!}{
    \begin{tabular}{|p{3.8cm} |p{9.2cm}|}
    \hline
    \bf Question &\bf Answer \\\midrule 
         \emph{What might be the future impact of widespread AI integration in daily life?} &  In a future with ubiquitous AI, we could experience personalized and efficient services, enhanced decision-making support, and seamless automation across various aspects of daily life, transforming the way we work and interact with technology.\\
         \bottomrule
    \end{tabular}
    }
    \caption{Dialogue with ChatGPT to illustrate the generative approach where a model generates answers without relying on a predefined knowledge base.}
    \label{tab:chatgpt}
\end{table*}

These systems can further be classified based on various criteria, such as the nature of the questions and the types of answers they require (e.g., factoid, refers to simple factual questions, vs. non-factoid). Additionally, classification can be based on the breadth of domain coverage and the sources of knowledge employed to generate answers (e.g., closed-domain vs. open-domain).\cite{chen-yih-2020-open}

\paragraph{\gls{qa} Evaluation} A \gls{qa} system is usually trained and evaluated on an appropriate dataset, tackling tasks such as \gls{nlu}, \gls{ir}, reasoning, and world modeling.\cite{Rogers2023,schlegel2020framework,sugawara2017evaluation} These evaluations serve to measure the system's performance across various aspects of language understanding and response generation.

\subsection{Question Answering in Greek: Language Models and Methods}
Existing \gls{qa} systems for Greek use extractive algorithms that extract answers from given knowledge bases. \citet{mountantonakis2022} introduced an open-domain, factoid cross-lingual \gls{qa} system. Specifically, they translated user questions from Greek to English and used English BERT-based \gls{qa} models to retrieve answers from DBpedia abstracts.\cite{dbpedia} The answers were then translated back into Greek before being returned to the user. Additionally, the context from which answers are retrieved, known as the knowledge base, may also be structured data. An example of this is the closed-domain, factoid \gls{qa} system by \citeauthor{Marakakis2017},\cite{Marakakis2017} which converted the user's question into a database query, searched the database, and combined the noun and verb phrases of the question with the response to form an answer in natural language.

\paragraph{Off-the-shelf} Services have been used to build \gls{qa} systems, providing conversational interfaces even without the need for programming knowledge.\cite{braun2017evaluating} Specifically, \citet{Malamas2022} and \citet{Ventoura2021} developed e-healthcare assistants, with the latter providing information and support specifically for the Covid19 pandemic. Both studies used an \gls{nlu} service, Rasa,\cite{rasa} in order to generate the answer, which first extracted the intent (e.g., pharmacy finder) and the entities (e.g., ``city'', ``time'') from the question (e.g., ``What pharmacies will be open in Athens tomorrow morning at 8?'').

\subsection{Question Answering in Greek: Language Resources}
Table~\ref{tab:qa-datasets} presents the only two Greek \gls{qa} \glspl{lr} developed in the studies reviewed. Both resources are monolingual and have limited accessibility. First, \citet{mountantonakis2022} provided an evaluation dataset for \gls{qa} systems, comprising 20 texts on diverse subjects, derived from a Greek text bank.\cite{greek_language_certification_teachers} For each of these 20 texts, they crafted ten questions along with their respective correct answers, resulting in a total of 200 questions and answers. \citet{lopes2016} offered a dataset of audio recordings and the corresponding transcriptions of 200 dialogues collected from call center interactions regarding movie inquiries. This dataset, which is available upon request, is also manually annotated with gender, task success, anger, satisfaction, and miscommunication annotations, making it suitable for a range of \gls{nlp} tasks beyond \gls{qa}. 
    
\begin{table}[H]
    \centering
    	\caption{
    \gls{qa} datasets, indicating availability (Yes: publicly available, Lmt: limited  availability; see Table~\ref{tab:lr-availability} for details; the citation point to URL), annotation type (see Table~\ref{tab:lr-ann-status} for details), size, and size unit.}
    \label{tab:qa-datasets}
    \resizebox{\textwidth}{!}{
    \begin{tabular}{|l | l | l | l| l |}	 
    \hline
    \textbf{Authors} & \textbf{Availability} & \textbf{Ann. type} & \textbf{Size}  &\textbf{Size unit}\\  
    \hline
    \citet{mountantonakis2022} & Lmt\cite{tiresias_evaluation_results} &  manual & 200 & question-answer pair \\  
    \hline
    \citet{lopes2016} & Lmt &  manual & 200 & dialogue \\ 
    \bottomrule
    \end{tabular}}
\end{table}

\subsection{Summary of Question Answering in Greek}

Reflecting on the available \gls{qa} \glspl{lr} and architectures for Greek, several key observations arise. First, there is a clear scarcity of \gls{qa} resources specifically designed for Greek. To confirm this, we conducted a search in the Hugging Face repository for Greek \gls{qa} datasets, which revealed 20 datasets containing Greek text. Of these, 18 were multilingual, which makes reliance on such resources essential for advancing Greek \gls{qa} tasks. Moreover, these datasets are predominantly translations from English. However, a limitation of these multilingual, translated datasets is that they do not offer representative samples of Greek questions. This is consistent with the observation by \citeauthor{Rogers2023},\cite{Rogers2023} who noted that multilingual \gls{qa} benchmarks often prioritize uniformity across languages, potentially overlooking natural or representative, language-specific samples. Additionally, the development of \gls{qa} architectures for Greek remains relatively underexplored, with only a handful of studies addressing this area. Notably, no deep learning-based \gls{qa} solution has been proposed specifically for Greek, which may be due to the lack of dedicated monolingual \gls{qa} \glspl{lr} for the language.

\section{Track: Machine Translation, Multilingualism, and Cross-Lingual NLP}\label{sec:mt}

This section concerns studies handling multiple languages in \gls{nlp}, including Greek. The most canonical task in this thematic domain is \gls{mt}, i.e. the automated translation from one natural language to another. Emerging as one of the earliest tasks a computer could possibly solve, \gls{mt} was inspired by Warren Weaver's ``translation memorandum'' in 1947 and IBM's word-for-word translation system in 1954.\cite{hutchins1997first} 
For Greek, most work focuses on evaluating \gls{mt}, using \gls{mt} for data augmentation, or enabling \gls{nlp} tools in other languages through \gls{mt}.

\subsection{Machine Translation, Multilingualism, and Cross-Lingual NLP in Greek: Language Models and Methods}

\paragraph{\gls{mt}} \citet{kouremenos2018novel} developed a rule-based \gls{mt} system between Greek and \gls{gsl} for producing parallel corpora of Greek and \gls{gsl} glossed text. In this context, glossing is a method for representing the meaning and grammatical structure of signed language in written form. The authors opted for a rule-based approach due to the absence of a standardized writing system for \gls{gsl}, the scarcity of publicly available \gls{gsl} grammar resources, and the lack of Greek-\gls{gsl} parallel data.
\citet{beinborn2013cognate} focused on the automatic production of cognates using character-based \gls{mt}, applying their method on language pairs with different alphabets, specifically from English to Greek, Russian, and Farsi. Their aim was to identify not only genetic cognates, meaning two words in two languages that have the same etymological origin,\cite{crystal2011dictionary} but also words that are sufficiently similar to be associated by language learners. For example, the English word ``strange'' has the Italian correspondent ``strano''. The two words have different roots and are therefore genetically unrelated. For language learners, however, the similarity is more evident than for example the English-Italian genetic cognate father-padre. Their approach relied on phrase-based \gls{smt} using the MOSES framework,\cite{koehn2007moses} but instead of translating phrases, they transformed character sequences from one language into the other, using words instead of sentences and characters instead of words.
\citet{pecina2012domain} aimed to adapt a general-domain \gls{smt} system to specific domains by acquiring in-domain monolingual and parallel data through domain-focused Web crawling. The authors specifically focused on two language pairs: English–Greek and English-French; and two domains: Natural Environment and Labor Legislation.
 
Studies evaluating \gls{mt} systems have consistently demonstrated the superior performance of \gls{nmt} compared to \gls{smt}. \citet{mouratidis2021innovatively} tackled \gls{mt} evaluation specifically for the English-Greek and English-Italian language pairs. They developed a \gls{dl} framework that integrates a \gls{rnn} with linguistic features, word embeddings, and automatic \gls{mt} metrics. The evaluation used small, noisy datasets consisting of educational video subtitles. In an earlier study, \citet{stasimioti-2020-machine} also included human evaluation, considering factors such as post-editing effort, adequacy and fluency ratings, and error classification. Previous studies of the same team retrieved by our search protocol dealing with \gls{mt} evaluation were those by \citeauthor{mouratidis2020innovative},\cite{mouratidis2020innovative,mouratidis2019ensemble}. Additionally, \citet{castilho2018evaluating} conducted a quantitative and qualitative comparative evaluation of \gls{smt} and \gls{nmt} using automatic metrics and input from a small group of professional translators. The evaluation focused on the translation of educational texts in four language pairs: English to Greek, German, Portuguese and Russian. 

\paragraph{Multilingualism} The following studies investigate various aspects of language analysis and processing across multiple languages. \citet{Giorgi2021} developed a cognitive architecture based on a large-scale \gls{nn} for processing and producing four natural languages simultaneously, exhibiting disambiguation in semantic and grammatical levels. The architecture was evaluated using two approaches: individual training for each language and cumulative training across all languages, demonstrating competence in comprehending and responding to preschool literature questions. \citet{gamallo2020measuring} measured intra-linguistic distances between six isolated European languages (including Greek), and the inter distances with other European languages.
Various linguistic measures were deployed to assess the distance between language pairs. \citet{fragkou2014text} conducted language identification using forums containing Greek, English, and Greeklish content and employed techniques borrowed from topic change segmentation. \citet{bollegala2015cross} automatically detected translations for biomedical terms and introduced a dimensionality reduction technique. 

\paragraph{Cross-Lingual \gls{nlp}}

\gls{mt} helps address the scarcity of Greek \glspl{lr}. This is achieved either through data augmentation or by enabling the use of English \glspl{lr} and leveraging \gls{nmt} models to bridge the language gap. Moreover, multilingual \glspl{plm} facilitate the knowledge transfer across languages. \citet{papaioannou-etal-2022-cross} investigated cross-lingual knowledge-transfer strategies for clinical phenotyping. They evaluated three approaches: 1) translating Greek and Spanish notes into English before using a medical-specific English encoder, 2) using multilingual encoders,\cite{conneau2020unsupervised} and 3) expanding multi-lingual encoders with adapters.\cite{pfeiffer2020mad} 
Their findings indicated that the second approach lags behind the other two. They also recommended the most suitable method based on specific scenarios. The source code is available.\cite{neuron1682_crosslingual}

The remaining studies either used data augmentation or translated text into English (to use English systems) for task resolution. To augment their data and address class imbalance, \citet{patsiouras2023greekpolitics} (see \S\ref{sec:sentiment}) translated tweets from the minority class from Greek into English and back. They included back-translated sentences only if these retained the original meaning while differing in wording. Additionally, they augmented data by substituting synonyms using a lexicon. \citet{athanasiou2017} automatically translated original Greek data into English and used both the original and translated data for \gls{sa} (\S\ref{sec:sentiment}). 
\citet{Manias2020} examined the impact of \gls{nmt} on \gls{sa} by comparing English \gls{dnn} \glspl{lm} applied to both source (English) and target (Greek, German) languages for \gls{sa} (\S\ref{sec:sentiment}).
\citet{Singh2019} did not use augmentation but proposed a cross-lingual augmentation approach by replacing part of the natural language input with its translation into another language. This was assessed across 14 languages, including Greek, in \gls{nli} (\S\ref{sec:semantics}) and \gls{qa} (\S\ref{sec:qa}). 
For the latter task, \citet{mountantonakis2022} translated user questions from Greek to English, employing English \gls{qa} \glspl{lm} for answer retrieval and then translating the answers back to Greek. 
Similarly, \citet{Papadopoulos2021penelopie} developed a Transformer-based \gls{nmt} system for English-Greek and from Greek-English translation.\cite{lighteternal_hf} By using, then, the English translations as input, they performed \gls{oie} (\S\ref{sec:ie}) and the responses were translated back into Greek.
A similar strategy was followed by \citeauthor{shukla2022dimsum},\cite{shukla2022dimsum} the top team in a financial narrative summarization shared task,\cite{el-haj-etal-2022-financial} who used text classification for English reports and then translated the top-rated narrative sections into Greek (and Spanish) (\S\ref{sec:summarization}). \gls{mt} has also been applied in Greek \gls{sa} to enable English lexicon-based feature extraction.\cite{Chatzakou2017}

\subsection{Machine Translation, Multilingualism, and Cross-Lingual NLP in Greek: Language Resources}\label{ssec:mt-lrs}

Table~\ref{tab:mt-datasets} presents multilingual \glspl{lr} created by the studies discussed in this track. \citet{prokopidis-2016-parallel} created multilingual corpora (756 language pairs) from content available on Global Voices,\cite{global_voices} a platform where volunteers publish and translate news stories in 41 languages. The Greek documents number 3,629 and are translated into 40 languages; however, not all documents were translated into every language, resulting in 17,018 document pairs in total. The resource of \citet{bollegala2015cross} consists of pairs of biomedical terms in multiple languages, comprising also the corresponding character n-gram and contextual features. \citet{Giorgi2021} created a dataset simulating parent-child verbal interactions, featuring a fictional four-year-old. Initially in English, the dataset includes around 700 sentences (declarative and interrogative), translated into Greek, Italian, and Albanian.

\begin{table}[H]
    \centering
    \caption{Multilingual \glspl{lr} containing Greek, with information on availability (Yes: publicly available, Lmt: limited  availability, No: no information provided; see Table~\ref{tab:lr-availability} for details; the citations point to URLs), annotation type (detailed in Table~\ref{tab:lr-ann-status}), size and size unit of the Greek part of the datasets, document pairs (for parallel datasets), and number of languages.}
    \label{tab:mt-datasets}
    \resizebox{\textwidth}{!}{
    \begin{tabular}{|l | l | l | l | l | l | l |}	 
    \hline
    \textbf{Authors} & \textbf{Availability} & \textbf{Ann. type} & \textbf{Size} & \textbf{Size unit} & \textbf{Document pairs} & \textbf{Languages} \\  
    \hline
    \citet{prokopidis-2016-parallel} & Yes\cite{pgv_ilsp} &  user-generated & 3,581 & document & 17,018 & 41 \\  
    \hline
    \citet{bollegala2015cross} & Lmt\cite{plos_one_supplementary_material} &  automatic &- & {term} & n/a & 5 \\  
    \hline
    \citet{Giorgi2021} & No &  manual & 700 & {sentence} & n/a & 4 \\  
    \hline
    \end{tabular}
    }
\end{table}

Multilingual datasets for \gls{mt} comprising Greek can be found in online repositories, such as Hugging Face\cite{hugging_face} or the CLARIN:EL repository.\cite{clarin_greece_inventory} Hugging Face, as accessed on June 15th, 2023, hosts a collection of 35 multilingual parallel corpora,\cite{huggingface_datasets_translation} and 49 multilingual \glspl{lm},\cite{huggingface_models_translation} created by only seven teams, where Greek is one of the many hosted languages. CLARIN:EL hosts 25 multilingual and 200 bilingual corpora, with many of them being derived from the multilingual datasets within the repository. 
Therefore, the availability of potential \glspl{lr} for \gls{mt} in Greek (e.g., in Hugging Face)\cite{huggingface_datasets} is not reflected in the corresponding number of studies in our survey. This is mainly due to the fact that our search protocol (\S\ref{sec:search}, Appendix~\ref{app:quality_assurance_round}) could not capture recent \gls{nmt} research papers encompassing numerous languages that were not explicitly mentioned in the title or the abstract.

\subsection{Summary of Machine Translation, Multilingualism, and Cross-Lingual NLP in Greek}
Work related to \gls{mt}, Multilingualism, and Cross-Lingual \gls{nlp} in Greek primarily focuses on evaluation, augmentation, and transformation purposes. Although the \glspl{lr} in our survey (see Table~\ref{tab:mt-datasets}) are limited in size, they still offer valuable insights. Additionally, useful multilingual resources, including Greek, are available online (see \S\ref{ssec:mt-lrs}). However, these resources might not always be accompanied by published papers. Even when such papers exist, they may not explicitly mention all the languages in their titles or abstracts, making these resources less visible and harder to document. This issue arises because research related to \gls{nmt} often involves multilingual \gls{nmt} models that enable translation between multiple language pairs using a single model.\cite{dabre2020survey} Building \gls{nmt} models for each language pair is impractical, even when parallel data are available.\cite{haddow2022survey} As a result, papers on multilingual research rarely include all the languages in their titles or abstracts, making it challenging for our search protocol to identify relevant studies. 

\section{Track: NLP Applications}\label{sec:nlp-apps}

In addition to the thematic topics discussed thus far, this section presents studies with \gls{nlp} applications across various domains. 

\subsection{NLP Applications in Greek: Language Models and Methods}

The studies we identified leverage \gls{nlp} tasks for applications in the legal, business, educational, clinical, and media domains in Greek.

\paragraph{Legal} \citet{papaloukas2021} conducted multi-class legal topic classification using a range of techniques, including traditional \gls{ml} methods, \gls{rnn}-based methods, and multilingual and monolingual Transformer-based methods, highlighting the efficacy of monolingual Transformer-based models. \citet{Lachana2020} developed an information retrieval tool for legal documents that allows users to search for documents based on their unique number, keywords, or individual articles. The system identifies correlations among Greek laws based on their number format, extracts tags from legal documents using \gls{tf-idf}, and decomposes laws into articles using regular expressions.
\citet{Garofalakis2016} proposed a system that uses regular expressions and pattern matching to locate and update laws, consolidating the historical revisions of legal documents. 

\paragraph{Business} \citet{paraskevopoulos2022multimodal} classified safety reports and workplace images. They employed a multimodal fusion pipeline, experimenting with Transformer-based text encoders, a visual Transformer-based model, and a visual \gls{cnn}-based architecture to predict workplace safety audits. \citet{boskou2018assessing} evaluated internal audits by combining 26 internal audit criteria into a single quality measure and applying Linear Regression with \gls{tf-idf}.

\paragraph{Education} \citet{chatzipanagiotidis2021broad} focused on readability classification for Greek. The task aims to assess whether a text is appropriate for a given group of readers with varying education levels, such as first (L1) or second language proficiency (L2), or in terms of special needs (e.g., due to cognitive disabilities). Using handcrafted features and conventional \gls{ml} algorithms, the authors classified textbooks covering various school subjects, as well as Greek as a second language.

\paragraph{Media} \citet{piskorski2023semeval} overviewed SemEval-2023 Task 3, which focuses on detecting genre category, framing, and persuasion techniques in online news across nine languages, including Greek. Participants tackled three subtasks: categorizing articles into opinion, reporting, or satire; identifying frames among 14 generic options; and detecting persuasion techniques within paragraphs using a taxonomy of 23 techniques. 41 teams submitted entries for evaluation, with Greek, Georgian, and Spanish data used only for testing. Participants predominantly used Transformer-based models across all subtasks. In the News Genre Categorization subtask, they addressed data scarcity by combining multilingual datasets, using automatic translation, or finding similar datasets, with ensemble methods being popular. In both Framing Detection and Persuasion Techniques Detection subtasks, what differentiated the participating systems were the pre-processing and data augmentation techniques.

\paragraph{Clinical} 
\citet{athanasiou2023long} developed prediction models for influenza-like illness (ILI) outbreaks using ILI surveillance, weather, and Twitter data with \gls{lstm} neural networks. They employed transfer learning to combine features from separate \gls{lstm} models for each data category. Results showed the transfer learning model integrating all three data types outperformed models using individual sources.
\citet{stamouli2023web} analyzed transcribed spoken narrative discourses using the ILSP Neural \gls{nlp} toolkit (see \S\ref{sec:syntax}) to assess linguistic features relevant to language impairments. The results indicated no significant linguistic differences between remote and in-person data collection, which validates the feasibility of remote assessment. 

\subsection{NLP Applications in Greek: Language Resources}
Table \ref{tab:nlp-app-datasets} displays datasets related to Greek \gls{nlp} Applications. The dataset presented by \citeauthor{papaloukas2021},\cite{papaloukas2021} consisting of legal resources from Greek legislation, is the only publicly available one. The corpus comes from a collection of Greek legislative documents titled ``Permanent Greek Legislation Code - Raptarchis,''\cite{raptarchis} classified from broader to more specialized categories. \citet{paraskevopoulos2022multimodal} shared a multi-modal dataset, which is available upon request, containing 5,344 safety-related observations reported by 86 Safety Officers after inspecting 486 sites. Each observation includes a brief description of the issue, accompanying images, relevant metadata, and a priority score. \citet{Garofalakis2016} provided an alpha version of the historical consolidation of 320 laws from the period 2004-2015, including the history of their revisions.

\begin{table}[H]
    \centering
    \caption{Datasets related to \gls{nlp} Applications with information on availability (Yes: publicly available, Lmt: limited  availability, No: no information provided; see Table~\ref{tab:lr-availability} for details; the citations point to URLs), annotation type (see Table~\ref{tab:lr-ann-status} for details; size, size unit, domain, and annotation type.}
\label{tab:nlp-app-datasets}
    \resizebox{\textwidth}{!}{
    \begin{tabular}{|l | l | l | l| l | l | p{5cm} |}	 
	\hline
	\textbf{Authors} & \textbf{Availability} & \textbf{Ann. type} & \textbf{Size}  &\textbf{Size unit} & \textbf{Domain} & \textbf{Annotation type}\\  
	\hline
        \citet{papaloukas2021} & Yes\cite{greek_legal_code_dataset} & curated\textsuperscript{a} &47,563 & document & legal & topic class\\  
	\hline
        \citet{paraskevopoulos2022multimodal} & Lmt &  manual & 5,344 & issue & industry & safety observation\\  
	\hline
        \citet{Garofalakis2016} & Lmt\cite{openlawsgr_greek_laws} &  automatic & 3,209& document& legal & law revision\\  
	\hline
        \citet{stamouli2023web} & No &  hybrid & 28,238 & token & general & token, lemma, pos, syntax \\  
	\hline
        \citet{piskorski2023semeval} & No &  manual & 64 & document & globally discussed topics & genre, framing and persuasion techniques\\  
	\hline
        \citet{Lachana2020} & No &  automatic & 70& document & legal & law revision \\  
	\hline
        \citet{boskou2018assessing} & No &  manual & 133 & document & finance & internal audit criteria\\  
	\hline
        \end{tabular}
    }
\textsuperscript{a} Unclear annotation process.  
\end{table}

The following datasets are not publicly available. \citet{stamouli2023web} transcribed 139 spoken narrative discourses from ten humans, while \citet{piskorski2023semeval} described the dataset for the SemEval-2023 Task 3, which includes human annotations on genre, framing, and persuasion techniques in online news across nine languages. \citet{Lachana2020} obtained a subset of Greek laws from the Government Gazette, establishing connections between the laws and categorizing the modification type of each law. \citet{boskou2018assessing} compiled a corpus from the internal audit texts found in the annual reports of 133 publicly traded Greek companies for the year 2014. The texts were manually evaluated against 26 internal audit quality criteria.

\subsection{Summary of NLP Applications in Greek}
Greek \gls{nlp} Applications span various domains, including legal, business, education, clinical, and media, with approaches often tailored to specific fields, such as consolidating historical revisions in the legal domain. 
Alternatively, they involve text classification tasks, such as assessing text readability education purposes. Additionally, approaches that are relatively uncommon in Greek \gls{nlp} literature were employed, including multimodal approaches, information retrieval, and topic classification.

\section{Discussion}\label{sec:discussion}

\subsection{Challenges for Monolingual NLP Surveys}\label{ssec:challenges}
By undertaking a monolingual survey for Greek, we were able to identify two major challenges. We describe them below to assist future monolingual surveys for any language. 

\subsubsection{Missed Multilingual Entries}
Monolingual surveys may operate in a reduced search space because the name of the language in question is part of the query. This poses particular challenges for studies focusing on specific languages within multilingual contexts, as such studies do not always explicitly mention all the languages involved in their titles or abstracts. However, the findings of our work reveal that 41\% of the surveyed papers address multilingual research. While multilingual research may not be as extensively represented as monolingual research, it is sufficiently represented within the corpus examined.

\subsubsection{False Positives}
On the other hand, the language name may be mentioned in the title or abstract not because it was the focus of the study, but for a variety of other reasons (e.g., in the case of Greek, the etymology of a word, terminology, classical studies, etc.). \citeauthor{liu2021multi},\cite{liu2021multi} for example, while mentioning an English term derived from a Greek one, explicitly mention the language in the abstract, which confused our search. The exact sentence was ``Meme is derived from the Greek word ``Mimema'' and refers to the idea being imitated.''  
Another example is the work of \citeauthor{das2016computational},\cite{das2016computational} who referenced Ancient Greek, which falls outside the scope of this study. The exact sentence mentioned in the abstract was: ``Indian epics have not been analyzed computationally to the extent that Greek epics have.''

The challenge of retrieving false positives was particularly pronounced for Greek. This is illustrated by the following observation: we compared the search results in Google Scholar for specific languages, sampled from well-supported (i.e., first tier) or moderately-supported (i.e., second tier) languages (Figure~\ref{fig:acl_languages}). A search for the German language, which belongs to the first tier, yielded 5,100,000 results, for Arabic (first tier) 3,010,000, for Finnish (second tier) 2,440,000, and for Latin (second tier) 4,700,000. Greek (second tier) returned 4,840,000 results, a number that is comparable to or higher than languages with stronger support. 

Our search protocol limited the search for the language name, i.e., ``Greek'', to the title or abstract of the papers. Meanwhile, the term ``Natural Language Processing'' was searched across the entire paper (see \S\ref{sec:search} for details). This approach retrieved 1,135 unique papers, 142 of which were relevant to the survey's purpose. Although restricting the language name to the title or abstract might not capture all published papers for Greek \gls{nlp}, it was a necessary measure to reduce the high number of false positives that could have been returned without this restriction.

\subsubsection{{Monolingual NLP Survey Scarcity}} The level of the challenge ultimately depends on the language, but both issues mentioned above (missed multilingual entries and false positives) apply to some extent to any language. This added difficulty may explain the general scarcity of monolingual \gls{nlp} surveys. By retrieving language-specific \gls{nlp} surveys from Google Scholar, published between 2012 and 2023, we found that only 19\% of well- and moderately-supported languages have monolingual surveys for \gls{nlp} (see \S\ref{sec:other-lang-surveys}). These surveys, however, do not provide the search protocol used, hindering reproducibility and interpretability, nor do they classify the available \glspl{lr} for quality and licensing, which makes them of limited use for experiments with multilingual and \glspl{llm}.

\subsection{Greek NLP Trends}

\begin{figure}[H]
\centering
\includegraphics[trim={10 20 20 0},clip,width=.8\columnwidth]{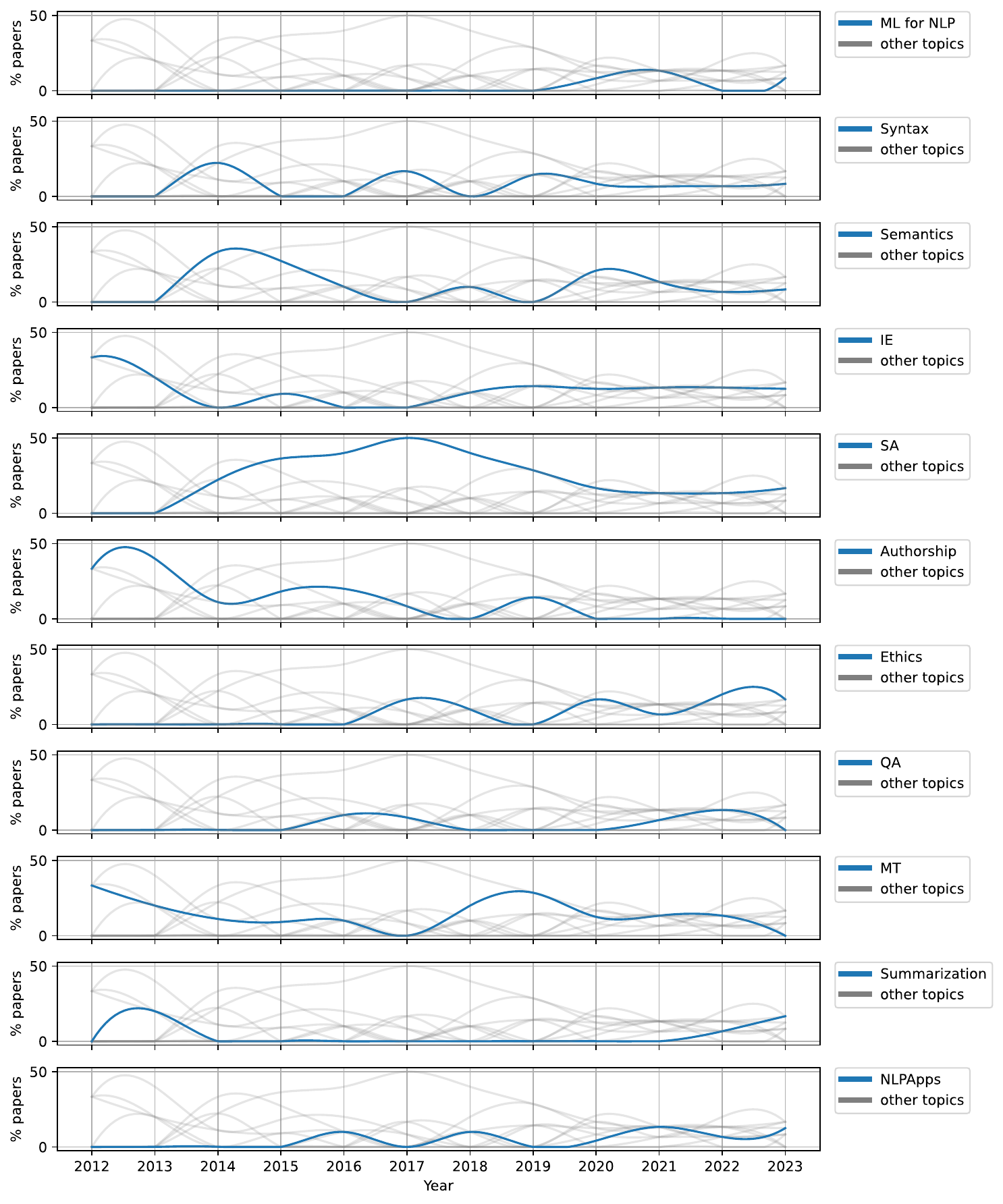}
\caption{Relative popularity of tracks over time. Each time-series illustrates the relative percentage of studies per year for a specific track (bold blue line) with time-series of other tracks appearing shadowed in the background.}
\label{fig:track-year-distr}
\end{figure}

\paragraph{NLP Tracks Trends}
From the outcomes of our survey, we were able to extract insights regarding trends in the \gls{nlp} tracks. These shed light on shifts in the prevalence of \gls{nlp} thematic areas in Greek from 2012 to 2023. Figure~\ref{fig:track-year-distr} illustrates the relative track popularity, with each presented as a time-series (bold blue line), as opposed to the rest of the tracks that appear shadowed in the background.
While we did not search for \gls{nlp} surveys in each of the thousands of less-supported languages, we acknowledge that the situation is likely worse for them, likely due to the limited attention they have received from the \gls{nlp} research community.

\gls{sa} emerged as the most prominent topic, constituting about a quarter of publications (see Figure~\ref{fig:track-year-distr}), yet shows a declining trend since 2018, echoing global patterns.\cite{rohatgi2023acl} 
In contrast, the tracks that have gained focus in recent years and align with global trends are Ethics and \gls{nlp}, Summarization, \gls{ie}, and \gls{nlp} Applications. Their gained focus is likely driven by the rise of Transformer-based \glspl{plm}, which also explains the recent relative interest in \gls{ml} for \gls{nlp}.

Semantics experiences fluctuations across the years, with three peaks . The latest peak, in 2020, is smaller than the initial peak in 2014, signaling a decline in interest after a slight rise around 2018. Notably, Semantics studies in Greek predominantly focus on Lexical Semantics, rather than Sentence-Level Semantics or Discourse Analysis. Syntax maintains interest within the Greek \gls{nlp} community. This contrasts with global trends, as reflected in the shrinking number of submissions to ACL tracks focused on Syntax: Tagging, Chunking, and Parsing.\cite{rohatgi2023acl}

\gls{mt} showed a decline in interest in 2023, following a slight peak in the preceding years. This aligns with global trends,\cite{rohatgi2023acl} where interest peaked in the early 2010s and has been steadily declining since. In this survey, we may have missed papers addressing research in a multilingual context (\S\ref{ssec:challenges}), which is primarily the case in \gls{mt}, so the trends may not be entirely accurate for this track. 

Authorship Analysis also shows a decline in interest, with the most recent paper dating to 2019 and the majority of publications concentrated between 2012 and 2017. Notably, studies within this topic primarily focus on traditional tasks like author profiling, authorship attribution and verification, rather than exploring newer tasks such as robots profiling or style change detection.

\paragraph{Shared Tasks for Greek} 
Shared tasks are a well-established method for advancing \gls{nlp} research, helping to define best practices and introduce new datasets.\cite{escartin2017ethical} Therefore, it is crucial to include less-supported languages in these task-specific events. For Greek, the tasks in shared tasks and workshops that address its context are Author Identification, Author Clustering, Offensive Language Detection, and Summarization. Specifically, the PAN Workshop series focused on Greek for Authorship Analysis tasks, such as author identification and clustering, from 2013 to 2016 (\S\ref{sec:authorship}). In 2020, the OffensEval-2020 shared task included Greek in its Offensive Language Detection challenge (\S\ref{sec:toxicity}). The most recent is the FNS shared task from the FNP workshop series, which, since 2022, targets Summarization for the Greek context. Summarization for Greek was also addressed in an earlier workshop, Multiling 2013 (\S\ref{sec:summarization}).

\subsection{Analyzing Greek Datasets}

\paragraph{Dataset Availability} 
We analyzed the Greek datasets developed by the authors of the surveyed studies with regard to their availability as defined in \S\ref{tab:lr-availability}. A total of 94 datasets were identified, with more than half (59.6\%) characterized as available by their creators. However, only 14.8\% of the total datasets are publicly available according to our availability schema, meaning they are accessible, licensed, and in machine-readable format. This highlights the limited proportion of truly open resources despite claims of availability

Among the remaining datasets, 33.3\% are of limited availability. These include datasets that either lack a license, require a fee for access, or are accessible only upon request. Notably, 11.7\% of the datasets yielded an HTTP error during our access attempts. This issue stems from storage on web pages that often lack proper maintenance and curation, such as institutional repositories or personal websites. These web pages often do not provide the same preservation guarantees as established trusted data repositories, such as GitHub\cite{github} (used by \citet{evdaimon2023greekbart,korre2021elerrant, kavros2022soundexgr,Bilianos2022,Perifanos2021, gari2021let}), Zenodo\cite{zenodo} (used by \citet{papantoniou2023automating,dritsa2022greek, fitsilis2021development, stamatatos:2016}), Hugging Face\cite{hugging_face} (used by \citet{koniaris2023evaluation,papaloukas2021,Zampieri2020}), or CLARIN:EL\cite{clarin_greece_inventory} (used by \citet{Pontiki2020}). 

The largest category, comprising 40.4\% of the datasets, includes those for which no availability information was provided. We observed that 31 of these 40 datasets are human-annotated, underscoring a significant missed opportunity to expand the pool of gold-standard resources for Greek \gls{nlp}.

\begin{figure}[H]
\centering
\includegraphics[width=.8\columnwidth]{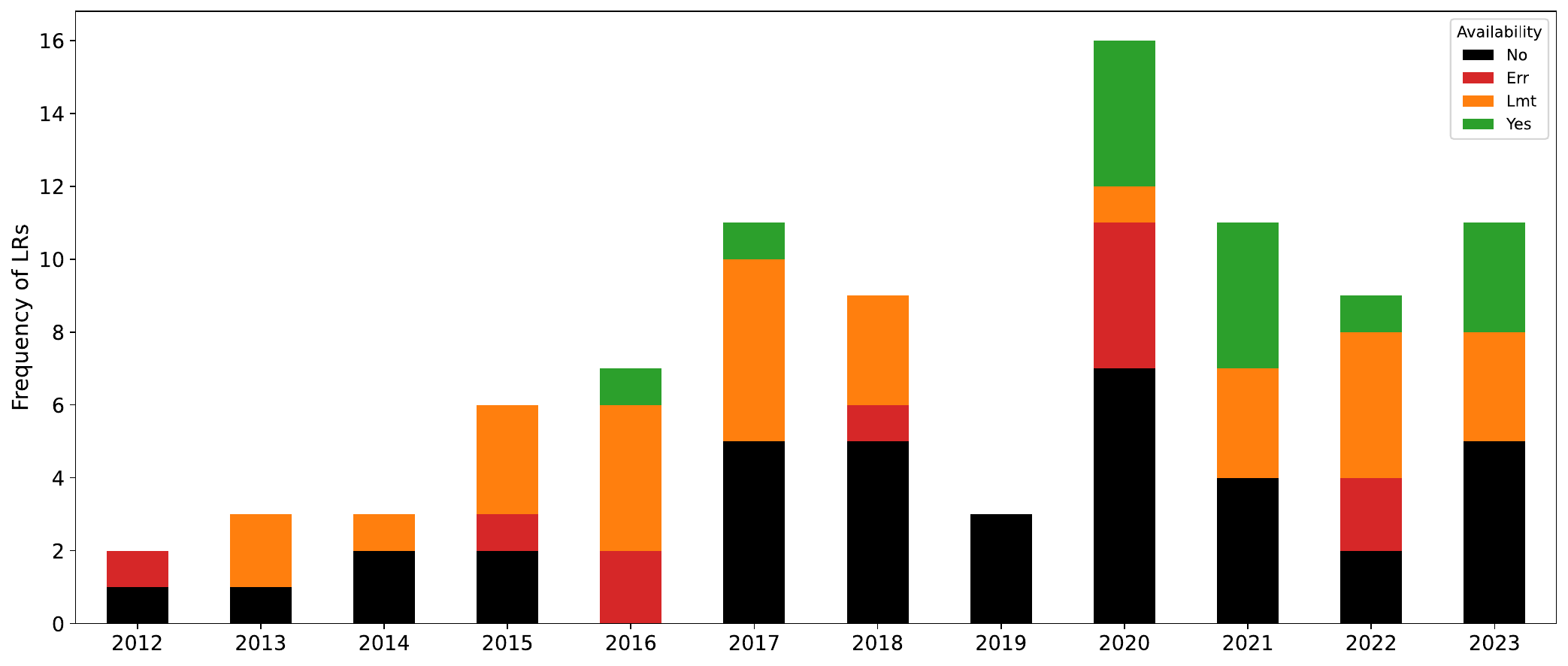}
\caption{Number of Greek datasets per year according to their availability classification (Yes: publicly available, Lmt: limited public availability, Err: unavailable, No: no information provided; see Table~\ref{tab:lr-availability} for more details).}
\label{fig:avail-data-per-year}
\end{figure}

Expanding this analysis further, Figure~\ref{fig:avail-data-per-year} presents the availability of Greek \glspl{lr} over the years. Publicly available datasets (depicted in green) have been consistently provided since 2020, reflecting recent trends favoring data sharing and open access.\cite{cao2023rise} 
In contrast, datasets with no availability information (in black) and those of limited availability (in orange) have persisted across all years, often representing a significant proportion of the datasets in each year. This observation indicates that issues of restricted access are systemic and not confined to earlier periods. Datasets inaccessible due to HTTP errors (in red) appear sporadically across the years, likely due to inadequate storage maintenance.

\paragraph{Standardization of Annotations}

Annotated datasets constitute the majority of datasets developed in the surveyed papers, underscoring the importance of standardized annotation schemes for methodological comparability, especially for tasks where consistency and comparability across methodologies are crucial. Standardized annotations enable consistent benchmarks, enhance comparability of methods, and foster integration with broader research efforts. For example, in Syntax and \gls{gec}, standardized annotations are vital. 
\citet{prokopidis2017universal} developed the Greek \gls{ud} treebank for dependency parsing. This is part of the \gls{ud} project\cite{nivre2016universal} that offers standardized treebanks and provides consistent and unified annotation practices across languages. \citet{korre2021elerrant} established an annotation schema for grammatical error types in Greek. Similarly, in \gls{ner}, \citet{Bartziokas2020} provided two annotated datasets, one with four label tags akin to the CONLL-2003 dataset,\cite{sang2003introduction} and the other incorporating 18 tags for entities, as in the OntoNotes 5 English dataset,\cite{pradhan2007ontonotes} enabling direct comparison across languages and studies. Standardized guidelines are also critical in Toxicity Detection and in \gls{sa}, where annotation consistency is needed to compare models addressing sensitive or subjective content. For example, \citet{Zampieri2020} provided a hierarchical three-level annotation schema for offensive language identification. In \gls{sa}, the majority of studies performing emotion detection were based on Ekman's six basic emotions as a standard framework. 

\paragraph{Multilingual Nature of Datasets}
In terms of linguality (number of languages covered, i.e., mono- or multi-lingual), approximately 86\% of the datasets are monolingual, while the remaining portion comprises bilingual or multilingual datasets. Furthermore, only a small fraction (5.3\%) of all datasets comprises translations into Greek. This limited reliance on translations is advantageous, as translations may not faithfully capture the natural linguistic features and nuances of native Greek speakers.\cite{Rogers2023} The emphasis on native Greek datasets enhances the representativeness and quality of language resources for \gls{nlp} tasks.

\subsection{Greek NLP Datasets: Raw and Multi-Task}\label{sec:res-and-eval}
The \glspl{lr} presented thus far include annotations for specific \gls{nlp} tasks. Additionally, we identified corpora that provide raw text without any meta-information, as well as those with meta-information that can be used for various \gls{nlp} tasks. Table~\ref{tab:res-and-eval} shows all these datasets in Greek, classified by their availability, annotation type, and any automatically-extracted meta-information included.
\begin{table}[ht!]
    \centering
    \caption{Raw or annotated for various \gls{nlp} tasks Greek datasets classified according to their availability (Yes: publicly available, Lmt: limited availability, Err: unavailable; see Table~\ref{tab:lr-availability} for details; the citations point to URLs), annotation type (see Table~\ref{tab:lr-ann-status} for details), size, size unit, and automatically-extracted meta-information.}
    \label{tab:res-and-eval}
    \resizebox{\textwidth}{!}{
    \begin{tabular}{|l|l|l|l|l|p{0.3\linewidth}|} 
    \hline
    \textbf{Authors} & \textbf{Availability} & \textbf{Ann. type} & \textbf{Size} & \textbf{Size unit} & \textbf{Meta-information} \\  
    \hline
    \citet{dritsa2022greek} & Yes\cite{greek_parliament_proceedings_dataset} & \ automatic & 1.28M & \ political speech & \ member name, sitting date, parliamentary period, parliamentary session, parliamentary sitting, political party, government, member region, roles, member gender, speech\\  
    \hline
    \citet{fitsilis2021development} & Yes\cite{parliamentary_questions_corpus} & \ automatic & 2,000 & \ parliamentary question & \ question type, subject, parliamentary period, parliamentary session, political party, submitter, ministers, ministers \\ 
    \hline
    \citet{prokopidis2020neural}& Yes\cite{neural_nlp_toolkit_greek} & \ no annotation & 101,857 & \ web page & \ n/a\\ 
    \hline
    \citet{Barzokas2020}& Yes\cite{greek_words_evolution} & curated & 34.88M & \ token & \ literature type, author, pub. year, isbn\\ 
    \hline
    \citet{lopes2016} & \ Lmt & manual & 200 & \ dialogue & \ gender, task success, anger, satisfaction, and miscommunication \\ 
    \hline
    \citet{Lioudakis2019} & Err\cite{aueb-resources_Lioudakis2019} & \ curated\textsuperscript{a} & 8,005 & \ article & \ topic\\ 
    \hline
    \citet{iosif-etal-2016-cognitively}& Err\cite{adsms_actindex} & no annotation & 66M & \ web document snippet & n/a\\ 
    \hline
    \end{tabular}
    }
\textsuperscript{a} Unclear annotation process.
\end{table}

\subsubsection*{Greek datasets with no annotations} 
Datasets without annotations are versatile resources that are potentially applicable to various \gls{nlp} tasks beyond their original intended purposes. For instance, publicly available raw \glspl{lr} can facilitate pre-training purposes, such as (masked) language modeling. \citet{prokopidis2020neural} created a dataset of 101,857 open content web-pages, comprising online archives of Greek newspapers from 2003 to 2020 and the Greek part of the w2c corpus,\cite{majlivs2012language} scraped and pre-processed. \citet{iosif-etal-2016-cognitively} shared a dataset consisting of web document snippets in English, German, Italian, and Greek, with the Greek portion comprising 66M.

\subsubsection*{Greek datasets for various NLP tasks}
\citet{fitsilis2021development} assembled a corpus of 2,000 parliamentary questions from 2009 to 2019, corresponding to 638,865 tokens, while \citet{dritsa2022greek} processed 1.28M political speeches from Greek parliamentary records, spanning from July 1989 to July 2020. Both datasets are publicly available and they comprise automatically-extracted metadata related to parliamentary procedures (e.g., parliamentary period, political party). \citet{Barzokas2020} generated a corpus of 34.88M tokens by processing e-books from Project Gutenberg,\cite{project_gutenberg} and from OpenBook.\cite{openbook_gr} \citet{lopes2016} developed a dataset comprising audio recordings and transcripts of 200 dialogues from call-center interactions regarding movie inquiries, annotated with gender, task success, anger, satisfaction, and miscommunication. Additionally, \citet{Lioudakis2019} compiled a dataset sourced from the online corpus of the newspaper ``Macedonia'',\cite{makedonia_corpus} consisting of 8,005 articles annotated with their respective topics.

\subsection{Emerging Greek Benchmark Datasets}\label{ssec:leak}
Table~\ref{tab:leakage} summarizes Greek datasets from our survey that qualify as benchmarks for \gls{nlp} research. These datasets meet strict criteria. First, they should be publicly available (see the definition in \S\ref{sec:lr-avail-taxonomy}). That is, they should be accessible, licensed, free of charge, and in a machine-readable format. Second, they should include human-generated annotations (see the definition in \S\ref{sec:lr-ann-status-taxonomy}), which are classified as ``manual'', ``curated'', ``user-generated'' or ``hybrid'' in Tables~\ref{tab:syntax-datasets}-\ref{tab:nlp-app-datasets}.

\begin{table}[ht!]
    \centering
    \caption{Emerging Greek NLP benchmark datasets: Publicly-available Greek datasets with human-generated annotations. Train/test splits are indicated. The dataset citations point to URLs.}
    \label{tab:leakage}
    \resizebox{\textwidth}{!}{
    \begin{tabular}{|l | l | l | l|}	 
	\hline
	\textbf{Authors} & \textbf{Dataset} & \textbf{Task} & \textbf{Split} \\  
	\hline
        \citet{koniaris2023evaluation} & DominusTea/GreekLegalSum\cite{greeklegalsum} & summarization & \ yes \\ 
        \hline
        \citet{rizou2023efficient} & Uniway\cite{uniway_dataset} & ner, intent cl. & \ no\\ 
        \hline
        \citet{papaloukas2021} & AI-team-UoA/greek\_legal\_code\cite{greek_legal_code_dataset} & topic cl. & \ yes \\ 
        \hline
        \citet{korre2021elerrant} & GNC,\cite{elerrant} GWE\cite{elerrant} & gec & \ no \\ 
        \hline
        \citet{Bartziokas2020} & elNER\cite{elner} & ner & \ yes\textsuperscript{a} \\ 
        \hline
        \citet{Zampieri2020} & strombergnlp/offenseval\_2020\cite{strombergnlp_offenseval_2020} & toxicity det. & \ yes \\ 
        \hline
        \citet{Barzokas2020} & openbook, project\_gutenberg\cite{greek_words_evolution} & text cl. & \ no \\ 
        \hline
        \citet{prokopidis2017universal} & UD\_Greek-GDT\cite{ud-greek-gdt} & syntax& \ yes \\ 
        \hline
        \citet{prokopidis-2016-parallel} & PGV\cite{pgv_ilsp} & mt & \ no \\
        \hline
    \end{tabular}
    }
\textsuperscript{a} Only training splits.
\end{table}

Of the 91 annotated datasets identified across all \gls{nlp} track sections, only nine meet these criteria. While most provide train-test splits, reducing data leakage risks, four datasets lack split specifications,\cite{rizou2023efficient,korre2021elerrant,Barzokas2020,prokopidis-2016-parallel} necessitating caution in evaluation use. These datasets span across nine different \gls{nlp} tasks, including Summarization, \gls{ner}, Intent Classification, Topic Classification, \gls{gec}, Toxicity Detection, Syntactical and Morphological Analysis, \gls{mt}, and Text Classification.

By revisiting the \glspl{lr} tables of the \gls{nlp} track sections (Tables \ref{tab:syntax-datasets} - \ref{tab:nlp-app-datasets}), certain resources marked as ``Lmt'' in the availability type column could potentially become publicly available through proper licensing. Similarly, resources marked as ``Err'' could be converted into accessible benchmark datasets by curating their storage pages. These steps could transform 17 additional datasets into actionable benchmarks, accelerating progress in under-supported tasks such as \gls{sa} and Authorship Analysis.

\subsection{Challenges and Opportunities for Greek in the \gls{llm} Era}\label{ssec:future-nlp}

Our work offers a detailed account of the landscape of Greek \gls{nlp}, highlighting not only the evolution of research themes but also the status of available \glspl{lr} and licensing practices. By identifying openly available datasets and models (see Tables~\ref{tab:leakage} and \ref{tab:pfms}), this study lays important groundwork for the future tuning and training of \glspl{llm} tailored to Greek. However, this forward-looking potential must be approached with caution, especially in light of the unique challenges posed by limited linguistic representation in current \glspl{llm}.

\glspl{llm} perform best on languages that are well-represented in their training data. For Greek — likely constituting only a small fraction of such datasets — models have limited exposure to its syntax, morphology, and usage patterns, as well as to idiomatic expressions, named entities, and culturally specific references. As a result, generated text in Greek may be grammatically awkward, semantically imprecise, contextually or culturally inappropriate.\cite{fokides2025comparing} Furthermore, \glspl{llm} trained predominantly on English and other high-resource languages may misinterpret polysemous words, struggle with dialectal variation and code-switching, and default to English-centric assumptions even when interacting in Greek.\cite{papadimitriou2023multilingual} Consequently, responses to Greek queries often reflect cultural norms and worldviews rooted in English-language data, while Greek-specific historical, legal, or societal knowledge may be omitted or distorted.

As \glspl{llm} become increasingly embedded in everyday applications, the risk of excluding speakers of variations of the Greek language (including regional variants) grows. Such users may experience miscommunication, reduced access to services, or feel culturally invisible within AI systems. Over time, this could reinforce social and linguistic hierarchies, further marginalizing non-standard language communities. There is thus a pressing need for more open, high-quality, and properly licensed Greek language resources—especially from communicative contexts—that can be ingested by \glspl{llm} to improve linguistic coverage, fairness, and inclusivity.

\section{Conclusions}
Our work achieves two primary goals. First, we introduce a generalizable methodology for conducting systematic monolingual \gls{nlp} surveys. By addressing the lack of standardized frameworks for monolingual surveys, we provide a replicable approach that minimizes selection bias, ensures reproducibility, and organizes findings into coherent thematic tracks. The second goal concerns our application of this methodology to create a comprehensive survey of Greek \gls{nlp} from 2012 to 2023. This methodology not only advances Greek \gls{nlp} but also serves as a blueprint for under-supported languages worldwide. 

A key contribution of our survey is the thorough cataloging of Greek \glspl{lr}, including nine publicly available, human-annotated datasets spanning nine \gls{nlp} tasks, such as Summarization, \gls{ner}, and \gls{mt}. These resources hold significant potential as benchmarks for advancing Greek \gls{nlp} research. While Greek remains resource-scarce in certain tasks (e.g., \gls{sa}), we have addressed \glspl{lr} that with licensing or maintainance resolution, can be converted easily to benchmarks. Our analysis of methodological shifts reveals that while \gls{dl} dominates post-2019, traditional \gls{ml} methods persist in certain tasks, signaling opportunities for more innovated approaches. Additionally, Greek \gls{nlp} favors monolingual language models (e.g., GreekBERT) over multilingual systems, achieving state-of-the-art results in tasks such as \gls{sa}. This preference underscores the importance of language-specific pretraining. Task-specific trends further illustrate Greek \gls{nlp}’s unique trajectory: while global interest in Syntax declines, Greek retains a strong focus, likely due to its morphological complexity. Conversely, \gls{sa} research declines locally, mirroring broader shifts toward emergent tasks like Ethics and \gls{nlp}.

To ensure accessibility and longevity, we host our survey results in an online repository, designed as a continuously evolving resource for the \gls{nlp} community. Our systematic methodology ensures unbiased and replicable results, setting a standard for future monolingual surveys. By addressing resource disparities in Greek \gls{nlp} and providing a replicable framework, our work bridges the gap between monolingual and multilingual \gls{nlp} research, promoting inclusivity and equitable progress for under-supported languages worldwide.

\section*{Resource Availability}


\subsection*{Lead contact}


Requests for further information and resources should be directed to and will be fulfilled by the lead contact, John Pavlopoulos (annis@aueb.gr).

\subsection*{Materials availability}


This study did not generate new materials.

\subsection*{Data and code availability}

The survey results are publicly available (\href{https://doi.org/10.5281/zenodo.15314882}{https://doi.org/10.5281/zenodo.15314882}),\cite{greek_nlp_survey_2025} comprising metadata of the surveyed papers/datasets and figures illustrating key findings on Greek \gls{nlp}.

\section*{Acknowledgments}


This work has been partially supported by project MIS 5154714 of the National Recovery and Resilience Plan Greece 2.0 funded by the European Union under the NextGenerationEU Program.

\section*{Author Contributions}


Conceptualization, J.B and J.P; methodology, J.B. and K.P. and M.G. and J.P.; investigation, J.B and K.P.; writing-–original draft, J.B.; writing-–review \& editing, J.B. and K.P. and M.G. and J.P.; project administration J.B. and J.P.

\section*{Declaration of Interests}


The authors declare no competing interests.

\bibliography{custom}

\begin{thebibliography}{466}
\providecommand{\natexlab}[1]{#1}
\providecommand{\url}[1]{\texttt{#1}}
\providecommand{\href}[2]{#2}
\providecommand{\path}[1]{#1}
\providecommand{\DOIprefix}{doi: }
\providecommand{\ArXivprefix}{arXiv: }
\providecommand{\URLprefix}{URL: }
\providecommand{\Pubmedprefix}{pmid: }
\providecommand{\doi}[1]{\href{http://dx.doi.org/#1}{\path{#1}}}
\providecommand{\Pubmed}[1]{\href{pmid:#1}{\path{#1}}}
\providecommand{\BIBand}{and}
\providecommand{\bibinfo}[2]{#2}
\ifx\xfnm\undefined \def\xfnm[#1]{\unskip,\space#1}\fi
\makeatletter\def\@biblabel#1{#1.}\makeatother
\bibitem[{Vaswani et~al.(2017)Vaswani, Shazeer, Parmar, Uszkoreit, Jones, Gomez, Kaiser and Polosukhin}]{vaswani2017attention}
\bibinfo{author}{Vaswani, A.}, \bibinfo{author}{Shazeer, N.}, \bibinfo{author}{Parmar, N.}, \bibinfo{author}{Uszkoreit, J.}, \bibinfo{author}{Jones, L.}, \bibinfo{author}{Gomez, A.N.}, \bibinfo{author}{Kaiser, L.u.}, and \bibinfo{author}{Polosukhin, I.} (\bibinfo{year}{2017}). \bibinfo{title}{Attention is all you need}.
\newblock In \bibinfo{editor}{ I.{ }Guyon}, \bibinfo{editor}{ U.V.{ }Luxburg}, \bibinfo{editor}{ S.{ }Bengio}, \bibinfo{editor}{ H.{ }Wallach}, \bibinfo{editor}{ R.{ }Fergus}, \bibinfo{editor}{ S.{ }Vishwanathan}, and \bibinfo{editor}{ R.{ }Garnett}, eds. \bibinfo{booktitle}{Advances in Neural Information Processing Systems} vol.~\bibinfo{volume}{30}. \bibinfo{publisher}{Curran Associates, Inc.} pp. \bibinfo{pages}{5998--6008}.
\newblock \URLprefix \url{https://proceedings.neurips.cc/paper_files/paper/2017/file/3f5ee243547dee91fbd053c1c4a845aa-Paper.pdf}.
\bibitem[{Brown et~al.(2020)Brown, Mann, Ryder, Subbiah, Kaplan, Dhariwal, Neelakantan, Shyam, Sastry, Askell, Agarwal, Herbert-Voss, Krueger, Henighan, Child, Ramesh, Ziegler, Wu, Winter, Hesse, Chen, Sigler, Litwin, Gray, Chess, Clark, Berner, McCandlish, Radford, Sutskever and Amodei}]{brown2020language}
\bibinfo{author}{Brown, T.B.}, \bibinfo{author}{Mann, B.}, \bibinfo{author}{Ryder, N.}, \bibinfo{author}{Subbiah, M.}, \bibinfo{author}{Kaplan, J.D.}, \bibinfo{author}{Dhariwal, P.}, \bibinfo{author}{Neelakantan, A.}, \bibinfo{author}{Shyam, P.}, \bibinfo{author}{Sastry, G.}, \bibinfo{author}{Askell, A.}, \bibinfo{author}{Agarwal, S.}, \bibinfo{author}{Herbert-Voss, A.}, \bibinfo{author}{Krueger, G.}, \bibinfo{author}{Henighan, T.}, \bibinfo{author}{Child, R.}, \bibinfo{author}{Ramesh, A.}, \bibinfo{author}{Ziegler, D.M.}, \bibinfo{author}{Wu, J.}, \bibinfo{author}{Winter, C.}, \bibinfo{author}{Hesse, C.}, \bibinfo{author}{Chen, M.}, \bibinfo{author}{Sigler, E.}, \bibinfo{author}{Litwin, M.}, \bibinfo{author}{Gray, S.}, \bibinfo{author}{Chess, B.}, \bibinfo{author}{Clark, J.}, \bibinfo{author}{Berner, C.}, \bibinfo{author}{McCandlish, S.}, \bibinfo{author}{Radford, A.}, \bibinfo{author}{Sutskever, I.}, and \bibinfo{author}{Amodei, D.} (\bibinfo{year}{2020}). \bibinfo{title}{Language models are few-shot
  learners}.
\newblock In \bibinfo{booktitle}{Advances in Neural Information Processing Systems} vol.~\bibinfo{volume}{33}. \bibinfo{publisher}{Curran Associates, Inc.} pp. \bibinfo{pages}{1877--1901}.
\newblock \URLprefix \url{https://proceedings.neurips.cc/paper/2020/file/1457c0d6bfcb4967418bfb8ac142f64a-Paper.pdf}.
\bibitem[{Pires et~al.(2019)Pires, Schlinger and Garrette}]{pires2019multilingual}
\bibinfo{author}{Pires, T.}, \bibinfo{author}{Schlinger, E.}, and \bibinfo{author}{Garrette, D.} (\bibinfo{year}{2019}). \bibinfo{title}{How multilingual is multilingual bert?}
\newblock In \bibinfo{booktitle}{Proceedings of the 57th Annual Meeting of the Association for Computational Linguistics}. pp. \bibinfo{pages}{4996--5001}.
\bibitem[{Qin et~al.(2025)Qin, Chen, Zhou, Chen, Li, Liao, Li, Che and Philip}]{qin2025survey}
\bibinfo{author}{Qin, L.}, \bibinfo{author}{Chen, Q.}, \bibinfo{author}{Zhou, Y.}, \bibinfo{author}{Chen, Z.}, \bibinfo{author}{Li, Y.}, \bibinfo{author}{Liao, L.}, \bibinfo{author}{Li, M.}, \bibinfo{author}{Che, W.}, and \bibinfo{author}{Philip, S.Y.} (\bibinfo{year}{2025}). \bibinfo{title}{A survey of multilingual large language models}.
\newblock \bibinfo{journal}{Patterns} \emph{\bibinfo{volume}{6}}, \bibinfo{pages}{101118}. \DOIprefix\doi{doi: 10.1016/j.patter.2024.101118}.
\bibitem[{Blasi et~al.(2022)Blasi, Anastasopoulos and Neubig}]{blasi-etal-2022-systematic}
\bibinfo{author}{Blasi, D.}, \bibinfo{author}{Anastasopoulos, A.}, and \bibinfo{author}{Neubig, G.} (\bibinfo{year}{2022}). \bibinfo{title}{Systematic inequalities in language technology performance across the world`s languages}.
\newblock In \bibinfo{editor}{ S.{ }Muresan}, \bibinfo{editor}{ P.{ }Nakov}, and \bibinfo{editor}{ A.{ }Villavicencio}, eds. \bibinfo{booktitle}{Proceedings of the 60th Annual Meeting of the Association for Computational Linguistics (Volume 1: Long Papers)}. \bibinfo{address}{Dublin, Ireland}: \bibinfo{publisher}{Association for Computational Linguistics} pp. \bibinfo{pages}{5486--5505}.
\newblock \DOIprefix\doi{10.18653/v1/2022.acl-long.376}.
\bibitem[{Bender et~al.(2021)Bender, Gebru, McMillan-Major and Shmitchell}]{bender2021dangers}
\bibinfo{author}{Bender, E.M.}, \bibinfo{author}{Gebru, T.}, \bibinfo{author}{McMillan-Major, A.}, and \bibinfo{author}{Shmitchell, S.} (\bibinfo{year}{2021}). \bibinfo{title}{On the dangers of stochastic parrots: Can language models be too big?}
\newblock In \bibinfo{booktitle}{Proceedings of the 2021 ACM conference on fairness, accountability, and transparency}. pp. \bibinfo{pages}{610--623}.
\bibitem[{Oflazer(2014)}]{oflazer2014turkish}
\bibinfo{author}{Oflazer, K.} (\bibinfo{year}{2014}). \bibinfo{title}{Turkish and its challenges for language processing}.
\newblock \bibinfo{journal}{Language resources and evaluation} \emph{\bibinfo{volume}{48}}, \bibinfo{pages}{639--653}. \DOIprefix\doi{https://doi.org/10.1007/s10579-014-9267-2}.
\bibitem[{Shoufan and Alameri(2015)}]{shoufan2015natural}
\bibinfo{author}{Shoufan, A.}, and \bibinfo{author}{Alameri, S.} (\bibinfo{year}{2015}). \bibinfo{title}{Natural language processing for dialectical arabic: A survey}.
\newblock In \bibinfo{booktitle}{Proceedings of the second workshop on Arabic natural language processing}. pp. \bibinfo{pages}{36--48}.
\bibitem[{Al-Ayyoub et~al.(2018)Al-Ayyoub, Nuseir, Alsmearat, Jararweh and Gupta}]{al2018deep}
\bibinfo{author}{Al-Ayyoub, M.}, \bibinfo{author}{Nuseir, A.}, \bibinfo{author}{Alsmearat, K.}, \bibinfo{author}{Jararweh, Y.}, and \bibinfo{author}{Gupta, B.} (\bibinfo{year}{2018}). \bibinfo{title}{Deep learning for arabic nlp: A survey}.
\newblock \bibinfo{journal}{Journal of computational science} \emph{\bibinfo{volume}{26}}, \bibinfo{pages}{522--531}. \DOIprefix\doi{https://doi.org/10.1016/j.jocs.2017.11.011}.
\bibitem[{Larabi Marie-Sainte et~al.(2019)Larabi Marie-Sainte, Alalyani, Alotaibi, Ghouzali and Abunadi}]{Marie2019}
\bibinfo{author}{Larabi Marie-Sainte, S.}, \bibinfo{author}{Alalyani, N.}, \bibinfo{author}{Alotaibi, S.}, \bibinfo{author}{Ghouzali, S.}, and \bibinfo{author}{Abunadi, I.} (\bibinfo{year}{2019}). \bibinfo{title}{Arabic natural language processing and machine learning-based systems}.
\newblock \bibinfo{journal}{IEEE Access} \emph{\bibinfo{volume}{7}}, \bibinfo{pages}{7011--7020}. \DOIprefix\doi{10.1109/ACCESS.2018.2890076}.
\bibitem[{Papantoniou and Tzitzikas(2020)}]{papantoniou2020nlp}
\bibinfo{author}{Papantoniou, K.}, and \bibinfo{author}{Tzitzikas, Y.} (\bibinfo{year}{2020}). \bibinfo{title}{{NLP} for the {G}reek language: a brief survey}.
\newblock In \bibinfo{booktitle}{11th Hellenic Conference on Artificial Intelligence}. pp. \bibinfo{pages}{101--109}.
\newblock \DOIprefix\doi{https://doi.org/10.1145/3411408.3411410}.
\bibitem[{Alam et~al.(2021)Alam, Hasan, Alam, Khan, Tajrin, Khan and Chowdhury}]{alam2021review}
\bibinfo{author}{Alam, F.}, \bibinfo{author}{Hasan, A.}, \bibinfo{author}{Alam, T.}, \bibinfo{author}{Khan, A.}, \bibinfo{author}{Tajrin, J.}, \bibinfo{author}{Khan, N.}, and \bibinfo{author}{Chowdhury, S.A.} (\bibinfo{year}{2021}). \bibinfo{title}{A review of bangla natural language processing tasks and the utility of transformer models}.
\newblock \bibinfo{journal}{Preprint at arXiv \url{https://doi.org/10.48550/arXiv.2107.03844}}.
\bibitem[{H{\"a}m{\"a}l{\"a}inen and Alnajjar(2021)}]{hamalainen2021current}
\bibinfo{author}{H{\"a}m{\"a}l{\"a}inen, M.}, and \bibinfo{author}{Alnajjar, K.} (\bibinfo{year}{2021}). \bibinfo{title}{The {C}urrent {S}tate of {F}innish {NLP}}.
\newblock In \bibinfo{booktitle}{Proceedings of the Seventh International Workshop on Computational Linguistics of Uralic Languages}. pp. \bibinfo{pages}{65--72}.
\bibitem[{Desai and Dabhi(2021)}]{desai2021taxonomic}
\bibinfo{author}{Desai, N.P.}, and \bibinfo{author}{Dabhi, V.K.} (\bibinfo{year}{2021}). \bibinfo{title}{Taxonomic survey of {H}indi {L}anguage {NLP} systems}.
\newblock \bibinfo{journal}{Preprint at arXiv \url{https://doi.org/10.48550/arXiv.2102.00214}}.
\bibitem[{Guellil et~al.(2021)Guellil, Sa{\^a}dane, Azouaou, Gueni and Nouvel}]{guellil2021arabic}
\bibinfo{author}{Guellil, I.}, \bibinfo{author}{Sa{\^a}dane, H.}, \bibinfo{author}{Azouaou, F.}, \bibinfo{author}{Gueni, B.}, and \bibinfo{author}{Nouvel, D.} (\bibinfo{year}{2021}). \bibinfo{title}{Arabic natural language processing: {A}n overview}.
\newblock \bibinfo{journal}{Journal of King Saud University-Computer and Information Sciences} \emph{\bibinfo{volume}{33}}, \bibinfo{pages}{497--507}. \DOIprefix\doi{https://doi.org/10.1016/j.jksuci.2019.02.006}.
\bibitem[{Darwish et~al.(2021)Darwish, Habash, Abbas, Al-Khalifa, Al-Natsheh, Bouamor, Bouzoubaa, Cavalli-Sforza, El-Beltagy, El-Hajj et~al.}]{darwish2021panoramic}
\bibinfo{author}{Darwish, K.}, \bibinfo{author}{Habash, N.}, \bibinfo{author}{Abbas, M.}, \bibinfo{author}{Al-Khalifa, H.}, \bibinfo{author}{Al-Natsheh, H.T.}, \bibinfo{author}{Bouamor, H.}, \bibinfo{author}{Bouzoubaa, K.}, \bibinfo{author}{Cavalli-Sforza, V.}, \bibinfo{author}{El-Beltagy, S.R.}, \bibinfo{author}{El-Hajj, W.} et~al. (\bibinfo{year}{2021}). \bibinfo{title}{A panoramic survey of natural language processing in the arab world}.
\newblock \bibinfo{journal}{Communications of the ACM} \emph{\bibinfo{volume}{64}}, \bibinfo{pages}{72--81}. \DOIprefix\doi{http://dx.doi.org/10.1145/3447735}.
\bibitem[{Rajendran et~al.(2022)Rajendran, Anand~Kumar, Rajalakshmi, Dhanalakshmi, Balasubramanian and Soman}]{rajendran2022tamil}
\bibinfo{author}{Rajendran, S.}, \bibinfo{author}{Anand~Kumar, M.}, \bibinfo{author}{Rajalakshmi, R.}, \bibinfo{author}{Dhanalakshmi, V.}, \bibinfo{author}{Balasubramanian, P.}, and \bibinfo{author}{Soman, K.} (\bibinfo{year}{2022}). \bibinfo{title}{Tamil {NLP} {T}echnologies: {C}hallenges, {S}tate of the {A}rt, {T}rends and {F}uture {S}cope}.
\newblock In \bibinfo{booktitle}{International Conference on Speech and Language Technologies for Low-resource Languages}. \bibinfo{organization}{Springer} pp. \bibinfo{pages}{73--98}.
\bibitem[{Gonzalez-Dios and Altuna(2022)}]{gonzalez2022natural}
\bibinfo{author}{Gonzalez-Dios, I.}, and \bibinfo{author}{Altuna, B.} (\bibinfo{year}{2022}). \bibinfo{title}{{N}atural {L}anguage {P}rocessing and {L}anguage {T}echnologies for the {B}asque {L}anguage}.
\newblock \bibinfo{journal}{Cuadernos Europeos de Deusto} \emph{\bibinfo{volume}{20}}, \bibinfo{pages}{203--230}. \DOIprefix\doi{https://doi.org/10.18543/ced.2477}.
\bibitem[{Athanasiou and Maragoudakis(2017)}]{athanasiou2017}
\bibinfo{author}{Athanasiou, V.}, and \bibinfo{author}{Maragoudakis, M.} (\bibinfo{year}{2017}). \bibinfo{title}{A novel, gradient boosting framework for sentiment analysis in languages where nlp resources are not plentiful: A case study for modern greek}.
\newblock \bibinfo{journal}{Algorithms} \emph{\bibinfo{volume}{10}}. \DOIprefix\doi{10.3390/a10010034}.
\bibitem[{Papadopoulos et~al.(2021{\natexlab{a}})Papadopoulos, Papadakis and Matsatsinis}]{Papadopoulos2021penelopie}
\bibinfo{author}{Papadopoulos, D.}, \bibinfo{author}{Papadakis, N.}, and \bibinfo{author}{Matsatsinis, N.F.} (\bibinfo{year}{2021}{\natexlab{a}}). \bibinfo{title}{{PENELOPIE:} enabling open information extraction for the greek language through machine translation}.
\newblock \bibinfo{journal}{Preprint at arXiv \url{https://doi.org/10.48550/arXiv.2103.15075}}.
\bibitem[{Mountantonakis et~al.(2022)Mountantonakis, Bastakis, Mertzanis and Tzitzikas}]{mountantonakis2022}
\bibinfo{author}{Mountantonakis, M.}, \bibinfo{author}{Bastakis, M.}, \bibinfo{author}{Mertzanis, L.}, and \bibinfo{author}{Tzitzikas, Y.} (\bibinfo{year}{2022}). \bibinfo{title}{Tiresias: Bilingual question answering over dbpedia}.
\newblock In \bibinfo{booktitle}{Workshop at ISWC}. CEUR Workshop Proceedings. \bibinfo{publisher}{CEUR-WS.org}.
\bibitem[{Papaioannou et~al.(2022{\natexlab{a}})Papaioannou, Grundmann, van Aken, Samaras, Kyparissidis, Giannakoulas, Gers and Loeser}]{papaioannou-etal-2022-cross}
\bibinfo{author}{Papaioannou, J.M.}, \bibinfo{author}{Grundmann, P.}, \bibinfo{author}{van Aken, B.}, \bibinfo{author}{Samaras, A.}, \bibinfo{author}{Kyparissidis, I.}, \bibinfo{author}{Giannakoulas, G.}, \bibinfo{author}{Gers, F.}, and \bibinfo{author}{Loeser, A.} (\bibinfo{year}{2022}{\natexlab{a}}). \bibinfo{title}{Cross-lingual knowledge transfer for clinical phenotyping}.
\newblock In \bibinfo{editor}{ N.{ }Calzolari}, \bibinfo{editor}{ F.{ }B{\'e}chet}, \bibinfo{editor}{ P.{ }Blache}, \bibinfo{editor}{ K.{ }Choukri}, \bibinfo{editor}{ C.{ }Cieri}, \bibinfo{editor}{ T.{ }Declerck}, \bibinfo{editor}{ S.{ }Goggi}, \bibinfo{editor}{ H.{ }Isahara}, \bibinfo{editor}{ B.{ }Maegaard}, \bibinfo{editor}{ J.{ }Mariani}, \bibinfo{editor}{ H.{ }Mazo}, \bibinfo{editor}{ J.{ }Odijk}, and \bibinfo{editor}{ S.{ }Piperidis}, eds. \bibinfo{booktitle}{Proceedings of the Thirteenth Language Resources and Evaluation Conference}. \bibinfo{address}{Marseille, France}: \bibinfo{publisher}{European Language Resources Association}{\natexlab{a}} pp. \bibinfo{pages}{900--909}.
\bibitem[{Evdaimon et~al.(2024)Evdaimon, Abdine, Xypolopoulos, Outsios, Vazirgiannis and Stamou}]{evdaimon2023greekbart}
\bibinfo{author}{Evdaimon, I.}, \bibinfo{author}{Abdine, H.}, \bibinfo{author}{Xypolopoulos, C.}, \bibinfo{author}{Outsios, S.}, \bibinfo{author}{Vazirgiannis, M.}, and \bibinfo{author}{Stamou, G.} (\bibinfo{year}{2024}). \bibinfo{title}{Greekbart: The first pretrained greek sequence-to-sequence model}.
\newblock In \bibinfo{booktitle}{Proceedings of the 2024 Joint International Conference on Computational Linguistics, Language Resources and Evaluation (LREC-COLING 2024)}. pp. \bibinfo{pages}{7949--7962}.
\bibitem[{Ranasinghe and Zampieri(2021)}]{Ranasinghe2021}
\bibinfo{author}{Ranasinghe, T.}, and \bibinfo{author}{Zampieri, M.} (\bibinfo{year}{2021}). \bibinfo{title}{Multilingual offensive language identification for low-resource languages}.
\newblock \bibinfo{journal}{ACM Trans. Asian Low-Resour. Lang. Inf. Process.} \emph{\bibinfo{volume}{22}}, \bibinfo{pages}{1--13}. \DOIprefix\doi{10.1145/3457610}.
\bibitem[{Koutsikakis et~al.(2020{\natexlab{a}})Koutsikakis, Chalkidis, Malakasiotis and Androutsopoulos}]{koutsikakis2020}
\bibinfo{author}{Koutsikakis, J.}, \bibinfo{author}{Chalkidis, I.}, \bibinfo{author}{Malakasiotis, P.}, and \bibinfo{author}{Androutsopoulos, I.} (\bibinfo{year}{2020}{\natexlab{a}}). \bibinfo{title}{Greek-bert: The greeks visiting sesame street}.
\newblock In \bibinfo{booktitle}{11th Hellenic Conference on Artificial Intelligence}. SETN 2020. \bibinfo{address}{New York, NY, USA}: \bibinfo{publisher}{Association for Computing Machinery}{\natexlab{a}}.
\newblock ISBN \bibinfo{isbn}{9781450388788} pp. \bibinfo{pages}{110--117}.
\newblock \DOIprefix\doi{10.1145/3411408.3411440}.
\bibitem[{Ahn et~al.(2020)Ahn, Sun, Park and Seo}]{ahn2020nlpdove}
\bibinfo{author}{Ahn, H.}, \bibinfo{author}{Sun, J.}, \bibinfo{author}{Park, C.Y.}, and \bibinfo{author}{Seo, J.} (\bibinfo{year}{2020}). \bibinfo{title}{Nlpdove at semeval-2020 task 12: Improving offensive language detection with cross-lingual transfer}.
\newblock \bibinfo{journal}{Preprint at arXiv \url{https://doi.org/10.48550/arXiv.2008.01354}}.
\bibitem[{Zampieri et~al.(2023)Zampieri, Rosenthal, Nakov, Dmonte and Ranasinghe}]{zampieri2023offenseval}
\bibinfo{author}{Zampieri, M.}, \bibinfo{author}{Rosenthal, S.}, \bibinfo{author}{Nakov, P.}, \bibinfo{author}{Dmonte, A.}, and \bibinfo{author}{Ranasinghe, T.} (\bibinfo{year}{2023}). \bibinfo{title}{Offenseval 2023: Offensive language identification in the age of large language models}.
\newblock \bibinfo{journal}{Natural Language Engineering} \emph{\bibinfo{volume}{29}}, \bibinfo{pages}{1416--1435}. \DOIprefix\doi{https://doi.org/10.1017/S1351324923000517}.
\bibitem[{Rohatgi et~al.(2023)Rohatgi, Qin, Aw, Unnithan and Kan}]{rohatgi2023acl}
\bibinfo{author}{Rohatgi, S.}, \bibinfo{author}{Qin, Y.}, \bibinfo{author}{Aw, B.}, \bibinfo{author}{Unnithan, N.}, and \bibinfo{author}{Kan, M.Y.} (\bibinfo{year}{2023}). \bibinfo{title}{The {ACL} {OCL} corpus: Advancing open science in computational linguistics}.
\newblock In \bibinfo{editor}{ H.{ }Bouamor}, \bibinfo{editor}{ J.{ }Pino}, and \bibinfo{editor}{ K.{ }Bali}, eds. \bibinfo{booktitle}{Proceedings of the 2023 Conference on Empirical Methods in Natural Language Processing}. \bibinfo{address}{Singapore}: \bibinfo{publisher}{Association for Computational Linguistics} pp. \bibinfo{pages}{10348--10361}.
\newblock \DOIprefix\doi{10.18653/v1/2023.emnlp-main.640}.
\bibitem[{Bender(????)}]{bender2019rule}
\bibinfo{author}{Bender, E.M.}
\newblock \bibinfo{title}{The \#benderrule: On naming the languages we study and why it matters}.
\newblock \bibinfo{note}{\url{https://thegradient.pub/the-benderrule-on-naming-the-languages-we-study-and-why-it-matters/}}.
\bibitem[{Phillips and Davis(2009)}]{rfc5646}
\bibinfo{author}{Phillips, A.}, and \bibinfo{author}{Davis, M.} (\bibinfo{year}{2009}).
\newblock \bibinfo{title}{Tags for identifying languages (rfc 5646)}. \bibinfo{publisher}{Internet Engineering Task Force (IETF)}.
\newblock \bibinfo{note}{\url{https://datatracker.ietf.org/doc/html/rfc5646.html}}.
\bibitem[{{SIL International}(????)}]{ethnologue}
\bibinfo{author}{{SIL International}}.
\newblock \bibinfo{title}{Ethnologue: Languages of the world}.
\newblock \bibinfo{note}{\url{https://www.ethnologue.com/}}.
\bibitem[{Gavriilidou et~al.(2023)Gavriilidou, Giagkou, Loizidou and Piperidis}]{gavriilidou2023language}
\bibinfo{author}{Gavriilidou, M.}, \bibinfo{author}{Giagkou, M.}, \bibinfo{author}{Loizidou, D.}, and \bibinfo{author}{Piperidis, S.} (\bibinfo{year}{2023}). \bibinfo{title}{Language report greek}.
\newblock In \bibinfo{booktitle}{European Language Equality: A Strategic Agenda for Digital Language Equality} pp. \bibinfo{pages}{151--154}.. \bibinfo{publisher}{Springer} pp. \bibinfo{pages}{151--154}.
\newblock \DOIprefix\doi{https://doi.org/10.1007/978-3-031-28819-7_19}.
\bibitem[{Davies(2015)}]{Davies2015}
\bibinfo{author}{Davies, A.M.} (\bibinfo{year}{2015}).
\newblock \bibinfo{title}{Greek language}. \bibinfo{publisher}{Oxford University Press}.
\newblock \DOIprefix\doi{10.1093/acrefore/9780199381135.013.2895}.
\bibitem[{Gavriilidou et~al.(2012)Gavriilidou, Koutsombogera, Patrikakos and Piperidis}]{gavrilidou2012metanet}
\bibinfo{author}{Gavriilidou, M.}, \bibinfo{author}{Koutsombogera, M.}, \bibinfo{author}{Patrikakos, A.}, and \bibinfo{author}{Piperidis, S.} (\bibinfo{year}{2012}). \bibinfo{title}{{T}he {G}reek {L}anguage in the {D}igital {A}ge}.
\newblock META-NET White Paper Series. Georg Rehm and Hans Uszkoreit (Series Editors). \bibinfo{publisher}{Springer}.
\newblock ISBN \bibinfo{isbn}{978-3-642-28935-4}.
\bibitem[{Tzanidaki(1995)}]{tzanidaki1995greek}
\bibinfo{author}{Tzanidaki, D.I.} (\bibinfo{year}{1995}). \bibinfo{title}{Greek word order: towards a new approach}.
\newblock \bibinfo{journal}{UCL Working Papers in Linguistics} \emph{\bibinfo{volume}{7}}, \bibinfo{pages}{247--277}.
\bibitem[{Manning(2015)}]{manning-2015-last}
\bibinfo{author}{Manning, C.D.} (\bibinfo{year}{2015}). \bibinfo{title}{Last words: Computational linguistics and deep learning}.
\newblock \bibinfo{journal}{Computational Linguistics} \emph{\bibinfo{volume}{41}}, \bibinfo{pages}{701--707}. \DOIprefix\doi{doi:10.1162/COLI_a_00239}.
\bibitem[{Qiu et~al.(2020)Qiu, Sun, Xu, Shao, Dai and Huang}]{qiu2020pre}
\bibinfo{author}{Qiu, X.}, \bibinfo{author}{Sun, T.}, \bibinfo{author}{Xu, Y.}, \bibinfo{author}{Shao, Y.}, \bibinfo{author}{Dai, N.}, and \bibinfo{author}{Huang, X.} (\bibinfo{year}{2020}). \bibinfo{title}{Pre-trained models for natural language processing: A survey}.
\newblock \bibinfo{journal}{Science China technological sciences} \emph{\bibinfo{volume}{63}}, \bibinfo{pages}{1872--1897}. \DOIprefix\doi{https://doi.org/10.1007/s11431-020-1647-3}.
\bibitem[{Bommasani et~al.(2021)Bommasani, Hudson, Adeli, Altman, Arora, von Arx, Bernstein, Bohg, Bosselut, Brunskill et~al.}]{bommasani2021opportunities}
\bibinfo{author}{Bommasani, R.}, \bibinfo{author}{Hudson, D.A.}, \bibinfo{author}{Adeli, E.}, \bibinfo{author}{Altman, R.}, \bibinfo{author}{Arora, S.}, \bibinfo{author}{von Arx, S.}, \bibinfo{author}{Bernstein, M.S.}, \bibinfo{author}{Bohg, J.}, \bibinfo{author}{Bosselut, A.}, \bibinfo{author}{Brunskill, E.} et~al. (\bibinfo{year}{2021}). \bibinfo{title}{On the opportunities and risks of foundation models}.
\newblock \bibinfo{journal}{Preprint at arXiv \url{https://doi.org/10.48550/arXiv.2108.07258}}.
\bibitem[{Minaee et~al.(2024)Minaee, Mikolov, Nikzad, Chenaghlu, Socher, Amatriain and Gao}]{minaee2024large}
\bibinfo{author}{Minaee, S.}, \bibinfo{author}{Mikolov, T.}, \bibinfo{author}{Nikzad, N.}, \bibinfo{author}{Chenaghlu, M.}, \bibinfo{author}{Socher, R.}, \bibinfo{author}{Amatriain, X.}, and \bibinfo{author}{Gao, J.} (\bibinfo{year}{2024}). \bibinfo{title}{Large language models: A survey}.
\newblock \bibinfo{journal}{Preprint at arXiv \url{https://doi.org/10.48550/arXiv.2402.06196}}.
\bibitem[{Pascanu et~al.(2013)Pascanu, Mikolov and Bengio}]{pascanu2013difficulty}
\bibinfo{author}{Pascanu, R.}, \bibinfo{author}{Mikolov, T.}, and \bibinfo{author}{Bengio, Y.} (\bibinfo{year}{2013}). \bibinfo{title}{On the difficulty of training recurrent neural networks}.
\newblock In \bibinfo{booktitle}{International conference on machine learning}. \bibinfo{organization}{Pmlr} pp. \bibinfo{pages}{1310--1318}.
\bibitem[{Rumelhart et~al.(1986)Rumelhart, Hinton and Williams}]{rumelhart1986learning}
\bibinfo{author}{Rumelhart, D.E.}, \bibinfo{author}{Hinton, G.E.}, and \bibinfo{author}{Williams, R.J.} (\bibinfo{year}{1986}). \bibinfo{title}{Learning representations by back-propagating errors}.
\newblock \bibinfo{journal}{nature} \emph{\bibinfo{volume}{323}}, \bibinfo{pages}{533--536}. \DOIprefix\doi{https://doi.org/10.1038/323533a0}.
\bibitem[{Elman(1990)}]{elman1990finding}
\bibinfo{author}{Elman, J.L.} (\bibinfo{year}{1990}). \bibinfo{title}{Finding structure in time}.
\newblock \bibinfo{journal}{Cognitive science} \emph{\bibinfo{volume}{14}}, \bibinfo{pages}{179--211}. \DOIprefix\doi{https://doi.org/10.1207/s15516709cog1402_1}.
\bibitem[{Werbos(1988)}]{werbos1988generalization}
\bibinfo{author}{Werbos, P.J.} (\bibinfo{year}{1988}). \bibinfo{title}{Generalization of backpropagation with application to a recurrent gas market model}.
\newblock \bibinfo{journal}{Neural networks} \emph{\bibinfo{volume}{1}}, \bibinfo{pages}{339--356}. \DOIprefix\doi{https://doi.org/10.1016/0893-6080(88)90007-X}.
\bibitem[{Devlin et~al.(2019)Devlin, Chang, Lee and Toutanova}]{devlin-etal-2019-bert}
\bibinfo{author}{Devlin, J.}, \bibinfo{author}{Chang, M.W.}, \bibinfo{author}{Lee, K.}, and \bibinfo{author}{Toutanova, K.} (\bibinfo{year}{2019}). \bibinfo{title}{{BERT}: Pre-training of deep bidirectional transformers for language understanding}.
\newblock In \bibinfo{editor}{ J.{ }Burstein}, \bibinfo{editor}{ C.{ }Doran}, and \bibinfo{editor}{ T.{ }Solorio}, eds. \bibinfo{booktitle}{Proceedings of the 2019 Conference of the North {A}merican Chapter of the Association for Computational Linguistics: Human Language Technologies, Volume 1 (Long and Short Papers)}. \bibinfo{address}{Minneapolis, Minnesota}: \bibinfo{publisher}{Association for Computational Linguistics} pp. \bibinfo{pages}{4171--4186}.
\newblock \DOIprefix\doi{10.18653/v1/N19-1423}.
\bibitem[{Liu et~al.(2019)Liu, Ott, Goyal, Du, Joshi, Chen, Levy, Lewis, Zettlemoyer and Stoyanov}]{liu2019roberta}
\bibinfo{author}{Liu, Y.}, \bibinfo{author}{Ott, M.}, \bibinfo{author}{Goyal, N.}, \bibinfo{author}{Du, J.}, \bibinfo{author}{Joshi, M.}, \bibinfo{author}{Chen, D.}, \bibinfo{author}{Levy, O.}, \bibinfo{author}{Lewis, M.}, \bibinfo{author}{Zettlemoyer, L.}, and \bibinfo{author}{Stoyanov, V.} (\bibinfo{year}{2019}). \bibinfo{title}{Roberta: A robustly optimized bert pretraining approach}.
\newblock \bibinfo{journal}{Preprint at arXiv \url{https://doi.org/10.48550/arXiv.1907.11692}}.
\bibitem[{Lan et~al.(2020)Lan, Chen, Goodman, Gimpel, Sharma and Soricut}]{lan2020albertlitebertselfsupervised}
\bibinfo{author}{Lan, Z.}, \bibinfo{author}{Chen, M.}, \bibinfo{author}{Goodman, S.}, \bibinfo{author}{Gimpel, K.}, \bibinfo{author}{Sharma, P.}, and \bibinfo{author}{Soricut, R.} (\bibinfo{year}{2020}). \bibinfo{title}{Albert: A lite bert for self-supervised learning of language representations}.
\newblock \bibinfo{journal}{Preprint at arXiv \url{https://doi.org/10.48550/arXiv.1909.11942}}.
\bibitem[{He et~al.(2021)He, Liu, Gao and Chen}]{he2021debertadecodingenhancedbertdisentangled}
\bibinfo{author}{He, P.}, \bibinfo{author}{Liu, X.}, \bibinfo{author}{Gao, J.}, and \bibinfo{author}{Chen, W.} (\bibinfo{year}{2021}). \bibinfo{title}{Deberta: Decoding-enhanced bert with disentangled attention}.
\newblock \bibinfo{journal}{Preprint at arXiv \url{https://doi.org/10.48550/arXiv.2006.03654}}.
\bibitem[{Lample and Conneau(2019)}]{lample2019crosslinguallanguagemodelpretraining}
\bibinfo{author}{Lample, G.}, and \bibinfo{author}{Conneau, A.} (\bibinfo{year}{2019}). \bibinfo{title}{Cross-lingual language model pretraining}.
\newblock \bibinfo{journal}{Preprint at arXiv \url{https://doi.org/10.48550/arXiv.1901.07291}}.
\bibitem[{Conneau et~al.(2020{\natexlab{a}})Conneau, Khandelwal, Goyal, Chaudhary, Wenzek, Guzm{\'a}n, Grave, Ott, Zettlemoyer and Stoyanov}]{conneau2019unsupervised}
\bibinfo{author}{Conneau, A.}, \bibinfo{author}{Khandelwal, K.}, \bibinfo{author}{Goyal, N.}, \bibinfo{author}{Chaudhary, V.}, \bibinfo{author}{Wenzek, G.}, \bibinfo{author}{Guzm{\'a}n, F.}, \bibinfo{author}{Grave, {\'E}.}, \bibinfo{author}{Ott, M.}, \bibinfo{author}{Zettlemoyer, L.}, and \bibinfo{author}{Stoyanov, V.} (\bibinfo{year}{2020}{\natexlab{a}}). \bibinfo{title}{Unsupervised cross-lingual representation learning at scale}.
\newblock In \bibinfo{booktitle}{Proceedings of the 58th Annual Meeting of the Association for Computational Linguistics}. pp. \bibinfo{pages}{8440--8451}.
\bibitem[{Yang et~al.(2020)Yang, Dai, Yang, Carbonell, Salakhutdinov and Le}]{yang2020xlnetgeneralizedautoregressivepretraining}
\bibinfo{author}{Yang, Z.}, \bibinfo{author}{Dai, Z.}, \bibinfo{author}{Yang, Y.}, \bibinfo{author}{Carbonell, J.}, \bibinfo{author}{Salakhutdinov, R.}, and \bibinfo{author}{Le, Q.V.} (\bibinfo{year}{2020}). \bibinfo{title}{Xlnet: Generalized autoregressive pretraining for language understanding}.
\newblock \bibinfo{journal}{Preprint at arXiv \url{https://doi.org/10.48550/arXiv.1906.08237}}.
\bibitem[{{Hugging Face}(????{\natexlab{a}})}]{bertology_hf}
\bibinfo{author}{{Hugging Face}}.
\newblock \bibinfo{title}{Bertology documentation}.
\newblock \bibinfo{note}{\url{https://huggingface.co/docs/transformers/bertology}}.
\bibitem[{Radford et~al.(2018)Radford, Narasimhan, Salimans, Sutskever et~al.}]{radford2018improving}
\bibinfo{author}{Radford, A.}, \bibinfo{author}{Narasimhan, K.}, \bibinfo{author}{Salimans, T.}, \bibinfo{author}{Sutskever, I.} et~al. (\bibinfo{year}{2018}).
\newblock \bibinfo{title}{Improving language understanding by generative pre-training}. \bibinfo{publisher}{OpenAI}.
\bibitem[{Radford et~al.(2019)Radford, Wu, Child, Luan, Amodei, Sutskever et~al.}]{radford2019language}
\bibinfo{author}{Radford, A.}, \bibinfo{author}{Wu, J.}, \bibinfo{author}{Child, R.}, \bibinfo{author}{Luan, D.}, \bibinfo{author}{Amodei, D.}, \bibinfo{author}{Sutskever, I.} et~al. (\bibinfo{year}{2019}).
\newblock \bibinfo{title}{Language models are unsupervised multitask learners}. \bibinfo{publisher}{OpenAI}.
\newblock \bibinfo{note}{\url{https://openai.com/blog/language-unsupervised}}.
\bibitem[{Raffel et~al.(2020)Raffel, Shazeer, Roberts, Lee, Narang, Matena, Zhou, Li and Liu}]{raffel2020exploring}
\bibinfo{author}{Raffel, C.}, \bibinfo{author}{Shazeer, N.}, \bibinfo{author}{Roberts, A.}, \bibinfo{author}{Lee, K.}, \bibinfo{author}{Narang, S.}, \bibinfo{author}{Matena, M.}, \bibinfo{author}{Zhou, Y.}, \bibinfo{author}{Li, W.}, and \bibinfo{author}{Liu, P.J.} (\bibinfo{year}{2020}). \bibinfo{title}{Exploring the limits of transfer learning with a unified text-to-text transformer}.
\newblock \bibinfo{journal}{The Journal of Machine Learning Research} \emph{\bibinfo{volume}{21}}, \bibinfo{pages}{5485--5551}.
\bibitem[{Xue et~al.(2021)Xue, Constant, Roberts, Kale, Al-Rfou, Siddhant, Barua and Raffel}]{xue2021mt5}
\bibinfo{author}{Xue, L.}, \bibinfo{author}{Constant, N.}, \bibinfo{author}{Roberts, A.}, \bibinfo{author}{Kale, M.}, \bibinfo{author}{Al-Rfou, R.}, \bibinfo{author}{Siddhant, A.}, \bibinfo{author}{Barua, A.}, and \bibinfo{author}{Raffel, C.} (\bibinfo{year}{2021}). \bibinfo{title}{mt5: A massively multilingual pre-trained text-to-text transformer}.
\newblock In \bibinfo{booktitle}{Proceedings of the 2021 Conference of the North American Chapter of the Association for Computational Linguistics: Human Language Technologies}. pp. \bibinfo{pages}{483--498}.
\bibitem[{Lewis et~al.(2019)Lewis, Liu, Goyal, Ghazvininejad, Mohamed, Levy, Stoyanov and Zettlemoyer}]{lewis2019bart}
\bibinfo{author}{Lewis, M.}, \bibinfo{author}{Liu, Y.}, \bibinfo{author}{Goyal, N.}, \bibinfo{author}{Ghazvininejad, M.}, \bibinfo{author}{Mohamed, A.}, \bibinfo{author}{Levy, O.}, \bibinfo{author}{Stoyanov, V.}, and \bibinfo{author}{Zettlemoyer, L.} (\bibinfo{year}{2019}). \bibinfo{title}{Bart: Denoising sequence-to-sequence pre-training for natural language generation, translation, and comprehension}.
\newblock \bibinfo{journal}{Preprint at arXiv \url{https://doi.org/10.48550/arXiv.1910.13461}}.
\bibitem[{Wei et~al.(2021)Wei, Bosma, Zhao, Guu, Yu, Lester, Du, Dai and Le}]{wei2021finetuned}
\bibinfo{author}{Wei, J.}, \bibinfo{author}{Bosma, M.}, \bibinfo{author}{Zhao, V.Y.}, \bibinfo{author}{Guu, K.}, \bibinfo{author}{Yu, A.W.}, \bibinfo{author}{Lester, B.}, \bibinfo{author}{Du, N.}, \bibinfo{author}{Dai, A.M.}, and \bibinfo{author}{Le, Q.V.} (\bibinfo{year}{2021}). \bibinfo{title}{Finetuned language models are zero-shot learners}.
\newblock \bibinfo{journal}{Preprint at arXiv \url{https://doi.org/10.48550/arXiv.2109.01652}}.
\bibitem[{Rae et~al.(2021)Rae, Borgeaud, Cai, Millican, Hoffmann, Song, Aslanides, Henderson, Ring, Young et~al.}]{rae2021scaling}
\bibinfo{author}{Rae, J.W.}, \bibinfo{author}{Borgeaud, S.}, \bibinfo{author}{Cai, T.}, \bibinfo{author}{Millican, K.}, \bibinfo{author}{Hoffmann, J.}, \bibinfo{author}{Song, F.}, \bibinfo{author}{Aslanides, J.}, \bibinfo{author}{Henderson, S.}, \bibinfo{author}{Ring, R.}, \bibinfo{author}{Young, S.} et~al. (\bibinfo{year}{2021}). \bibinfo{title}{Scaling language models: Methods, analysis \& insights from training gopher}.
\newblock \bibinfo{journal}{Preprint at arXiv \url{https://doi.org/10.48550/arXiv.2112.11446}}.
\bibitem[{Sanh et~al.(2021)Sanh, Webson, Raffel, Bach, Sutawika, Alyafeai, Chaffin, Stiegler, Scao, Raja et~al.}]{sanh2021multitask}
\bibinfo{author}{Sanh, V.}, \bibinfo{author}{Webson, A.}, \bibinfo{author}{Raffel, C.}, \bibinfo{author}{Bach, S.H.}, \bibinfo{author}{Sutawika, L.}, \bibinfo{author}{Alyafeai, Z.}, \bibinfo{author}{Chaffin, A.}, \bibinfo{author}{Stiegler, A.}, \bibinfo{author}{Scao, T.L.}, \bibinfo{author}{Raja, A.} et~al. (\bibinfo{year}{2021}). \bibinfo{title}{Multitask prompted training enables zero-shot task generalization}.
\newblock \bibinfo{journal}{Preprint at arXiv \url{https://doi.org/10.48550/arXiv.2110.08207}}.
\bibitem[{Du et~al.(2022)Du, Huang, Dai, Tong, Lepikhin, Xu, Krikun, Zhou, Yu, Firat et~al.}]{du2022glam}
\bibinfo{author}{Du, N.}, \bibinfo{author}{Huang, Y.}, \bibinfo{author}{Dai, A.M.}, \bibinfo{author}{Tong, S.}, \bibinfo{author}{Lepikhin, D.}, \bibinfo{author}{Xu, Y.}, \bibinfo{author}{Krikun, M.}, \bibinfo{author}{Zhou, Y.}, \bibinfo{author}{Yu, A.W.}, \bibinfo{author}{Firat, O.} et~al. (\bibinfo{year}{2022}). \bibinfo{title}{Glam: Efficient scaling of language models with mixture-of-experts}.
\newblock In \bibinfo{booktitle}{International Conference on Machine Learning}. \bibinfo{organization}{PMLR} pp. \bibinfo{pages}{5547--5569}.
\bibitem[{Giarelis et~al.(2023{\natexlab{a}})Giarelis, Mastrokostas, Siachos and Karacapilidis}]{giarelis2023review}
\bibinfo{author}{Giarelis, N.}, \bibinfo{author}{Mastrokostas, C.}, \bibinfo{author}{Siachos, I.}, and \bibinfo{author}{Karacapilidis, N.} (\bibinfo{year}{2023}{\natexlab{a}}). \bibinfo{title}{A review of greek nlp technologies for chatbot development}.
\newblock In \bibinfo{booktitle}{Proceedings of the 27th Pan-Hellenic Conference on Progress in Computing and Informatics}. pp. \bibinfo{pages}{15--20}.
\bibitem[{Nikiforos et~al.(2021)Nikiforos, Voutos, Drougani, Mylonas and Kermanidis}]{Nikiforos2021}
\bibinfo{author}{Nikiforos, M.N.}, \bibinfo{author}{Voutos, Y.}, \bibinfo{author}{Drougani, A.}, \bibinfo{author}{Mylonas, P.}, and \bibinfo{author}{Kermanidis, K.L.} (\bibinfo{year}{2021}). \bibinfo{title}{The modern greek language on the social web: A survey of data sets and mining applications}.
\newblock \bibinfo{journal}{Data} \emph{\bibinfo{volume}{6}}, \bibinfo{pages}{52}. \DOIprefix\doi{https://doi.org/10.3390/data6050052}.
\bibitem[{Alexandridis et~al.(2021{\natexlab{a}})Alexandridis, Varlamis, Korovesis, Caridakis and Tsantilas}]{Alexandridis2021Survey}
\bibinfo{author}{Alexandridis, G.}, \bibinfo{author}{Varlamis, I.}, \bibinfo{author}{Korovesis, K.}, \bibinfo{author}{Caridakis, G.}, and \bibinfo{author}{Tsantilas, P.} (\bibinfo{year}{2021}{\natexlab{a}}). \bibinfo{title}{A survey on sentiment analysis and opinion mining in greek social media}.
\newblock \bibinfo{journal}{Information} \emph{\bibinfo{volume}{12}}. \DOIprefix\doi{10.3390/info12080331}.
\bibitem[{Krasadakis et~al.(2022)Krasadakis, Sakkopoulos and Verykios}]{Krasadakis2021}
\bibinfo{author}{Krasadakis, P.}, \bibinfo{author}{Sakkopoulos, E.}, and \bibinfo{author}{Verykios, V.S.} (\bibinfo{year}{2022}). \bibinfo{title}{A natural language processing survey on legislative and greek documents}.
\newblock In \bibinfo{booktitle}{25th Pan-Hellenic Conference on Informatics}. PCI 2021. \bibinfo{address}{New York, NY, USA}: \bibinfo{publisher}{Association for Computing Machinery}.
\newblock ISBN \bibinfo{isbn}{9781450395557} pp. \bibinfo{pages}{407--412}.
\newblock \DOIprefix\doi{10.1145/3503823.3503898}.
\bibitem[{{Association for Computational Linguistics}(????{\natexlab{a}})}]{acl_anthology}
\bibinfo{author}{{Association for Computational Linguistics}}.
\newblock \bibinfo{title}{Acl anthology}.
\newblock \bibinfo{note}{\url{https://aclanthology.org/}}.
\bibitem[{{Allen Institute for AI}(????)}]{semantic_scholar}
\bibinfo{author}{{Allen Institute for AI}}.
\newblock \bibinfo{title}{Semantic scholar}.
\newblock \bibinfo{note}{\url{https://www.semanticscholar.org/}}.
\bibitem[{{Elsevier}(????)}]{scopus}
\bibinfo{author}{{Elsevier}}.
\newblock \bibinfo{title}{Scopus}.
\newblock \bibinfo{note}{\url{https://www.scopus.com/home.uri}}.
\bibitem[{{ACL}(????)}]{acl_anthology_repo}
\bibinfo{author}{{ACL}}.
\newblock \bibinfo{title}{Data and software for building the acl anthology}. \bibinfo{publisher}{GitHub}.
\newblock \bibinfo{note}{\url{https://github.com/acl-org/acl-anthology}}.
\bibitem[{{Google}(????)}]{google_scholar}
\bibinfo{author}{{Google}}.
\newblock \bibinfo{title}{Google scholar}.
\newblock \bibinfo{note}{\url{https://scholar.google.com/}}.
\bibitem[{Bommasani et~al.(2023)Bommasani, Liang and Lee}]{bommasani2023holistic}
\bibinfo{author}{Bommasani, R.}, \bibinfo{author}{Liang, P.}, and \bibinfo{author}{Lee, T.} (\bibinfo{year}{2023}). \bibinfo{title}{Holistic evaluation of language models}.
\newblock \bibinfo{journal}{Annals of the New York Academy of Sciences} \emph{\bibinfo{volume}{1525}}, \bibinfo{pages}{140--146}.
\bibitem[{{Association for Computational Linguistics}(????{\natexlab{b}})}]{acl2023_call}
\bibinfo{author}{{Association for Computational Linguistics}}.
\newblock \bibinfo{title}{Acl 2023 main conference call for papers}.
\newblock \bibinfo{note}{\url{https://2023.aclweb.org/calls/main_conference/}}.
\bibitem[{Wilkinson et~al.(2016)Wilkinson, Dumontier, Aalbersberg, Appleton, Axton et~al.}]{wilkinson2016fair}
\bibinfo{author}{Wilkinson, M.D.}, \bibinfo{author}{Dumontier, M.}, \bibinfo{author}{Aalbersberg, I.J.}, \bibinfo{author}{Appleton, G.}, \bibinfo{author}{Axton, M.} et~al. (\bibinfo{year}{2016}). \bibinfo{title}{The {FAIR} {G}uiding {P}rinciples for scientific data management and stewardship}.
\newblock \bibinfo{journal}{Scientific data} \emph{\bibinfo{volume}{3}}, \bibinfo{pages}{1--9}. \DOIprefix\doi{10.1038/sdata.2016.18}.
\bibitem[{Belinkov et~al.(2020)Belinkov, Gehrmann and Pavlick}]{belinkov2020interpretability}
\bibinfo{author}{Belinkov, Y.}, \bibinfo{author}{Gehrmann, S.}, and \bibinfo{author}{Pavlick, E.} (\bibinfo{year}{2020}). \bibinfo{title}{Interpretability and analysis in neural nlp}.
\newblock In \bibinfo{booktitle}{Proceedings of the 58th annual meeting of the association for computational linguistics: tutorial abstracts}. pp. \bibinfo{pages}{1--5}.
\bibitem[{Prokopidis and Piperidis(2020{\natexlab{a}})}]{prokopidis2020neural}
\bibinfo{author}{Prokopidis, P.}, and \bibinfo{author}{Piperidis, S.} (\bibinfo{year}{2020}{\natexlab{a}}). \bibinfo{title}{A neural nlp toolkit for greek}.
\newblock In \bibinfo{booktitle}{11th Hellenic Conference on Artificial Intelligence}. pp. \bibinfo{pages}{125--128}.
\newblock \DOIprefix\doi{https://doi.org/10.1145/3411408.3411430}.
\bibitem[{Bojanowski et~al.(2017)Bojanowski, Grave, Joulin and Mikolov}]{bojanowski2017enriching}
\bibinfo{author}{Bojanowski, P.}, \bibinfo{author}{Grave, E.}, \bibinfo{author}{Joulin, A.}, and \bibinfo{author}{Mikolov, T.} (\bibinfo{year}{2017}). \bibinfo{title}{Enriching word vectors with subword information}.
\newblock \bibinfo{journal}{Transactions of the Association for Computational Linguistics} \emph{\bibinfo{volume}{5}}, \bibinfo{pages}{135--146}. \DOIprefix\doi{https://doi.org/10.1162/tacl_a_00051}.
\bibitem[{Tsakalidis et~al.(2018{\natexlab{a}})Tsakalidis, Papadopoulos, Voskaki, Ioannidou, Boididou, Cristea, Liakata and Kompatsiaris}]{tsakalidis2018building}
\bibinfo{author}{Tsakalidis, A.}, \bibinfo{author}{Papadopoulos, S.}, \bibinfo{author}{Voskaki, R.}, \bibinfo{author}{Ioannidou, K.}, \bibinfo{author}{Boididou, C.}, \bibinfo{author}{Cristea, A.I.}, \bibinfo{author}{Liakata, M.}, and \bibinfo{author}{Kompatsiaris, Y.} (\bibinfo{year}{2018}{\natexlab{a}}). \bibinfo{title}{Building and evaluating resources for sentiment analysis in the greek language}.
\newblock \bibinfo{journal}{Language resources and evaluation} \emph{\bibinfo{volume}{52}}, \bibinfo{pages}{1021--1044}. \DOIprefix\doi{https://doi.org/10.1007/s10579-018-9420-4}.
\bibitem[{Mikolov et~al.(2013{\natexlab{a}})Mikolov, Sutskever, Chen, Corrado and Dean}]{mikolov2013distributed}
\bibinfo{author}{Mikolov, T.}, \bibinfo{author}{Sutskever, I.}, \bibinfo{author}{Chen, K.}, \bibinfo{author}{Corrado, G.S.}, and \bibinfo{author}{Dean, J.} (\bibinfo{year}{2013}{\natexlab{a}}). \bibinfo{title}{Distributed representations of words and phrases and their compositionality}.
\newblock In \bibinfo{editor}{ C.{ }Burges}, \bibinfo{editor}{ L.{ }Bottou}, \bibinfo{editor}{ M.{ }Welling}, \bibinfo{editor}{ Z.{ }Ghahramani}, and \bibinfo{editor}{ K.{ }Weinberger}, eds. \bibinfo{booktitle}{Advances in Neural Information Processing Systems} vol.~\bibinfo{volume}{26}. \bibinfo{publisher}{Curran Associates, Inc.}{\natexlab{a}} pp. \bibinfo{pages}{3111--3119}.
\newblock \URLprefix \url{https://proceedings.neurips.cc/paper_files/paper/2013/file/9aa42b31882ec039965f3c4923ce901b-Paper.pdf}.
\bibitem[{Pavlopoulos et~al.(2017{\natexlab{a}})Pavlopoulos, Malakasiotis and Androutsopoulos}]{Pavlopoulos2017deep}
\bibinfo{author}{Pavlopoulos, J.}, \bibinfo{author}{Malakasiotis, P.}, and \bibinfo{author}{Androutsopoulos, I.} (\bibinfo{year}{2017}{\natexlab{a}}). \bibinfo{title}{Deep learning for user comment moderation}.
\newblock \bibinfo{journal}{Preprint at arXiv \url{https://doi.org/10.48550/arXiv.1705.09993}}.
\bibitem[{Pavlopoulos et~al.(2017{\natexlab{b}})Pavlopoulos, Malakasiotis, Bakagianni and Androutsopoulos}]{Pavlopoulos2017improved}
\bibinfo{author}{Pavlopoulos, J.}, \bibinfo{author}{Malakasiotis, P.}, \bibinfo{author}{Bakagianni, J.}, and \bibinfo{author}{Androutsopoulos, I.} (\bibinfo{year}{2017}{\natexlab{b}}). \bibinfo{title}{Improved abusive comment moderation with user embeddings}.
\newblock \bibinfo{journal}{Preprint at arXiv \url{https://doi.org/10.48550/arXiv.1708.03699}}.
\bibitem[{Medrouk and Pappa(2017)}]{Medrouk2017}
\bibinfo{author}{Medrouk, L.}, and \bibinfo{author}{Pappa, A.} (\bibinfo{year}{2017}). \bibinfo{title}{Deep learning model for sentiment analysis in multi-lingual corpus}.
\newblock In \bibinfo{editor}{ D.{ }Liu}, \bibinfo{editor}{ S.{ }Xie}, \bibinfo{editor}{ Y.{ }Li}, \bibinfo{editor}{ D.{ }Zhao}, and \bibinfo{editor}{ E.S.M.{ }El-Alfy}, eds. \bibinfo{booktitle}{Neural Information Processing}. \bibinfo{address}{Cham}: \bibinfo{publisher}{Springer International Publishing}.
\newblock ISBN \bibinfo{isbn}{978-3-319-70087-8} pp. \bibinfo{pages}{205--212}.
\newblock \DOIprefix\doi{https://doi.org/10.1007/978-3-319-70087-8_22}.
\bibitem[{Pitenis et~al.(2020)Pitenis, Zampieri and Ranasinghe}]{Pitenis2020}
\bibinfo{author}{Pitenis, Z.}, \bibinfo{author}{Zampieri, M.}, and \bibinfo{author}{Ranasinghe, T.} (\bibinfo{year}{2020}). \bibinfo{title}{Offensive language identification in greek}.
\newblock \bibinfo{journal}{Preprint at arXiv \url{https://doi.org/10.48550/arXiv.2003.07459}}.
\bibitem[{Arcila-Calderón et~al.(2022)Arcila-Calderón, Amores, Sánchez-Holgado, Vrysis, Vryzas and Oller~Alonso}]{Calderón2022}
\bibinfo{author}{Arcila-Calderón, C.}, \bibinfo{author}{Amores, J.J.}, \bibinfo{author}{Sánchez-Holgado, P.}, \bibinfo{author}{Vrysis, L.}, \bibinfo{author}{Vryzas, N.}, and \bibinfo{author}{Oller~Alonso, M.} (\bibinfo{year}{2022}). \bibinfo{title}{How to detect online hate towards migrants and refugees? developing and evaluating a classifier of racist and xenophobic hate speech using shallow and deep learning}.
\newblock \bibinfo{journal}{Sustainability} \emph{\bibinfo{volume}{14}}. \DOIprefix\doi{10.3390/su142013094}.
\bibitem[{Giarelis et~al.(2024{\natexlab{a}})Giarelis, Mastrokostas and Karacapilidis}]{giarelis2024greekt5}
\bibinfo{author}{Giarelis, N.}, \bibinfo{author}{Mastrokostas, C.}, and \bibinfo{author}{Karacapilidis, N.} (\bibinfo{year}{2024}{\natexlab{a}}). \bibinfo{title}{Greekt5: Sequence-to-sequence models for greek news summarization}.
\newblock In \bibinfo{booktitle}{IFIP International Conference on Artificial Intelligence Applications and Innovations}. \bibinfo{organization}{Springer} pp. \bibinfo{pages}{60--73}.
\newblock \DOIprefix\doi{https://doi.org/10.1007/978-3-031-63215-0_5}.
\bibitem[{Giarelis et~al.(2024{\natexlab{b}})Giarelis, Mastrokostas and Karacapilidis}]{imis_greekt5_mt5_small}
\bibinfo{author}{Giarelis, N.}, \bibinfo{author}{Mastrokostas, C.}, and \bibinfo{author}{Karacapilidis, N.} (\bibinfo{year}{2024}{\natexlab{b}}).
\newblock \bibinfo{title}{Greekt5-mt5-small-greeksum}. \bibinfo{publisher}{Hugging Face}{\natexlab{b}}.
\newblock \bibinfo{note}{\url{https://huggingface.co/IMISLab/GreekT5-mt5-small-greeksum}}.
\bibitem[{Giarelis et~al.(2024{\natexlab{c}})Giarelis, Mastrokostas and Karacapilidis}]{imis_greekt5_umt5_small}
\bibinfo{author}{Giarelis, N.}, \bibinfo{author}{Mastrokostas, C.}, and \bibinfo{author}{Karacapilidis, N.} (\bibinfo{year}{2024}{\natexlab{c}}).
\newblock \bibinfo{title}{Greekt5-umt5-small-greeksum}. \bibinfo{publisher}{Hugging Face}{\natexlab{c}}.
\newblock \bibinfo{note}{\url{https://huggingface.co/IMISLab/GreekT5-umt5-small-greeksum}}.
\bibitem[{Giarelis et~al.(2024{\natexlab{d}})Giarelis, Mastrokostas and Karacapilidis}]{imis_greekt5_umt5_base}
\bibinfo{author}{Giarelis, N.}, \bibinfo{author}{Mastrokostas, C.}, and \bibinfo{author}{Karacapilidis, N.} (\bibinfo{year}{2024}{\natexlab{d}}).
\newblock \bibinfo{title}{Greekt5-umt5-base-greeksum}. \bibinfo{publisher}{Hugging Face}{\natexlab{d}}.
\newblock \bibinfo{note}{\url{https://huggingface.co/IMISLab/GreekT5-umt5-base-greeksum}}.
\bibitem[{Evdaimon et~al.(2023)Evdaimon, Abdine, Xypolopoulos, Outsios, Vazirgiannis and Stamou}]{GreekBART}
\bibinfo{author}{Evdaimon, I.}, \bibinfo{author}{Abdine, H.}, \bibinfo{author}{Xypolopoulos, C.}, \bibinfo{author}{Outsios, S.}, \bibinfo{author}{Vazirgiannis, M.}, and \bibinfo{author}{Stamou, G.} (\bibinfo{year}{2023}).
\newblock \bibinfo{title}{Greekbart: The first pretrained greek sequence-to-sequence model}. \bibinfo{publisher}{GitHub}.
\newblock \bibinfo{note}{\url{https://github.com/iakovosevdaimon/GreekBART}}.
\bibitem[{Koutsikakis et~al.(2020{\natexlab{b}})Koutsikakis, Chalkidis, Malakasiotis and Androutsopoulos}]{GreekBERT}
\bibinfo{author}{Koutsikakis, J.}, \bibinfo{author}{Chalkidis, I.}, \bibinfo{author}{Malakasiotis, P.}, and \bibinfo{author}{Androutsopoulos, I.} (\bibinfo{year}{2020}{\natexlab{b}}).
\newblock \bibinfo{title}{Greekbert: bert-base-greek-uncased-v1}. \bibinfo{publisher}{Hugging Face}{\natexlab{b}}.
\newblock \bibinfo{note}{\url{https://huggingface.co/nlpaueb/bert-base-greek-uncased-v1}}.
\bibitem[{Zaikis et~al.(2023)Zaikis, Stylianou and Vlahavas}]{zaikis2023pima}
\bibinfo{author}{Zaikis, D.}, \bibinfo{author}{Stylianou, N.}, and \bibinfo{author}{Vlahavas, I.} (\bibinfo{year}{2023}). \bibinfo{title}{Pima: Parameter-shared intelligent media analytics framework for low resource languages}.
\newblock \bibinfo{journal}{Applied Sciences} \emph{\bibinfo{volume}{13}}, \bibinfo{pages}{3265}. \DOIprefix\doi{https://doi.org/10.3390/app13053265}.
\bibitem[{Zaikis et~al.(2021)Zaikis, Stylianou and Vlahavas}]{GreekMediaBERT}
\bibinfo{author}{Zaikis, D.}, \bibinfo{author}{Stylianou, N.}, and \bibinfo{author}{Vlahavas, I.} (\bibinfo{year}{2021}).
\newblock \bibinfo{title}{Greek media bert base uncased}. \bibinfo{publisher}{Hugging Face}.
\newblock \bibinfo{note}{\url{https://huggingface.co/dimitriz/greek-media-bert-base-uncased}}.
\bibitem[{Alexandridis et~al.(2021{\natexlab{b}})Alexandridis, Varlamis, Korovesis, Caridakis and Tsantilas}]{GreekSocialBERT2023}
\bibinfo{author}{Alexandridis, G.}, \bibinfo{author}{Varlamis, I.}, \bibinfo{author}{Korovesis, K.}, \bibinfo{author}{Caridakis, G.}, and \bibinfo{author}{Tsantilas, P.} (\bibinfo{year}{2021}{\natexlab{b}}).
\newblock \bibinfo{title}{Greeksocialbert base greek uncased v1}. \bibinfo{publisher}{Hugging Face}{\natexlab{b}}.
\newblock \bibinfo{note}{\url{https://huggingface.co/gealexandri/greeksocialbert-base-greek-uncased-v1}}.
\bibitem[{Alexandridis et~al.(2021{\natexlab{c}})Alexandridis, Varlamis, Korovesis, Caridakis and Tsantilas}]{PaloBERT2023}
\bibinfo{author}{Alexandridis, G.}, \bibinfo{author}{Varlamis, I.}, \bibinfo{author}{Korovesis, K.}, \bibinfo{author}{Caridakis, G.}, and \bibinfo{author}{Tsantilas, P.} (\bibinfo{year}{2021}{\natexlab{c}}).
\newblock \bibinfo{title}{Palobert base greek uncased v1}. \bibinfo{publisher}{Hugging Face}{\natexlab{c}}.
\newblock \bibinfo{note}{\url{https://huggingface.co/gealexandri/palobert-base-greek-uncased-v1}}.
\bibitem[{Perifanos and Goutsos(2021{\natexlab{a}})}]{Perifanos2021}
\bibinfo{author}{Perifanos, K.}, and \bibinfo{author}{Goutsos, D.} (\bibinfo{year}{2021}{\natexlab{a}}). \bibinfo{title}{Multimodal hate speech detection in greek social media}.
\newblock \bibinfo{journal}{Multimodal Technologies and Interaction} \emph{\bibinfo{volume}{5}}, \bibinfo{pages}{34}. \DOIprefix\doi{10.3390/mti5070034}.
\bibitem[{Perifanos and Goutsos(2021{\natexlab{b}})}]{BERTaTweetGR2023}
\bibinfo{author}{Perifanos, K.}, and \bibinfo{author}{Goutsos, D.} (\bibinfo{year}{2021}{\natexlab{b}}).
\newblock \bibinfo{title}{Bertatweetgr}. \bibinfo{publisher}{Hugging Face}{\natexlab{b}}.
\newblock \bibinfo{note}{\url{https://huggingface.co/Konstantinos/BERTaTweetGR}}.
\bibitem[{Alexandridis et~al.(2022)Alexandridis, Korovesis, Varlamis, Tsantilas and Caridakis}]{Alexandridis2021}
\bibinfo{author}{Alexandridis, G.}, \bibinfo{author}{Korovesis, K.}, \bibinfo{author}{Varlamis, I.}, \bibinfo{author}{Tsantilas, P.}, and \bibinfo{author}{Caridakis, G.} (\bibinfo{year}{2022}). \bibinfo{title}{Emotion detection on greek social media using bidirectional encoder representations from transformers}.
\newblock In \bibinfo{booktitle}{25th Pan-Hellenic Conference on Informatics}. PCI 2021. \bibinfo{address}{New York, NY, USA}: \bibinfo{publisher}{Association for Computing Machinery}.
\newblock ISBN \bibinfo{isbn}{9781450395557} pp. \bibinfo{pages}{28--32}.
\newblock \DOIprefix\doi{10.1145/3503823.3503829}.
\bibitem[{Rizou et~al.(2022{\natexlab{a}})Rizou, Paflioti, Theofilatos, Vakali, Sarigiannidis and Chatzisavvas}]{Rizou2022}
\bibinfo{author}{Rizou, S.}, \bibinfo{author}{Paflioti, A.}, \bibinfo{author}{Theofilatos, A.}, \bibinfo{author}{Vakali, A.}, \bibinfo{author}{Sarigiannidis, G.}, and \bibinfo{author}{Chatzisavvas, K.} (\bibinfo{year}{2022}{\natexlab{a}}). \bibinfo{title}{Multilingual name entity recognition and intent classification employing deep learning architectures}.
\newblock \bibinfo{journal}{Simulation Modelling Practice and Theory} \emph{\bibinfo{volume}{120}}, \bibinfo{pages}{102620}. \DOIprefix\doi{https://doi.org/10.1016/j.simpat.2022.102620}.
\bibitem[{Bilianos(2022)}]{Bilianos2022}
\bibinfo{author}{Bilianos, D.} (\bibinfo{year}{2022}). \bibinfo{title}{Experiments in text classification: Analyzing the sentiment of electronic product reviews in greek}.
\newblock \bibinfo{journal}{Journal of Quantitative Linguistics} \emph{\bibinfo{volume}{29}}, \bibinfo{pages}{374--386}. \DOIprefix\doi{10.1080/09296174.2021.1885872}.
\bibitem[{Kapoteli et~al.(2022)Kapoteli, Koukaras and Tjortjis}]{Kapoteli2022}
\bibinfo{author}{Kapoteli, E.}, \bibinfo{author}{Koukaras, P.}, and \bibinfo{author}{Tjortjis, C.} (\bibinfo{year}{2022}). \bibinfo{title}{Social media sentiment analysis related to covid-19 vaccines: Case studies in english and greek language}.
\newblock In \bibinfo{editor}{ I.{ }Maglogiannis}, \bibinfo{editor}{ L.{ }Iliadis}, \bibinfo{editor}{ J.{ }Macintyre}, and \bibinfo{editor}{ P.{ }Cortez}, eds. \bibinfo{booktitle}{Artificial Intelligence Applications and Innovations}. \bibinfo{address}{Cham}: \bibinfo{publisher}{Springer International Publishing}.
\newblock ISBN \bibinfo{isbn}{978-3-031-08337-2} pp. \bibinfo{pages}{360--372}.
\newblock \DOIprefix\doi{https://doi.org/10.1007/978-3-031-08337-2_30}.
\bibitem[{{Wikimedia Foundation}(????{\natexlab{a}})}]{greek_wikipedia_dumps}
\bibinfo{author}{{Wikimedia Foundation}}.
\newblock \bibinfo{title}{Greek wikipedia dumps}.
\newblock \bibinfo{note}{\url{https://dumps.wikimedia.org/elwiki/}}.
\bibitem[{Koehn(2005)}]{koehn2005europarl}
\bibinfo{author}{Koehn, P.} (\bibinfo{year}{2005}). \bibinfo{title}{Europarl: A parallel corpus for statistical machine translation}.
\newblock In \bibinfo{booktitle}{Proceedings of machine translation summit x: papers}. pp. \bibinfo{pages}{79--86}.
\bibitem[{Su{\'a}rez et~al.(2019)Su{\'a}rez, Sagot and Romary}]{suarez2019asynchronous}
\bibinfo{author}{Su{\'a}rez, P.J.O.}, \bibinfo{author}{Sagot, B.}, and \bibinfo{author}{Romary, L.} (\bibinfo{year}{2019}). \bibinfo{title}{Asynchronous pipeline for processing huge corpora on medium to low resource infrastructures}.
\newblock In \bibinfo{booktitle}{7th Workshop on the Challenges in the Management of Large Corpora (CMLC-7)}. \bibinfo{organization}{Leibniz-Institut f{\"u}r Deutsche Sprache} pp. \bibinfo{pages}{9--16}.
\newblock \DOIprefix\doi{10.14618/IDS-PUB-9021}.
\bibitem[{{Common Crawl Foundation}(2025)}]{common_crawl}
\bibinfo{author}{{Common Crawl Foundation}} (\bibinfo{year}{2025}).
\newblock \bibinfo{title}{Common crawl}.
\newblock \bibinfo{note}{\url{https://commoncrawl.org/}}.
\bibitem[{Lewis et~al.(2020{\natexlab{a}})Lewis, Liu, Goyal, Ghazvininejad, Mohamed, Levy, Stoyanov and Zettlemoyer}]{lewis2020bart}
\bibinfo{author}{Lewis, M.}, \bibinfo{author}{Liu, Y.}, \bibinfo{author}{Goyal, N.}, \bibinfo{author}{Ghazvininejad, M.}, \bibinfo{author}{Mohamed, A.}, \bibinfo{author}{Levy, O.}, \bibinfo{author}{Stoyanov, V.}, and \bibinfo{author}{Zettlemoyer, L.} (\bibinfo{year}{2020}{\natexlab{a}}). \bibinfo{title}{Bart: Denoising sequence-to-sequence pre-training for natural language generation, translation, and comprehension}.
\newblock In \bibinfo{booktitle}{Proceedings of the 58th Annual Meeting of the Association for Computational Linguistics}. pp. \bibinfo{pages}{7871--7880}.
\bibitem[{Outsios et~al.(2018)Outsios, Skianis, Meladianos, Xypolopoulos and Vazirgiannis}]{outsios2018word}
\bibinfo{author}{Outsios, S.}, \bibinfo{author}{Skianis, K.}, \bibinfo{author}{Meladianos, P.}, \bibinfo{author}{Xypolopoulos, C.}, and \bibinfo{author}{Vazirgiannis, M.} (\bibinfo{year}{2018}). \bibinfo{title}{Word embeddings from large-scale greek web content}.
\newblock \bibinfo{journal}{Preprint at arXiv \url{https://doi.org/10.48550/arXiv.1810.06694}}.
\bibitem[{Chung et~al.(2024{\natexlab{a}})Chung, Garcia, Roberts, Tay, Firat, Narang and Constant}]{chungunimax}
\bibinfo{author}{Chung, H.W.}, \bibinfo{author}{Garcia, X.}, \bibinfo{author}{Roberts, A.}, \bibinfo{author}{Tay, Y.}, \bibinfo{author}{Firat, O.}, \bibinfo{author}{Narang, S.}, and \bibinfo{author}{Constant, N.} (\bibinfo{year}{2024}{\natexlab{a}}). \bibinfo{title}{Unimax: Fairer and more effective language sampling for large-scale multilingual pretraining}.
\newblock In \bibinfo{booktitle}{The Eleventh International Conference on Learning Representations}.
\bibitem[{Papadimitriou et~al.(2023)Papadimitriou, Lopez and Jurafsky}]{papadimitriou2023multilingual}
\bibinfo{author}{Papadimitriou, I.}, \bibinfo{author}{Lopez, K.}, and \bibinfo{author}{Jurafsky, D.} (\bibinfo{year}{2023}). \bibinfo{title}{Multilingual bert has an accent: Evaluating english influences on fluency in multilingual models}.
\newblock In \bibinfo{booktitle}{Findings of the Association for Computational Linguistics: EACL 2023}. pp. \bibinfo{pages}{1194--1200}.
\bibitem[{Ahn and Oh(2021)}]{ahn2021mitigating}
\bibinfo{author}{Ahn, J.}, and \bibinfo{author}{Oh, A.} (\bibinfo{year}{2021}). \bibinfo{title}{Mitigating language-dependent ethnic bias in bert}.
\newblock \bibinfo{journal}{Preprint at arXiv \url{https://doi.org/10.48550/arXiv.2109.05704}}.
\bibitem[{Gar{\'\i}~Soler and Apidianaki(2021)}]{gari2021let}
\bibinfo{author}{Gar{\'\i}~Soler, A.}, and \bibinfo{author}{Apidianaki, M.} (\bibinfo{year}{2021}). \bibinfo{title}{Let’s play mono-poly: Bert can reveal words’ polysemy level and partitionability into senses}.
\newblock \bibinfo{journal}{Transactions of the Association for Computational Linguistics} \emph{\bibinfo{volume}{9}}, \bibinfo{pages}{825--844}. \DOIprefix\doi{https://doi.org/10.1162/tacl_a_00400}.
\bibitem[{Garí~Soler and Apidianaki(2021)}]{monopoly_repo}
\bibinfo{author}{Garí~Soler, A.}, and \bibinfo{author}{Apidianaki, M.} (\bibinfo{year}{2021}).
\newblock \bibinfo{title}{Let’s play mono-poly}. \bibinfo{publisher}{GitHub}.
\newblock \bibinfo{note}{\url{https://github.com/ainagari/monopoly}}.
\bibitem[{Gonen et~al.(2020{\natexlab{a}})Gonen, Ravfogel, Elazar and Goldberg}]{Gonen2020}
\bibinfo{author}{Gonen, H.}, \bibinfo{author}{Ravfogel, S.}, \bibinfo{author}{Elazar, Y.}, and \bibinfo{author}{Goldberg, Y.} (\bibinfo{year}{2020}{\natexlab{a}}). \bibinfo{title}{It's not greek to mbert: Inducing word-level translations from multilingual {BERT}}.
\newblock \bibinfo{journal}{Preprint at arXiv \url{https://doi.org/10.48550/arXiv.2010.08275}}.
\bibitem[{Gonen et~al.(2020{\natexlab{b}})Gonen, Ravfogel, Elazar and Goldberg}]{mbert_repo}
\bibinfo{author}{Gonen, H.}, \bibinfo{author}{Ravfogel, S.}, \bibinfo{author}{Elazar, Y.}, and \bibinfo{author}{Goldberg, Y.} (\bibinfo{year}{2020}{\natexlab{b}}).
\newblock \bibinfo{title}{It's not greek to mbert: Inducing word-level translations from multilingual bert}. \bibinfo{publisher}{GitHub}{\natexlab{b}}.
\newblock \bibinfo{note}{\url{https://github.com/gonenhila/mbert}}.
\bibitem[{Balloccu et~al.(2024)Balloccu, Schmidtov{\'a}, Lango and Dusek}]{balloccu-etal-2024-leak}
\bibinfo{author}{Balloccu, S.}, \bibinfo{author}{Schmidtov{\'a}, P.}, \bibinfo{author}{Lango, M.}, and \bibinfo{author}{Dusek, O.} (\bibinfo{year}{2024}). \bibinfo{title}{Leak, {C}heat, {R}epeat: {D}ata {C}ontamination and {E}valuation {M}alpractices in {C}losed-{S}ource {LLM}s}.
\newblock In \bibinfo{editor}{ Y.{ }Graham}, and \bibinfo{editor}{ M.{ }Purver}, eds. \bibinfo{booktitle}{Proceedings of the 18th Conference of the European Chapter of the Association for Computational Linguistics (Volume 1: Long Papers)}. \bibinfo{address}{St. Julian{'}s, Malta}: \bibinfo{publisher}{Association for Computational Linguistics} pp. \bibinfo{pages}{67--93}.
\bibitem[{Achiam et~al.(2023)Achiam, Adler, Agarwal, Ahmad, Akkaya, Aleman, Almeida, Altenschmidt, Altman, Anadkat et~al.}]{achiam2023gpt}
\bibinfo{author}{Achiam, J.}, \bibinfo{author}{Adler, S.}, \bibinfo{author}{Agarwal, S.}, \bibinfo{author}{Ahmad, L.}, \bibinfo{author}{Akkaya, I.}, \bibinfo{author}{Aleman, F.L.}, \bibinfo{author}{Almeida, D.}, \bibinfo{author}{Altenschmidt, J.}, \bibinfo{author}{Altman, S.}, \bibinfo{author}{Anadkat, S.} et~al. (\bibinfo{year}{2023}). \bibinfo{title}{Gpt-4 technical report}.
\newblock \bibinfo{journal}{Preprint at arXiv \url{https://doi.org/10.48550/arXiv.2303.08774}}.
\bibitem[{Chung et~al.(2024{\natexlab{b}})Chung, Hou, Longpre, Zoph, Tay, Fedus, Li, Wang, Dehghani, Brahma et~al.}]{chung2024scaling}
\bibinfo{author}{Chung, H.W.}, \bibinfo{author}{Hou, L.}, \bibinfo{author}{Longpre, S.}, \bibinfo{author}{Zoph, B.}, \bibinfo{author}{Tay, Y.}, \bibinfo{author}{Fedus, W.}, \bibinfo{author}{Li, Y.}, \bibinfo{author}{Wang, X.}, \bibinfo{author}{Dehghani, M.}, \bibinfo{author}{Brahma, S.} et~al. (\bibinfo{year}{2024}{\natexlab{b}}). \bibinfo{title}{Scaling instruction-finetuned language models}.
\newblock \bibinfo{journal}{Journal of Machine Learning Research} \emph{\bibinfo{volume}{25}}, \bibinfo{pages}{1--53}. \DOIprefix\doi{https://doi.org/10.5555/3722577.3722647}.
\bibitem[{Loukas et~al.(2025)Loukas, Smyrnioudis, Dikonomaki, Barbakos, Toumazatos, Koutsikakis, Kyriakakis, Georgiou, Vassos, Pavlopoulos and Androutsopoulos}]{loukas-etal-2025-gr}
\bibinfo{author}{Loukas, L.}, \bibinfo{author}{Smyrnioudis, N.}, \bibinfo{author}{Dikonomaki, C.}, \bibinfo{author}{Barbakos, S.}, \bibinfo{author}{Toumazatos, A.}, \bibinfo{author}{Koutsikakis, J.}, \bibinfo{author}{Kyriakakis, M.}, \bibinfo{author}{Georgiou, M.}, \bibinfo{author}{Vassos, S.}, \bibinfo{author}{Pavlopoulos, J.}, and \bibinfo{author}{Androutsopoulos, I.} (\bibinfo{year}{2025}). \bibinfo{title}{{GR}-{NLP}-{TOOLKIT}: An open-source {NLP} toolkit for {M}odern {G}reek}.
\newblock In \bibinfo{editor}{ O.{ }Rambow}, \bibinfo{editor}{ L.{ }Wanner}, \bibinfo{editor}{ M.{ }Apidianaki}, \bibinfo{editor}{ H.{ }Al-Khalifa}, \bibinfo{editor}{ B.D.{ }Eugenio}, \bibinfo{editor}{ S.{ }Schockaert}, \bibinfo{editor}{ B.{ }Mather}, and \bibinfo{editor}{ M.{ }Dras}, eds. \bibinfo{booktitle}{Proceedings of the 31st International Conference on Computational Linguistics: System Demonstrations}. \bibinfo{address}{Abu Dhabi, UAE}: \bibinfo{publisher}{Association for Computational Linguistics} pp. \bibinfo{pages}{174--182}.
\bibitem[{Voukoutis et~al.(2024)Voukoutis, Roussis, Paraskevopoulos, Sofianopoulos, Prokopidis, Papavasileiou, Katsamanis, Piperidis and Katsouros}]{voukoutis2024meltemi}
\bibinfo{author}{Voukoutis, L.}, \bibinfo{author}{Roussis, D.}, \bibinfo{author}{Paraskevopoulos, G.}, \bibinfo{author}{Sofianopoulos, S.}, \bibinfo{author}{Prokopidis, P.}, \bibinfo{author}{Papavasileiou, V.}, \bibinfo{author}{Katsamanis, A.}, \bibinfo{author}{Piperidis, S.}, and \bibinfo{author}{Katsouros, V.} (\bibinfo{year}{2024}). \bibinfo{title}{Meltemi: The first open large language model for greek}.
\newblock \bibinfo{journal}{Preprint at arXiv \url{https://doi.org/10.48550/arXiv.2407.20743}}.
\bibitem[{{Institute for Language and Speech Processing (ILSP)}(2024)}]{LlamaKrikri2024}
\bibinfo{author}{{Institute for Language and Speech Processing (ILSP)}} (\bibinfo{year}{2024}).
\newblock \bibinfo{title}{Llama-krikri-8b-base}. \bibinfo{publisher}{Hugging Face}.
\newblock \bibinfo{note}{\url{https://huggingface.co/ilsp/Llama-Krikri-8B-Base}}.
\bibitem[{Liang et~al.(2020)Liang, Duan, Gong, Wu, Guo, Qi, Gong, Shou, Jiang, Cao et~al.}]{liang2020xglue}
\bibinfo{author}{Liang, Y.}, \bibinfo{author}{Duan, N.}, \bibinfo{author}{Gong, Y.}, \bibinfo{author}{Wu, N.}, \bibinfo{author}{Guo, F.}, \bibinfo{author}{Qi, W.}, \bibinfo{author}{Gong, M.}, \bibinfo{author}{Shou, L.}, \bibinfo{author}{Jiang, D.}, \bibinfo{author}{Cao, G.} et~al. (\bibinfo{year}{2020}). \bibinfo{title}{Xglue: A new benchmark dataset for cross-lingual pre-training, understanding and generation}.
\newblock In \bibinfo{booktitle}{Proceedings of the 2020 Conference on Empirical Methods in Natural Language Processing (EMNLP)}. pp. \bibinfo{pages}{6008--6018}.
\bibitem[{Hu et~al.(2020)Hu, Ruder, Siddhant, Neubig, Firat and Johnson}]{hu2020xtreme}
\bibinfo{author}{Hu, J.}, \bibinfo{author}{Ruder, S.}, \bibinfo{author}{Siddhant, A.}, \bibinfo{author}{Neubig, G.}, \bibinfo{author}{Firat, O.}, and \bibinfo{author}{Johnson, M.} (\bibinfo{year}{2020}). \bibinfo{title}{Xtreme: A massively multilingual multi-task benchmark for evaluating cross-lingual generalisation}.
\newblock In \bibinfo{booktitle}{International conference on machine learning}. \bibinfo{organization}{PMLR} pp. \bibinfo{pages}{4411--4421}.
\bibitem[{Woolf(2010)}]{woolf2010building}
\bibinfo{author}{Woolf, B.P.} (\bibinfo{year}{2010}). \bibinfo{title}{Building intelligent interactive tutors: Student-centered strategies for revolutionizing e-learning}. \bibinfo{publisher}{Morgan Kaufmann}.
\bibitem[{Cambria et~al.(2017)Cambria, Poria, Gelbukh and Thelwall}]{cambria2017sentiment}
\bibinfo{author}{Cambria, E.}, \bibinfo{author}{Poria, S.}, \bibinfo{author}{Gelbukh, A.}, and \bibinfo{author}{Thelwall, M.} (\bibinfo{year}{2017}). \bibinfo{title}{Sentiment analysis is a big suitcase}.
\newblock \bibinfo{journal}{IEEE Intelligent Systems} \emph{\bibinfo{volume}{32}}, \bibinfo{pages}{74--80}. \DOIprefix\doi{10.1109/MIS.2017.4531228}.
\bibitem[{Zhang et~al.(2023)Zhang, Mao and Cambria}]{zhang2023survey}
\bibinfo{author}{Zhang, X.}, \bibinfo{author}{Mao, R.}, and \bibinfo{author}{Cambria, E.} (\bibinfo{year}{2023}). \bibinfo{title}{A survey on syntactic processing techniques}.
\newblock \bibinfo{journal}{Artificial Intelligence Review} \emph{\bibinfo{volume}{56}}, \bibinfo{pages}{5645--5728}. \DOIprefix\doi{https://doi.org/10.1007/s10462-022-10300-7}.
\bibitem[{Wang et~al.(2021)Wang, Wang, Dang, Liu and Liu}]{wang2021comprehensive}
\bibinfo{author}{Wang, Y.}, \bibinfo{author}{Wang, Y.}, \bibinfo{author}{Dang, K.}, \bibinfo{author}{Liu, J.}, and \bibinfo{author}{Liu, Z.} (\bibinfo{year}{2021}). \bibinfo{title}{A comprehensive survey of grammatical error correction}.
\newblock \bibinfo{journal}{ACM Transactions on Intelligent Systems and Technology (TIST)} \emph{\bibinfo{volume}{12}}, \bibinfo{pages}{1--51}. \DOIprefix\doi{https://doi.org/10.1145/3474840}.
\bibitem[{Kiss and Strunk(2006)}]{kiss2006unsupervised}
\bibinfo{author}{Kiss, T.}, and \bibinfo{author}{Strunk, J.} (\bibinfo{year}{2006}). \bibinfo{title}{Unsupervised multilingual sentence boundary detection}.
\newblock \bibinfo{journal}{Computational linguistics} \emph{\bibinfo{volume}{32}}, \bibinfo{pages}{485--525}. \DOIprefix\doi{https://doi.org/10.1162/coli.2006.32.4.485}.
\bibitem[{Qi et~al.(2019)Qi, Dozat, Zhang and Manning}]{qi2019universal}
\bibinfo{author}{Qi, P.}, \bibinfo{author}{Dozat, T.}, \bibinfo{author}{Zhang, Y.}, and \bibinfo{author}{Manning, C.D.} (\bibinfo{year}{2019}). \bibinfo{title}{Universal dependency parsing from scratch}.
\newblock \bibinfo{journal}{Preprint at arXiv \url{https://doi.org/10.48550/arXiv.1901.10457}}.
\bibitem[{Dozat and Manning(2016)}]{dozat2016deep}
\bibinfo{author}{Dozat, T.}, and \bibinfo{author}{Manning, C.D.} (\bibinfo{year}{2016}). \bibinfo{title}{Deep biaffine attention for neural dependency parsing}.
\newblock \bibinfo{journal}{Preprint at arXiv \url{https://doi.org/10.48550/arXiv.1611.01734}}.
\bibitem[{Prokopidis and Papageorgiou(2017{\natexlab{a}})}]{prokopidis2017universal}
\bibinfo{author}{Prokopidis, P.}, and \bibinfo{author}{Papageorgiou, H.} (\bibinfo{year}{2017}{\natexlab{a}}). \bibinfo{title}{Universal dependencies for greek}.
\newblock In \bibinfo{booktitle}{Proceedings of the nodalida 2017 workshop on universal dependencies (udw 2017)}. pp. \bibinfo{pages}{102--106}.
\bibitem[{Partalidou et~al.(2019)Partalidou, Spyromitros-Xioufis, Doropoulos, Vologiannidis and Diamantaras}]{Partalidou2019}
\bibinfo{author}{Partalidou, E.}, \bibinfo{author}{Spyromitros-Xioufis, E.}, \bibinfo{author}{Doropoulos, S.}, \bibinfo{author}{Vologiannidis, S.}, and \bibinfo{author}{Diamantaras, K.} (\bibinfo{year}{2019}). \bibinfo{title}{Design and implementation of an open source greek pos tagger and entity recognizer using spacy}.
\newblock In \bibinfo{booktitle}{IEEE/WIC/ACM International Conference on Web Intelligence}. WI '19 \bibinfo{organization}{IEEE}. \bibinfo{address}{New York, NY, USA}: \bibinfo{publisher}{Association for Computing Machinery}.
\newblock ISBN \bibinfo{isbn}{9781450369343} pp. \bibinfo{pages}{337--341}.
\newblock \DOIprefix\doi{10.1145/3350546.3352543}.
\bibitem[{Honnibal and Montani(2017)}]{honnibal2017spacy}
\bibinfo{author}{Honnibal, M.}, and \bibinfo{author}{Montani, I.} (\bibinfo{year}{2017}). \bibinfo{title}{spacy 2: Natural language understanding with bloom embeddings, convolutional neural networks and incremental parsing}.
\newblock \bibinfo{note}{Preprint at Sentometrics Research \url{https://sentometrics-research.com/publication/72/}}.
\bibitem[{Prokopidis et~al.(2011)Prokopidis, Georgantopoulos and Papageorgiou}]{prokopidis2011suite}
\bibinfo{author}{Prokopidis, P.}, \bibinfo{author}{Georgantopoulos, B.}, and \bibinfo{author}{Papageorgiou, H.} (\bibinfo{year}{2011}). \bibinfo{title}{A suite of natural language processing tools for greek}.
\newblock In \bibinfo{booktitle}{The 10th international conference of Greek linguistics}.
\bibitem[{Bird(2006)}]{bird2006nltk}
\bibinfo{author}{Bird, S.} (\bibinfo{year}{2006}). \bibinfo{title}{Nltk: the natural language toolkit}.
\newblock In \bibinfo{booktitle}{Proceedings of the COLING/ACL 2006 Interactive Presentation Sessions}. pp. \bibinfo{pages}{69--72}.
\bibitem[{Al-Rfou(2015)}]{Polyglot2015}
\bibinfo{author}{Al-Rfou, R.} (\bibinfo{year}{2015}).
\newblock \bibinfo{title}{Polyglot: Natural language pipeline supporting massive multilingual applications}. \bibinfo{publisher}{GitHub}.
\newblock \bibinfo{note}{\url{https://github.com/aboSamoor/polyglot}}.
\bibitem[{{Explosion AI}(2025)}]{spacy_greek_models}
\bibinfo{author}{{Explosion AI}} (\bibinfo{year}{2025}).
\newblock \bibinfo{title}{spacy greek language models}. \bibinfo{publisher}{spaCy}.
\newblock \bibinfo{note}{\url{https://spacy.io/models/el}}.
\bibitem[{{GFOSS - Open Technologies Alliance}(2018)}]{gsoc2018_spacy_greek}
\bibinfo{author}{{GFOSS - Open Technologies Alliance}} (\bibinfo{year}{2018}).
\newblock \bibinfo{title}{Greek language support for spacy (gsoc 2018 project)}. \bibinfo{publisher}{GitHub}.
\newblock \bibinfo{note}{\url{https://github.com/eellak/gsoc2018-spacy}}.
\bibitem[{Qi et~al.(2020)Qi, Zhang, Zhang, Bolton and Manning}]{Qi2020}
\bibinfo{author}{Qi, P.}, \bibinfo{author}{Zhang, Y.}, \bibinfo{author}{Zhang, Y.}, \bibinfo{author}{Bolton, J.}, and \bibinfo{author}{Manning, C.D.} (\bibinfo{year}{2020}). \bibinfo{title}{Stanza: {A} python natural language processing toolkit for many human languages}.
\newblock \bibinfo{journal}{Preprint at arXiv \url{https://doi.org/10.48550/arXiv.2003.07082}}.
\bibitem[{Straka and Strakov{\'a}(2017)}]{straka2017tokenizing}
\bibinfo{author}{Straka, M.}, and \bibinfo{author}{Strakov{\'a}, J.} (\bibinfo{year}{2017}). \bibinfo{title}{Tokenizing, pos tagging, lemmatizing and parsing ud 2.0 with udpipe}.
\newblock In \bibinfo{booktitle}{Proceedings of the CoNLL 2017 shared task: Multilingual parsing from raw text to universal dependencies}. pp. \bibinfo{pages}{88--99}.
\bibitem[{Zuhra and Saleem(2023)}]{zuhra2023hybrid}
\bibinfo{author}{Zuhra, F.T.}, and \bibinfo{author}{Saleem, K.} (\bibinfo{year}{2023}). \bibinfo{title}{Hybrid embeddings for transition-based dependency parsing of free word order languages}.
\newblock \bibinfo{journal}{Information Processing \& Management} \emph{\bibinfo{volume}{60}}, \bibinfo{pages}{103334}. \DOIprefix\doi{https://doi.org/10.1016/j.ipm.2023.103334}.
\bibitem[{Wong et~al.(2014)Wong, Chao and Zeng}]{wong2014isentenizer}
\bibinfo{author}{Wong, D.F.}, \bibinfo{author}{Chao, L.S.}, and \bibinfo{author}{Zeng, X.} (\bibinfo{year}{2014}). \bibinfo{title}{isentenizer-: Multilingual sentence boundary detection model}.
\newblock \bibinfo{journal}{The Scientific World Journal} \emph{\bibinfo{volume}{2014}}, \bibinfo{pages}{196574}. \DOIprefix\doi{https://doi.org/10.1155/2014/196574}.
\bibitem[{Fotopoulou and Giouli(2015)}]{fotopoulou2015mwes}
\bibinfo{author}{Fotopoulou, A.}, and \bibinfo{author}{Giouli, V.} (\bibinfo{year}{2015}). \bibinfo{title}{Mwes: support/light verb constructions vs fixed expressions in modern greek and french}.
\newblock In \bibinfo{booktitle}{Workshop on Multiword units in machine translation and translation technology}. \bibinfo{organization}{Tradulex} pp. \bibinfo{pages}{68--73}.
\bibitem[{Samaridi and Markantonatou(2014)}]{samaridi2014}
\bibinfo{author}{Samaridi, N.}, and \bibinfo{author}{Markantonatou, S.} (\bibinfo{year}{2014}). \bibinfo{title}{Parsing {M}odern {G}reek verb {MWE}s with {LFG}/{XLE} grammars}.
\newblock In \bibinfo{booktitle}{Proceedings of the 10th Workshop on Multiword Expressions ({MWE})}. \bibinfo{address}{Gothenburg, Sweden}: \bibinfo{publisher}{Association for Computational Linguistics} pp. \bibinfo{pages}{33--37}.
\bibitem[{Korre et~al.(2021{\natexlab{a}})Korre, Chatzipanagiotou and Pavlopoulos}]{korre2021elerrant}
\bibinfo{author}{Korre, K.}, \bibinfo{author}{Chatzipanagiotou, M.}, and \bibinfo{author}{Pavlopoulos, J.} (\bibinfo{year}{2021}{\natexlab{a}}). \bibinfo{title}{Elerrant: Automatic grammatical error type classification for greek}.
\newblock In \bibinfo{booktitle}{Proceedings of the International Conference on Recent Advances in Natural Language Processing (RANLP 2021)}. pp. \bibinfo{pages}{708--717}.
\bibitem[{Gakis et~al.(2016)Gakis, Panagiotakopoulos, Sgarbas, Tsalidis and Verykios}]{gakis:2017}
\bibinfo{author}{Gakis, P.}, \bibinfo{author}{Panagiotakopoulos, C.}, \bibinfo{author}{Sgarbas, K.}, \bibinfo{author}{Tsalidis, C.}, and \bibinfo{author}{Verykios, V.} (\bibinfo{year}{2016}). \bibinfo{title}{Design and construction of the greek grammar checker}.
\newblock \bibinfo{journal}{Digital Scholarship in the Humanities} \emph{\bibinfo{volume}{32}}, \bibinfo{pages}{554--576}. \DOIprefix\doi{10.1093/llc/fqw025}.
\bibitem[{{Neurolingo L.P.}(2017)}]{neurolingo_ggc}
\bibinfo{author}{{Neurolingo L.P.}} (\bibinfo{year}{2017}).
\newblock \bibinfo{title}{Grammar checker}.
\newblock \bibinfo{note}{\url{http://www.neurolingo.gr/en/online_tools/ggc}}.
\bibitem[{Kavros and Tzitzikas(2022{\natexlab{a}})}]{kavros2022soundexgr}
\bibinfo{author}{Kavros, A.}, and \bibinfo{author}{Tzitzikas, Y.} (\bibinfo{year}{2022}{\natexlab{a}}). \bibinfo{title}{Soundexgr: An algorithm for phonetic matching for the greek language}.
\newblock \bibinfo{journal}{Natural Language Engineering} \emph{\bibinfo{volume}{29}}, \bibinfo{pages}{1--36}. \DOIprefix\doi{https://doi.org/10.1017/S1351324922000018}.
\bibitem[{Kavros and Tzitzikas(2022{\natexlab{b}})}]{soundexgr_repo}
\bibinfo{author}{Kavros, A.}, and \bibinfo{author}{Tzitzikas, Y.} (\bibinfo{year}{2022}{\natexlab{b}}).
\newblock \bibinfo{title}{Soundexgr: An algorithm for phonetic matching for the greek language}. \bibinfo{publisher}{GitHub}{\natexlab{b}}.
\newblock \bibinfo{note}{\url{https://github.com/YannisTzitzikas/SoundexGR}}.
\bibitem[{Hua et~al.(2018)Hua, Danescu-Niculescu-Mizil, Taraborelli, Thain, Sorensen and Dixon}]{hua-2018-wikiconv}
\bibinfo{author}{Hua, Y.}, \bibinfo{author}{Danescu-Niculescu-Mizil, C.}, \bibinfo{author}{Taraborelli, D.}, \bibinfo{author}{Thain, N.}, \bibinfo{author}{Sorensen, J.}, and \bibinfo{author}{Dixon, L.} (\bibinfo{year}{2018}). \bibinfo{title}{{W}iki{C}onv: A corpus of the complete conversational history of a large online collaborative community}.
\newblock In \bibinfo{booktitle}{Proceedings of the 2018 Conference on Empirical Methods in Natural Language Processing}. \bibinfo{address}{Brussels, Belgium}: \bibinfo{publisher}{Association for Computational Linguistics} pp. \bibinfo{pages}{2818--2823}.
\bibitem[{Nivre et~al.(2016)Nivre, De~Marneffe, Ginter, Goldberg, Hajic, Manning, McDonald, Petrov, Pyysalo, Silveira et~al.}]{nivre2016universal}
\bibinfo{author}{Nivre, J.}, \bibinfo{author}{De~Marneffe, M.C.}, \bibinfo{author}{Ginter, F.}, \bibinfo{author}{Goldberg, Y.}, \bibinfo{author}{Hajic, J.}, \bibinfo{author}{Manning, C.D.}, \bibinfo{author}{McDonald, R.}, \bibinfo{author}{Petrov, S.}, \bibinfo{author}{Pyysalo, S.}, \bibinfo{author}{Silveira, N.} et~al. (\bibinfo{year}{2016}). \bibinfo{title}{Universal dependencies v1: A multilingual treebank collection}.
\newblock In \bibinfo{booktitle}{Proceedings of the Tenth International Conference on Language Resources and Evaluation (LREC'16)}. pp. \bibinfo{pages}{1659--1666}.
\bibitem[{Prokopidis et~al.(2005)Prokopidis, Desipri, Koutsombogera, Papageorgiou and Piperidis}]{prokopidis2005theoretical}
\bibinfo{author}{Prokopidis, P.}, \bibinfo{author}{Desipri, E.}, \bibinfo{author}{Koutsombogera, M.}, \bibinfo{author}{Papageorgiou, H.}, and \bibinfo{author}{Piperidis, S.} (\bibinfo{year}{2005}). \bibinfo{title}{Theoretical and practical issues in the construction of a greek dependency corpus}.
\newblock In \bibinfo{booktitle}{Proceedings of the Fourth Workshop on Treebanks and Linguistic Theories:(TLT 2005): 9-10 december 2005, Barcelona}. \bibinfo{organization}{Publicaciones y Ediciones= Publicacions i Edicions} pp. \bibinfo{pages}{149--160}.
\bibitem[{Gakis et~al.(2015)Gakis, Panagiotakopoulos, Sgarbas and Tsalidis}]{gakis2015analysis}
\bibinfo{author}{Gakis, P.}, \bibinfo{author}{Panagiotakopoulos, C.}, \bibinfo{author}{Sgarbas, K.}, and \bibinfo{author}{Tsalidis, C.} (\bibinfo{year}{2015}). \bibinfo{title}{Analysis of lexical ambiguity in modern greek using a computational lexicon}.
\newblock \bibinfo{journal}{Digital Scholarship in the Humanities} \emph{\bibinfo{volume}{30}}, \bibinfo{pages}{20--38}. \DOIprefix\doi{https://doi.org/10.1093/llc/fqt035}.
\bibitem[{Gakis et~al.(2012)Gakis, Panagiotakopoulos, Sgarbas and Tsalidis}]{gakis2012}
\bibinfo{author}{Gakis, P.}, \bibinfo{author}{Panagiotakopoulos, C.}, \bibinfo{author}{Sgarbas, K.}, and \bibinfo{author}{Tsalidis, C.} (\bibinfo{year}{2012}). \bibinfo{title}{Design and implementation of an electronic lexicon for modern greek}.
\newblock \bibinfo{journal}{Literary and Linguistic Computing} \emph{\bibinfo{volume}{27}}, \bibinfo{pages}{155--169}. \DOIprefix\doi{10.1093/llc/fqs002}.
\bibitem[{Korre et~al.(2021{\natexlab{b}})Korre, Chatzipanagiotou and Pavlopoulos}]{elerrant}
\bibinfo{author}{Korre, K.}, \bibinfo{author}{Chatzipanagiotou, M.}, and \bibinfo{author}{Pavlopoulos, J.} (\bibinfo{year}{2021}{\natexlab{b}}).
\newblock \bibinfo{title}{Elerrant: Greek version of errant}. \bibinfo{publisher}{GitHub}{\natexlab{b}}.
\newblock \bibinfo{note}{\url{https://github.com/katkorre/elerrant}}.
\bibitem[{Prokopidis and Papageorgiou(2017{\natexlab{b}})}]{ud-greek-gdt}
\bibinfo{author}{Prokopidis, P.}, and \bibinfo{author}{Papageorgiou, H.} (\bibinfo{year}{2017}{\natexlab{b}}).
\newblock \bibinfo{title}{Ud\_greek-gdt: Universal dependencies treebank}. \bibinfo{publisher}{GitHub}{\natexlab{b}}.
\newblock \bibinfo{note}{\url{https://github.com/UniversalDependencies/UD_Greek-GDT}}.
\bibitem[{Goddard(2011)}]{goddard2011semantic}
\bibinfo{author}{Goddard, C.} (\bibinfo{year}{2011}). \bibinfo{title}{Semantic analysis: A practical introduction}. \bibinfo{publisher}{Oxford University Press, USA}.
\bibitem[{Cruse(1986)}]{cruse1986lexical}
\bibinfo{author}{Cruse, D.A.} (\bibinfo{year}{1986}). \bibinfo{title}{Lexical semantics}. \bibinfo{publisher}{Cambridge university press}.
\bibitem[{Prasad et~al.(2008)Prasad, Dinesh, Lee, Miltsakaki, Robaldo, Joshi and Webber}]{prasad2008penn}
\bibinfo{author}{Prasad, R.}, \bibinfo{author}{Dinesh, N.}, \bibinfo{author}{Lee, A.}, \bibinfo{author}{Miltsakaki, E.}, \bibinfo{author}{Robaldo, L.}, \bibinfo{author}{Joshi, A.K.}, and \bibinfo{author}{Webber, B.L.} (\bibinfo{year}{2008}). \bibinfo{title}{The penn discourse treebank 2.0.}
\newblock In \bibinfo{booktitle}{Proceedings of the Sixth International Language Resources and Evaluation (LREC'08)}. \bibinfo{organization}{European Language Resources Association (ELRA)} pp. \bibinfo{pages}{2961--2968}.
\bibitem[{Lenci et~al.(2022)Lenci, Sahlgren, Jeuniaux, Cuba~Gyllensten and Miliani}]{lenci2022comparative}
\bibinfo{author}{Lenci, A.}, \bibinfo{author}{Sahlgren, M.}, \bibinfo{author}{Jeuniaux, P.}, \bibinfo{author}{Cuba~Gyllensten, A.}, and \bibinfo{author}{Miliani, M.} (\bibinfo{year}{2022}). \bibinfo{title}{A comparative evaluation and analysis of three generations of distributional semantic models}.
\newblock \bibinfo{journal}{Language resources and evaluation} \emph{\bibinfo{volume}{56}}, \bibinfo{pages}{1269--1313}. \DOIprefix\doi{https://doi.org/10.1007/s10579-021-09575-z}.
\bibitem[{Harris(1954)}]{harris1954distributional}
\bibinfo{author}{Harris, Z.S.} (\bibinfo{year}{1954}). \bibinfo{title}{Distributional structure}.
\newblock \bibinfo{journal}{Word} \emph{\bibinfo{volume}{10}}, \bibinfo{pages}{146--162}. \DOIprefix\doi{https://doi.org/10.1080/00437956.1954.11659520}.
\bibitem[{Sahlgren(2008)}]{sahlgren2008distributional}
\bibinfo{author}{Sahlgren, M.} (\bibinfo{year}{2008}). \bibinfo{title}{The distributional hypothesis}.
\newblock \bibinfo{journal}{Italian Journal of Linguistics} \emph{\bibinfo{volume}{20}}, \bibinfo{pages}{33--53}.
\bibitem[{Zervanou et~al.(2014)Zervanou, Iosif and Potamianos}]{zervanou2014word}
\bibinfo{author}{Zervanou, K.}, \bibinfo{author}{Iosif, E.}, and \bibinfo{author}{Potamianos, A.} (\bibinfo{year}{2014}). \bibinfo{title}{Word semantic similarity for morphologically rich languages.}
\newblock In \bibinfo{booktitle}{LREC}. pp. \bibinfo{pages}{1642--1648}.
\bibitem[{Palogiannidi et~al.(2015)Palogiannidi, Losif, Koutsakis and Potamianos}]{Palogiannidi2015}
\bibinfo{author}{Palogiannidi, E.}, \bibinfo{author}{Losif, E.}, \bibinfo{author}{Koutsakis, P.}, and \bibinfo{author}{Potamianos, A.} (\bibinfo{year}{2015}). \bibinfo{title}{Valence, arousal and dominance estimation for english, german, greek,portuguese and spanish lexica using semantic models}.
\newblock In \bibinfo{booktitle}{INTERSPEECH 2015}. pp. \bibinfo{pages}{1527--1531}.
\bibitem[{Palogiannidi et~al.(2016{\natexlab{a}})Palogiannidi, Koutsakis, Iosif and Potamianos}]{palogiannidi2016}
\bibinfo{author}{Palogiannidi, E.}, \bibinfo{author}{Koutsakis, P.}, \bibinfo{author}{Iosif, E.}, and \bibinfo{author}{Potamianos, A.} (\bibinfo{year}{2016}{\natexlab{a}}). \bibinfo{title}{Affective lexicon creation for the {G}reek language}.
\newblock In \bibinfo{booktitle}{Proceedings of the Tenth International Conference on Language Resources and Evaluation ({LREC}'16)}. \bibinfo{address}{Portoro{\v{z}}, Slovenia}: \bibinfo{publisher}{European Language Resources Association (ELRA)}{\natexlab{a}} pp. \bibinfo{pages}{2867--2872}.
\bibitem[{Iosif et~al.(2016{\natexlab{a}})Iosif, Georgiladakis and Potamianos}]{iosif-etal-2016-cognitively}
\bibinfo{author}{Iosif, E.}, \bibinfo{author}{Georgiladakis, S.}, and \bibinfo{author}{Potamianos, A.} (\bibinfo{year}{2016}{\natexlab{a}}). \bibinfo{title}{Cognitively motivated distributional representations of meaning}.
\newblock In \bibinfo{booktitle}{Proceedings of the Tenth International Conference on Language Resources and Evaluation ({LREC}'16)}. \bibinfo{address}{Portoro{\v{z}}, Slovenia}: \bibinfo{publisher}{European Language Resources Association (ELRA)}{\natexlab{a}} pp. \bibinfo{pages}{1226--1232}.
\bibitem[{Kahneman(2011)}]{kahneman2013thinking}
\bibinfo{author}{Kahneman, D.} (\bibinfo{year}{2011}). \bibinfo{title}{Thinking, Fast and Slow} vol. \bibinfo{volume}{499}. \bibinfo{publisher}{Farrar, Straus and Giroux}.
\newblock ISBN \bibinfo{isbn}{978-0374275631}.
\bibitem[{Lioudakis et~al.(2019{\natexlab{a}})Lioudakis, Outsios and Vazirgiannis}]{Lioudakis2019}
\bibinfo{author}{Lioudakis, M.}, \bibinfo{author}{Outsios, S.}, and \bibinfo{author}{Vazirgiannis, M.} (\bibinfo{year}{2019}{\natexlab{a}}). \bibinfo{title}{An ensemble method for producing word representations for the greek language}.
\newblock \bibinfo{journal}{Preprint at arXiv \url{https://doi.org/10.18653/v1/2020.loresmt-1.13}}.
\bibitem[{Lioudakis et~al.(2019{\natexlab{b}})Lioudakis, Outsios and Vazirgiannis}]{mikeliou_greek_word_embeddings}
\bibinfo{author}{Lioudakis, M.}, \bibinfo{author}{Outsios, S.}, and \bibinfo{author}{Vazirgiannis, M.} (\bibinfo{year}{2019}{\natexlab{b}}).
\newblock \bibinfo{title}{An ensemble method for producing word representations for the greek language}. \bibinfo{publisher}{GitHub}{\natexlab{b}}.
\newblock \bibinfo{note}{\url{https://github.com/mikeliou/greek_word_embeddings}}.
\bibitem[{Outsios et~al.(2020{\natexlab{a}})Outsios, Karatsalos, Skianis and Vazirgiannis}]{outsios2020-evaluation}
\bibinfo{author}{Outsios, S.}, \bibinfo{author}{Karatsalos, C.}, \bibinfo{author}{Skianis, K.}, and \bibinfo{author}{Vazirgiannis, M.} (\bibinfo{year}{2020}{\natexlab{a}}). \bibinfo{title}{Evaluation of {G}reek word embeddings}.
\newblock In \bibinfo{booktitle}{Proceedings of the Twelfth Language Resources and Evaluation Conference}. \bibinfo{address}{Marseille, France}: \bibinfo{publisher}{European Language Resources Association}{\natexlab{a}}.
\newblock ISBN \bibinfo{isbn}{979-10-95546-34-4} pp. \bibinfo{pages}{2543--2551}.
\bibitem[{Dritsa et~al.(2022{\natexlab{a}})Dritsa, Thoma, Pavlopoulos and Louridas}]{dritsa2022greek}
\bibinfo{author}{Dritsa, K.}, \bibinfo{author}{Thoma, A.}, \bibinfo{author}{Pavlopoulos, I.}, and \bibinfo{author}{Louridas, P.} (\bibinfo{year}{2022}{\natexlab{a}}). \bibinfo{title}{A greek parliament proceedings dataset for computational linguistics and political analysis}.
\newblock In \bibinfo{editor}{ S.{ }Koyejo}, \bibinfo{editor}{ S.{ }Mohamed}, \bibinfo{editor}{ A.{ }Agarwal}, \bibinfo{editor}{ D.{ }Belgrave}, \bibinfo{editor}{ K.{ }Cho}, and \bibinfo{editor}{ A.{ }Oh}, eds. \bibinfo{booktitle}{Advances in Neural Information Processing Systems} vol.~\bibinfo{volume}{35}. \bibinfo{publisher}{Curran Associates, Inc.}{\natexlab{a}} pp. \bibinfo{pages}{28874--28888}.
\newblock \URLprefix \url{https://proceedings.neurips.cc/paper_files/paper/2022/file/b96ce67b2f2d45e4ab315e13a6b5b9c5-Paper-Datasets_and_Benchmarks.pdf}.
\bibitem[{Barzokas et~al.(2020)Barzokas, Papagiannopoulou and Tsoumakas}]{Barzokas2020}
\bibinfo{author}{Barzokas, V.}, \bibinfo{author}{Papagiannopoulou, E.}, and \bibinfo{author}{Tsoumakas, G.} (\bibinfo{year}{2020}). \bibinfo{title}{Studying the evolution of greek words via word embeddings}.
\newblock In \bibinfo{booktitle}{11th Hellenic Conference on Artificial Intelligence}. SETN 2020. \bibinfo{address}{New York, NY, USA}: \bibinfo{publisher}{Association for Computing Machinery}.
\newblock ISBN \bibinfo{isbn}{9781450388788} pp. \bibinfo{pages}{118--124}.
\newblock \DOIprefix\doi{10.1145/3411408.3411425}.
\bibitem[{Hamilton et~al.(2016{\natexlab{a}})Hamilton, Leskovec and Jurafsky}]{hamilton2016diachronic}
\bibinfo{author}{Hamilton, W.L.}, \bibinfo{author}{Leskovec, J.}, and \bibinfo{author}{Jurafsky, D.} (\bibinfo{year}{2016}{\natexlab{a}}). \bibinfo{title}{Diachronic word embeddings reveal statistical laws of semantic change}.
\newblock In \bibinfo{booktitle}{Proceedings of the 54th Annual Meeting of the Association for Computational Linguistics (Volume 1: Long Papers)}. pp. \bibinfo{pages}{1489--1501}.
\bibitem[{Di~Carlo et~al.(2019)Di~Carlo, Bianchi and Palmonari}]{di2019training}
\bibinfo{author}{Di~Carlo, V.}, \bibinfo{author}{Bianchi, F.}, and \bibinfo{author}{Palmonari, M.} (\bibinfo{year}{2019}). \bibinfo{title}{Training temporal word embeddings with a compass}.
\newblock In \bibinfo{booktitle}{Proceedings of the AAAI conference on artificial intelligence} vol.~\bibinfo{volume}{33}. pp. \bibinfo{pages}{6326--6334}.
\bibitem[{Gonen et~al.(2020{\natexlab{c}})Gonen, Jawahar, Seddah and Goldberg}]{gonen-etal-2020-simple}
\bibinfo{author}{Gonen, H.}, \bibinfo{author}{Jawahar, G.}, \bibinfo{author}{Seddah, D.}, and \bibinfo{author}{Goldberg, Y.} (\bibinfo{year}{2020}{\natexlab{c}}). \bibinfo{title}{Simple, interpretable and stable method for detecting words with usage change across corpora}.
\newblock In \bibinfo{editor}{ D.{ }Jurafsky}, \bibinfo{editor}{ J.{ }Chai}, \bibinfo{editor}{ N.{ }Schluter}, and \bibinfo{editor}{ J.{ }Tetreault}, eds. \bibinfo{booktitle}{Proceedings of the 58th Annual Meeting of the Association for Computational Linguistics}. \bibinfo{address}{Online}: \bibinfo{publisher}{Association for Computational Linguistics}{\natexlab{c}} pp. \bibinfo{pages}{538--555}.
\newblock \DOIprefix\doi{10.18653/v1/2020.acl-main.51}.
\bibitem[{Hamilton et~al.(2016{\natexlab{b}})Hamilton, Leskovec and Jurafsky}]{hamilton2016cultural}
\bibinfo{author}{Hamilton, W.L.}, \bibinfo{author}{Leskovec, J.}, and \bibinfo{author}{Jurafsky, D.} (\bibinfo{year}{2016}{\natexlab{b}}). \bibinfo{title}{Cultural shift or linguistic drift? comparing two computational measures of semantic change}.
\newblock In \bibinfo{booktitle}{Proceedings of the conference on empirical methods in natural language processing. Conference on empirical methods in natural language processing} vol. \bibinfo{volume}{2016}. pp. \bibinfo{pages}{2116--2121}.
\bibitem[{Florou et~al.(2018)Florou, Perifanos and Goutsos}]{Florou2018}
\bibinfo{author}{Florou, E.}, \bibinfo{author}{Perifanos, K.}, and \bibinfo{author}{Goutsos, D.} (\bibinfo{year}{2018}). \bibinfo{title}{Neural embeddings for metaphor detection in a corpus of greek texts}.
\newblock In \bibinfo{booktitle}{2018 9th International Conference on Information, Intelligence, Systems and Applications (IISA)}. \bibinfo{organization}{IEEE} pp. \bibinfo{pages}{1--4}.
\newblock \DOIprefix\doi{10.1109/IISA.2018.8633668}.
\bibitem[{Steen(2007)}]{steen2007finding}
\bibinfo{author}{Steen, G.} (\bibinfo{year}{2007}). \bibinfo{title}{Finding Metaphor in Grammar and Usage: A Methodological Analysis of Theory and Research}.
\newblock Converging evidence in language and communication research. \bibinfo{publisher}{J. Benjamins Publishing Company}.
\newblock ISBN \bibinfo{isbn}{9789027238979}.
\bibitem[{Chowdhury et~al.(2014)Chowdhury, Ghosh, Stepanov, Bayer, Riccardi, Klasinas et~al.}]{chowdhury2014cross}
\bibinfo{author}{Chowdhury, S.A.}, \bibinfo{author}{Ghosh, A.}, \bibinfo{author}{Stepanov, E.A.}, \bibinfo{author}{Bayer, A.O.}, \bibinfo{author}{Riccardi, G.}, \bibinfo{author}{Klasinas, I.} et~al. (\bibinfo{year}{2014}). \bibinfo{title}{Cross-language transfer of semantic annotation via targeted crowdsourcing.}
\newblock In \bibinfo{booktitle}{INTERSPEECH}. pp. \bibinfo{pages}{2108--2112}.
\newblock \DOIprefix\doi{10.21437/Interspeech.2014-478}.
\bibitem[{Kamath and Das(2018)}]{kamath2018survey}
\bibinfo{author}{Kamath, A.}, and \bibinfo{author}{Das, R.} (\bibinfo{year}{2018}). \bibinfo{title}{A survey on semantic parsing}.
\newblock \bibinfo{journal}{Preprint at arXiv \url{https://doi.org/10.48550/arXiv.1812.00978}}.
\bibitem[{Li et~al.(2015)Li, Zhu, Lu and Zhou}]{li2015improving}
\bibinfo{author}{Li, J.}, \bibinfo{author}{Zhu, M.}, \bibinfo{author}{Lu, W.}, and \bibinfo{author}{Zhou, G.} (\bibinfo{year}{2015}). \bibinfo{title}{Improving semantic parsing with enriched synchronous context-free grammar}.
\newblock In \bibinfo{booktitle}{Proceedings of the 2015 Conference on Empirical Methods in Natural Language Processing}. pp. \bibinfo{pages}{1455--1465}.
\bibitem[{Conneau et~al.(2018)Conneau, Rinott, Lample, Williams, Bowman, Schwenk and Stoyanov}]{conneau2018xnli}
\bibinfo{author}{Conneau, A.}, \bibinfo{author}{Rinott, R.}, \bibinfo{author}{Lample, G.}, \bibinfo{author}{Williams, A.}, \bibinfo{author}{Bowman, S.R.}, \bibinfo{author}{Schwenk, H.}, and \bibinfo{author}{Stoyanov, V.} (\bibinfo{year}{2018}). \bibinfo{title}{Xnli: Evaluating cross-lingual sentence representations}.
\newblock In \bibinfo{booktitle}{Proceedings of the 2018 Conference on Empirical Methods in Natural Language Processing}. \bibinfo{publisher}{Association for Computational Linguistics} pp. \bibinfo{pages}{2475--2485}.
\bibitem[{Parikh et~al.(2016)Parikh, T{\"a}ckstr{\"o}m, Das and Uszkoreit}]{parikh2016decomposable}
\bibinfo{author}{Parikh, A.}, \bibinfo{author}{T{\"a}ckstr{\"o}m, O.}, \bibinfo{author}{Das, D.}, and \bibinfo{author}{Uszkoreit, J.} (\bibinfo{year}{2016}). \bibinfo{title}{A decomposable attention model for natural language inference}.
\newblock In \bibinfo{booktitle}{Proceedings of the 2016 Conference on Empirical Methods in Natural Language Processing}. pp. \bibinfo{pages}{2249--2255}.
\bibitem[{Giachos et~al.(2023)Giachos, Papakitsos, Antonopoulos and Laskaris}]{giachos2023systemic}
\bibinfo{author}{Giachos, I.}, \bibinfo{author}{Papakitsos, E.C.}, \bibinfo{author}{Antonopoulos, I.}, and \bibinfo{author}{Laskaris, N.} (\bibinfo{year}{2023}). \bibinfo{title}{Systemic and hole semantics in human-machine language interfaces}.
\newblock In \bibinfo{booktitle}{2023 17th International Conference on Engineering of Modern Electric Systems (EMES)}. \bibinfo{organization}{IEEE} pp. \bibinfo{pages}{1--4}.
\newblock \DOIprefix\doi{https://doi.org/10.1109/EMES58375.2023.10171635}.
\bibitem[{Ganitkevitch and Callison-Burch(2014)}]{ganitkevitch2014multilingual}
\bibinfo{author}{Ganitkevitch, J.}, and \bibinfo{author}{Callison-Burch, C.} (\bibinfo{year}{2014}). \bibinfo{title}{The multilingual paraphrase database.}
\newblock In \bibinfo{booktitle}{LREC}. \bibinfo{organization}{Citeseer} pp. \bibinfo{pages}{4276--4283}.
\bibitem[{Ganitkevitch and Callison-Burch(2013)}]{paraphrase}
\bibinfo{author}{Ganitkevitch, J.}, and \bibinfo{author}{Callison-Burch, C.} (\bibinfo{year}{2013}).
\newblock \bibinfo{title}{The paraphrase database}.
\newblock \bibinfo{note}{\url{http://paraphrase.org/}}.
\bibitem[{Outsios et~al.(2020{\natexlab{b}})Outsios, Karatsalos, Skianis and Vazirgiannis}]{aueb_outsios2020ev_353tr}
\bibinfo{author}{Outsios, S.}, \bibinfo{author}{Karatsalos, C.}, \bibinfo{author}{Skianis, K.}, and \bibinfo{author}{Vazirgiannis, M.} (\bibinfo{year}{2020}{\natexlab{b}}).
\newblock \bibinfo{title}{Wordsim353 dataset translated}.
\newblock \bibinfo{note}{\url{http://archive.aueb.gr:7000/resources/}}.
\bibitem[{Outsios et~al.(2020{\natexlab{c}})Outsios, Karatsalos, Skianis and Vazirgiannis}]{aueb_outsios2020ev_Test}
\bibinfo{author}{Outsios, S.}, \bibinfo{author}{Karatsalos, C.}, \bibinfo{author}{Skianis, K.}, and \bibinfo{author}{Vazirgiannis, M.} (\bibinfo{year}{2020}{\natexlab{c}}).
\newblock \bibinfo{title}{Greek word analogy test set}.
\newblock \bibinfo{note}{\url{http://archive.aueb.gr:7000/resources/}}.
\bibitem[{Pilitsidou and Giouli(2021)}]{pilitsidou2021frame}
\bibinfo{author}{Pilitsidou, V.}, and \bibinfo{author}{Giouli, V.} (\bibinfo{year}{2021}). \bibinfo{title}{Frame semantics in the specialized domain of finance: Building a termbase to aid translation}.
\newblock In \bibinfo{editor}{ Z.{ }Gavriilidou}, \bibinfo{editor}{ M.{ }Mitsiaki}, and \bibinfo{editor}{ A.{ }Fliatouras}, eds. \bibinfo{booktitle}{Lexicography for Inclusion: Proceedings of the 19th EURALEX International Congress, 7--9 September 2021, Alexandroupolis} vol.~\bibinfo{volume}{1}. \bibinfo{publisher}{Democritus University of Thrace}.
\newblock ISBN \bibinfo{isbn}{978-618-85138-1-5} pp. \bibinfo{pages}{263--271}.
\newblock \URLprefix \url{https://euralex.org/publications/frame-semantics-in-the-specialized-domain-of-finance-building-a-termbase-to-translation/}.
\bibitem[{Giouli et~al.(2020)Giouli, Pilitsidou and Christopoulos}]{giouli2020greek}
\bibinfo{author}{Giouli, V.}, \bibinfo{author}{Pilitsidou, V.}, and \bibinfo{author}{Christopoulos, H.} (\bibinfo{year}{2020}). \bibinfo{title}{{G}reek within the global {F}rame{N}et initiative: Challenges and conclusions so far}.
\newblock In \bibinfo{booktitle}{Proceedings of the International FrameNet Workshop 2020: Towards a Global, Multilingual FrameNet}. \bibinfo{address}{Marseille, France}: \bibinfo{publisher}{European Language Resources Association}.
\newblock ISBN \bibinfo{isbn}{979-10-95546-58-0} pp. \bibinfo{pages}{48--55}.
\bibitem[{Ganitkevitch et~al.(2013)Ganitkevitch, Van~Durme and Callison-Burch}]{ganitkevitch2013ppdb}
\bibinfo{author}{Ganitkevitch, J.}, \bibinfo{author}{Van~Durme, B.}, and \bibinfo{author}{Callison-Burch, C.} (\bibinfo{year}{2013}). \bibinfo{title}{Ppdb: The paraphrase database}.
\newblock In \bibinfo{booktitle}{Proceedings of the 2013 Conference of the North American Chapter of the Association for Computational Linguistics: Human Language Technologies}. pp. \bibinfo{pages}{758--764}.
\bibitem[{Bovi et~al.(2017)Bovi, Camacho-Collados, Raganato and Navigli}]{bovi2017eurosense}
\bibinfo{author}{Bovi, C.D.}, \bibinfo{author}{Camacho-Collados, J.}, \bibinfo{author}{Raganato, A.}, and \bibinfo{author}{Navigli, R.} (\bibinfo{year}{2017}). \bibinfo{title}{Eurosense: Automatic harvesting of multilingual sense annotations from parallel text}.
\newblock In \bibinfo{booktitle}{Proceedings of the 55th Annual Meeting of the Association for Computational Linguistics (Volume 2: Short Papers)}. pp. \bibinfo{pages}{594--600}.
\bibitem[{Navigli and Ponzetto(2012)}]{navigli2012babelnet}
\bibinfo{author}{Navigli, R.}, and \bibinfo{author}{Ponzetto, S.P.} (\bibinfo{year}{2012}). \bibinfo{title}{Babelnet: The automatic construction, evaluation and application of a wide-coverage multilingual semantic network}.
\newblock \bibinfo{journal}{Artificial intelligence} \emph{\bibinfo{volume}{193}}, \bibinfo{pages}{217--250}. \DOIprefix\doi{https://doi.org/10.1016/j.artint.2012.07.001}.
\bibitem[{Finkelstein et~al.(2001)Finkelstein, Gabrilovich, Matias, Rivlin, Solan, Wolfman and Ruppin}]{finkelstein2001placing}
\bibinfo{author}{Finkelstein, L.}, \bibinfo{author}{Gabrilovich, E.}, \bibinfo{author}{Matias, Y.}, \bibinfo{author}{Rivlin, E.}, \bibinfo{author}{Solan, Z.}, \bibinfo{author}{Wolfman, G.}, and \bibinfo{author}{Ruppin, E.} (\bibinfo{year}{2001}). \bibinfo{title}{Placing search in context: The concept revisited}.
\newblock In \bibinfo{booktitle}{Proceedings of the 10th international conference on World Wide Web}. pp. \bibinfo{pages}{406--414}.
\bibitem[{Mikolov et~al.(2013{\natexlab{b}})Mikolov, Chen, Corrado and Dean}]{mikolov2013efficient}
\bibinfo{author}{Mikolov, T.}, \bibinfo{author}{Chen, K.}, \bibinfo{author}{Corrado, G.}, and \bibinfo{author}{Dean, J.} (\bibinfo{year}{2013}{\natexlab{b}}). \bibinfo{title}{Efficient estimation of word representations in vector space}.
\newblock \bibinfo{journal}{Preprint at arXiv \url{https://doi.org/10.48550/arXiv.1301.3781}}.
\bibitem[{Nikolaev and Pad{\'o}(2023{\natexlab{a}})}]{nikolaev2023universe}
\bibinfo{author}{Nikolaev, D.}, and \bibinfo{author}{Pad{\'o}, S.} (\bibinfo{year}{2023}{\natexlab{a}}). \bibinfo{title}{The universe of utterances according to bert}.
\newblock In \bibinfo{booktitle}{Proceedings of the 15th International Conference on Computational Semantics}. pp. \bibinfo{pages}{99--105}.
\bibitem[{Nikolaev and Pad{\'o}(2023{\natexlab{b}})}]{pado2023investigating}
\bibinfo{author}{Nikolaev, D.}, and \bibinfo{author}{Pad{\'o}, S.} (\bibinfo{year}{2023}{\natexlab{b}}). \bibinfo{title}{Investigating semantic subspaces of transformer sentence embeddings through linear structural probing}.
\newblock In \bibinfo{editor}{ Y.{ }Belinkov}, \bibinfo{editor}{ S.{ }Hao}, \bibinfo{editor}{ J.{ }Jumelet}, \bibinfo{editor}{ N.{ }Kim}, \bibinfo{editor}{ A.{ }McCarthy}, and \bibinfo{editor}{ H.{ }Mohebbi}, eds. \bibinfo{booktitle}{Proceedings of the 6th BlackboxNLP Workshop: Analyzing and Interpreting Neural Networks for NLP}. \bibinfo{address}{Singapore}: \bibinfo{publisher}{Association for Computational Linguistics}{\natexlab{b}} pp. \bibinfo{pages}{142--154}.
\newblock \DOIprefix\doi{10.18653/v1/2023.blackboxnlp-1.11}.
\bibitem[{Yang et~al.(2022)Yang, Wu, Yang, Lian, Guo and Wang}]{yang2022survey}
\bibinfo{author}{Yang, Y.}, \bibinfo{author}{Wu, Z.}, \bibinfo{author}{Yang, Y.}, \bibinfo{author}{Lian, S.}, \bibinfo{author}{Guo, F.}, and \bibinfo{author}{Wang, Z.} (\bibinfo{year}{2022}). \bibinfo{title}{A survey of information extraction based on deep learning}.
\newblock \bibinfo{journal}{Applied Sciences} \emph{\bibinfo{volume}{12}}, \bibinfo{pages}{9691}. \DOIprefix\doi{https://doi.org/10.3390/app12199691}.
\bibitem[{Mouratidis et~al.(2023)Mouratidis, Kermanidis and Kanavos}]{mouratidis2023comparative}
\bibinfo{author}{Mouratidis, D.}, \bibinfo{author}{Kermanidis, K.}, and \bibinfo{author}{Kanavos, A.} (\bibinfo{year}{2023}). \bibinfo{title}{Comparative study of recurrent and dense neural networks for classifying maritime terms}.
\newblock In \bibinfo{booktitle}{2023 14th International Conference on Information, Intelligence, Systems \& Applications (IISA)}. \bibinfo{organization}{IEEE} pp. \bibinfo{pages}{1--6}.
\newblock \DOIprefix\doi{https://doi.org/10.1109/IISA59645.2023.10345925}.
\bibitem[{Mouratidis et~al.(2022)Mouratidis, Mathe, Voutos, Stamou, Kermanidis, Mylonas and Kanavos}]{Mouratidis2022}
\bibinfo{author}{Mouratidis, D.}, \bibinfo{author}{Mathe, E.}, \bibinfo{author}{Voutos, Y.}, \bibinfo{author}{Stamou, K.}, \bibinfo{author}{Kermanidis, K.L.}, \bibinfo{author}{Mylonas, P.}, and \bibinfo{author}{Kanavos, A.} (\bibinfo{year}{2022}). \bibinfo{title}{Domain-specific term extraction: A case study on greek maritime legal texts}.
\newblock In \bibinfo{booktitle}{Proceedings of the 12th Hellenic Conference on Artificial Intelligence}. SETN '22. \bibinfo{address}{New York, NY, USA}: \bibinfo{publisher}{Association for Computing Machinery}.
\newblock ISBN \bibinfo{isbn}{9781450395977} pp. \bibinfo{pages}{1--6}.
\bibitem[{Papadopoulos et~al.(2021{\natexlab{b}})Papadopoulos, Papadakis and Matsatsinis}]{penelopie}
\bibinfo{author}{Papadopoulos, D.}, \bibinfo{author}{Papadakis, N.}, and \bibinfo{author}{Matsatsinis, N.} (\bibinfo{year}{2021}{\natexlab{b}}).
\newblock \bibinfo{title}{Penelopie: Enabling open information extraction for the greek language through machine translation}. \bibinfo{publisher}{GitHub}{\natexlab{b}}.
\newblock \bibinfo{note}{\url{https://github.com/lighteternal/PENELOPIE}}.
\bibitem[{Barbaresi and Lejeune(2020)}]{barbaresi-lejeune-2020-box}
\bibinfo{author}{Barbaresi, A.}, and \bibinfo{author}{Lejeune, G.} (\bibinfo{year}{2020}). \bibinfo{title}{Out-of-the-box and into the ditch? multilingual evaluation of generic text extraction tools}.
\newblock In \bibinfo{booktitle}{Proceedings of the 12th Web as Corpus Workshop}. \bibinfo{address}{Marseille, France}: \bibinfo{publisher}{European Language Resources Association}.
\newblock ISBN \bibinfo{isbn}{979-10-95546-68-9} pp. \bibinfo{pages}{5--13}.
\bibitem[{Lejeune et~al.(2015)Lejeune, Brixtel, Doucet and Lucas}]{lejeune2015multilingual}
\bibinfo{author}{Lejeune, G.}, \bibinfo{author}{Brixtel, R.}, \bibinfo{author}{Doucet, A.}, and \bibinfo{author}{Lucas, N.} (\bibinfo{year}{2015}). \bibinfo{title}{Multilingual event extraction for epidemic detection}.
\newblock \bibinfo{journal}{Artificial intelligence in medicine} \emph{\bibinfo{volume}{65}}, \bibinfo{pages}{131--143}. \DOIprefix\doi{https://doi.org/10.1016/j.artmed.2015.06.005}.
\bibitem[{Brixtel et~al.(2013)Brixtel, Lejeune, Doucet and Lucas}]{brixtel2013any}
\bibinfo{author}{Brixtel, R.}, \bibinfo{author}{Lejeune, G.}, \bibinfo{author}{Doucet, A.}, and \bibinfo{author}{Lucas, N.} (\bibinfo{year}{2013}). \bibinfo{title}{Any language early detection of epidemic diseases from web news streams}.
\newblock In \bibinfo{booktitle}{2013 IEEE International Conference on Healthcare Informatics}. \bibinfo{organization}{IEEE} pp. \bibinfo{pages}{159--168}.
\newblock \DOIprefix\doi{10.1109/ICHI.2013.94}.
\bibitem[{Lejeune et~al.(2012{\natexlab{a}})Lejeune, Brixtel, Doucet and Lucas}]{lejeune2012daniel}
\bibinfo{author}{Lejeune, G.}, \bibinfo{author}{Brixtel, R.}, \bibinfo{author}{Doucet, A.}, and \bibinfo{author}{Lucas, N.} (\bibinfo{year}{2012}{\natexlab{a}}). \bibinfo{title}{Daniel: Language independent character-based news surveillance}.
\newblock In \bibinfo{booktitle}{International Conference on NLP}. \bibinfo{organization}{Springer} pp. \bibinfo{pages}{64--75}.
\newblock \DOIprefix\doi{https://doi.org/10.1007/978-3-642-33983-7_7}.
\bibitem[{Papantoniou et~al.(2023)Papantoniou, Efthymiou and Plexousakis}]{papantoniou2023automating}
\bibinfo{author}{Papantoniou, K.}, \bibinfo{author}{Efthymiou, V.}, and \bibinfo{author}{Plexousakis, D.} (\bibinfo{year}{2023}). \bibinfo{title}{Automating benchmark generation for named entity recognition and entity linking}.
\newblock In \bibinfo{booktitle}{European Semantic Web Conference}. \bibinfo{organization}{Springer} pp. \bibinfo{pages}{143--148}.
\newblock \DOIprefix\doi{https://doi.org/10.1007/978-3-031-43458-7_27}.
\bibitem[{Rizou et~al.(2023{\natexlab{a}})Rizou, Theofilatos, Paflioti, Pissari, Varlamis, Sarigiannidis and Chatzisavvas}]{rizou2023efficient}
\bibinfo{author}{Rizou, S.}, \bibinfo{author}{Theofilatos, A.}, \bibinfo{author}{Paflioti, A.}, \bibinfo{author}{Pissari, E.}, \bibinfo{author}{Varlamis, I.}, \bibinfo{author}{Sarigiannidis, G.}, and \bibinfo{author}{Chatzisavvas, K.C.} (\bibinfo{year}{2023}{\natexlab{a}}). \bibinfo{title}{Efficient intent classification and entity recognition for university administrative services employing deep learning models}.
\newblock \bibinfo{journal}{Intelligent Systems with Applications} \emph{\bibinfo{volume}{19}}, \bibinfo{pages}{200247}. \DOIprefix\doi{https://doi.org/10.1016/j.iswa.2023.200247}.
\bibitem[{Hemphill et~al.(1990)Hemphill, Godfrey and Doddington}]{hemphill-etal-1990-atis}
\bibinfo{author}{Hemphill, C.T.}, \bibinfo{author}{Godfrey, J.J.}, and \bibinfo{author}{Doddington, G.R.} (\bibinfo{year}{1990}). \bibinfo{title}{The {ATIS} spoken language systems pilot corpus}.
\newblock In \bibinfo{booktitle}{Speech and Natural Language: Proceedings of a Workshop Held at Hidden Valley, {P}ennsylvania, June 24-27,1990}. pp. \bibinfo{pages}{96--101}.
\newblock \DOIprefix\doi{https://doi.org/10.3115/116580.116613}.
\bibitem[{Bartziokas et~al.(2020)Bartziokas, Mavropoulos and Kotropoulos}]{Bartziokas2020}
\bibinfo{author}{Bartziokas, N.}, \bibinfo{author}{Mavropoulos, T.}, and \bibinfo{author}{Kotropoulos, C.} (\bibinfo{year}{2020}). \bibinfo{title}{Datasets and performance metrics for greek named entity recognition}.
\newblock In \bibinfo{booktitle}{11th Hellenic Conference on Artificial Intelligence}. SETN 2020. \bibinfo{address}{New York, NY, USA}: \bibinfo{publisher}{Association for Computing Machinery}.
\newblock ISBN \bibinfo{isbn}{9781450388788} pp. \bibinfo{pages}{160--167}.
\newblock \DOIprefix\doi{10.1145/3411408.3411437}.
\bibitem[{Sang and De~Meulder(2003)}]{sang2003introduction}
\bibinfo{author}{Sang, E.F.}, and \bibinfo{author}{De~Meulder, F.} (\bibinfo{year}{2003}). \bibinfo{title}{Introduction to the conll-2003 shared task: Language-independent named entity recognition}.
\newblock \bibinfo{journal}{Preprint at arXiv \url{https://doi.org/10.48550/arXiv.cs/0306050}}.
\bibitem[{Pradhan et~al.(2007)Pradhan, Hovy, Marcus, Palmer, Ramshaw and Weischedel}]{pradhan2007ontonotes}
\bibinfo{author}{Pradhan, S.S.}, \bibinfo{author}{Hovy, E.}, \bibinfo{author}{Marcus, M.}, \bibinfo{author}{Palmer, M.}, \bibinfo{author}{Ramshaw, L.}, and \bibinfo{author}{Weischedel, R.} (\bibinfo{year}{2007}). \bibinfo{title}{Ontonotes: A unified relational semantic representation}.
\newblock In \bibinfo{booktitle}{International Conference on Semantic Computing (ICSC 2007)}. \bibinfo{organization}{IEEE} pp. \bibinfo{pages}{517--526}.
\newblock \DOIprefix\doi{10.1109/ICSC.2007.83}.
\bibitem[{Angelidis et~al.(2018{\natexlab{a}})Angelidis, Chalkidis and Koubarakis}]{Angelidis2018}
\bibinfo{author}{Angelidis, I.}, \bibinfo{author}{Chalkidis, I.}, and \bibinfo{author}{Koubarakis, M.} (\bibinfo{year}{2018}{\natexlab{a}}). \bibinfo{title}{Named entity recognition, linking and generation for greek legislation.}
\newblock In \bibinfo{booktitle}{JURIX}. pp. \bibinfo{pages}{1--10}.
\newblock \DOIprefix\doi{10.3233/978-1-61499-935-5-1}.
\bibitem[{Papantoniou et~al.(2021)Papantoniou, Efthymiou and Flouris}]{papantoniou2021}
\bibinfo{author}{Papantoniou, K.}, \bibinfo{author}{Efthymiou, V.}, and \bibinfo{author}{Flouris, G.} (\bibinfo{year}{2021}). \bibinfo{title}{El-nel: Entity linking for greek news articles.}
\newblock In \bibinfo{booktitle}{ISWC (Posters/Demos/Industry)}.
\bibitem[{Papantoniou et~al.(2022)Papantoniou, Efthymiou and Plexousakis}]{ner_nel_greek_wikipedia}
\bibinfo{author}{Papantoniou, K.}, \bibinfo{author}{Efthymiou, V.}, and \bibinfo{author}{Plexousakis, D.} (\bibinfo{year}{2022}).
\newblock \bibinfo{title}{Dataset for named entity recognition and entity linking from greek wikipedia events}. \bibinfo{publisher}{Zenodo}.
\newblock \bibinfo{note}{\url{https://doi.org/10.5281/zenodo.7429037}}.
\bibitem[{Rizou et~al.(2023{\natexlab{b}})Rizou, Theofilatos, Paflioti, Pissari, Varlamis, Sarigiannidis and Chatzisavvas}]{uniway_dataset}
\bibinfo{author}{Rizou, S.}, \bibinfo{author}{Theofilatos, A.}, \bibinfo{author}{Paflioti, A.}, \bibinfo{author}{Pissari, E.}, \bibinfo{author}{Varlamis, I.}, \bibinfo{author}{Sarigiannidis, G.}, and \bibinfo{author}{Chatzisavvas, K.C.} (\bibinfo{year}{2023}{\natexlab{b}}).
\newblock \bibinfo{title}{Uniway dataset}. \bibinfo{publisher}{mSensis}{\natexlab{b}}.
\newblock \bibinfo{note}{\url{https://msensis.com/wp-content/uploads/2023/06/uniway.zip}}.
\bibitem[{Bartziokas et~al.(2021)Bartziokas, Mavropoulos and Kotropoulos}]{elner}
\bibinfo{author}{Bartziokas, N.}, \bibinfo{author}{Mavropoulos, T.}, and \bibinfo{author}{Kotropoulos, C.} (\bibinfo{year}{2021}).
\newblock \bibinfo{title}{elner: Greek named entity recognition}. \bibinfo{publisher}{GitHub}.
\newblock \bibinfo{note}{\url{https://github.com/nmpartzio/elNER}}.
\bibitem[{Rizou et~al.(2022{\natexlab{b}})Rizou, Paflioti, Theofilatos, Vakali, Sarigiannidis and Chatzisavvas}]{msensis_downloads}
\bibinfo{author}{Rizou, S.}, \bibinfo{author}{Paflioti, A.}, \bibinfo{author}{Theofilatos, A.}, \bibinfo{author}{Vakali, A.}, \bibinfo{author}{Sarigiannidis, G.}, and \bibinfo{author}{Chatzisavvas, K.} (\bibinfo{year}{2022}{\natexlab{b}}).
\newblock \bibinfo{title}{Atis gr dataset}.
\newblock \bibinfo{note}{\url{https://msensis.com/research-and-development/downloads}}.
\bibitem[{Lioudakis et~al.(2019{\natexlab{c}})Lioudakis, Outsios and Vazirgiannis}]{aueb-resources_Lioudakis2019}
\bibinfo{author}{Lioudakis, M.}, \bibinfo{author}{Outsios, S.}, and \bibinfo{author}{Vazirgiannis, M.} (\bibinfo{year}{2019}{\natexlab{c}}).
\newblock \bibinfo{title}{Greek ner spacy dataset}. \bibinfo{publisher}{AUEB}{\natexlab{c}}.
\newblock \bibinfo{note}{\url{http://archive.aueb.gr:7000/resources/}}.
\bibitem[{Angelidis et~al.(2018{\natexlab{b}})Angelidis, Chalkidis and Koubarakis}]{uoa_legislation}
\bibinfo{author}{Angelidis, I.}, \bibinfo{author}{Chalkidis, I.}, and \bibinfo{author}{Koubarakis, M.} (\bibinfo{year}{2018}{\natexlab{b}}).
\newblock \bibinfo{title}{Publications – legislation mining project}. \bibinfo{publisher}{National and Kapodistrian University of Athens}{\natexlab{b}}.
\newblock \bibinfo{note}{\url{https://legislation.di.uoa.gr/publications}}.
\bibitem[{Lejeune et~al.(2012{\natexlab{b}})Lejeune, Brixtel, Doucet and Lucas}]{daniel_lejeune}
\bibinfo{author}{Lejeune, G.}, \bibinfo{author}{Brixtel, R.}, \bibinfo{author}{Doucet, A.}, and \bibinfo{author}{Lucas, N.} (\bibinfo{year}{2012}{\natexlab{b}}).
\newblock \bibinfo{title}{Daniel: Language independent character-based news surveillance}. \bibinfo{publisher}{Université de Caen Normandie}{\natexlab{b}}.
\newblock \bibinfo{note}{\url{https://lejeuneg.users.greyc.fr/daniel/}}.
\bibitem[{{Explosion AI}(????)}]{prodigy}
\bibinfo{author}{{Explosion AI}}.
\newblock \bibinfo{title}{Prodigy: An annotation tool for ai, machine learning \& nlp}.
\newblock \bibinfo{note}{\url{https://prodi.gy/}}.
\bibitem[{Lawrence and Reed(2020)}]{lawrence2020argument}
\bibinfo{author}{Lawrence, J.}, and \bibinfo{author}{Reed, C.} (\bibinfo{year}{2020}). \bibinfo{title}{Argument mining: A survey}.
\newblock \bibinfo{journal}{Computational Linguistics} \emph{\bibinfo{volume}{45}}, \bibinfo{pages}{765--818}. \DOIprefix\doi{https://doi.org/10.1162/coli_a_00364}.
\bibitem[{Liu(2020)}]{liu2020sentiment}
\bibinfo{author}{Liu, B.} (\bibinfo{year}{2020}). \bibinfo{title}{Sentiment analysis: Mining opinions, sentiments, and emotions}. \bibinfo{publisher}{Cambridge university press}.
\bibitem[{Al{-}Ghuribi and Noah(2021)}]{Ghuribi2021}
\bibinfo{author}{Al{-}Ghuribi, S.M.}, and \bibinfo{author}{Noah, S.A.M.} (\bibinfo{year}{2021}). \bibinfo{title}{A comprehensive overview of recommender system and sentiment analysis}.
\newblock \bibinfo{journal}{Preprint at arXiv \url{https://doi.org/10.48550/arXiv.2109.08794}}.
\bibitem[{Rokade and Aruna(2019)}]{rokade2019business}
\bibinfo{author}{Rokade, P.P.}, and \bibinfo{author}{Aruna, K.D.} (\bibinfo{year}{2019}). \bibinfo{title}{Business intelligence analytics using sentiment analysis-a survey}.
\newblock \bibinfo{journal}{International Journal of Electrical and Computer Engineering} \emph{\bibinfo{volume}{9}}, \bibinfo{pages}{613}. \DOIprefix\doi{10.11591/ijece.v9i1.pp613-620}.
\bibitem[{Mudinas et~al.(2019)Mudinas, Zhang and Levene}]{mudinas2019market}
\bibinfo{author}{Mudinas, A.}, \bibinfo{author}{Zhang, D.}, and \bibinfo{author}{Levene, M.} (\bibinfo{year}{2019}). \bibinfo{title}{Market trend prediction using sentiment analysis: lessons learned and paths forward}.
\newblock \bibinfo{journal}{Preprint at arXiv \url{https://doi.org/10.48550/arXiv.1903.05440}}.
\bibitem[{Chauhan et~al.(2021)Chauhan, Sharma and Sikka}]{chauhan2021emergence}
\bibinfo{author}{Chauhan, P.}, \bibinfo{author}{Sharma, N.}, and \bibinfo{author}{Sikka, G.} (\bibinfo{year}{2021}). \bibinfo{title}{The emergence of social media data and sentiment analysis in election prediction}.
\newblock \bibinfo{journal}{Journal of Ambient Intelligence and Humanized Computing} \emph{\bibinfo{volume}{12}}, \bibinfo{pages}{2601--2627}. \DOIprefix\doi{https://doi.org/10.1007/s12652-020-02423-y}.
\bibitem[{Turney(2002)}]{Turney2002thumbs}
\bibinfo{author}{Turney, P.D.} (\bibinfo{year}{2002}). \bibinfo{title}{Thumbs up or thumbs down? semantic orientation applied to unsupervised classification of reviews}.
\newblock \bibinfo{journal}{Preprint at arXiv \url{https://doi.org/10.48550/arXiv.cs/0212032}}.
\bibitem[{Pang et~al.(2002)Pang, Lee and Vaithyanathan}]{Pang2002thumbs}
\bibinfo{author}{Pang, B.}, \bibinfo{author}{Lee, L.}, and \bibinfo{author}{Vaithyanathan, S.} (\bibinfo{year}{2002}). \bibinfo{title}{Thumbs up? sentiment classification using machine learning techniques}.
\newblock \bibinfo{journal}{Preprint at arXiv \url{https://doi.org/10.48550/arXiv.cs/0205070}}.
\bibitem[{Fragkis(2022)}]{greek_movies_kaggle}
\bibinfo{author}{Fragkis, N.} (\bibinfo{year}{2022}).
\newblock \bibinfo{title}{Greek movies dataset}. \bibinfo{publisher}{Kaggle}.
\newblock \bibinfo{note}{\url{https://www.kaggle.com/datasets/nikosfragkis/greek-movies-dataset}}.
\bibitem[{Braoudaki et~al.(2020)Braoudaki, Kanellou, Kozanitis and Fatourou}]{Braoudaki2020}
\bibinfo{author}{Braoudaki, A.}, \bibinfo{author}{Kanellou, E.}, \bibinfo{author}{Kozanitis, C.}, and \bibinfo{author}{Fatourou, P.} (\bibinfo{year}{2020}). \bibinfo{title}{Hybrid data driven and rule based sentiment analysis on greek text}.
\newblock \bibinfo{journal}{Procedia Computer Science} \emph{\bibinfo{volume}{178}}, \bibinfo{pages}{234--243}. \DOIprefix\doi{https://doi.org/10.1016/j.procs.2020.11.025}.
\newblock \bibinfo{note}{9th International Young Scientists Conference in Computational Science, YSC2020, 05-12 September 2020}.
\bibitem[{Medrouk and Pappa(2018)}]{Medrouk2018}
\bibinfo{author}{Medrouk, L.}, and \bibinfo{author}{Pappa, A.} (\bibinfo{year}{2018}). \bibinfo{title}{Do deep networks really need complex modules for multilingual sentiment polarity detection and domain classification?}
\newblock In \bibinfo{booktitle}{2018 International Joint Conference on Neural Networks (IJCNN)}. pp. \bibinfo{pages}{1--6}.
\newblock \DOIprefix\doi{10.1109/IJCNN.2018.8489613}.
\bibitem[{Manias et~al.(2020)Manias, Mavrogiorgou, Kiourtis and Kyriazis}]{Manias2020}
\bibinfo{author}{Manias, G.}, \bibinfo{author}{Mavrogiorgou, A.}, \bibinfo{author}{Kiourtis, A.}, and \bibinfo{author}{Kyriazis, D.} (\bibinfo{year}{2020}). \bibinfo{title}{An evaluation of neural machine translation and pre-trained word embeddings in multilingual neural sentiment analysis}.
\newblock In \bibinfo{booktitle}{2020 IEEE International Conference on Progress in Informatics and Computing (PIC)}. \bibinfo{organization}{IEEE} pp. \bibinfo{pages}{274--283}.
\newblock \DOIprefix\doi{10.1109/PIC50277.2020.9350849}.
\bibitem[{Utathya(2018)}]{imdb_review_kaggle}
\bibinfo{author}{Utathya} (\bibinfo{year}{2018}).
\newblock \bibinfo{title}{Imdb review dataset}. \bibinfo{publisher}{Kaggle}.
\newblock \bibinfo{note}{\url{https://www.kaggle.com/utathya/imdb-review-dataset}}.
\bibitem[{Markopoulos et~al.(2015)Markopoulos, Mikros, Iliadi and Liontos}]{markopoulos2015}
\bibinfo{author}{Markopoulos, G.}, \bibinfo{author}{Mikros, G.}, \bibinfo{author}{Iliadi, A.}, and \bibinfo{author}{Liontos, M.} (\bibinfo{year}{2015}). \bibinfo{title}{Sentiment analysis of hotel reviews in greek: A comparison of unigram features}.
\newblock In \bibinfo{editor}{ V.{ }Katsoni}, ed. \bibinfo{booktitle}{Cultural Tourism in a Digital Era}. \bibinfo{address}{Cham}: \bibinfo{publisher}{Springer International Publishing}.
\newblock ISBN \bibinfo{isbn}{978-3-319-15859-4} pp. \bibinfo{pages}{373--383}.
\bibitem[{Spatiotis et~al.(2016)Spatiotis, Mporas, Paraskevas and Perikos}]{Spatiotis2016}
\bibinfo{author}{Spatiotis, N.}, \bibinfo{author}{Mporas, I.}, \bibinfo{author}{Paraskevas, M.}, and \bibinfo{author}{Perikos, I.} (\bibinfo{year}{2016}). \bibinfo{title}{Sentiment analysis for the greek language}.
\newblock In \bibinfo{booktitle}{Proceedings of the 20th Pan-Hellenic Conference on Informatics}. PCI '16. \bibinfo{address}{New York, NY, USA}: \bibinfo{publisher}{Association for Computing Machinery}.
\newblock ISBN \bibinfo{isbn}{9781450347891} pp. \bibinfo{pages}{1--4}.
\bibitem[{Spatiotis et~al.(2017)Spatiotis, Paraskevas, Perikos and Mporas}]{spatiotis2017examining}
\bibinfo{author}{Spatiotis, N.}, \bibinfo{author}{Paraskevas, M.}, \bibinfo{author}{Perikos, I.}, and \bibinfo{author}{Mporas, I.} (\bibinfo{year}{2017}). \bibinfo{title}{Examining the impact of feature selection on sentiment analysis for the greek language}.
\newblock In \bibinfo{booktitle}{Speech and Computer: 19th International Conference, SPECOM 2017, Hatfield, UK, September 12-16, 2017, Proceedings 19}. \bibinfo{organization}{Springer} pp. \bibinfo{pages}{353--361}.
\newblock \DOIprefix\doi{https://doi.org/10.1007/978-3-319-66429-3_34}.
\bibitem[{Spatiotis et~al.(2019)Spatiotis, Perikos, Mporas and Paraskevas}]{Spatiotis2019}
\bibinfo{author}{Spatiotis, N.}, \bibinfo{author}{Perikos, I.}, \bibinfo{author}{Mporas, I.}, and \bibinfo{author}{Paraskevas, M.} (\bibinfo{year}{2019}). \bibinfo{title}{Examining the impact of discretization technique on sentiment analysis for the greek language}.
\newblock In \bibinfo{booktitle}{2019 10th International Conference on Information, Intelligence, Systems and Applications (IISA)}. pp. \bibinfo{pages}{1--6}.
\newblock \DOIprefix\doi{10.1109/IISA.2019.8900699}.
\bibitem[{Beleveslis et~al.(2019)Beleveslis, Tjortjis, Psaradelis and Nikoglou}]{Beleveslis2019}
\bibinfo{author}{Beleveslis, D.}, \bibinfo{author}{Tjortjis, C.}, \bibinfo{author}{Psaradelis, D.}, and \bibinfo{author}{Nikoglou, D.} (\bibinfo{year}{2019}). \bibinfo{title}{A hybrid method for sentiment analysis of election related tweets}.
\newblock In \bibinfo{booktitle}{2019 4th South-East Europe Design Automation, Computer Engineering, Computer Networks and Social Media Conference (SEEDA-CECNSM)}. pp. \bibinfo{pages}{1--6}.
\newblock \DOIprefix\doi{10.1109/SEEDA-CECNSM.2019.8908289}.
\bibitem[{Spatiotis et~al.(2020)Spatiotis, Perikos, Mporas and Paraskevas}]{spatiotis2020}
\bibinfo{author}{Spatiotis, N.}, \bibinfo{author}{Perikos, I.}, \bibinfo{author}{Mporas, I.}, and \bibinfo{author}{Paraskevas, M.} (\bibinfo{year}{2020}). \bibinfo{title}{Sentiment analysis of teachers using social information in educational platform environments}.
\newblock \bibinfo{journal}{International Journal on Artificial Intelligence Tools} \emph{\bibinfo{volume}{29}}, \bibinfo{pages}{1--29}. \DOIprefix\doi{https://doi.org/10.1142/S0218213020400047}.
\bibitem[{Giatsoglou et~al.(2017)Giatsoglou, Vozalis, Diamantaras, Vakali, Sarigiannidis and Chatzisavvas}]{GIATSOGLOU2017}
\bibinfo{author}{Giatsoglou, M.}, \bibinfo{author}{Vozalis, M.G.}, \bibinfo{author}{Diamantaras, K.}, \bibinfo{author}{Vakali, A.}, \bibinfo{author}{Sarigiannidis, G.}, and \bibinfo{author}{Chatzisavvas, K.C.} (\bibinfo{year}{2017}). \bibinfo{title}{Sentiment analysis leveraging emotions and word embeddings}.
\newblock \bibinfo{journal}{Expert Systems with Applications} \emph{\bibinfo{volume}{69}}, \bibinfo{pages}{214--224}. \DOIprefix\doi{https://doi.org/10.1016/j.eswa.2016.10.043}.
\bibitem[{Patsiouras et~al.(2023{\natexlab{a}})Patsiouras, Koroni, Mademlis and Pitas}]{patsiouras2023greekpolitics}
\bibinfo{author}{Patsiouras, E.}, \bibinfo{author}{Koroni, I.}, \bibinfo{author}{Mademlis, I.}, and \bibinfo{author}{Pitas, I.} (\bibinfo{year}{2023}{\natexlab{a}}). \bibinfo{title}{Greekpolitics: Sentiment analysis on greek politically charged tweets}.
\newblock In \bibinfo{booktitle}{2023 31st European Signal Processing Conference (EUSIPCO)}. \bibinfo{organization}{IEEE} pp. \bibinfo{pages}{1320--1324}.
\newblock \DOIprefix\doi{https://doi.org/10.23919/EUSIPCO58844.2023.10289909}.
\bibitem[{Katika et~al.(2023)Katika, Zoulias, Koufi and Malamateniou}]{katika2023mining}
\bibinfo{author}{Katika, A.}, \bibinfo{author}{Zoulias, E.}, \bibinfo{author}{Koufi, V.}, and \bibinfo{author}{Malamateniou, F.} (\bibinfo{year}{2023}). \bibinfo{title}{Mining greek tweets on long covid using sentiment analysis and topic modeling}.
\newblock In \bibinfo{booktitle}{Healthcare Transformation with Informatics and Artificial Intelligence} pp. \bibinfo{pages}{545--548}.. \bibinfo{publisher}{IOS Press} pp. \bibinfo{pages}{545--548}.
\newblock \DOIprefix\doi{10.3233/SHTI230554}.
\bibitem[{Konstantinidis(2023)}]{gpt2_greek}
\bibinfo{author}{Konstantinidis, N.} (\bibinfo{year}{2023}).
\newblock \bibinfo{title}{Gpt-2 model for greek}. \bibinfo{publisher}{Hugging Face}.
\newblock \bibinfo{note}{\url{https://huggingface.co/nikokons/gpt2-greek}}.
\bibitem[{Drakopoulos et~al.(2020)Drakopoulos, Giannoukou, Mylonas and Sioutas}]{Drakopoulos2020}
\bibinfo{author}{Drakopoulos, G.}, \bibinfo{author}{Giannoukou, I.}, \bibinfo{author}{Mylonas, P.}, and \bibinfo{author}{Sioutas, S.} (\bibinfo{year}{2020}). \bibinfo{title}{A graph neural network for assessing the affective coherence of twitter graphs}.
\newblock In \bibinfo{booktitle}{2020 IEEE International Conference on Big Data (Big Data)}. \bibinfo{organization}{IEEE} pp. \bibinfo{pages}{3618--3627}.
\newblock \DOIprefix\doi{10.1109/BigData50022.2020.9378492}.
\bibitem[{Charalampakis et~al.(2016)Charalampakis, Spathis, Kouslis and Kermanidis}]{Charalampakis2016}
\bibinfo{author}{Charalampakis, B.}, \bibinfo{author}{Spathis, D.}, \bibinfo{author}{Kouslis, E.}, and \bibinfo{author}{Kermanidis, K.} (\bibinfo{year}{2016}). \bibinfo{title}{A comparison between semi-supervised and supervised text mining techniques on detecting irony in greek political tweets}.
\newblock \bibinfo{journal}{Engineering Applications of Artificial Intelligence} \emph{\bibinfo{volume}{51}}, \bibinfo{pages}{50--57}. \DOIprefix\doi{https://doi.org/10.1016/j.engappai.2016.01.007}.
\newblock \bibinfo{note}{Mining the Humanities: Technologies and Applications}.
\bibitem[{Charalampakis et~al.(2015{\natexlab{a}})Charalampakis, Spathis, Kouslis and Kermanidis}]{Charalampakis2015}
\bibinfo{author}{Charalampakis, B.}, \bibinfo{author}{Spathis, D.}, \bibinfo{author}{Kouslis, E.}, and \bibinfo{author}{Kermanidis, K.} (\bibinfo{year}{2015}{\natexlab{a}}). \bibinfo{title}{Detecting irony on greek political tweets: A text mining approach}.
\newblock In \bibinfo{booktitle}{Proceedings of the 16th International Conference on Engineering Applications of Neural Networks (INNS)}. EANN '15. \bibinfo{address}{New York, NY, USA}: \bibinfo{publisher}{Association for Computing Machinery}{\natexlab{a}}.
\newblock ISBN \bibinfo{isbn}{9781450335805} pp. \bibinfo{pages}{1--5}.
\bibitem[{Solakidis et~al.(2014)Solakidis, Vavliakis and Mitkas}]{Solakidis2014}
\bibinfo{author}{Solakidis, G.S.}, \bibinfo{author}{Vavliakis, K.N.}, and \bibinfo{author}{Mitkas, P.A.} (\bibinfo{year}{2014}). \bibinfo{title}{Multilingual sentiment analysis using emoticons and keywords}.
\newblock In \bibinfo{booktitle}{2014 IEEE/WIC/ACM International Joint Conferences on Web Intelligence (WI) and Intelligent Agent Technologies (IAT)} vol.~\bibinfo{volume}{2}. \bibinfo{organization}{IEEE} pp. \bibinfo{pages}{102--109}.
\newblock \DOIprefix\doi{10.1109/WI-IAT.2014.86}.
\bibitem[{Chatzakou et~al.(2017)Chatzakou, Vakali and Kafetsios}]{Chatzakou2017}
\bibinfo{author}{Chatzakou, D.}, \bibinfo{author}{Vakali, A.}, and \bibinfo{author}{Kafetsios, K.} (\bibinfo{year}{2017}). \bibinfo{title}{Detecting variation of emotions in online activities}.
\newblock \bibinfo{journal}{Expert Systems with Applications} \emph{\bibinfo{volume}{89}}, \bibinfo{pages}{318--332}. \DOIprefix\doi{https://doi.org/10.1016/j.eswa.2017.07.044}.
\bibitem[{Kydros et~al.(2021)Kydros, Argyropoulou and Vrana}]{Kydros2021}
\bibinfo{author}{Kydros, D.}, \bibinfo{author}{Argyropoulou, M.}, and \bibinfo{author}{Vrana, V.} (\bibinfo{year}{2021}). \bibinfo{title}{A content and sentiment analysis of greek tweets during the pandemic}.
\newblock \bibinfo{journal}{Sustainability} \emph{\bibinfo{volume}{13}}, \bibinfo{pages}{6150}. \DOIprefix\doi{10.3390/su13116150}.
\bibitem[{Antonakaki et~al.(2017{\natexlab{a}})Antonakaki, Spiliotopoulos, V.~Samaras, Pratikakis, Ioannidis and Fragopoulou}]{antonakaki2017}
\bibinfo{author}{Antonakaki, D.}, \bibinfo{author}{Spiliotopoulos, D.}, \bibinfo{author}{V.~Samaras, C.}, \bibinfo{author}{Pratikakis, P.}, \bibinfo{author}{Ioannidis, S.}, and \bibinfo{author}{Fragopoulou, P.} (\bibinfo{year}{2017}{\natexlab{a}}). \bibinfo{title}{Social media analysis during political turbulence}.
\newblock \bibinfo{journal}{PloS one} \emph{\bibinfo{volume}{12}}, \bibinfo{pages}{e0186836}. \DOIprefix\doi{https://doi.org/10.1371/journal.pone.0186836}.
\bibitem[{Antonakaki et~al.(2016)Antonakaki, Spiliotopoulos, Samaras, Ioannidis and Fragopoulou}]{Antonakaki2016}
\bibinfo{author}{Antonakaki, D.}, \bibinfo{author}{Spiliotopoulos, D.}, \bibinfo{author}{Samaras, C.V.}, \bibinfo{author}{Ioannidis, S.}, and \bibinfo{author}{Fragopoulou, P.} (\bibinfo{year}{2016}). \bibinfo{title}{Investigating the complete corpus of referendum and elections tweets}.
\newblock In \bibinfo{booktitle}{2016 IEEE/ACM International Conference on Advances in Social Networks Analysis and Mining (ASONAM)}. \bibinfo{organization}{IEEE} pp. \bibinfo{pages}{100--105}.
\newblock \DOIprefix\doi{10.1109/ASONAM.2016.7752220}.
\bibitem[{Antonakaki et~al.(2017{\natexlab{b}})Antonakaki, Spiliotopoulos, V.~Samaras, Pratikakis, Ioannidis and Fragopoulou}]{antonakaki_elections_study_2017}
\bibinfo{author}{Antonakaki, D.}, \bibinfo{author}{Spiliotopoulos, D.}, \bibinfo{author}{V.~Samaras, C.}, \bibinfo{author}{Pratikakis, P.}, \bibinfo{author}{Ioannidis, S.}, and \bibinfo{author}{Fragopoulou, P.} (\bibinfo{year}{2017}{\natexlab{b}}).
\newblock \bibinfo{title}{Elections study: Release 0.1}. \bibinfo{publisher}{Zenodo}{\natexlab{b}}.
\newblock \bibinfo{note}{\url{https://doi.org/10.5281/zenodo.820555}}.
\bibitem[{Petasis et~al.(2014)Petasis, Spiliotopoulos, Tsirakis and Tsantilas}]{petasis2014sentiment}
\bibinfo{author}{Petasis, G.}, \bibinfo{author}{Spiliotopoulos, D.}, \bibinfo{author}{Tsirakis, N.}, and \bibinfo{author}{Tsantilas, P.} (\bibinfo{year}{2014}). \bibinfo{title}{Sentiment analysis for reputation management: Mining the greek web}.
\newblock In \bibinfo{booktitle}{Artificial Intelligence: Methods and Applications: 8th Hellenic Conference on AI, SETN 2014, Ioannina, Greece, May 15-17, 2014. Proceedings 8}. \bibinfo{organization}{Springer} pp. \bibinfo{pages}{327--340}.
\newblock \DOIprefix\doi{https://doi.org/10.1007/978-3-319-07064-3_26}.
\bibitem[{Tsakalidis et~al.(2018{\natexlab{b}})Tsakalidis, Aletras, Cristea and Liakata}]{Tsakalidis2018}
\bibinfo{author}{Tsakalidis, A.}, \bibinfo{author}{Aletras, N.}, \bibinfo{author}{Cristea, A.I.}, and \bibinfo{author}{Liakata, M.} (\bibinfo{year}{2018}{\natexlab{b}}). \bibinfo{title}{Nowcasting the stance of social media users in a sudden vote: The case of the greek referendum}.
\newblock In \bibinfo{booktitle}{Proceedings of the 27th ACM International Conference on Information and Knowledge Management}. CIKM '18. \bibinfo{address}{New York, NY, USA}: \bibinfo{publisher}{Association for Computing Machinery}{\natexlab{b}}.
\newblock ISBN \bibinfo{isbn}{9781450360142} pp. \bibinfo{pages}{367--376}.
\newblock \DOIprefix\doi{10.1145/3269206.3271783}.
\bibitem[{Sliwa et~al.(2018)Sliwa, Ma, Liu, Borad, Ziyaei, Ghobadi, Sabbah and Aker}]{sliwa-etal-2018-multi}
\bibinfo{author}{Sliwa, A.}, \bibinfo{author}{Ma, Y.}, \bibinfo{author}{Liu, R.}, \bibinfo{author}{Borad, N.}, \bibinfo{author}{Ziyaei, S.}, \bibinfo{author}{Ghobadi, M.}, \bibinfo{author}{Sabbah, F.}, and \bibinfo{author}{Aker, A.} (\bibinfo{year}{2018}). \bibinfo{title}{Multi-lingual argumentative corpora in {E}nglish, {T}urkish, {G}reek, {A}lbanian, {C}roatian, {S}erbian, {M}acedonian, {B}ulgarian, {R}omanian and {A}rabic}.
\newblock In \bibinfo{booktitle}{Proceedings of the Eleventh International Conference on Language Resources and Evaluation ({LREC} 2018)}. \bibinfo{address}{Miyazaki, Japan}: \bibinfo{publisher}{European Language Resources Association (ELRA)}.
\bibitem[{Sardianos et~al.(2015)Sardianos, Katakis, Petasis and Karkaletsis}]{sardianos2015argument}
\bibinfo{author}{Sardianos, C.}, \bibinfo{author}{Katakis, I.M.}, \bibinfo{author}{Petasis, G.}, and \bibinfo{author}{Karkaletsis, V.} (\bibinfo{year}{2015}). \bibinfo{title}{Argument extraction from news}.
\newblock In \bibinfo{booktitle}{Proceedings of the 2nd Workshop on Argumentation Mining}. pp. \bibinfo{pages}{56--66}.
\bibitem[{Lafferty et~al.(2001)Lafferty, McCallum and Pereira}]{lafferty2001conditional}
\bibinfo{author}{Lafferty, J.D.}, \bibinfo{author}{McCallum, A.}, and \bibinfo{author}{Pereira, F.C.} (\bibinfo{year}{2001}). \bibinfo{title}{Conditional random fields: Probabilistic models for segmenting and labeling sequence data}.
\newblock In \bibinfo{booktitle}{Proceedings of the Eighteenth International Conference on Machine Learning}. pp. \bibinfo{pages}{282--289}.
\bibitem[{Chen and Skiena(2014)}]{chen2014building}
\bibinfo{author}{Chen, Y.}, and \bibinfo{author}{Skiena, S.} (\bibinfo{year}{2014}). \bibinfo{title}{Building sentiment lexicons for all major languages}.
\newblock In \bibinfo{booktitle}{Proceedings of the 52nd Annual Meeting of the Association for Computational Linguistics (Volume 2: Short Papers)}. pp. \bibinfo{pages}{383--389}.
\newblock \DOIprefix\doi{10.3115/v1/P14-2063}.
\bibitem[{Patsiouras et~al.(2023{\natexlab{b}})Patsiouras, Koroni, Mademlis and Pitas}]{auth_greekpolitics}
\bibinfo{author}{Patsiouras, E.}, \bibinfo{author}{Koroni, I.}, \bibinfo{author}{Mademlis, I.}, and \bibinfo{author}{Pitas, I.} (\bibinfo{year}{2023}{\natexlab{b}}).
\newblock \bibinfo{title}{Auth greek politics dataset}. \bibinfo{publisher}{Artificial Intelligence and Information Analysis Group, Aristotle University of Thessaloniki}{\natexlab{b}}.
\newblock \bibinfo{note}{\url{https://aiia.csd.auth.gr/auth-greekpolitics-dataset}}.
\bibitem[{Bilianos(2020)}]{greek_sentiment_analysis}
\bibinfo{author}{Bilianos, D.} (\bibinfo{year}{2020}).
\newblock \bibinfo{title}{Greek sentiment analysis}. \bibinfo{publisher}{GitHub}.
\newblock \bibinfo{note}{\url{https://github.com/DimitrisBil/greek-sentiment-analysis}}.
\bibitem[{Chatzakou et~al.(2016)Chatzakou, Vakali and Kafetsios}]{chatzakou2016dataset}
\bibinfo{author}{Chatzakou, D.}, \bibinfo{author}{Vakali, A.}, and \bibinfo{author}{Kafetsios, K.} (\bibinfo{year}{2016}).
\newblock \bibinfo{title}{Annotated twitter dataset for emotion detection}. \bibinfo{publisher}{Dropbox}.
\newblock \bibinfo{note}{\url{http://bit.ly/2bLgVUP}}.
\bibitem[{Antonakaki et~al.(2017{\natexlab{c}})Antonakaki, Spiliotopoulos, V.~Samaras, Pratikakis, Ioannidis and Fragopoulou}]{social_media_turbulence}
\bibinfo{author}{Antonakaki, D.}, \bibinfo{author}{Spiliotopoulos, D.}, \bibinfo{author}{V.~Samaras, C.}, \bibinfo{author}{Pratikakis, P.}, \bibinfo{author}{Ioannidis, S.}, and \bibinfo{author}{Fragopoulou, P.} (\bibinfo{year}{2017}{\natexlab{c}}).
\newblock \bibinfo{title}{Social media analysis during political turbulence}. \bibinfo{publisher}{Figshare}{\natexlab{c}}.
\newblock \bibinfo{note}{\url{https://figshare.com/articles/dataset/Social_media_analysis_during_political_turbulence_DATA/5492443/1}}.
\bibitem[{Makrynioti and Vassalos(2015)}]{Makrynioti2015}
\bibinfo{author}{Makrynioti, N.}, and \bibinfo{author}{Vassalos, V.} (\bibinfo{year}{2015}). \bibinfo{title}{Sentiment extraction from tweets: Multilingual challenges}.
\newblock In \bibinfo{editor}{ S.{ }Madria}, and \bibinfo{editor}{ T.{ }Hara}, eds. \bibinfo{booktitle}{Big Data Analytics and Knowledge Discovery}. \bibinfo{address}{Cham}: \bibinfo{publisher}{Springer International Publishing}.
\newblock ISBN \bibinfo{isbn}{978-3-319-22729-0} pp. \bibinfo{pages}{136--148}.
\newblock \DOIprefix\doi{https://doi.org/10.1007/978-3-319-22729-0_11}.
\bibitem[{Charalampakis et~al.(2015{\natexlab{b}})Charalampakis, Spathis, Kouslis and Kermanidis}]{websent}
\bibinfo{author}{Charalampakis, B.}, \bibinfo{author}{Spathis, D.}, \bibinfo{author}{Kouslis, E.}, and \bibinfo{author}{Kermanidis, K.} (\bibinfo{year}{2015}{\natexlab{b}}).
\newblock \bibinfo{title}{Websent: A web-based sentiment analysis tool}. \bibinfo{publisher}{Humanistic Informatics Laboratory, Ionian University}{\natexlab{b}}.
\newblock \bibinfo{note}{\url{https://di.ionio.gr/hilab/doku.php}}.
\bibitem[{{Apache Software Foundation}(2004)}]{apache_license}
\bibinfo{author}{{Apache Software Foundation}} (\bibinfo{year}{2004}).
\newblock \bibinfo{title}{Apache license, version 2.0}.
\newblock \bibinfo{note}{\url{https://www.apache.org/licenses/LICENSE-2.0.html}}.
\bibitem[{{X, Inc.}(2024)}]{twitter_tos}
\bibinfo{author}{{X, Inc.}} (\bibinfo{year}{2024}).
\newblock \bibinfo{title}{X terms of service}.
\newblock \bibinfo{note}{\url{https://x.com/en/tos}}.
\bibitem[{{Avgi SA}(????)}]{avgi}
\bibinfo{author}{{Avgi SA}}.
\newblock \bibinfo{title}{Avgi: the morning newspaper of the left}.
\newblock \bibinfo{note}{\url{https://www.avgi.gr/}}.
\bibitem[{Tsakalidis et~al.(2017)Tsakalidis, Papadopoulos and Kompatsiaris}]{tsakalidis2017building}
\bibinfo{author}{Tsakalidis, A.}, \bibinfo{author}{Papadopoulos, S.}, and \bibinfo{author}{Kompatsiaris, Y.} (\bibinfo{year}{2017}).
\newblock \bibinfo{title}{Building and evaluating resources for sentiment analysis in the greek language}. \bibinfo{publisher}{Multimedia Knowledge and Social Media Analytics Laboratory, CERTH}.
\newblock \bibinfo{note}{\url{http://mklab.iti.gr/resources/tsakalidis2017building.zip}}.
\bibitem[{{mSensis}(????)}]{msensis_demon}
\bibinfo{author}{{mSensis}}.
\newblock \bibinfo{title}{Demon: Deep emotion understanding}.
\newblock \bibinfo{note}{\url{https://msensis.com/research-and-development/demon}}.
\bibitem[{Palogiannidi et~al.(2016{\natexlab{b}})Palogiannidi, Koutsakis, Iosif and Potamianos}]{greek_affective_lexicon}
\bibinfo{author}{Palogiannidi, E.}, \bibinfo{author}{Koutsakis, P.}, \bibinfo{author}{Iosif, E.}, and \bibinfo{author}{Potamianos, A.} (\bibinfo{year}{2016}{\natexlab{b}}).
\newblock \bibinfo{title}{Greek affective lexicon automatically created}. \bibinfo{publisher}{Technical University of Crete}{\natexlab{b}}.
\newblock \bibinfo{note}{\url{https://www.telecom.tuc.gr/~epalogiannidi/docs/resources/greek_affective_lexicon_automatically_created.zip}}.
\bibitem[{{Manolis Triantafyllidis Foundation}(????)}]{triantafyllides_dictionary}
\bibinfo{author}{{Manolis Triantafyllidis Foundation}}.
\newblock \bibinfo{title}{Dictionary of standard modern greek}. \bibinfo{publisher}{Institute for Modern Greek Studies, Aristotle University of Thessaloniki}.
\newblock \bibinfo{note}{\url{https://www.greek-language.gr/greekLang/modern_greek/tools/lexica/triantafyllides/}}.
\bibitem[{Tsakalidis et~al.(2014)Tsakalidis, Papadopoulos and Kompatsiaris}]{tsakalidis2014ensemble}
\bibinfo{author}{Tsakalidis, A.}, \bibinfo{author}{Papadopoulos, S.}, and \bibinfo{author}{Kompatsiaris, I.} (\bibinfo{year}{2014}). \bibinfo{title}{An ensemble model for cross-domain polarity classification on twitter}.
\newblock In \bibinfo{booktitle}{Web Information Systems Engineering--WISE 2014: 15th International Conference, Thessaloniki, Greece, October 12-14, 2014, Proceedings, Part II 15}. \bibinfo{organization}{Springer} pp. \bibinfo{pages}{168--177}.
\newblock \DOIprefix\doi{https://doi.org/10.1007/978-3-319-11746-1_12}.
\bibitem[{Bradley and Lang(1999)}]{bradley1999affective}
\bibinfo{author}{Bradley, M.M.}, and \bibinfo{author}{Lang, P.J.}
\newblock \bibinfo{title}{Affective norms for english words (anew): Instruction manual and affective ratings}.
\newblock \bibinfo{type}{Tech. Rep.} Technical report C-1, the center for research in psychophysiology~… (\bibinfo{year}{1999}).
\bibitem[{Wankhade et~al.(2022)Wankhade, Rao and Kulkarni}]{wankhade2022survey}
\bibinfo{author}{Wankhade, M.}, \bibinfo{author}{Rao, A.C.S.}, and \bibinfo{author}{Kulkarni, C.} (\bibinfo{year}{2022}). \bibinfo{title}{A survey on sentiment analysis methods, applications, and challenges}.
\newblock \bibinfo{journal}{Artificial Intelligence Review} \emph{\bibinfo{volume}{55}}, \bibinfo{pages}{5731--5780}. \DOIprefix\doi{https://doi.org/10.1007/s10462-022-10144-1}.
\bibitem[{El and Kassou(2014)}]{el2014authorship}
\bibinfo{author}{El, S.E.M.}, and \bibinfo{author}{Kassou, I.} (\bibinfo{year}{2014}). \bibinfo{title}{Authorship analysis studies: A survey}.
\newblock \bibinfo{journal}{International Journal of Computer Applications} \emph{\bibinfo{volume}{86}}, \bibinfo{pages}{22--29}. \DOIprefix\doi{10.5120/15038-3384}.
\bibitem[{Kestemont et~al.(2018)Kestemont, Tschuggnall, Stamatatos, Daelemans, Specht, Stein and Potthast}]{kestemont2018overview}
\bibinfo{author}{Kestemont, M.}, \bibinfo{author}{Tschuggnall, M.}, \bibinfo{author}{Stamatatos, E.}, \bibinfo{author}{Daelemans, W.}, \bibinfo{author}{Specht, G.}, \bibinfo{author}{Stein, B.}, and \bibinfo{author}{Potthast, M.} (\bibinfo{year}{2018}). \bibinfo{title}{Overview of the author identification task at pan-2018: cross-domain authorship attribution and style change detection}.
\newblock In \bibinfo{booktitle}{Working Notes Papers of the CLEF 2018 Evaluation Labs. Avignon, France, September 10-14, 2018/Cappellato, Linda [edit.]; et al.} pp. \bibinfo{pages}{1--25}.
\bibitem[{Rosso et~al.(2016)Rosso, Rangel, Potthast, Stamatatos, Tschuggnall and Stein}]{rosso-2016-overview}
\bibinfo{author}{Rosso, P.}, \bibinfo{author}{Rangel, F.}, \bibinfo{author}{Potthast, M.}, \bibinfo{author}{Stamatatos, E.}, \bibinfo{author}{Tschuggnall, M.}, and \bibinfo{author}{Stein, B.} (\bibinfo{year}{2016}). \bibinfo{title}{Overview of pan'16}.
\newblock In \bibinfo{editor}{ N.{ }Fuhr}, \bibinfo{editor}{ P.{ }Quaresma}, \bibinfo{editor}{ T.{ }Gon{\c{c}}alves}, \bibinfo{editor}{ B.{ }Larsen}, \bibinfo{editor}{ K.{ }Balog}, \bibinfo{editor}{ C.{ }Macdonald}, \bibinfo{editor}{ L.{ }Cappellato}, and \bibinfo{editor}{ N.{ }Ferro}, eds. \bibinfo{booktitle}{Experimental IR Meets Multilinguality, Multimodality, and Interaction}. \bibinfo{address}{Cham}: \bibinfo{publisher}{Springer International Publishing}.
\newblock ISBN \bibinfo{isbn}{978-3-319-44564-9} pp. \bibinfo{pages}{332--350}.
\newblock \DOIprefix\doi{https://doi.org/10.1007/978-3-319-44564-9_28}.
\bibitem[{{Webis Group}(????)}]{pan_webis}
\bibinfo{author}{{Webis Group}}.
\newblock \bibinfo{title}{Pan: Digital text forensics and stylometry}.
\newblock \bibinfo{note}{\url{https://pan.webis.de/}}.
\bibitem[{Bevendorff et~al.(2023)Bevendorff, Chinea-R{\'\i}os, Franco-Salvador, Heini, K{\"o}rner, Kredens, Mayerl, P{\k{e}}zik, Potthast, Rangel et~al.}]{bevendorff2023overview}
\bibinfo{author}{Bevendorff, J.}, \bibinfo{author}{Chinea-R{\'\i}os, M.}, \bibinfo{author}{Franco-Salvador, M.}, \bibinfo{author}{Heini, A.}, \bibinfo{author}{K{\"o}rner, E.}, \bibinfo{author}{Kredens, K.}, \bibinfo{author}{Mayerl, M.}, \bibinfo{author}{P{\k{e}}zik, P.}, \bibinfo{author}{Potthast, M.}, \bibinfo{author}{Rangel, F.} et~al. (\bibinfo{year}{2023}). \bibinfo{title}{Overview of pan 2023: Authorship verification, multi-author writing style analysis, profiling cryptocurrency influencers, and trigger detection}.
\newblock In \bibinfo{booktitle}{European Conference on Information Retrieval}. \bibinfo{organization}{Springer} pp. \bibinfo{pages}{518--526}.
\bibitem[{Juola and Stamatatos(2013{\natexlab{a}})}]{juola2013overview}
\bibinfo{author}{Juola, P.}, and \bibinfo{author}{Stamatatos, E.} (\bibinfo{year}{2013}{\natexlab{a}}). \bibinfo{title}{Overview of the author identification task at pan 2013}.
\newblock In \bibinfo{editor}{ P.{ }Forner}, \bibinfo{editor}{ R.{ }Navigli}, \bibinfo{editor}{ D.{ }Tufis}, and \bibinfo{editor}{ N.{ }Ferro}, eds. \bibinfo{booktitle}{Working Notes for CLEF 2013 Conference} vol. \bibinfo{volume}{1179}. \bibinfo{publisher}{CEUR-WS.org}{\natexlab{a}} pp. \bibinfo{pages}{1--20}.
\newblock \URLprefix \url{https://ceur-ws.org/Vol-1179/CLEF2013wn-PAN-JuolaEt2013.pdf}.
\bibitem[{Stamatatos et~al.(2014{\natexlab{a}})Stamatatos, Daelemans, Verhoeven, Potthast, Stein, Juola, Sanchez-Perez, Barr{\'o}n-Cede{\~n}o et~al.}]{stamatatos2014overview}
\bibinfo{author}{Stamatatos, E.}, \bibinfo{author}{Daelemans, W.}, \bibinfo{author}{Verhoeven, B.}, \bibinfo{author}{Potthast, M.}, \bibinfo{author}{Stein, B.}, \bibinfo{author}{Juola, P.}, \bibinfo{author}{Sanchez-Perez, M.A.}, \bibinfo{author}{Barr{\'o}n-Cede{\~n}o, A.} et~al. (\bibinfo{year}{2014}{\natexlab{a}}). \bibinfo{title}{Overview of the author identification task at pan 2014}.
\newblock In \bibinfo{booktitle}{CEUR Workshop Proceedings} vol. \bibinfo{volume}{1180}. \bibinfo{organization}{CEUR-WS} pp. \bibinfo{pages}{877--897}.
\bibitem[{Stamatatos et~al.(2015{\natexlab{a}})Stamatatos, Daelemans, Verhoeven, Juola, {L{\'o}pez L{\'o}pez}, Potthast and Stein}]{stamatatos2015b}
\bibinfo{author}{Stamatatos, E.}, \bibinfo{author}{Daelemans, W.}, \bibinfo{author}{Verhoeven, B.}, \bibinfo{author}{Juola, P.}, \bibinfo{author}{{L{\'o}pez L{\'o}pez}, A.}, \bibinfo{author}{Potthast, M.}, and \bibinfo{author}{Stein, B.} (\bibinfo{year}{2015}{\natexlab{a}}). \bibinfo{title}{{Overview of the Author Identification Task at PAN 2015}}.
\newblock In \bibinfo{editor}{ L.{ }Cappellato}, \bibinfo{editor}{ N.{ }Ferro}, \bibinfo{editor}{ G.{ }Jones}, and \bibinfo{editor}{ E.{ }{San Juan}}, eds. \bibinfo{booktitle}{Working Notes Papers of the CLEF 2015 Evaluation Labs} vol. \bibinfo{volume}{1391} of \emph{\bibinfo{series}{Lecture Notes in Computer Science}}. pp. \bibinfo{pages}{877--897}.
\bibitem[{Stamatatos et~al.(2016{\natexlab{a}})Stamatatos, Tschuggnall, Verhoeven, Daelemans, Specht, Stein and Potthast}]{stamatatos:2016}
\bibinfo{author}{Stamatatos, E.}, \bibinfo{author}{Tschuggnall, M.}, \bibinfo{author}{Verhoeven, B.}, \bibinfo{author}{Daelemans, W.}, \bibinfo{author}{Specht, G.}, \bibinfo{author}{Stein, B.}, and \bibinfo{author}{Potthast, M.} (\bibinfo{year}{2016}{\natexlab{a}}). \bibinfo{title}{{Clustering by Authorship Within and Across Documents}}.
\newblock In \bibinfo{editor}{ K.{ }Balog}, \bibinfo{editor}{ L.{ }Cappellato}, \bibinfo{editor}{ N.{ }Ferro}, and \bibinfo{editor}{ C.{ }Macdonald}, eds. \bibinfo{booktitle}{Working Notes Papers of the CLEF 2016 Evaluation Labs} vol. \bibinfo{volume}{1609} of \emph{\bibinfo{series}{Lecture Notes in Computer Science}}. pp. \bibinfo{pages}{691--715}.
\bibitem[{Juola et~al.(2019)Juola, Mikros and Vinsick}]{juola2019comparative}
\bibinfo{author}{Juola, P.}, \bibinfo{author}{Mikros, G.K.}, and \bibinfo{author}{Vinsick, S.} (\bibinfo{year}{2019}). \bibinfo{title}{A comparative assessment of the difficulty of authorship attribution in greek and in english}.
\newblock \bibinfo{journal}{Journal of the Association for Information Science and Technology} \emph{\bibinfo{volume}{70}}, \bibinfo{pages}{61--70}. \DOIprefix\doi{https://doi.org/10.1002/asi.24073}.
\bibitem[{Juola(2020)}]{jgaap}
\bibinfo{author}{Juola, P.} (\bibinfo{year}{2020}).
\newblock \bibinfo{title}{Jgaap: Java graphical authorship attribution program}. \bibinfo{publisher}{GitHub}.
\newblock \bibinfo{note}{\url{https://github.com/evllabs/JGAAP}}.
\bibitem[{Kocher and Savoy(2017)}]{kocher2017simple}
\bibinfo{author}{Kocher, M.}, and \bibinfo{author}{Savoy, J.} (\bibinfo{year}{2017}). \bibinfo{title}{A simple and efficient algorithm for authorship verification}.
\newblock \bibinfo{journal}{Journal of the Association for Information Science and Technology} \emph{\bibinfo{volume}{68}}, \bibinfo{pages}{259--269}. \DOIprefix\doi{https://doi.org/10.1002/asi.23648}.
\bibitem[{H{\"u}rlimann et~al.(2015)H{\"u}rlimann, Weck, van~den Berg, Suster and Nissim}]{hurlimann2015glad}
\bibinfo{author}{H{\"u}rlimann, M.}, \bibinfo{author}{Weck, B.}, \bibinfo{author}{van~den Berg, E.}, \bibinfo{author}{Suster, S.}, and \bibinfo{author}{Nissim, M.} (\bibinfo{year}{2015}). \bibinfo{title}{Glad: Groningen lightweight authorship detection.}
\newblock In \bibinfo{booktitle}{Proceedings of CLEF 2015 Labs and Workshops, Notebook Papers, CEUR Workshop}. pp. \bibinfo{pages}{1--12}.
\bibitem[{H{\""u}rlimann et~al.(2015)H{\""u}rlimann, Weck, van~den Berg, Suster and Nissim}]{huerlimann15_glad}
\bibinfo{author}{H{\""u}rlimann, M.}, \bibinfo{author}{Weck, B.}, \bibinfo{author}{van~den Berg, E.}, \bibinfo{author}{Suster, S.}, and \bibinfo{author}{Nissim, M.} (\bibinfo{year}{2015}).
\newblock \bibinfo{title}{Glad: Groningen lightweight authorship detection}. \bibinfo{publisher}{GitHub}.
\newblock \URLprefix \url{https://github.com/pan-webis-de/huerlimann15}.
\bibitem[{Halvani et~al.(2016)Halvani, Winter and Pflug}]{halvani2016authorship}
\bibinfo{author}{Halvani, O.}, \bibinfo{author}{Winter, C.}, and \bibinfo{author}{Pflug, A.} (\bibinfo{year}{2016}). \bibinfo{title}{Authorship verification for different languages, genres and topics}.
\newblock \bibinfo{journal}{Digital Investigation} \emph{\bibinfo{volume}{16}}, \bibinfo{pages}{S33--S43}. \DOIprefix\doi{https://doi.org/10.1016/j.diin.2016.01.006}.
\bibitem[{Halvani et~al.(2020)Halvani, Graner and Regev}]{halvani2020taveer}
\bibinfo{author}{Halvani, O.}, \bibinfo{author}{Graner, L.}, and \bibinfo{author}{Regev, R.} (\bibinfo{year}{2020}). \bibinfo{title}{Taveer: an interpretable topic-agnostic authorship verification method}.
\newblock In \bibinfo{booktitle}{Proceedings of the 15th International Conference on Availability, Reliability and Security}. pp. \bibinfo{pages}{1--10}.
\bibitem[{Kestemont et~al.(2020)Kestemont, Manjavacas, Markov, Bevendorff, Wiegmann, Stamatatos, Potthast and Stein}]{kestemont2020overview}
\bibinfo{author}{Kestemont, M.}, \bibinfo{author}{Manjavacas, E.}, \bibinfo{author}{Markov, I.}, \bibinfo{author}{Bevendorff, J.}, \bibinfo{author}{Wiegmann, M.}, \bibinfo{author}{Stamatatos, E.}, \bibinfo{author}{Potthast, M.}, and \bibinfo{author}{Stein, B.} (\bibinfo{year}{2020}). \bibinfo{title}{Overview of the cross-domain authorship verification task at pan 2020}.
\newblock In \bibinfo{booktitle}{Working notes of CLEF 2020-Conference and Labs of the Evaluation Forum, 22-25 September, Thessaloniki, Greece}. pp. \bibinfo{pages}{1--14}.
\bibitem[{Mikros(2013)}]{mikros2013systematic}
\bibinfo{author}{Mikros, G.K.} (\bibinfo{year}{2013}). \bibinfo{title}{Systematic stylometric differences in men and women authors: A corpus-based study}.
\newblock In \bibinfo{editor}{ R.{ }Köhler}, and \bibinfo{editor}{ G.{ }Altmann}, eds. \bibinfo{booktitle}{Issues in Quantitative Linguistics 3} vol.~\bibinfo{volume}{13} of \emph{\bibinfo{series}{Studies in Quantitative Linguistics}} pp. \bibinfo{pages}{206--223}.. \bibinfo{publisher}{RAM-Verlag}.
\newblock ISBN \bibinfo{isbn}{978-3-942303-12-5} pp. \bibinfo{pages}{206--223}.
\newblock \URLprefix \url{https://www.isbn.de/buch/9783942303125/issues-in-quantitative-linguistics-3}.
\bibitem[{Perifanos(2015)}]{perifanos2015gender}
\bibinfo{author}{Perifanos, G.K.M.K.} (\bibinfo{year}{2015}). \bibinfo{title}{Gender identification in modern greek tweets}.
\newblock \bibinfo{journal}{Recent Contributions to Quantitative Linguistics} \emph{\bibinfo{volume}{70}}, \bibinfo{pages}{75}. \DOIprefix\doi{https://doi.org/10.1515/9783110420296-008}.
\bibitem[{Mikros(2012)}]{mikros2012authorship}
\bibinfo{author}{Mikros, G.K.} (\bibinfo{year}{2012}). \bibinfo{title}{Authorship attribution and gender identification in greek blogs}.
\newblock \bibinfo{journal}{Methods and Applications of Quantitative Linguistics} \emph{\bibinfo{volume}{21}}, \bibinfo{pages}{21--32}.
\bibitem[{Platt(1999)}]{Platt1999}
\bibinfo{author}{Platt, J.C.} (\bibinfo{year}{1999}). \bibinfo{title}{Fast training of support vector machines using sequential minimal optimization}.
\newblock In \bibinfo{editor}{ B.{ }Schölkopf}, \bibinfo{editor}{ C.J.{ }Burges}, and \bibinfo{editor}{ A.J.{ }Smola}, eds. \bibinfo{booktitle}{Advances in Kernel Methods: Support Vector Learning} pp. \bibinfo{pages}{185--208}.. \bibinfo{publisher}{MIT Press}.
\newblock ISBN \bibinfo{isbn}{0-262-19416-3} pp. \bibinfo{pages}{185--208}.
\newblock \DOIprefix\doi{https://doi.org/10.7551/mitpress/1130.003.0016}.
\bibitem[{{Alter Ego Media S.A.}(????)}]{tovima}
\bibinfo{author}{{Alter Ego Media S.A.}}
\newblock \bibinfo{title}{To vima: Greek news portal}.
\newblock \bibinfo{note}{\url{https://www.tovima.gr/}}.
\bibitem[{Stamatatos et~al.(2016{\natexlab{b}})Stamatatos, Tschuggnall, Verhoeven, Daelemans, Specht, Stein and Potthast}]{pan16_clustering}
\bibinfo{author}{Stamatatos, E.}, \bibinfo{author}{Tschuggnall, M.}, \bibinfo{author}{Verhoeven, B.}, \bibinfo{author}{Daelemans, W.}, \bibinfo{author}{Specht, G.}, \bibinfo{author}{Stein, B.}, and \bibinfo{author}{Potthast, M.} (\bibinfo{year}{2016}{\natexlab{b}}).
\newblock \bibinfo{title}{Pan16 author identification: Clustering}. \bibinfo{publisher}{Zenodo}{\natexlab{b}}.
\newblock \bibinfo{note}{\url{https://doi.org/10.5281/zenodo.3737586}}.
\bibitem[{Stamatatos et~al.(2015{\natexlab{b}})Stamatatos, Daelemans, Verhoeven, Juola, López-López, Potthast and Stein}]{pan15_verification}
\bibinfo{author}{Stamatatos, E.}, \bibinfo{author}{Daelemans, W.}, \bibinfo{author}{Verhoeven, B.}, \bibinfo{author}{Juola, P.}, \bibinfo{author}{López-López, A.}, \bibinfo{author}{Potthast, M.}, and \bibinfo{author}{Stein, B.} (\bibinfo{year}{2015}{\natexlab{b}}).
\newblock \bibinfo{title}{Pan15 author identification: Verification}. \bibinfo{publisher}{Zenodo}{\natexlab{b}}.
\newblock \bibinfo{note}{\url{https://doi.org/10.5281/zenodo.3737562}}.
\bibitem[{Stamatatos et~al.(2014{\natexlab{b}})Stamatatos, Daelemans, Verhoeven, Potthast, Stein, Juola, Sanchez-Perez and Barrón-Cedeño}]{stamatatos2014pan14}
\bibinfo{author}{Stamatatos, E.}, \bibinfo{author}{Daelemans, W.}, \bibinfo{author}{Verhoeven, B.}, \bibinfo{author}{Potthast, M.}, \bibinfo{author}{Stein, B.}, \bibinfo{author}{Juola, P.}, \bibinfo{author}{Sanchez-Perez, M.A.}, and \bibinfo{author}{Barrón-Cedeño, A.} (\bibinfo{year}{2014}{\natexlab{b}}).
\newblock \bibinfo{title}{Pan14 author identification: Verification}. \bibinfo{publisher}{Zenodo}{\natexlab{b}}.
\newblock \URLprefix \url{https://doi.org/10.5281/zenodo.3716033}. \DOIprefix\doi{10.5281/zenodo.3716033}.
\bibitem[{Juola and Stamatatos(2013{\natexlab{b}})}]{juola2013pan13}
\bibinfo{author}{Juola, P.}, and \bibinfo{author}{Stamatatos, E.} (\bibinfo{year}{2013}{\natexlab{b}}).
\newblock \bibinfo{title}{Pan13 author identification: Verification}. \bibinfo{publisher}{Zenodo}{\natexlab{b}}.
\newblock \URLprefix \url{https://doi.org/10.5281/zenodo.3715999}. \DOIprefix\doi{10.5281/zenodo.3715999}.
\bibitem[{Halvani et~al.(2015)Halvani, Winter and Pflug}]{bitly_1OjFRhJ}
\bibinfo{author}{Halvani, O.}, \bibinfo{author}{Winter, C.}, and \bibinfo{author}{Pflug, A.} (\bibinfo{year}{2015}).
\newblock \bibinfo{title}{Halvani, winter and plug dataset}.
\newblock \bibinfo{note}{\url{http://bit.ly/1OjFRhJ}}.
\bibitem[{Savoy(2020)}]{savoy2020machine}
\bibinfo{author}{Savoy, J.} (\bibinfo{year}{2020}). \bibinfo{title}{Machine Learning Methods for Stylometry: Authorship Attribution and Author Profiling}. \bibinfo{address}{Cham}: \bibinfo{publisher}{Springer}.
\newblock ISBN \bibinfo{isbn}{978-3-030-53359-5}.
\newblock \URLprefix \url{https://link.springer.com/book/10.1007/978-3-030-53360-1}. \DOIprefix\doi{10.1007/978-3-030-53360-1}.
\bibitem[{Potthast et~al.(2019)Potthast, Rosso, Stamatatos and Stein}]{potthast2019decade}
\bibinfo{author}{Potthast, M.}, \bibinfo{author}{Rosso, P.}, \bibinfo{author}{Stamatatos, E.}, and \bibinfo{author}{Stein, B.} (\bibinfo{year}{2019}). \bibinfo{title}{A decade of shared tasks in digital text forensics at pan}.
\newblock In \bibinfo{booktitle}{Advances in Information Retrieval: 41st European Conference on IR Research, ECIR 2019, Cologne, Germany, April 14--18, 2019, Proceedings, Part II 41}. \bibinfo{organization}{Springer} pp. \bibinfo{pages}{291--300}.
\bibitem[{Hovy et~al.(2017)Hovy, Spruit, Mitchell, Bender, Strube and Wallach}]{hovy2017proceedings}
\bibinfo{editor}{Hovy, D.}, \bibinfo{editor}{Spruit, S.}, \bibinfo{editor}{Mitchell, M.}, \bibinfo{editor}{Bender, E.M.}, \bibinfo{editor}{Strube, M.}, and \bibinfo{editor}{Wallach, H.}, eds.
\newblock \bibinfo{title}{Proceedings of the First {ACL} Workshop on Ethics in Natural Language Processing}. \bibinfo{address}{Valencia, Spain}: \bibinfo{publisher}{Association for Computational Linguistics} (\bibinfo{year}{2017}).
\newblock \DOIprefix\doi{10.18653/v1/W17-16}.
\bibitem[{Schmidt and Wiegand(2017)}]{schmidt2017survey}
\bibinfo{author}{Schmidt, A.}, and \bibinfo{author}{Wiegand, M.} (\bibinfo{year}{2017}). \bibinfo{title}{A survey on hate speech detection using natural language processing}.
\newblock In \bibinfo{booktitle}{Proceedings of the fifth international workshop on natural language processing for social media}. pp. \bibinfo{pages}{1--10}.
\bibitem[{D{\'\i}az et~al.(2022)D{\'\i}az, Amironesei, Weidinger and Gabriel}]{diaz2022accounting}
\bibinfo{author}{D{\'\i}az, M.}, \bibinfo{author}{Amironesei, R.}, \bibinfo{author}{Weidinger, L.}, and \bibinfo{author}{Gabriel, I.} (\bibinfo{year}{2022}). \bibinfo{title}{Accounting for offensive speech as a practice of resistance}.
\newblock In \bibinfo{booktitle}{Proceedings of the Sixth Workshop on Online Abuse and Harms (WOAH)}. pp. \bibinfo{pages}{192--202}.
\bibitem[{Welbl et~al.(2021)Welbl, Glaese, Uesato, Dathathri, Mellor, Hendricks, Anderson, Kohli, Coppin and Huang}]{welbl2021challenges}
\bibinfo{author}{Welbl, J.}, \bibinfo{author}{Glaese, A.}, \bibinfo{author}{Uesato, J.}, \bibinfo{author}{Dathathri, S.}, \bibinfo{author}{Mellor, J.}, \bibinfo{author}{Hendricks, L.A.}, \bibinfo{author}{Anderson, K.}, \bibinfo{author}{Kohli, P.}, \bibinfo{author}{Coppin, B.}, and \bibinfo{author}{Huang, P.S.} (\bibinfo{year}{2021}). \bibinfo{title}{Challenges in detoxifying language models}.
\newblock \bibinfo{journal}{Preprint at arXiv \url{https://doi.org/10.48550/arXiv.2109.07445}}.
\bibitem[{Pavlopoulos et~al.(2020)Pavlopoulos, Sorensen, Dixon, Thain and Androutsopoulos}]{pavlopoulos2020toxicity}
\bibinfo{author}{Pavlopoulos, J.}, \bibinfo{author}{Sorensen, J.}, \bibinfo{author}{Dixon, L.}, \bibinfo{author}{Thain, N.}, and \bibinfo{author}{Androutsopoulos, I.} (\bibinfo{year}{2020}). \bibinfo{title}{Toxicity detection: Does context really matter?}
\newblock \bibinfo{journal}{Preprint at arXiv \url{https://doi.org/10.48550/arXiv.2006.00998}}.
\bibitem[{Hovy and Yang(2021)}]{hovy2021importance}
\bibinfo{author}{Hovy, D.}, and \bibinfo{author}{Yang, D.} (\bibinfo{year}{2021}). \bibinfo{title}{The importance of modeling social factors of language: Theory and practice}.
\newblock In \bibinfo{booktitle}{Proceedings of the 2021 Conference of the North American Chapter of the Association for Computational Linguistics: Human Language Technologies}. pp. \bibinfo{pages}{588--602}.
\bibitem[{Sap et~al.(2019)Sap, Card, Gabriel, Choi and Smith}]{sap2019risk}
\bibinfo{author}{Sap, M.}, \bibinfo{author}{Card, D.}, \bibinfo{author}{Gabriel, S.}, \bibinfo{author}{Choi, Y.}, and \bibinfo{author}{Smith, N.A.} (\bibinfo{year}{2019}). \bibinfo{title}{The risk of racial bias in hate speech detection}.
\newblock In \bibinfo{booktitle}{Proceedings of the 57th annual meeting of the association for computational linguistics}. pp. \bibinfo{pages}{1668--1678}.
\bibitem[{Gordon et~al.(2022)Gordon, Lam, Park, Patel, Hancock, Hashimoto and Bernstein}]{gordon2022jury}
\bibinfo{author}{Gordon, M.L.}, \bibinfo{author}{Lam, M.S.}, \bibinfo{author}{Park, J.S.}, \bibinfo{author}{Patel, K.}, \bibinfo{author}{Hancock, J.}, \bibinfo{author}{Hashimoto, T.}, and \bibinfo{author}{Bernstein, M.S.} (\bibinfo{year}{2022}). \bibinfo{title}{Jury learning: Integrating dissenting voices into machine learning models}.
\newblock In \bibinfo{booktitle}{Proceedings of the 2022 CHI Conference on Human Factors in Computing Systems}. pp. \bibinfo{pages}{1--19}.
\newblock \DOIprefix\doi{https://doi.org/10.1145/3491102.350200}.
\bibitem[{Pavlopoulos and Likas(2024)}]{pavlopoulos2024polarized}
\bibinfo{author}{Pavlopoulos, J.}, and \bibinfo{author}{Likas, A.} (\bibinfo{year}{2024}). \bibinfo{title}{Polarized opinion detection improves the detection of toxic language}.
\newblock In \bibinfo{booktitle}{Proceedings of the 18th Conference of the European Chapter of the Association for Computational Linguistics (Volume 1: Long Papers)}. pp. \bibinfo{pages}{1946--1958}.
\bibitem[{Zampieri et~al.(2020)Zampieri, Nakov, Rosenthal, Atanasova, Karadzhov, Mubarak, Derczynski, Pitenis and {\c{C}}{\"{o}}ltekin}]{Zampieri2020}
\bibinfo{author}{Zampieri, M.}, \bibinfo{author}{Nakov, P.}, \bibinfo{author}{Rosenthal, S.}, \bibinfo{author}{Atanasova, P.}, \bibinfo{author}{Karadzhov, G.}, \bibinfo{author}{Mubarak, H.}, \bibinfo{author}{Derczynski, L.}, \bibinfo{author}{Pitenis, Z.}, and \bibinfo{author}{{\c{C}}{\"{o}}ltekin, {\c{C}}.} (\bibinfo{year}{2020}). \bibinfo{title}{Semeval-2020 task 12: Multilingual offensive language identification in social media (offenseval 2020)}.
\newblock \bibinfo{journal}{Preprint at arXiv \url{https://doi.org/10.48550/arXiv.2006.07235}}.
\bibitem[{Wang et~al.(2020)Wang, Liu, Ouyang and Sun}]{wang2020galileo}
\bibinfo{author}{Wang, S.}, \bibinfo{author}{Liu, J.}, \bibinfo{author}{Ouyang, X.}, and \bibinfo{author}{Sun, Y.} (\bibinfo{year}{2020}). \bibinfo{title}{Galileo at semeval-2020 task 12: Multi-lingual learning for offensive language identification using pre-trained language models}.
\newblock \bibinfo{journal}{Preprint at arXiv \url{https://doi.org/10.48550/arXiv.2010.03542}}.
\bibitem[{Socha(2020)}]{socha2020ks}
\bibinfo{author}{Socha, K.} (\bibinfo{year}{2020}). \bibinfo{title}{Ks@ lth at semeval-2020 task 12: Fine-tuning multi-and monolingual transformer models for offensive language detection}.
\newblock In \bibinfo{booktitle}{Proceedings of the Fourteenth Workshop on Semantic Evaluation}. pp. \bibinfo{pages}{2045--2053}.
\bibitem[{Plum et~al.(2019)Plum, Ranasinghe, Orasan and Mitkov}]{plum2019rgcl}
\bibinfo{author}{Plum, A.}, \bibinfo{author}{Ranasinghe, T.}, \bibinfo{author}{Orasan, C.}, and \bibinfo{author}{Mitkov, R.} (\bibinfo{year}{2019}).
\newblock \bibinfo{title}{Rgcl at germeval 2019: Offensive language detection with deep learning}. \bibinfo{publisher}{German Society for Computational Linguistics \& Language Technology}.
\bibitem[{Zampieri et~al.(2019{\natexlab{a}})Zampieri, Malmasi, Nakov, Rosenthal, Farra and Kumar}]{zampieri2019semeval}
\bibinfo{author}{Zampieri, M.}, \bibinfo{author}{Malmasi, S.}, \bibinfo{author}{Nakov, P.}, \bibinfo{author}{Rosenthal, S.}, \bibinfo{author}{Farra, N.}, and \bibinfo{author}{Kumar, R.} (\bibinfo{year}{2019}{\natexlab{a}}). \bibinfo{title}{Semeval-2019 task 6: Identifying and categorizing offensive language in social media (offenseval)}.
\newblock In \bibinfo{booktitle}{Proceedings of the 13th International Workshop on Semantic Evaluation}. pp. \bibinfo{pages}{75--86}.
\bibitem[{Zampieri et~al.(2019{\natexlab{b}})Zampieri, Malmasi, Nakov, Rosenthal, Farra and Kumar}]{zampieri2019predicting}
\bibinfo{author}{Zampieri, M.}, \bibinfo{author}{Malmasi, S.}, \bibinfo{author}{Nakov, P.}, \bibinfo{author}{Rosenthal, S.}, \bibinfo{author}{Farra, N.}, and \bibinfo{author}{Kumar, R.} (\bibinfo{year}{2019}{\natexlab{b}}). \bibinfo{title}{Predicting the type and target of offensive posts in social media}.
\newblock \bibinfo{journal}{Preprint at arXiv \url{https://doi.org/10.48550/arXiv.1902.09666}}.
\bibitem[{{Liquid Media}(????)}]{gazzetta}
\bibinfo{author}{{Liquid Media}}.
\newblock \bibinfo{title}{Gazzetta}.
\newblock \bibinfo{note}{\url{http://www.gazzetta.gr/}}.
\bibitem[{Arvanitidis et~al.(2021)Arvanitidis, Papagiannitsis, Desli, Vergou and Gourgouliani}]{Arvanitidis_2021}
\bibinfo{author}{Arvanitidis, P.}, \bibinfo{author}{Papagiannitsis, G.}, \bibinfo{author}{Desli, A.Z.}, \bibinfo{author}{Vergou, P.}, and \bibinfo{author}{Gourgouliani, S.} (\bibinfo{year}{2021}). \bibinfo{title}{Attitudes towards refugees \& immigrants in greece: A national-local comparative analysis}.
\newblock \bibinfo{journal}{European Journal of Geography} \emph{\bibinfo{volume}{12}}, \bibinfo{pages}{39--55}. \DOIprefix\doi{10.48088/ejg.p.arv.12.3.39.55}.
\bibitem[{Pontiki et~al.(2020)Pontiki, Gavriilidou, Gkoumas and Piperidis}]{Pontiki2020}
\bibinfo{author}{Pontiki, M.}, \bibinfo{author}{Gavriilidou, M.}, \bibinfo{author}{Gkoumas, D.}, and \bibinfo{author}{Piperidis, S.} (\bibinfo{year}{2020}). \bibinfo{title}{Verbal aggression as an indicator of xenophobic attitudes in {G}reek {T}witter during and after the financial crisis}.
\newblock In \bibinfo{booktitle}{Proceedings of the Workshop about Language Resources for the SSH Cloud}. \bibinfo{address}{Marseille, France}: \bibinfo{publisher}{European Language Resources Association}.
\newblock ISBN \bibinfo{isbn}{979-10-95546-43-6} pp. \bibinfo{pages}{19--26}.
\bibitem[{Kotsakis et~al.(2023)Kotsakis, Vrysis, Vryzas, Saridou, Matsiola, Veglis and Dimoulas}]{kotsakis2023web}
\bibinfo{author}{Kotsakis, R.}, \bibinfo{author}{Vrysis, L.}, \bibinfo{author}{Vryzas, N.}, \bibinfo{author}{Saridou, T.}, \bibinfo{author}{Matsiola, M.}, \bibinfo{author}{Veglis, A.}, and \bibinfo{author}{Dimoulas, C.} (\bibinfo{year}{2023}). \bibinfo{title}{A web framework for information aggregation and management of multilingual hate speech}.
\newblock \bibinfo{journal}{Heliyon} \emph{\bibinfo{volume}{9}}, \bibinfo{pages}{e16084}. \DOIprefix\doi{https://doi.org/10.1016/j.heliyon.2023.e16084}.
\bibitem[{{The PHARM Project Team}(2021)}]{pharmproject_scripts}
\bibinfo{author}{{The PHARM Project Team}} (\bibinfo{year}{2021}).
\newblock \bibinfo{title}{Pharm scripts}. \bibinfo{publisher}{GitHub}.
\newblock \bibinfo{note}{\url{https://github.com/thepharmproject/set_of_scripts}}.
\bibitem[{Lekea and Karampelas(2018)}]{lekea2018detecting}
\bibinfo{author}{Lekea, I.K.}, and \bibinfo{author}{Karampelas, P.} (\bibinfo{year}{2018}). \bibinfo{title}{Detecting hate speech within the terrorist argument: a greek case}.
\newblock In \bibinfo{booktitle}{2018 IEEE/ACM International Conference on Advances in Social Networks Analysis and Mining (ASONAM)}. \bibinfo{organization}{IEEE} pp. \bibinfo{pages}{1084--1091}.
\newblock \DOIprefix\doi{10.1109/ASONAM.2018.8508270}.
\bibitem[{Nikiforos et~al.(2020)Nikiforos, Tzanavaris and Kermanidis}]{nikiforos2020}
\bibinfo{author}{Nikiforos, S.}, \bibinfo{author}{Tzanavaris, S.}, and \bibinfo{author}{Kermanidis, K.L.} (\bibinfo{year}{2020}). \bibinfo{title}{Virtual learning communities (vlcs) rethinking: influence on behavior modification—bullying detection through machine learning and natural language processing}.
\newblock \bibinfo{journal}{Journal of Computers in Education} \emph{\bibinfo{volume}{7}}, \bibinfo{pages}{531--551}. \DOIprefix\doi{https://doi.org/10.1007/s40692-020-00166-5}.
\bibitem[{Strømberg(2020)}]{strombergnlp_offenseval_2020}
\bibinfo{author}{Strømberg, A.} (\bibinfo{year}{2020}).
\newblock \bibinfo{title}{Offenseval 2020 dataset}. \bibinfo{publisher}{Hugging Face}.
\newblock \bibinfo{note}{\url{https://huggingface.co/datasets/strombergnlp/offenseval_2020}}.
\bibitem[{Perifanos and Goutsos(2021{\natexlab{c}})}]{multimodal_hate_speech_detection}
\bibinfo{author}{Perifanos, K.}, and \bibinfo{author}{Goutsos, D.} (\bibinfo{year}{2021}{\natexlab{c}}).
\newblock \bibinfo{title}{Multimodal hate speech detection}. \bibinfo{publisher}{GitHub}{\natexlab{c}}.
\newblock \bibinfo{note}{\url{https://github.com/kperi/MultimodalHateSpeechDetection}}.
\bibitem[{Pontiki(????)}]{clarin_xenophobia}
\bibinfo{author}{Pontiki, M.}
\newblock \bibinfo{title}{Clarin: Xenophobia resources}. \bibinfo{publisher}{CLARIN:EL}.
\newblock \bibinfo{note}{\url{https://inventory.clarin.gr/search/xenophobia}}.
\bibitem[{Pavlopoulos et~al.(2015)Pavlopoulos, Malakasiotis and Androutsopoulos}]{gazzetta_comments_dataset}
\bibinfo{author}{Pavlopoulos, J.}, \bibinfo{author}{Malakasiotis, P.}, and \bibinfo{author}{Androutsopoulos, I.} (\bibinfo{year}{2015}).
\newblock \bibinfo{title}{Gazzetta comments dataset}.
\newblock \bibinfo{note}{\url{https://archive.org/details/gazzetta-comments-dataset.tar}}.
\bibitem[{{PHARM Project Consortium}(????)}]{pharmproject}
\bibinfo{author}{{PHARM Project Consortium}}.
\newblock \bibinfo{title}{Pharm project}. \bibinfo{publisher}{University of Salamanca}.
\newblock \bibinfo{note}{\url{https://pharmproject.usal.es/}}.
\bibitem[{Alfano et~al.(2018)Alfano, Hovy, Mitchell and Strube}]{ws-2018-acl}
\bibinfo{editor}{Alfano, M.}, \bibinfo{editor}{Hovy, D.}, \bibinfo{editor}{Mitchell, M.}, and \bibinfo{editor}{Strube, M.}, eds.
\newblock \bibinfo{title}{Proceedings of the Second {ACL} Workshop on Ethics in Natural Language Processing}. \bibinfo{address}{New Orleans, Louisiana, USA}: \bibinfo{publisher}{Association for Computational Linguistics} (\bibinfo{year}{2018}).
\newblock \DOIprefix\doi{10.18653/v1/W18-08}.
\bibitem[{El-Kassas et~al.(2021)El-Kassas, Salama, Rafea and Mohamed}]{el2021automatic}
\bibinfo{author}{El-Kassas, W.S.}, \bibinfo{author}{Salama, C.R.}, \bibinfo{author}{Rafea, A.A.}, and \bibinfo{author}{Mohamed, H.K.} (\bibinfo{year}{2021}). \bibinfo{title}{Automatic text summarization: A comprehensive survey}.
\newblock \bibinfo{journal}{Expert systems with applications} \emph{\bibinfo{volume}{165}}, \bibinfo{pages}{113679}. \DOIprefix\doi{https://doi.org/10.1016/j.eswa.2020.113679}.
\bibitem[{Maybury(1995)}]{maybury1995generating}
\bibinfo{author}{Maybury, M.T.} (\bibinfo{year}{1995}). \bibinfo{title}{Generating summaries from event data}.
\newblock \bibinfo{journal}{Information Processing \& Management} \emph{\bibinfo{volume}{31}}, \bibinfo{pages}{735--751}. \DOIprefix\doi{https://doi.org/10.1016/0306-4573(95)00025-C}.
\bibitem[{Luhn(1958)}]{luhn1958automatic}
\bibinfo{author}{Luhn, H.P.} (\bibinfo{year}{1958}). \bibinfo{title}{The automatic creation of literature abstracts}.
\newblock \bibinfo{journal}{IBM Journal of research and development} \emph{\bibinfo{volume}{2}}, \bibinfo{pages}{159--165}. \DOIprefix\doi{https://doi.org/10.1147/rd.22.0159}.
\bibitem[{Giarelis et~al.(2023{\natexlab{b}})Giarelis, Mastrokostas and Karacapilidis}]{giarelis2023abstractive}
\bibinfo{author}{Giarelis, N.}, \bibinfo{author}{Mastrokostas, C.}, and \bibinfo{author}{Karacapilidis, N.} (\bibinfo{year}{2023}{\natexlab{b}}). \bibinfo{title}{Abstractive vs. extractive summarization: An experimental review}.
\newblock \bibinfo{journal}{Applied Sciences} \emph{\bibinfo{volume}{13}}, \bibinfo{pages}{7620}. \DOIprefix\doi{https://doi.org/10.3390/app13137620}.
\bibitem[{Alomari et~al.(2022)Alomari, Idris, Sabri and Alsmadi}]{alomari2022deep}
\bibinfo{author}{Alomari, A.}, \bibinfo{author}{Idris, N.}, \bibinfo{author}{Sabri, A.Q.M.}, and \bibinfo{author}{Alsmadi, I.} (\bibinfo{year}{2022}). \bibinfo{title}{Deep reinforcement and transfer learning for abstractive text summarization: A review}.
\newblock \bibinfo{journal}{Computer Speech \& Language} \emph{\bibinfo{volume}{71}}, \bibinfo{pages}{101276}.
\bibitem[{El-Haj et~al.(2022)El-Haj, Zmandar, Rayson, AbuRa{'}ed, Litvak, Pittaras, Giannakopoulos, Kosmopoulos, Carbajo-Coronado and Moreno-Sandoval}]{el-haj-etal-2022-financial}
\bibinfo{author}{El-Haj, M.}, \bibinfo{author}{Zmandar, N.}, \bibinfo{author}{Rayson, P.}, \bibinfo{author}{AbuRa{'}ed, A.}, \bibinfo{author}{Litvak, M.}, \bibinfo{author}{Pittaras, N.}, \bibinfo{author}{Giannakopoulos, G.}, \bibinfo{author}{Kosmopoulos, A.}, \bibinfo{author}{Carbajo-Coronado, B.}, and \bibinfo{author}{Moreno-Sandoval, A.} (\bibinfo{year}{2022}). \bibinfo{title}{The financial narrative summarisation shared task ({FNS} 2022)}.
\newblock In \bibinfo{booktitle}{Proceedings of the 4th Financial Narrative Processing Workshop @LREC2022}. \bibinfo{address}{Marseille, France}: \bibinfo{publisher}{European Language Resources Association} pp. \bibinfo{pages}{43--52}.
\bibitem[{Zavitsanos et~al.(2023{\natexlab{a}})Zavitsanos, Kosmopoulos, Giannakopoulos, Litvak, Carbajo-Coronado, Moreno-Sandoval and El-Haj}]{zavitsanos2023financial}
\bibinfo{author}{Zavitsanos, E.}, \bibinfo{author}{Kosmopoulos, A.}, \bibinfo{author}{Giannakopoulos, G.}, \bibinfo{author}{Litvak, M.}, \bibinfo{author}{Carbajo-Coronado, B.}, \bibinfo{author}{Moreno-Sandoval, A.}, and \bibinfo{author}{El-Haj, M.} (\bibinfo{year}{2023}{\natexlab{a}}). \bibinfo{title}{The financial narrative summarisation shared task (fns 2023)}.
\newblock In \bibinfo{booktitle}{2023 IEEE International Conference on Big Data (BigData)}. \bibinfo{organization}{IEEE} pp. \bibinfo{pages}{2890--2896}.
\newblock \DOIprefix\doi{https://doi.org/10.1109/BigData59044.2023.10386228}.
\bibitem[{Shukla et~al.(2023)Shukla, Katikeri, Raja, Sivam, Yadav, Vaid and Prabhakararao}]{shukla2023generative}
\bibinfo{author}{Shukla, N.K.}, \bibinfo{author}{Katikeri, R.}, \bibinfo{author}{Raja, M.}, \bibinfo{author}{Sivam, G.}, \bibinfo{author}{Yadav, S.}, \bibinfo{author}{Vaid, A.}, and \bibinfo{author}{Prabhakararao, S.} (\bibinfo{year}{2023}). \bibinfo{title}{Generative ai approach to distributed summarization of financial narratives}.
\newblock In \bibinfo{booktitle}{2023 IEEE International Conference on Big Data (BigData)}. \bibinfo{organization}{IEEE} pp. \bibinfo{pages}{2872--2876}.
\newblock \DOIprefix\doi{https://doi.org/10.1109/BigData59044.2023.10386313}.
\bibitem[{Vanetik et~al.(2023)Vanetik, Podkaminer and Litvak}]{vanetik2023summarizing}
\bibinfo{author}{Vanetik, N.}, \bibinfo{author}{Podkaminer, E.}, and \bibinfo{author}{Litvak, M.} (\bibinfo{year}{2023}). \bibinfo{title}{Summarizing financial reports with positional language model}.
\newblock In \bibinfo{booktitle}{2023 IEEE International Conference on Big Data (BigData)}. \bibinfo{organization}{IEEE} pp. \bibinfo{pages}{2877--2883}.
\newblock \DOIprefix\doi{https://doi.org/10.1109/BigData59044.2023.10386704}.
\bibitem[{Shukla et~al.(2022)Shukla, Vaid, Katikeri, Keeriyadath and Raja}]{shukla2022dimsum}
\bibinfo{author}{Shukla, N.}, \bibinfo{author}{Vaid, A.}, \bibinfo{author}{Katikeri, R.}, \bibinfo{author}{Keeriyadath, S.}, and \bibinfo{author}{Raja, M.} (\bibinfo{year}{2022}). \bibinfo{title}{Dimsum: Distributed and multilingual summarization of financial narratives}.
\newblock In \bibinfo{booktitle}{Proceedings of the 4th Financial Narrative Processing Workshop@ LREC2022}. pp. \bibinfo{pages}{65--72}.
\bibitem[{Liu(2019)}]{liu2019fine}
\bibinfo{author}{Liu, Y.} (\bibinfo{year}{2019}). \bibinfo{title}{Fine-tune bert for extractive summarization}.
\newblock \bibinfo{journal}{Preprint at arXiv \url{https://doi.org/10.48550/arXiv.1903.10318}}.
\bibitem[{Lv and Zhai(2009)}]{lv2009positional}
\bibinfo{author}{Lv, Y.}, and \bibinfo{author}{Zhai, C.} (\bibinfo{year}{2009}). \bibinfo{title}{Positional language models for information retrieval}.
\newblock In \bibinfo{booktitle}{Proceedings of the 32nd international ACM SIGIR conference on Research and development in information retrieval}. pp. \bibinfo{pages}{299--306}.
\newblock \DOIprefix\doi{https://doi.org/10.1145/1571941.1571994}.
\bibitem[{Giannakopoulos(2013)}]{giannakopoulos2013multi}
\bibinfo{author}{Giannakopoulos, G.} (\bibinfo{year}{2013}). \bibinfo{title}{Multi-document multilingual summarization and evaluation tracks in acl 2013 multiling workshop}.
\newblock In \bibinfo{booktitle}{Proceedings of the multiling 2013 workshop on multilingual multi-document summarization}. pp. \bibinfo{pages}{20--28}.
\bibitem[{Giarelis et~al.(2024{\natexlab{e}})Giarelis, Mastrokostas and Karacapilidis}]{nc0der_greekt5}
\bibinfo{author}{Giarelis, N.}, \bibinfo{author}{Mastrokostas, C.}, and \bibinfo{author}{Karacapilidis, N.} (\bibinfo{year}{2024}{\natexlab{e}}).
\newblock \bibinfo{title}{Greekt5: A series of greek sequence-to-sequence models for news summarization}. \bibinfo{publisher}{GitHub}{\natexlab{e}}.
\newblock \bibinfo{note}{\url{https://github.com/NC0DER/GreekT5}}.
\bibitem[{Koniaris et~al.(2023)Koniaris, Galanis, Giannini and Tsanakas}]{koniaris2023evaluation}
\bibinfo{author}{Koniaris, M.}, \bibinfo{author}{Galanis, D.}, \bibinfo{author}{Giannini, E.}, and \bibinfo{author}{Tsanakas, P.} (\bibinfo{year}{2023}). \bibinfo{title}{Evaluation of automatic legal text summarization techniques for greek case law}.
\newblock \bibinfo{journal}{Information} \emph{\bibinfo{volume}{14}}, \bibinfo{pages}{250}. \DOIprefix\doi{https://doi.org/10.3390/info14040250}.
\bibitem[{Erkan and Radev(2004)}]{erkan2004lexrank}
\bibinfo{author}{Erkan, G.}, and \bibinfo{author}{Radev, D.R.} (\bibinfo{year}{2004}). \bibinfo{title}{Lexrank: Graph-based lexical centrality as salience in text summarization}.
\newblock \bibinfo{journal}{Journal of Artificial Intelligence Research} \emph{\bibinfo{volume}{22}}, \bibinfo{pages}{457--479}. \DOIprefix\doi{https://doi.org/10.1613/jair.1523}.
\bibitem[{Otterbacher et~al.(2009)Otterbacher, Erkan and Radev}]{otterbacher2009biased}
\bibinfo{author}{Otterbacher, J.}, \bibinfo{author}{Erkan, G.}, and \bibinfo{author}{Radev, D.R.} (\bibinfo{year}{2009}). \bibinfo{title}{Biased lexrank: Passage retrieval using random walks with question-based priors}.
\newblock \bibinfo{journal}{Information Processing \& Management} \emph{\bibinfo{volume}{45}}, \bibinfo{pages}{42--54}. \DOIprefix\doi{https://doi.org/10.1016/j.ipm.2008.06.004}.
\bibitem[{{Areios Pagos}(????)}]{areiospagos}
\bibinfo{author}{{Areios Pagos}}.
\newblock \bibinfo{title}{Areios pagos: Supreme civil and criminal court of greece}.
\newblock \bibinfo{note}{\url{https://www.areiospagos.gr/}}.
\bibitem[{{DominusTea}(2023)}]{greeklegalsum}
\bibinfo{author}{{DominusTea}} (\bibinfo{year}{2023}).
\newblock \bibinfo{title}{Greeklegalsum: A greek legal summarization dataset}. \bibinfo{publisher}{Hugging Face}.
\newblock \bibinfo{note}{\url{https://huggingface.co/datasets/DominusTea/GreekLegalSum}}.
\bibitem[{Evdaimon(2022)}]{greeksum}
\bibinfo{author}{Evdaimon, I.} (\bibinfo{year}{2022}).
\newblock \bibinfo{title}{Greeksum: A greek news summarization dataset}. \bibinfo{publisher}{GitHub}.
\newblock \bibinfo{note}{\url{https://github.com/iakovosevdaimon/GreekSUM}}.
\bibitem[{Zavitsanos et~al.(2023{\natexlab{b}})Zavitsanos, Kosmopoulos, Giannakopoulos, Litvak, Carbajo-Coronado, Moreno-Sandoval and El-Haj}]{fns2023}
\bibinfo{author}{Zavitsanos, E.}, \bibinfo{author}{Kosmopoulos, A.}, \bibinfo{author}{Giannakopoulos, G.}, \bibinfo{author}{Litvak, M.}, \bibinfo{author}{Carbajo-Coronado, B.}, \bibinfo{author}{Moreno-Sandoval, A.}, and \bibinfo{author}{El-Haj, M.} (\bibinfo{year}{2023}{\natexlab{b}}).
\newblock \bibinfo{title}{The financial narrative summarisation shared task (fns 2023)}. \bibinfo{publisher}{IEEE Computer Society Press}{\natexlab{b}}.
\newblock \bibinfo{note}{\url{https://doi.org/10.1109/BigData59044.2023.10386228}}.
\bibitem[{{European Language Resources Association}(2022)}]{fnp2022}
\bibinfo{author}{{European Language Resources Association}} (\bibinfo{year}{2022}).
\newblock \bibinfo{title}{4th financial narrative processing workshop (fnp 2022)}. \bibinfo{publisher}{European Language Resources Association}.
\newblock \bibinfo{note}{\url{https://aclanthology.org/2022.fnp-1.0.pdf}}.
\bibitem[{Li et~al.(2013)Li, For{\u{a}}scu, El-Haj and Giannakopoulos}]{li2013multi}
\bibinfo{author}{Li, L.}, \bibinfo{author}{For{\u{a}}scu, C.}, \bibinfo{author}{El-Haj, M.}, and \bibinfo{author}{Giannakopoulos, G.} (\bibinfo{year}{2013}). \bibinfo{title}{Multi-document multilingual summarization corpus preparation, part 1: Arabic, english, greek, chinese, romanian}.
\newblock In \bibinfo{booktitle}{Proceedings of the multiling 2013 workshop on multilingual multi-document summarization}. pp. \bibinfo{pages}{1--12}.
\bibitem[{Li et~al.(2014)Li, For{\u{a}}scu, El-Haj and Giannakopoulos}]{multiling_datasets}
\bibinfo{author}{Li, L.}, \bibinfo{author}{For{\u{a}}scu, C.}, \bibinfo{author}{El-Haj, M.}, and \bibinfo{author}{Giannakopoulos, G.} (\bibinfo{year}{2014}).
\newblock \bibinfo{title}{Multiling community datasets}. \bibinfo{publisher}{NCSR Demokritos}.
\newblock \bibinfo{note}{\url{http://multiling.iit.demokritos.gr/pages/view/1571/datasets}}.
\bibitem[{{Wikimedia Foundation}(????{\natexlab{b}})}]{wikinews}
\bibinfo{author}{{Wikimedia Foundation}}.
\newblock \bibinfo{title}{Wikinews: Free news source}.
\newblock \bibinfo{note}{\url{https://www.wikinews.org/}}.
\bibitem[{Zhu et~al.(2021)Zhu, Lei, Wang, Zheng, Poria and Chua}]{Zhu2021}
\bibinfo{author}{Zhu, F.}, \bibinfo{author}{Lei, W.}, \bibinfo{author}{Wang, C.}, \bibinfo{author}{Zheng, J.}, \bibinfo{author}{Poria, S.}, and \bibinfo{author}{Chua, T.} (\bibinfo{year}{2021}). \bibinfo{title}{Retrieving and reading: {A} comprehensive survey on open-domain question answering}.
\newblock \bibinfo{journal}{Preprint at arXiv \url{https://doi.org/10.48550/arXiv.2101.00774}}.
\bibitem[{Goodwin and Harabagiu(2016)}]{goodwin2016medical}
\bibinfo{author}{Goodwin, T.R.}, and \bibinfo{author}{Harabagiu, S.M.} (\bibinfo{year}{2016}). \bibinfo{title}{Medical question answering for clinical decision support}.
\newblock In \bibinfo{booktitle}{Proceedings of the 25th ACM international on conference on information and knowledge management}. pp. \bibinfo{pages}{297--306}.
\bibitem[{Li et~al.(2018)Li, Miao, Geng, Alt, Schwarzenberg, Hennig, Hu and Xu}]{li2018question}
\bibinfo{author}{Li, Y.}, \bibinfo{author}{Miao, Q.}, \bibinfo{author}{Geng, J.}, \bibinfo{author}{Alt, C.}, \bibinfo{author}{Schwarzenberg, R.}, \bibinfo{author}{Hennig, L.}, \bibinfo{author}{Hu, C.}, and \bibinfo{author}{Xu, F.} (\bibinfo{year}{2018}). \bibinfo{title}{Question answering for technical customer support}.
\newblock In \bibinfo{booktitle}{Natural Language Processing and Chinese Computing: 7th CCF International Conference, NLPCC 2018, Hohhot, China, August 26--30, 2018, Proceedings, Part I 7}. \bibinfo{organization}{Springer} pp. \bibinfo{pages}{3--15}.
\bibitem[{Adamopoulou and Moussiades(2020)}]{adamopoulou2020chatbots}
\bibinfo{author}{Adamopoulou, E.}, and \bibinfo{author}{Moussiades, L.} (\bibinfo{year}{2020}). \bibinfo{title}{Chatbots: History, technology, and applications}.
\newblock \bibinfo{journal}{Machine Learning with Applications} \emph{\bibinfo{volume}{2}}, \bibinfo{pages}{100006}. \DOIprefix\doi{10.1016/j.mlwa.2020.100006}.
\bibitem[{{de Barcelos Silva} et~al.(2020){de Barcelos Silva}, Gomes, {da Costa}, {da Rosa Righi}, Barbosa, Pessin, {De Doncker} and Federizzi}]{DEBARCELOSSILVA2020}
\bibinfo{author}{{de Barcelos Silva}, A.}, \bibinfo{author}{Gomes, M.M.}, \bibinfo{author}{{da Costa}, C.A.}, \bibinfo{author}{{da Rosa Righi}, R.}, \bibinfo{author}{Barbosa, J.L.V.}, \bibinfo{author}{Pessin, G.}, \bibinfo{author}{{De Doncker}, G.}, and \bibinfo{author}{Federizzi, G.} (\bibinfo{year}{2020}). \bibinfo{title}{Intelligent personal assistants: A systematic literature review}.
\newblock \bibinfo{journal}{Expert Systems with Applications} \emph{\bibinfo{volume}{147}}, \bibinfo{pages}{113193}. \DOIprefix\doi{https://doi.org/10.1016/j.eswa.2020.113193}.
\bibitem[{Gardner et~al.(2019)Gardner, Berant, Hajishirzi, Talmor and Min}]{Gardner2019}
\bibinfo{author}{Gardner, M.}, \bibinfo{author}{Berant, J.}, \bibinfo{author}{Hajishirzi, H.}, \bibinfo{author}{Talmor, A.}, and \bibinfo{author}{Min, S.} (\bibinfo{year}{2019}). \bibinfo{title}{Question answering is a format; when is it useful?}
\newblock \bibinfo{journal}{Preprint at arXiv \url{https://doi.org/10.48550/arXiv.1909.11291}}.
\bibitem[{{OpenAI}(????)}]{chatgpt}
\bibinfo{author}{{OpenAI}}.
\newblock \bibinfo{title}{Chatgpt by openai}.
\newblock \bibinfo{note}{\url{https://chat.openai.com}}.
\bibitem[{Chen and Yih(2020)}]{chen-yih-2020-open}
\bibinfo{author}{Chen, D.}, and \bibinfo{author}{Yih, W.t.} (\bibinfo{year}{2020}). \bibinfo{title}{Open-domain question answering}.
\newblock In \bibinfo{editor}{ A.{ }Savary}, and \bibinfo{editor}{ Y.{ }Zhang}, eds. \bibinfo{booktitle}{Proceedings of the 58th Annual Meeting of the Association for Computational Linguistics: Tutorial Abstracts}. \bibinfo{address}{Online}: \bibinfo{publisher}{Association for Computational Linguistics} pp. \bibinfo{pages}{34--37}.
\newblock \DOIprefix\doi{10.18653/v1/2020.acl-tutorials.8}.
\bibitem[{Rogers et~al.(2023)Rogers, Gardner and Augenstein}]{Rogers2023}
\bibinfo{author}{Rogers, A.}, \bibinfo{author}{Gardner, M.}, and \bibinfo{author}{Augenstein, I.} (\bibinfo{year}{2023}). \bibinfo{title}{Qa dataset explosion: A taxonomy of nlp resources for question answering and reading comprehension}.
\newblock \bibinfo{journal}{ACM Comput. Surv.} \emph{\bibinfo{volume}{55}}, \bibinfo{pages}{1--45}. \DOIprefix\doi{10.1145/3560260}.
\bibitem[{Schlegel et~al.(2020)Schlegel, Valentino, Freitas, Nenadic and Batista-Navarro}]{schlegel2020framework}
\bibinfo{author}{Schlegel, V.}, \bibinfo{author}{Valentino, M.}, \bibinfo{author}{Freitas, A.}, \bibinfo{author}{Nenadic, G.}, and \bibinfo{author}{Batista-Navarro, R.} (\bibinfo{year}{2020}). \bibinfo{title}{A framework for evaluation of machine reading comprehension gold standards}.
\newblock \bibinfo{journal}{Preprint at arXiv \url{https://doi.org/10.48550/arXiv.2003.04642}}.
\bibitem[{Sugawara et~al.(2017)Sugawara, Kido, Yokono and Aizawa}]{sugawara2017evaluation}
\bibinfo{author}{Sugawara, S.}, \bibinfo{author}{Kido, Y.}, \bibinfo{author}{Yokono, H.}, and \bibinfo{author}{Aizawa, A.} (\bibinfo{year}{2017}). \bibinfo{title}{Evaluation metrics for machine reading comprehension: Prerequisite skills and readability}.
\newblock In \bibinfo{booktitle}{Proceedings of the 55th Annual Meeting of the Association for Computational Linguistics (Volume 1: Long Papers)}. pp. \bibinfo{pages}{806--817}.
\bibitem[{{DBpedia Association}(????)}]{dbpedia}
\bibinfo{author}{{DBpedia Association}}.
\newblock \bibinfo{title}{Dbpedia: Structured information from wikipedia}.
\newblock \bibinfo{note}{\url{https://www.dbpedia.org/}}.
\bibitem[{Marakakis et~al.(2017)Marakakis, Kondylakis and Aris}]{Marakakis2017}
\bibinfo{author}{Marakakis, E.}, \bibinfo{author}{Kondylakis, H.}, and \bibinfo{author}{Aris, P.} (\bibinfo{year}{2017}). \bibinfo{title}{Apantisis: A greek question-answering system for knowledge-base exploriaton}.
\newblock In \bibinfo{editor}{ A.{ }Kavoura}, \bibinfo{editor}{ D.P.{ }Sakas}, and \bibinfo{editor}{ P.{ }Tomaras}, eds. \bibinfo{booktitle}{Strategic Innovative Marketing}. \bibinfo{address}{Cham}: \bibinfo{publisher}{Springer International Publishing}.
\newblock ISBN \bibinfo{isbn}{978-3-319-56288-9} pp. \bibinfo{pages}{501--510}.
\newblock \DOIprefix\doi{https://doi.org/10.1007/978-3-319-56288-9_67}.
\bibitem[{Braun et~al.(2017)Braun, Mendez, Matthes and Langen}]{braun2017evaluating}
\bibinfo{author}{Braun, D.}, \bibinfo{author}{Mendez, A.H.}, \bibinfo{author}{Matthes, F.}, and \bibinfo{author}{Langen, M.} (\bibinfo{year}{2017}). \bibinfo{title}{Evaluating natural language understanding services for conversational question answering systems}.
\newblock In \bibinfo{booktitle}{Proceedings of the 18th annual SIGdial meeting on discourse and dialogue}. pp. \bibinfo{pages}{174--185}.
\bibitem[{Malamas et~al.(2022)Malamas, Papangelou and Symeonidis}]{Malamas2022}
\bibinfo{author}{Malamas, N.}, \bibinfo{author}{Papangelou, K.}, and \bibinfo{author}{Symeonidis, A.L.} (\bibinfo{year}{2022}). \bibinfo{title}{Upon improving the performance of localized healthcare virtual assistants}.
\newblock \bibinfo{journal}{Healthcare} \emph{\bibinfo{volume}{10}}, \bibinfo{pages}{99}. \DOIprefix\doi{10.3390/healthcare10010099}.
\bibitem[{Ventoura et~al.(2021)Ventoura, Palios, Vasilakis, Paraskevopoulos, Katsamanis and Katsouros}]{Ventoura2021}
\bibinfo{author}{Ventoura, N.}, \bibinfo{author}{Palios, K.}, \bibinfo{author}{Vasilakis, Y.}, \bibinfo{author}{Paraskevopoulos, G.}, \bibinfo{author}{Katsamanis, N.}, and \bibinfo{author}{Katsouros, V.} (\bibinfo{year}{2021}). \bibinfo{title}{Theano: A {G}reek-speaking conversational agent for {COVID}-19}.
\newblock In \bibinfo{booktitle}{Proceedings of the 1st Workshop on NLP for Positive Impact}. \bibinfo{address}{Online}: \bibinfo{publisher}{Association for Computational Linguistics} pp. \bibinfo{pages}{36--46}.
\newblock \DOIprefix\doi{10.18653/v1/2021.nlp4posimpact-1.5}.
\bibitem[{{Rasa Technologies GmbH}(????)}]{rasa}
\bibinfo{author}{{Rasa Technologies GmbH}}.
\newblock \bibinfo{title}{Rasa: Open source conversational ai}.
\newblock \bibinfo{note}{\url{https://rasa.com/}}.
\bibitem[{{Centre for the Greek Language}(????)}]{greek_language_certification_teachers}
\bibinfo{author}{{Centre for the Greek Language}}.
\newblock \bibinfo{title}{Text bank}.
\newblock \bibinfo{note}{\url{https://www.greek-language.gr/certification/dbs/teachers/}}.
\bibitem[{Lopes et~al.(2016)Lopes, Chorianopoulou, Palogiannidi, Moniz, Abad, Louka, Iosif and Potamianos}]{lopes2016}
\bibinfo{author}{Lopes, J.}, \bibinfo{author}{Chorianopoulou, A.}, \bibinfo{author}{Palogiannidi, E.}, \bibinfo{author}{Moniz, H.}, \bibinfo{author}{Abad, A.}, \bibinfo{author}{Louka, K.}, \bibinfo{author}{Iosif, E.}, and \bibinfo{author}{Potamianos, A.} (\bibinfo{year}{2016}). \bibinfo{title}{The {S}pe{D}ial datasets: datasets for spoken dialogue systems analytics}.
\newblock In \bibinfo{booktitle}{Proceedings of the Tenth International Conference on Language Resources and Evaluation ({LREC}'16)}. \bibinfo{address}{Portoro{\v{z}}, Slovenia}: \bibinfo{publisher}{European Language Resources Association (ELRA)} pp. \bibinfo{pages}{104--110}.
\bibitem[{Bastakis(2023)}]{tiresias_evaluation_results}
\bibinfo{author}{Bastakis, M.} (\bibinfo{year}{2023}).
\newblock \bibinfo{title}{Tiresias evaluation results}. \bibinfo{publisher}{GitHub}.
\newblock \bibinfo{note}{\url{https://github.com/mbastakis/Tiresias/tree/master/evaluation_results}}.
\bibitem[{Hutchins(1997)}]{hutchins1997first}
\bibinfo{author}{Hutchins, J.} (\bibinfo{year}{1997}). \bibinfo{title}{From first conception to first demonstration: the nascent years of machine translation, 1947--1954. a chronology}.
\newblock \bibinfo{journal}{Machine Translation} \emph{\bibinfo{volume}{12}}, \bibinfo{pages}{195--252}. \DOIprefix\doi{https://doi.org/10.1023/A:1007969630568}.
\bibitem[{Kouremenos et~al.(2018)Kouremenos, Ntalianis and Kollias}]{kouremenos2018novel}
\bibinfo{author}{Kouremenos, D.}, \bibinfo{author}{Ntalianis, K.}, and \bibinfo{author}{Kollias, S.} (\bibinfo{year}{2018}). \bibinfo{title}{A novel rule based machine translation scheme from greek to greek sign language: Production of different types of large corpora and language models evaluation}.
\newblock \bibinfo{journal}{Computer Speech \& Language} \emph{\bibinfo{volume}{51}}, \bibinfo{pages}{110--135}. \DOIprefix\doi{https://doi.org/10.1016/j.csl.2018.04.001}.
\bibitem[{Beinborn et~al.(2013)Beinborn, Zesch and Gurevych}]{beinborn2013cognate}
\bibinfo{author}{Beinborn, L.}, \bibinfo{author}{Zesch, T.}, and \bibinfo{author}{Gurevych, I.} (\bibinfo{year}{2013}). \bibinfo{title}{Cognate production using character-based machine translation}.
\newblock In \bibinfo{booktitle}{Proceedings of the sixth international joint conference on natural language processing}. pp. \bibinfo{pages}{883--891}.
\bibitem[{Crystal(2011)}]{crystal2011dictionary}
\bibinfo{author}{Crystal, D.} (\bibinfo{year}{2011}). \bibinfo{title}{A dictionary of linguistics and phonetics}. \bibinfo{publisher}{John Wiley \& Sons}.
\bibitem[{Koehn et~al.(2007)Koehn, Hoang, Birch, Callison-Burch, Federico, Bertoldi, Cowan, Shen, Moran, Zens et~al.}]{koehn2007moses}
\bibinfo{author}{Koehn, P.}, \bibinfo{author}{Hoang, H.}, \bibinfo{author}{Birch, A.}, \bibinfo{author}{Callison-Burch, C.}, \bibinfo{author}{Federico, M.}, \bibinfo{author}{Bertoldi, N.}, \bibinfo{author}{Cowan, B.}, \bibinfo{author}{Shen, W.}, \bibinfo{author}{Moran, C.}, \bibinfo{author}{Zens, R.} et~al. (\bibinfo{year}{2007}). \bibinfo{title}{Moses: Open source toolkit for statistical machine translation}.
\newblock In \bibinfo{booktitle}{Proceedings of the 45th annual meeting of the association for computational linguistics companion volume proceedings of the demo and poster sessions}. pp. \bibinfo{pages}{177--180}.
\bibitem[{Pecina et~al.(2012)Pecina, Toral, Papavassiliou, Prokopidis and Van~Genabith}]{pecina2012domain}
\bibinfo{author}{Pecina, P.}, \bibinfo{author}{Toral, A.}, \bibinfo{author}{Papavassiliou, V.}, \bibinfo{author}{Prokopidis, P.}, and \bibinfo{author}{Van~Genabith, J.} (\bibinfo{year}{2012}). \bibinfo{title}{Domain adaptation of statistical machine translation using web-crawled resources: a case study}.
\newblock In \bibinfo{booktitle}{Proceedings of the 16th Annual Conference of the European Association for Machine Translation}. pp. \bibinfo{pages}{145--152}.
\bibitem[{Mouratidis et~al.(2021)Mouratidis, Kermanidis and Sosoni}]{mouratidis2021innovatively}
\bibinfo{author}{Mouratidis, D.}, \bibinfo{author}{Kermanidis, K.L.}, and \bibinfo{author}{Sosoni, V.} (\bibinfo{year}{2021}). \bibinfo{title}{Innovatively fused deep learning with limited noisy data for evaluating translations from poor into rich morphology}.
\newblock \bibinfo{journal}{Applied Sciences} \emph{\bibinfo{volume}{11}}, \bibinfo{pages}{639}. \DOIprefix\doi{https://doi.org/10.3390/app11020639}.
\bibitem[{Stasimioti et~al.(2020)Stasimioti, Sosoni, Kermanidis and Mouratidis}]{stasimioti-2020-machine}
\bibinfo{author}{Stasimioti, M.}, \bibinfo{author}{Sosoni, V.}, \bibinfo{author}{Kermanidis, K.}, and \bibinfo{author}{Mouratidis, D.} (\bibinfo{year}{2020}). \bibinfo{title}{Machine translation quality: A comparative evaluation of {SMT}, {NMT} and tailored-{NMT} outputs}.
\newblock In \bibinfo{booktitle}{Proceedings of the 22nd Annual Conference of the European Association for Machine Translation}. \bibinfo{address}{Lisboa, Portugal}: \bibinfo{publisher}{European Association for Machine Translation} pp. \bibinfo{pages}{441--450}.
\bibitem[{Mouratidis et~al.(2020)Mouratidis, Kermanidis and Sosoni}]{mouratidis2020innovative}
\bibinfo{author}{Mouratidis, D.}, \bibinfo{author}{Kermanidis, K.L.}, and \bibinfo{author}{Sosoni, V.} (\bibinfo{year}{2020}). \bibinfo{title}{Innovative deep neural network fusion for pairwise translation evaluation}.
\newblock In \bibinfo{booktitle}{Artificial Intelligence Applications and Innovations: 16th IFIP WG 12.5 International Conference, AIAI 2020, Neos Marmaras, Greece, June 5--7, 2020, Proceedings, Part II 16}. \bibinfo{organization}{Springer} pp. \bibinfo{pages}{76--87}.
\newblock \DOIprefix\doi{https://doi.org/10.1007/978-3-030-49186-4_7}.
\bibitem[{Mouratidis and Kermanidis(2019)}]{mouratidis2019ensemble}
\bibinfo{author}{Mouratidis, D.}, and \bibinfo{author}{Kermanidis, K.L.} (\bibinfo{year}{2019}). \bibinfo{title}{Ensemble and deep learning for language-independent automatic selection of parallel data}.
\newblock \bibinfo{journal}{Algorithms} \emph{\bibinfo{volume}{12}}, \bibinfo{pages}{26}. \DOIprefix\doi{https://doi.org/10.3390/a12010026}.
\bibitem[{Castilho et~al.(2018)Castilho, Moorkens, Gaspari, Sennrich, Way and Georgakopoulou}]{castilho2018evaluating}
\bibinfo{author}{Castilho, S.}, \bibinfo{author}{Moorkens, J.}, \bibinfo{author}{Gaspari, F.}, \bibinfo{author}{Sennrich, R.}, \bibinfo{author}{Way, A.}, and \bibinfo{author}{Georgakopoulou, P.} (\bibinfo{year}{2018}). \bibinfo{title}{Evaluating mt for massive open online courses: A multifaceted comparison between pbsmt and nmt systems}.
\newblock \bibinfo{journal}{Machine translation} \emph{\bibinfo{volume}{32}}, \bibinfo{pages}{255--278}. \DOIprefix\doi{https://doi.org/10.1007/s10590-018-9221-y}.
\bibitem[{Giorgi et~al.(2021)Giorgi, Golosio, Esposito, Cangelosi and Masala}]{Giorgi2021}
\bibinfo{author}{Giorgi, I.}, \bibinfo{author}{Golosio, B.}, \bibinfo{author}{Esposito, M.}, \bibinfo{author}{Cangelosi, A.}, and \bibinfo{author}{Masala, G.L.} (\bibinfo{year}{2021}). \bibinfo{title}{Modeling multiple language learning in a developmental cognitive architecture}.
\newblock \bibinfo{journal}{IEEE Transactions on Cognitive and Developmental Systems} \emph{\bibinfo{volume}{13}}, \bibinfo{pages}{922--933}. \DOIprefix\doi{10.1109/TCDS.2020.3033963}.
\bibitem[{Gamallo et~al.(2020)Gamallo, Pichel and Alegria}]{gamallo2020measuring}
\bibinfo{author}{Gamallo, P.}, \bibinfo{author}{Pichel, J.R.}, and \bibinfo{author}{Alegria, I.} (\bibinfo{year}{2020}). \bibinfo{title}{Measuring language distance of isolated european languages}.
\newblock \bibinfo{journal}{Information} \emph{\bibinfo{volume}{11}}, \bibinfo{pages}{181}. \DOIprefix\doi{https://doi.org/10.3390/info11040181}.
\bibitem[{Fragkou(2014)}]{fragkou2014text}
\bibinfo{author}{Fragkou, P.} (\bibinfo{year}{2014}). \bibinfo{title}{Text segmentation for language identification in greek forums}.
\newblock \bibinfo{journal}{Procedia-Social and Behavioral Sciences} \emph{\bibinfo{volume}{147}}, \bibinfo{pages}{160--166}. \DOIprefix\doi{https://doi.org/10.1016/j.sbspro.2014.07.140}.
\bibitem[{Bollegala et~al.(2015{\natexlab{a}})Bollegala, Kontonatsios and Ananiadou}]{bollegala2015cross}
\bibinfo{author}{Bollegala, D.}, \bibinfo{author}{Kontonatsios, G.}, and \bibinfo{author}{Ananiadou, S.} (\bibinfo{year}{2015}{\natexlab{a}}). \bibinfo{title}{A cross-lingual similarity measure for detecting biomedical term translations}.
\newblock \bibinfo{journal}{PloS one} \emph{\bibinfo{volume}{10}}, \bibinfo{pages}{e0126196}. \DOIprefix\doi{https://doi.org/10.1371/journal.pone.0126196}.
\bibitem[{Conneau et~al.(2020{\natexlab{b}})Conneau, Khandelwal, Goyal, Chaudhary, Wenzek, Guzm{\'a}n, Grave, Ott, Zettlemoyer and Stoyanov}]{conneau2020unsupervised}
\bibinfo{author}{Conneau, A.}, \bibinfo{author}{Khandelwal, K.}, \bibinfo{author}{Goyal, N.}, \bibinfo{author}{Chaudhary, V.}, \bibinfo{author}{Wenzek, G.}, \bibinfo{author}{Guzm{\'a}n, F.}, \bibinfo{author}{Grave, {\'E}.}, \bibinfo{author}{Ott, M.}, \bibinfo{author}{Zettlemoyer, L.}, and \bibinfo{author}{Stoyanov, V.} (\bibinfo{year}{2020}{\natexlab{b}}). \bibinfo{title}{Unsupervised cross-lingual representation learning at scale}.
\newblock In \bibinfo{booktitle}{Proceedings of the 58th Annual Meeting of the Association for Computational Linguistics}. pp. \bibinfo{pages}{8440--8451}.
\bibitem[{Pfeiffer et~al.(2020)Pfeiffer, Vuli{\'c}, Gurevych and Ruder}]{pfeiffer2020mad}
\bibinfo{author}{Pfeiffer, J.}, \bibinfo{author}{Vuli{\'c}, I.}, \bibinfo{author}{Gurevych, I.}, and \bibinfo{author}{Ruder, S.} (\bibinfo{year}{2020}). \bibinfo{title}{Mad-x: An adapter-based framework for multi-task cross-lingual transfer}.
\newblock \bibinfo{journal}{Preprint at arXiv \url{https://doi.org/10.48550/arXiv.2005.00052}}.
\bibitem[{Papaioannou et~al.(2022{\natexlab{b}})Papaioannou, Grundmann, van Aken, Samaras, Kyparissidis, Giannakoulas, Gers and Loeser}]{neuron1682_crosslingual}
\bibinfo{author}{Papaioannou, J.M.}, \bibinfo{author}{Grundmann, P.}, \bibinfo{author}{van Aken, B.}, \bibinfo{author}{Samaras, A.}, \bibinfo{author}{Kyparissidis, I.}, \bibinfo{author}{Giannakoulas, G.}, \bibinfo{author}{Gers, F.}, and \bibinfo{author}{Loeser, A.} (\bibinfo{year}{2022}{\natexlab{b}}).
\newblock \bibinfo{title}{Cross-lingual knowledge transfer for clinical phenotyping}. \bibinfo{publisher}{GitHub}{\natexlab{b}}.
\newblock \bibinfo{note}{\url{https://github.com/neuron1682/cross-lingual-phenotype-prediction/tree/main}}.
\bibitem[{Singh et~al.(2019)Singh, McCann, Keskar, Xiong and Socher}]{Singh2019}
\bibinfo{author}{Singh, J.}, \bibinfo{author}{McCann, B.}, \bibinfo{author}{Keskar, N.S.}, \bibinfo{author}{Xiong, C.}, and \bibinfo{author}{Socher, R.} (\bibinfo{year}{2019}). \bibinfo{title}{Xlda: Cross-lingual data augmentation for natural language inference and question answering}.
\newblock \bibinfo{journal}{Preprint at arXiv \url{https://doi.org/10.48550/arxiv.1905.11471}}.
\bibitem[{Papadopoulos et~al.(2021{\natexlab{c}})Papadopoulos, Papadakis and Matsatsinis}]{lighteternal_hf}
\bibinfo{author}{Papadopoulos, D.}, \bibinfo{author}{Papadakis, N.}, and \bibinfo{author}{Matsatsinis, N.F.} (\bibinfo{year}{2021}{\natexlab{c}}).
\newblock \bibinfo{title}{lighteternal models}. \bibinfo{publisher}{Hugging Face}{\natexlab{c}}.
\newblock \bibinfo{note}{\url{https://huggingface.co/lighteternal}}.
\bibitem[{Prokopidis et~al.(2016)Prokopidis, Papavassiliou and Piperidis}]{prokopidis-2016-parallel}
\bibinfo{author}{Prokopidis, P.}, \bibinfo{author}{Papavassiliou, V.}, and \bibinfo{author}{Piperidis, S.} (\bibinfo{year}{2016}). \bibinfo{title}{Parallel {G}lobal {V}oices: a collection of multilingual corpora with citizen media stories}.
\newblock In \bibinfo{booktitle}{Proceedings of the Tenth International Conference on Language Resources and Evaluation ({LREC}'16)}. \bibinfo{address}{Portoro{\v{z}}, Slovenia}: \bibinfo{publisher}{European Language Resources Association (ELRA)} pp. \bibinfo{pages}{900--905}.
\bibitem[{{Global Voices}(????)}]{global_voices}
\bibinfo{author}{{Global Voices}}.
\newblock \bibinfo{title}{Global voices}.
\newblock \bibinfo{note}{\url{https://globalvoices.org/}}.
\bibitem[{Prokopidis et~al.(????)Prokopidis, Papavassiliou and Piperidis}]{pgv_ilsp}
\bibinfo{author}{Prokopidis, P.}, \bibinfo{author}{Papavassiliou, V.}, and \bibinfo{author}{Piperidis, S.}
\newblock \bibinfo{title}{Pgv datasets)}.
\newblock \bibinfo{note}{\url{http://nlp.ilsp.gr/pgv/}}.
\bibitem[{Bollegala et~al.(2015{\natexlab{b}})Bollegala, Kontonatsios and Ananiadou}]{plos_one_supplementary_material}
\bibinfo{author}{Bollegala, D.}, \bibinfo{author}{Kontonatsios, G.}, and \bibinfo{author}{Ananiadou, S.} (\bibinfo{year}{2015}{\natexlab{b}}).
\newblock \bibinfo{title}{S1 dataset}. \bibinfo{publisher}{PLOS ONE}{\natexlab{b}}.
\newblock \bibinfo{note}{\url{https://doi.org/10.1371/journal.pone.0126196.s001}}.
\bibitem[{{Hugging Face, Inc.}(????)}]{hugging_face}
\bibinfo{author}{{Hugging Face, Inc.}}
\newblock \bibinfo{title}{Hugging face}.
\newblock \bibinfo{note}{\url{https://huggingface.co}}.
\bibitem[{{CLARIN:EL}(????)}]{clarin_greece_inventory}
\bibinfo{author}{{CLARIN:EL}}.
\newblock \bibinfo{title}{Clarin:el research infrastructure for language resources \& technologies}. \bibinfo{publisher}{CLARIN:EL}.
\newblock \bibinfo{note}{\url{https://inventory.clarin.gr/}}.
\bibitem[{{Hugging Face}(????{\natexlab{b}})}]{huggingface_datasets_translation}
\bibinfo{author}{{Hugging Face}}.
\newblock \bibinfo{title}{Hugging face datasets: Greek translation}.
\newblock \bibinfo{note}{\url{https://huggingface.co/datasets?task_categories=task_categories:translation&language=language:el&sort=downloads}}.
\bibitem[{{Hugging Face}(????{\natexlab{c}})}]{huggingface_models_translation}
\bibinfo{author}{{Hugging Face}}.
\newblock \bibinfo{title}{Hugging face models: Greek translation}.
\newblock \bibinfo{note}{\url{https://huggingface.co/models?pipeline_tag=translation&language=el&sort=downloads}}.
\bibitem[{{Hugging Face}(????{\natexlab{d}})}]{huggingface_datasets}
\bibinfo{author}{{Hugging Face}}.
\newblock \bibinfo{title}{Hugging face datasets}.
\newblock \bibinfo{note}{\url{https://huggingface.co/datasets}}.
\bibitem[{Dabre et~al.(2020)Dabre, Chu and Kunchukuttan}]{dabre2020survey}
\bibinfo{author}{Dabre, R.}, \bibinfo{author}{Chu, C.}, and \bibinfo{author}{Kunchukuttan, A.} (\bibinfo{year}{2020}). \bibinfo{title}{A survey of multilingual neural machine translation}.
\newblock \bibinfo{journal}{ACM Computing Surveys (CSUR)} \emph{\bibinfo{volume}{53}}, \bibinfo{pages}{1--38}. \DOIprefix\doi{https://doi.org/10.1145/3406095}.
\bibitem[{Haddow et~al.(2022)Haddow, Bawden, Barone, Helcl and Birch}]{haddow2022survey}
\bibinfo{author}{Haddow, B.}, \bibinfo{author}{Bawden, R.}, \bibinfo{author}{Barone, A.V.M.}, \bibinfo{author}{Helcl, J.}, and \bibinfo{author}{Birch, A.} (\bibinfo{year}{2022}). \bibinfo{title}{Survey of low-resource machine translation}.
\newblock \bibinfo{journal}{Computational Linguistics} \emph{\bibinfo{volume}{48}}, \bibinfo{pages}{673--732}. \DOIprefix\doi{https://doi.org/10.1162/coli_a_00446}.
\bibitem[{Papaloukas et~al.(2021{\natexlab{a}})Papaloukas, Chalkidis, Athinaios, Pantazi and Koubarakis}]{papaloukas2021}
\bibinfo{author}{Papaloukas, C.}, \bibinfo{author}{Chalkidis, I.}, \bibinfo{author}{Athinaios, K.}, \bibinfo{author}{Pantazi, D.}, and \bibinfo{author}{Koubarakis, M.} (\bibinfo{year}{2021}{\natexlab{a}}). \bibinfo{title}{Multi-granular legal topic classification on {G}reek legislation}.
\newblock In \bibinfo{booktitle}{Proceedings of the Natural Legal Language Processing Workshop 2021}. \bibinfo{address}{Punta Cana, Dominican Republic}: \bibinfo{publisher}{Association for Computational Linguistics}{\natexlab{a}} pp. \bibinfo{pages}{63--75}.
\bibitem[{Lachana et~al.(2020)Lachana, Loutsaris, Alexopoulos and Charalabidis}]{Lachana2020}
\bibinfo{author}{Lachana, Z.}, \bibinfo{author}{Loutsaris, M.A.}, \bibinfo{author}{Alexopoulos, C.}, and \bibinfo{author}{Charalabidis, Y.} (\bibinfo{year}{2020}). \bibinfo{title}{Automated analysis and interrelation of legal elements based on text mining}.
\newblock \bibinfo{journal}{International Journal of E-Services and Mobile Applications (IJESMA)} \emph{\bibinfo{volume}{12}}, \bibinfo{pages}{79--96}. \DOIprefix\doi{10.4018/IJESMA.2020040105}.
\bibitem[{Garofalakis et~al.(2016{\natexlab{a}})Garofalakis, Plessas and Plessas}]{Garofalakis2016}
\bibinfo{author}{Garofalakis, J.}, \bibinfo{author}{Plessas, K.}, and \bibinfo{author}{Plessas, A.} (\bibinfo{year}{2016}{\natexlab{a}}). \bibinfo{title}{A semi-automatic system for the consolidation of greek legislative texts}.
\newblock In \bibinfo{booktitle}{Proceedings of the 20th Pan-Hellenic Conference on Informatics}. PCI '16. \bibinfo{address}{New York, NY, USA}: \bibinfo{publisher}{Association for Computing Machinery}{\natexlab{a}}.
\newblock ISBN \bibinfo{isbn}{9781450347891} pp. \bibinfo{pages}{1--6}.
\bibitem[{Paraskevopoulos et~al.(2022)Paraskevopoulos, Pistofidis, Banoutsos, Georgiou and Katsouros}]{paraskevopoulos2022multimodal}
\bibinfo{author}{Paraskevopoulos, G.}, \bibinfo{author}{Pistofidis, P.}, \bibinfo{author}{Banoutsos, G.}, \bibinfo{author}{Georgiou, E.}, and \bibinfo{author}{Katsouros, V.} (\bibinfo{year}{2022}). \bibinfo{title}{Multimodal classification of safety-report observations}.
\newblock \bibinfo{journal}{Applied Sciences} \emph{\bibinfo{volume}{12}}, \bibinfo{pages}{5781}. \DOIprefix\doi{https://doi.org/10.3390/app12125781}.
\bibitem[{Boskou et~al.(2018)Boskou, Kirkos and Spathis}]{boskou2018assessing}
\bibinfo{author}{Boskou, G.}, \bibinfo{author}{Kirkos, E.}, and \bibinfo{author}{Spathis, C.} (\bibinfo{year}{2018}). \bibinfo{title}{Assessing internal audit with text mining}.
\newblock \bibinfo{journal}{Journal of Information \& Knowledge Management} \emph{\bibinfo{volume}{17}}, \bibinfo{pages}{1850020}. \DOIprefix\doi{https://doi.org/10.1142/S021964921850020X}.
\bibitem[{Chatzipanagiotidis et~al.(2021)Chatzipanagiotidis, Giagkou and Meurers}]{chatzipanagiotidis2021broad}
\bibinfo{author}{Chatzipanagiotidis, S.}, \bibinfo{author}{Giagkou, M.}, and \bibinfo{author}{Meurers, D.} (\bibinfo{year}{2021}). \bibinfo{title}{Broad linguistic complexity analysis for greek readability classification}.
\newblock In \bibinfo{booktitle}{Proceedings of the 16th Workshop on Innovative Use of NLP for Building Educational Applications}. pp. \bibinfo{pages}{48--58}.
\bibitem[{Piskorski et~al.(2023)Piskorski, Stefanovitch, Da~San~Martino and Nakov}]{piskorski2023semeval}
\bibinfo{author}{Piskorski, J.}, \bibinfo{author}{Stefanovitch, N.}, \bibinfo{author}{Da~San~Martino, G.}, and \bibinfo{author}{Nakov, P.} (\bibinfo{year}{2023}). \bibinfo{title}{Semeval-2023 task 3: Detecting the category, the framing, and the persuasion techniques in online news in a multi-lingual setup}.
\newblock In \bibinfo{booktitle}{Proceedings of the 17th International Workshop on Semantic Evaluation (SemEval-2023)}. pp. \bibinfo{pages}{2343--2361}.
\bibitem[{Athanasiou et~al.(2023)Athanasiou, Fragkozidis, Zarkogianni and Nikita}]{athanasiou2023long}
\bibinfo{author}{Athanasiou, M.}, \bibinfo{author}{Fragkozidis, G.}, \bibinfo{author}{Zarkogianni, K.}, and \bibinfo{author}{Nikita, K.S.} (\bibinfo{year}{2023}). \bibinfo{title}{Long short-term memory--based prediction of the spread of influenza-like illness leveraging surveillance, weather, and twitter data: Model development and validation}.
\newblock \bibinfo{journal}{Journal of Medical Internet Research} \emph{\bibinfo{volume}{25}}, \bibinfo{pages}{e42519}. \DOIprefix\doi{https://doi.org/10.2196/42519}.
\bibitem[{Stamouli et~al.(2023)Stamouli, Nerantzini, Papakyritsis, Katsamanis, Chatzoudis, Dimou, Plitsis, Katsouros, Varlokosta and Terzi}]{stamouli2023web}
\bibinfo{author}{Stamouli, S.}, \bibinfo{author}{Nerantzini, M.}, \bibinfo{author}{Papakyritsis, I.}, \bibinfo{author}{Katsamanis, A.}, \bibinfo{author}{Chatzoudis, G.}, \bibinfo{author}{Dimou, A.L.}, \bibinfo{author}{Plitsis, M.}, \bibinfo{author}{Katsouros, V.}, \bibinfo{author}{Varlokosta, S.}, and \bibinfo{author}{Terzi, A.} (\bibinfo{year}{2023}). \bibinfo{title}{A web-based application for eliciting narrative discourse from greek-speaking people with and without language impairments}.
\newblock \bibinfo{journal}{Frontiers in Communication} \emph{\bibinfo{volume}{8}}, \bibinfo{pages}{919617}. \DOIprefix\doi{https://doi.org/10.3389/fcomm.2023.919617}.
\bibitem[{{Hellenic Republic}(????)}]{raptarchis}
\bibinfo{author}{{Hellenic Republic}}.
\newblock \bibinfo{title}{Greek government gazette}.
\newblock \bibinfo{note}{\url{https://raptarchis.gov.gr}}.
\bibitem[{Papaloukas et~al.(2021{\natexlab{b}})Papaloukas, Chalkidis, Athinaios, Pantazi and Koubarakis}]{greek_legal_code_dataset}
\bibinfo{author}{Papaloukas, C.}, \bibinfo{author}{Chalkidis, I.}, \bibinfo{author}{Athinaios, K.}, \bibinfo{author}{Pantazi, D.}, and \bibinfo{author}{Koubarakis, M.} (\bibinfo{year}{2021}{\natexlab{b}}).
\newblock \bibinfo{title}{Greek legal code dataset}. \bibinfo{publisher}{Hugging Face}{\natexlab{b}}.
\newblock \bibinfo{note}{\url{https://huggingface.co/datasets/greek_legal_code}}.
\bibitem[{Garofalakis et~al.(2016{\natexlab{b}})Garofalakis, Plessas and Plessas}]{openlawsgr_greek_laws}
\bibinfo{author}{Garofalakis, J.}, \bibinfo{author}{Plessas, K.}, and \bibinfo{author}{Plessas, A.} (\bibinfo{year}{2016}{\natexlab{b}}).
\newblock \bibinfo{title}{Greek laws alpha}. \bibinfo{publisher}{GitHub}{\natexlab{b}}.
\newblock \bibinfo{note}{\url{https://github.com/OpenLawsGR/greek_laws_alpha}}.
\bibitem[{Liu et~al.(2021)Liu, Sun, Jiang, Jiang and Ming}]{liu2021multi}
\bibinfo{author}{Liu, Z.}, \bibinfo{author}{Sun, C.}, \bibinfo{author}{Jiang, Y.}, \bibinfo{author}{Jiang, S.}, and \bibinfo{author}{Ming, M.} (\bibinfo{year}{2021}). \bibinfo{title}{Multi-modal application: Image memes generation}.
\newblock \bibinfo{journal}{Preprint at arXiv \url{https://doi.org/10.48550/arXiv.2112.01651}}.
\bibitem[{Das et~al.(2016)Das, Das and Mahesh}]{das2016computational}
\bibinfo{author}{Das, D.}, \bibinfo{author}{Das, B.}, and \bibinfo{author}{Mahesh, K.} (\bibinfo{year}{2016}). \bibinfo{title}{A computational analysis of mahabharata}.
\newblock In \bibinfo{booktitle}{Proceedings of the 13th International Conference on Natural Language Processing}. pp. \bibinfo{pages}{219--228}.
\bibitem[{Escart{\'\i}n et~al.(2017)Escart{\'\i}n, Reijers, Lynn, Moorkens, Way and Liu}]{escartin2017ethical}
\bibinfo{author}{Escart{\'\i}n, C.P.}, \bibinfo{author}{Reijers, W.}, \bibinfo{author}{Lynn, T.}, \bibinfo{author}{Moorkens, J.}, \bibinfo{author}{Way, A.}, and \bibinfo{author}{Liu, C.H.} (\bibinfo{year}{2017}). \bibinfo{title}{Ethical considerations in nlp shared tasks}.
\newblock In \bibinfo{booktitle}{Proceedings of the First Workshop on Ethics in Natural Language Processing}. \bibinfo{address}{Valencia, Spain}: \bibinfo{publisher}{Association for Computational Linguistics} pp. \bibinfo{pages}{66--73}.
\newblock \DOIprefix\doi{10.18653/v1/W17-1608}.
\bibitem[{{GitHub, Inc.}(????)}]{github}
\bibinfo{author}{{GitHub, Inc.}}
\newblock \bibinfo{title}{Github}.
\newblock \bibinfo{note}{\url{https://github.com/}}.
\bibitem[{{CERN / OpenAIRE}(????)}]{zenodo}
\bibinfo{author}{{CERN / OpenAIRE}}.
\newblock \bibinfo{title}{Zenodo}.
\newblock \bibinfo{note}{\url{https://zenodo.org/}}.
\bibitem[{Fitsilis and Mikros(2021{\natexlab{a}})}]{fitsilis2021development}
\bibinfo{author}{Fitsilis, F.}, and \bibinfo{author}{Mikros, G.} (\bibinfo{year}{2021}{\natexlab{a}}). \bibinfo{title}{Development and validation of a corpus of written parliamentary questions in the hellenic parliament}.
\newblock \bibinfo{journal}{Journal of Open Humanities Data} \emph{\bibinfo{volume}{7}}, \bibinfo{pages}{18}. \DOIprefix\doi{https://doi.org/10.5334/johd.45}.
\bibitem[{Cao et~al.(2023)Cao, Dodge, Lo, McFarland and Wang}]{cao2023rise}
\bibinfo{author}{Cao, H.}, \bibinfo{author}{Dodge, J.}, \bibinfo{author}{Lo, K.}, \bibinfo{author}{McFarland, D.A.}, and \bibinfo{author}{Wang, L.L.} (\bibinfo{year}{2023}). \bibinfo{title}{The rise of open science: Tracking the evolution and perceived value of data and methods link-sharing practices}.
\newblock \bibinfo{journal}{Preprint at arXiv \url{https://doi.org/10.48550/arXiv.2310.03193}}.
\bibitem[{Dritsa et~al.(2022{\natexlab{b}})Dritsa, Thoma, Pavlopoulos and Louridas}]{greek_parliament_proceedings_dataset}
\bibinfo{author}{Dritsa, K.}, \bibinfo{author}{Thoma, K.}, \bibinfo{author}{Pavlopoulos, J.}, and \bibinfo{author}{Louridas, P.} (\bibinfo{year}{2022}{\natexlab{b}}).
\newblock \bibinfo{title}{A greek parliament proceedings dataset for computational linguistics and political analysis}. \bibinfo{publisher}{Zenodo}{\natexlab{b}}.
\newblock \bibinfo{note}{\url{https://doi.org/10.5281/zenodo.6626315}}.
\bibitem[{Fitsilis and Mikros(2021{\natexlab{b}})}]{parliamentary_questions_corpus}
\bibinfo{author}{Fitsilis, F.}, and \bibinfo{author}{Mikros, G.} (\bibinfo{year}{2021}{\natexlab{b}}).
\newblock \bibinfo{title}{Corpus of parliamentary questions in the hellenic parliament}. \bibinfo{publisher}{Zenodo}{\natexlab{b}}.
\newblock \bibinfo{note}{\url{https://doi.org/10.5281/zenodo.4747451}}.
\bibitem[{Prokopidis and Piperidis(2020{\natexlab{b}})}]{neural_nlp_toolkit_greek}
\bibinfo{author}{Prokopidis, P.}, and \bibinfo{author}{Piperidis, S.} (\bibinfo{year}{2020}{\natexlab{b}}).
\newblock \bibinfo{title}{Resources for the paper “a neural nlp toolkit for greek” (setn-2020)}. \bibinfo{publisher}{Institute for Language and Speech Processing (ILSP)}{\natexlab{b}}.
\newblock \bibinfo{note}{\url{http://nlp.ilsp.gr/setn-2020/}}.
\bibitem[{Barzokas et~al.(2024)Barzokas, Papagiannopoulou and Tsoumakas}]{greek_words_evolution}
\bibinfo{author}{Barzokas, V.}, \bibinfo{author}{Papagiannopoulou, E.}, and \bibinfo{author}{Tsoumakas, G.} (\bibinfo{year}{2024}).
\newblock \bibinfo{title}{Time-stamped greek corpus}. \bibinfo{publisher}{GitHub}.
\newblock \bibinfo{note}{\url{https://github.com/intelligence-csd-auth-gr/greek-words-evolution.git}}.
\bibitem[{Iosif et~al.(2016{\natexlab{b}})Iosif, Georgiladakis and Potamianos}]{adsms_actindex}
\bibinfo{author}{Iosif, E.}, \bibinfo{author}{Georgiladakis, S.}, and \bibinfo{author}{Potamianos, A.} (\bibinfo{year}{2016}{\natexlab{b}}).
\newblock \bibinfo{title}{Greek web document snippets dataset}. \bibinfo{publisher}{Technical University of Crete}{\natexlab{b}}.
\newblock \bibinfo{note}{\url{http://www.telecom.tuc.gr/~iosife/downloads/adsms/actindex.html/}}.
\bibitem[{Majli{\v{s}} and {\v{Z}}abokrtsk{\`y}(2012)}]{majlivs2012language}
\bibinfo{author}{Majli{\v{s}}, M.}, and \bibinfo{author}{{\v{Z}}abokrtsk{\`y}, Z.} (\bibinfo{year}{2012}). \bibinfo{title}{Language richness of the web}.
\newblock In \bibinfo{booktitle}{Proceedings of the Eighth International Conference on Language Resources and Evaluation (LREC'12)}. pp. \bibinfo{pages}{2927--2934}.
\bibitem[{{Project Gutenberg}(????)}]{project_gutenberg}
\bibinfo{author}{{Project Gutenberg}}.
\newblock \bibinfo{title}{Project gutenberg}.
\newblock \bibinfo{note}{\url{https://www.gutenberg.org/}}.
\bibitem[{{Openbook.gr Community}(????)}]{openbook_gr}
\bibinfo{author}{{Openbook.gr Community}}.
\newblock \bibinfo{title}{Openbook.gr}.
\newblock \bibinfo{note}{\url{https://www.openbook.gr/}}.
\bibitem[{{Center for the Greek Language}(2007)}]{makedonia_corpus}
\bibinfo{author}{{Center for the Greek Language}} (\bibinfo{year}{2007}).
\newblock \bibinfo{title}{Corpus of the newspaper “macedonia”}. \bibinfo{publisher}{Center for the Greek Language}.
\newblock \bibinfo{note}{\url{http://www.greek-language.gr/greekLang/modern_greek/tools/corpora/makedonia/content.html}}.
\bibitem[{Fokides and Peristeraki(2025)}]{fokides2025comparing}
\bibinfo{author}{Fokides, E.}, and \bibinfo{author}{Peristeraki, E.} (\bibinfo{year}{2025}). \bibinfo{title}{Comparing chatgpt's correction and feedback comments with that of educators in the context of primary students' short essays written in english and greek}.
\newblock \bibinfo{journal}{Education and Information Technologies} \emph{\bibinfo{volume}{30}}, \bibinfo{pages}{2577--2621}. \DOIprefix\doi{https://doi.org/10.1007/s10639-024-12912-8}.
\bibitem[{Bakagianni et~al.(2025)Bakagianni, Pouli, Gavriilidou and Pavlopoulos}]{greek_nlp_survey_2025}
\bibinfo{author}{Bakagianni, J.}, \bibinfo{author}{Pouli, K.}, \bibinfo{author}{Gavriilidou, M.}, and \bibinfo{author}{Pavlopoulos, J.} (\bibinfo{year}{2025}).
\newblock \bibinfo{title}{greek-nlp/survey: A systematic survey of natural language processing research for the greek language}. \bibinfo{publisher}{Zenodo}.
\newblock \URLprefix \url{https://doi.org/10.5281/zenodo.15314882}.
\bibitem[{Khurana et~al.(2023)Khurana, Koli, Khatter and Singh}]{khurana2023natural}
\bibinfo{author}{Khurana, D.}, \bibinfo{author}{Koli, A.}, \bibinfo{author}{Khatter, K.}, and \bibinfo{author}{Singh, S.} (\bibinfo{year}{2023}). \bibinfo{title}{Natural language processing: State of the art, current trends and challenges}.
\newblock \bibinfo{journal}{Multimedia tools and applications} \emph{\bibinfo{volume}{82}}, \bibinfo{pages}{3713--3744}. \DOIprefix\doi{https://doi.org/10.1007/s11042-022-13428-4}.
\bibitem[{Wang et~al.(2018)Wang, Singh, Michael, Hill, Levy and Bowman}]{wang2019glue}
\bibinfo{author}{Wang, A.}, \bibinfo{author}{Singh, A.}, \bibinfo{author}{Michael, J.}, \bibinfo{author}{Hill, F.}, \bibinfo{author}{Levy, O.}, and \bibinfo{author}{Bowman, S.} (\bibinfo{year}{2018}). \bibinfo{title}{{GLUE}: A multi-task benchmark and analysis platform for natural language understanding}.
\newblock In \bibinfo{editor}{ T.{ }Linzen}, \bibinfo{editor}{ G.{ }Chrupa{\l}a}, and \bibinfo{editor}{ A.{ }Alishahi}, eds. \bibinfo{booktitle}{Proceedings of the 2018 {EMNLP} Workshop {B}lackbox{NLP}: Analyzing and Interpreting Neural Networks for {NLP}}. \bibinfo{address}{Brussels, Belgium}: \bibinfo{publisher}{Association for Computational Linguistics} pp. \bibinfo{pages}{353--355}.
\newblock \DOIprefix\doi{10.18653/v1/W18-5446}.
\bibitem[{Bowman and Dahl(2021)}]{bowman-dahl-2021-will}
\bibinfo{author}{Bowman, S.R.}, and \bibinfo{author}{Dahl, G.} (\bibinfo{year}{2021}). \bibinfo{title}{What will it take to fix benchmarking in natural language understanding?}
\newblock In \bibinfo{editor}{ K.{ }Toutanova}, \bibinfo{editor}{ A.{ }Rumshisky}, \bibinfo{editor}{ L.{ }Zettlemoyer}, \bibinfo{editor}{ D.{ }Hakkani-Tur}, \bibinfo{editor}{ I.{ }Beltagy}, \bibinfo{editor}{ S.{ }Bethard}, \bibinfo{editor}{ R.{ }Cotterell}, \bibinfo{editor}{ T.{ }Chakraborty}, and \bibinfo{editor}{ Y.{ }Zhou}, eds. \bibinfo{booktitle}{Proceedings of the 2021 Conference of the North American Chapter of the Association for Computational Linguistics: Human Language Technologies}. \bibinfo{address}{Online}: \bibinfo{publisher}{Association for Computational Linguistics} pp. \bibinfo{pages}{4843--4855}.
\newblock \DOIprefix\doi{10.18653/v1/2021.naacl-main.385}.
\bibitem[{Chen and Gao(2022)}]{chen-gao-2022-curriculum}
\bibinfo{author}{Chen, Z.}, and \bibinfo{author}{Gao, Q.} (\bibinfo{year}{2022}). \bibinfo{title}{Curriculum: A broad-coverage benchmark for linguistic phenomena in natural language understanding}.
\newblock In \bibinfo{editor}{ M.{ }Carpuat}, \bibinfo{editor}{de~ M.C.{ }Marneffe}, and \bibinfo{editor}{ I.V.{ }Meza~Ruiz}, eds. \bibinfo{booktitle}{Proceedings of the 2022 Conference of the North American Chapter of the Association for Computational Linguistics: Human Language Technologies}. \bibinfo{address}{Seattle, United States}: \bibinfo{publisher}{Association for Computational Linguistics} pp. \bibinfo{pages}{3204--3219}.
\newblock \DOIprefix\doi{10.18653/v1/2022.naacl-main.234}.
\bibitem[{Wang et~al.(2019)Wang, Pruksachatkun, Nangia, Singh, Michael, Hill, Levy and Bowman}]{wang2019superglue}
\bibinfo{author}{Wang, A.}, \bibinfo{author}{Pruksachatkun, Y.}, \bibinfo{author}{Nangia, N.}, \bibinfo{author}{Singh, A.}, \bibinfo{author}{Michael, J.}, \bibinfo{author}{Hill, F.}, \bibinfo{author}{Levy, O.}, and \bibinfo{author}{Bowman, S.} (\bibinfo{year}{2019}). \bibinfo{title}{Superglue: A stickier benchmark for general-purpose language understanding systems}.
\newblock \bibinfo{journal}{Advances in neural information processing systems} \emph{\bibinfo{volume}{32}}.
\bibitem[{Dong et~al.(2019)Dong, Yang, Wang, Wei, Liu, Wang, Gao, Zhou and Hon}]{dong2019unified}
\bibinfo{author}{Dong, L.}, \bibinfo{author}{Yang, N.}, \bibinfo{author}{Wang, W.}, \bibinfo{author}{Wei, F.}, \bibinfo{author}{Liu, X.}, \bibinfo{author}{Wang, Y.}, \bibinfo{author}{Gao, J.}, \bibinfo{author}{Zhou, M.}, and \bibinfo{author}{Hon, H.W.} (\bibinfo{year}{2019}). \bibinfo{title}{Unified language model pre-training for natural language understanding and generation}.
\newblock \bibinfo{journal}{Advances in neural information processing systems} \emph{\bibinfo{volume}{32}}.
\bibitem[{Nie et~al.(2019)Nie, Williams, Dinan, Bansal, Weston and Kiela}]{nie2019adversarial}
\bibinfo{author}{Nie, Y.}, \bibinfo{author}{Williams, A.}, \bibinfo{author}{Dinan, E.}, \bibinfo{author}{Bansal, M.}, \bibinfo{author}{Weston, J.}, and \bibinfo{author}{Kiela, D.} (\bibinfo{year}{2019}). \bibinfo{title}{Adversarial nli: A new benchmark for natural language understanding}.
\newblock \bibinfo{journal}{Preprint at arXiv \url{https://doi.org/10.48550/arXiv.1910.14599}}.
\bibitem[{Urbizu et~al.(2022)Urbizu, San~Vicente, Saralegi, Agerri and Soroa}]{urbizu2022basqueglue}
\bibinfo{author}{Urbizu, G.}, \bibinfo{author}{San~Vicente, I.}, \bibinfo{author}{Saralegi, X.}, \bibinfo{author}{Agerri, R.}, and \bibinfo{author}{Soroa, A.} (\bibinfo{year}{2022}). \bibinfo{title}{Basqueglue: A natural language understanding benchmark for basque}.
\newblock In \bibinfo{booktitle}{Proceedings of the Thirteenth Language Resources and Evaluation Conference}. pp. \bibinfo{pages}{1603--1612}.
\bibitem[{Shavrina et~al.(2020)Shavrina, Fenogenova, Anton, Shevelev, Artemova, Malykh, Mikhailov, Tikhonova, Chertok and Evlampiev}]{shavrina-etal-2020-russiansuperglue}
\bibinfo{author}{Shavrina, T.}, \bibinfo{author}{Fenogenova, A.}, \bibinfo{author}{Anton, E.}, \bibinfo{author}{Shevelev, D.}, \bibinfo{author}{Artemova, E.}, \bibinfo{author}{Malykh, V.}, \bibinfo{author}{Mikhailov, V.}, \bibinfo{author}{Tikhonova, M.}, \bibinfo{author}{Chertok, A.}, and \bibinfo{author}{Evlampiev, A.} (\bibinfo{year}{2020}). \bibinfo{title}{{R}ussian{S}uper{GLUE}: A {R}ussian language understanding evaluation benchmark}.
\newblock In \bibinfo{editor}{ B.{ }Webber}, \bibinfo{editor}{ T.{ }Cohn}, \bibinfo{editor}{ Y.{ }He}, and \bibinfo{editor}{ Y.{ }Liu}, eds. \bibinfo{booktitle}{Proceedings of the 2020 Conference on Empirical Methods in Natural Language Processing (EMNLP)}. \bibinfo{address}{Online}: \bibinfo{publisher}{Association for Computational Linguistics} pp. \bibinfo{pages}{4717--4726}.
\newblock \DOIprefix\doi{10.18653/v1/2020.emnlp-main.381}.
\bibitem[{Wilie et~al.(2020)Wilie, Vincentio, Winata, Cahyawijaya, Li, Lim, Soleman, Mahendra, Fung, Bahar and Purwarianti}]{wilie-etal-2020-indonlu}
\bibinfo{author}{Wilie, B.}, \bibinfo{author}{Vincentio, K.}, \bibinfo{author}{Winata, G.I.}, \bibinfo{author}{Cahyawijaya, S.}, \bibinfo{author}{Li, X.}, \bibinfo{author}{Lim, Z.Y.}, \bibinfo{author}{Soleman, S.}, \bibinfo{author}{Mahendra, R.}, \bibinfo{author}{Fung, P.}, \bibinfo{author}{Bahar, S.}, and \bibinfo{author}{Purwarianti, A.} (\bibinfo{year}{2020}). \bibinfo{title}{{I}ndo{NLU}: Benchmark and resources for evaluating {I}ndonesian natural language understanding}.
\newblock In \bibinfo{editor}{ K.F.{ }Wong}, \bibinfo{editor}{ K.{ }Knight}, and \bibinfo{editor}{ H.{ }Wu}, eds. \bibinfo{booktitle}{Proceedings of the 1st Conference of the Asia-Pacific Chapter of the Association for Computational Linguistics and the 10th International Joint Conference on Natural Language Processing}. \bibinfo{address}{Suzhou, China}: \bibinfo{publisher}{Association for Computational Linguistics} pp. \bibinfo{pages}{843--857}.
\bibitem[{Xu et~al.(2020)Xu, Hu, Zhang, Li, Cao, Li, Xu, Sun, Yu, Yu et~al.}]{xu2020clue}
\bibinfo{author}{Xu, L.}, \bibinfo{author}{Hu, H.}, \bibinfo{author}{Zhang, X.}, \bibinfo{author}{Li, L.}, \bibinfo{author}{Cao, C.}, \bibinfo{author}{Li, Y.}, \bibinfo{author}{Xu, Y.}, \bibinfo{author}{Sun, K.}, \bibinfo{author}{Yu, D.}, \bibinfo{author}{Yu, C.} et~al. (\bibinfo{year}{2020}). \bibinfo{title}{Clue: A chinese language understanding evaluation benchmark}.
\newblock In \bibinfo{booktitle}{Proceedings of the 28th International Conference on Computational Linguistics}. pp. \bibinfo{pages}{4762--4772}.
\bibitem[{Rybak et~al.(2020)Rybak, Mroczkowski, Tracz and Gawlik}]{rybak-etal-2020-klej}
\bibinfo{author}{Rybak, P.}, \bibinfo{author}{Mroczkowski, R.}, \bibinfo{author}{Tracz, J.}, and \bibinfo{author}{Gawlik, I.} (\bibinfo{year}{2020}). \bibinfo{title}{{KLEJ}: Comprehensive benchmark for {P}olish language understanding}.
\newblock In \bibinfo{editor}{ D.{ }Jurafsky}, \bibinfo{editor}{ J.{ }Chai}, \bibinfo{editor}{ N.{ }Schluter}, and \bibinfo{editor}{ J.{ }Tetreault}, eds. \bibinfo{booktitle}{Proceedings of the 58th Annual Meeting of the Association for Computational Linguistics}. \bibinfo{address}{Online}: \bibinfo{publisher}{Association for Computational Linguistics} pp. \bibinfo{pages}{1191--1201}.
\newblock \DOIprefix\doi{10.18653/v1/2020.acl-main.111}.
\bibitem[{Ham et~al.(2020)Ham, Choe, Park, Choi and Soh}]{ham2020kornli}
\bibinfo{author}{Ham, J.}, \bibinfo{author}{Choe, Y.J.}, \bibinfo{author}{Park, K.}, \bibinfo{author}{Choi, I.}, and \bibinfo{author}{Soh, H.} (\bibinfo{year}{2020}). \bibinfo{title}{Kornli and korsts: New benchmark datasets for korean natural language understanding}.
\newblock In \bibinfo{booktitle}{Findings of the Association for Computational Linguistics: EMNLP 2020}. pp. \bibinfo{pages}{422--430}.
\bibitem[{Bandarkar et~al.(2023)Bandarkar, Liang, Muller, Artetxe, Shukla, Husa, Goyal, Krishnan, Zettlemoyer and Khabsa}]{bandarkar2023belebele}
\bibinfo{author}{Bandarkar, L.}, \bibinfo{author}{Liang, D.}, \bibinfo{author}{Muller, B.}, \bibinfo{author}{Artetxe, M.}, \bibinfo{author}{Shukla, S.N.}, \bibinfo{author}{Husa, D.}, \bibinfo{author}{Goyal, N.}, \bibinfo{author}{Krishnan, A.}, \bibinfo{author}{Zettlemoyer, L.}, and \bibinfo{author}{Khabsa, M.} (\bibinfo{year}{2023}). \bibinfo{title}{The belebele benchmark: a parallel reading comprehension dataset in 122 language variants}.
\newblock \bibinfo{journal}{Preprint at arXiv \url{https://doi.org/10.48550/arXiv.2308.16884}}.
\bibitem[{{ILSP/Athena RC}(????)}]{ilsp_arc_greek}
\bibinfo{author}{{ILSP/Athena RC}}.
\newblock \bibinfo{title}{Arc greek dataset}. \bibinfo{publisher}{Hugging Face}.
\newblock \bibinfo{note}{\url{https://huggingface.co/datasets/ilsp/arc_greek}}.
\bibitem[{Lewis et~al.(2020{\natexlab{b}})Lewis, Liu, Goyal, Ghazvininejad, Mohamed, Levy, Stoyanov and Zettlemoyer}]{lewis-etal-2020-bart}
\bibinfo{author}{Lewis, M.}, \bibinfo{author}{Liu, Y.}, \bibinfo{author}{Goyal, N.}, \bibinfo{author}{Ghazvininejad, M.}, \bibinfo{author}{Mohamed, A.}, \bibinfo{author}{Levy, O.}, \bibinfo{author}{Stoyanov, V.}, and \bibinfo{author}{Zettlemoyer, L.} (\bibinfo{year}{2020}{\natexlab{b}}). \bibinfo{title}{{BART}: Denoising sequence-to-sequence pre-training for natural language generation, translation, and comprehension}.
\newblock In \bibinfo{booktitle}{Proceedings of the 58th Annual Meeting of the Association for Computational Linguistics}. \bibinfo{address}{Online}: \bibinfo{publisher}{Association for Computational Linguistics}{\natexlab{b}} pp. \bibinfo{pages}{7871--7880}.
\bibitem[{Ji et~al.(2023)Ji, Lee, Frieske, Yu, Su, Xu, Ishii, Bang, Madotto and Fung}]{ji2023survey}
\bibinfo{author}{Ji, Z.}, \bibinfo{author}{Lee, N.}, \bibinfo{author}{Frieske, R.}, \bibinfo{author}{Yu, T.}, \bibinfo{author}{Su, D.}, \bibinfo{author}{Xu, Y.}, \bibinfo{author}{Ishii, E.}, \bibinfo{author}{Bang, Y.J.}, \bibinfo{author}{Madotto, A.}, and \bibinfo{author}{Fung, P.} (\bibinfo{year}{2023}). \bibinfo{title}{Survey of hallucination in natural language generation}.
\newblock \bibinfo{journal}{ACM Computing Surveys} \emph{\bibinfo{volume}{55}}, \bibinfo{pages}{1--38}. \DOIprefix\doi{https://doi.org/10.1145/3571730}.
\bibitem[{Holtzman et~al.(2019)Holtzman, Buys, Du, Forbes and Choi}]{holtzman2019curious}
\bibinfo{author}{Holtzman, A.}, \bibinfo{author}{Buys, J.}, \bibinfo{author}{Du, L.}, \bibinfo{author}{Forbes, M.}, and \bibinfo{author}{Choi, Y.} (\bibinfo{year}{2019}). \bibinfo{title}{The curious case of neural text degeneration}.
\newblock \bibinfo{journal}{Preprint at arXiv \url{https://doi.org/10.48550/arXiv.1904.09751}}.
\bibitem[{Welleck et~al.(2019)Welleck, Kulikov, Roller, Dinan, Cho and Weston}]{welleck2019neural}
\bibinfo{author}{Welleck, S.}, \bibinfo{author}{Kulikov, I.}, \bibinfo{author}{Roller, S.}, \bibinfo{author}{Dinan, E.}, \bibinfo{author}{Cho, K.}, and \bibinfo{author}{Weston, J.} (\bibinfo{year}{2019}). \bibinfo{title}{Neural text generation with unlikelihood training}.
\newblock \bibinfo{journal}{Preprint at arXiv \url{https://doi.org/10.48550/arXiv.1908.04319}}.
\bibitem[{Raunak et~al.(2021)Raunak, Menezes and Junczys-Dowmunt}]{raunak2021curious}
\bibinfo{author}{Raunak, V.}, \bibinfo{author}{Menezes, A.}, and \bibinfo{author}{Junczys-Dowmunt, M.} (\bibinfo{year}{2021}). \bibinfo{title}{The curious case of hallucinations in neural machine translation}.
\newblock \bibinfo{journal}{Preprint at arXiv \url{https://doi.org/10.48550/arXiv.2104.06683}}.
\bibitem[{Rohrbach et~al.(2018)Rohrbach, Hendricks, Burns, Darrell and Saenko}]{rohrbach2018object}
\bibinfo{author}{Rohrbach, A.}, \bibinfo{author}{Hendricks, L.A.}, \bibinfo{author}{Burns, K.}, \bibinfo{author}{Darrell, T.}, and \bibinfo{author}{Saenko, K.} (\bibinfo{year}{2018}). \bibinfo{title}{Object hallucination in image captioning}.
\newblock \bibinfo{journal}{Preprint at arXiv \url{https://doi.org/10.48550/arXiv.1809.02156}}.
\bibitem[{Maynez et~al.(2020)Maynez, Narayan, Bohnet and McDonald}]{maynez2020faithfulness}
\bibinfo{author}{Maynez, J.}, \bibinfo{author}{Narayan, S.}, \bibinfo{author}{Bohnet, B.}, and \bibinfo{author}{McDonald, R.} (\bibinfo{year}{2020}). \bibinfo{title}{On faithfulness and factuality in abstractive summarization}.
\newblock \bibinfo{journal}{Preprint at arXiv \url{https://doi.org/10.48550/arXiv.2005.00661}}.
\bibitem[{Kafetsios and Nezlek(2012)}]{kafetsios2012emotion}
\bibinfo{author}{Kafetsios, K.}, and \bibinfo{author}{Nezlek, J.B.} (\bibinfo{year}{2012}). \bibinfo{title}{Emotion and support perceptions in everyday social interaction: Testing the “less is more” hypothesis in two cultures}.
\newblock \bibinfo{journal}{Journal of Social and Personal Relationships} \emph{\bibinfo{volume}{29}}, \bibinfo{pages}{165--184}. \DOIprefix\doi{https://doi.org/10.1177/0265407511420194}.
\bibitem[{Parrot(2001)}]{parrot2001emotions}
\bibinfo{author}{Parrot, W.} (\bibinfo{year}{2001}). \bibinfo{title}{Emotions in social psychology}.
\newblock \bibinfo{journal}{Psychology, Philadelphia}.
\bibitem[{Ekman(1982)}]{ekman1982emotion}
\bibinfo{author}{Ekman, P.} (\bibinfo{year}{1982}). \bibinfo{title}{What emotion categories or dimensions can observers judge from facial behavior?}
\newblock \bibinfo{journal}{Emotions in the human face} pp. \bibinfo{pages}{39--55}.
\bibitem[{Arnold(1960)}]{arnold1960emotion}
\bibinfo{author}{Arnold, M.} (\bibinfo{year}{1960}). \bibinfo{title}{Emotion and Personality: Psychological aspects}.
\newblock Emotion and Personality. \bibinfo{publisher}{Columbia University Press}.
\newblock ISBN \bibinfo{isbn}{9780231089395}.
\newblock \URLprefix \url{https://books.google.gr/books?id=G2srAAAAIAAJ}.
\bibitem[{Schouten and Frasincar(2015)}]{schouten2015survey}
\bibinfo{author}{Schouten, K.}, and \bibinfo{author}{Frasincar, F.} (\bibinfo{year}{2015}). \bibinfo{title}{Survey on aspect-level sentiment analysis}.
\newblock \bibinfo{journal}{IEEE Transactions on Knowledge and Data Engineering} \emph{\bibinfo{volume}{28}}, \bibinfo{pages}{813--830}. \DOIprefix\doi{10.1109/TKDE.2015.2485209}.
\bibitem[{K{\"u}{\c{c}}{\"u}k and Can(2020)}]{kuccuk2020stance}
\bibinfo{author}{K{\"u}{\c{c}}{\"u}k, D.}, and \bibinfo{author}{Can, F.} (\bibinfo{year}{2020}). \bibinfo{title}{Stance detection: A survey}.
\newblock \bibinfo{journal}{ACM Computing Surveys (CSUR)} \emph{\bibinfo{volume}{53}}, \bibinfo{pages}{1--37}. \DOIprefix\doi{https://doi.org/10.1145/3369026}.
\bibitem[{Mohammad et~al.(2016)Mohammad, Kiritchenko, Sobhani, Zhu and Cherry}]{mohammad2016semeval}
\bibinfo{author}{Mohammad, S.}, \bibinfo{author}{Kiritchenko, S.}, \bibinfo{author}{Sobhani, P.}, \bibinfo{author}{Zhu, X.}, and \bibinfo{author}{Cherry, C.} (\bibinfo{year}{2016}). \bibinfo{title}{Semeval-2016 task 6: Detecting stance in tweets}.
\newblock In \bibinfo{booktitle}{Proceedings of the 10th international workshop on semantic evaluation (SemEval-2016)}. pp. \bibinfo{pages}{31--41}.
\newblock \DOIprefix\doi{10.18653/v1/S16-1003}.
\bibitem[{Sobhani(2017)}]{sobhani2017stance}
\bibinfo{author}{Sobhani, P.}
\newblock \bibinfo{title}{Stance detection and analysis in social media}.
\newblock Ph.D. thesis Universite d'Ottawa/University of Ottawa (\bibinfo{year}{2017}).
\bibitem[{Cabrio and Villata(2018)}]{cabrio2018five}
\bibinfo{author}{Cabrio, E.}, and \bibinfo{author}{Villata, S.} (\bibinfo{year}{2018}). \bibinfo{title}{Five years of argument mining: A data-driven analysis.}
\newblock In \bibinfo{booktitle}{IJCAI} vol.~\bibinfo{volume}{18}. pp. \bibinfo{pages}{5427--5433}.
\newblock \DOIprefix\doi{https://doi.org/10.24963/ijcai.2018/766}.
\bibitem[{Waseem et~al.(2017)Waseem, Davidson, Warmsley and Weber}]{waseem2017understanding}
\bibinfo{author}{Waseem, Z.}, \bibinfo{author}{Davidson, T.}, \bibinfo{author}{Warmsley, D.}, and \bibinfo{author}{Weber, I.} (\bibinfo{year}{2017}). \bibinfo{title}{Understanding abuse: A typology of abusive language detection subtasks}.
\newblock \bibinfo{journal}{Preprint at arXiv \url{https://doi.org/10.48550/arXiv.1705.09899}}.

\end{thebibliography}

\glsaddall
\printglossary[type=\acronymtype]

\appendix

\section{Quality Assurance Search Round}\label{app:quality_assurance_round}

The search strategy for retrieving publications on Greek Natural Language Processing (NLP) research involved querying three databases using their APIs, with the query terms Greek (or Modern Greek) and ``natural language processing'' for the period January 2012 to December 2023 (Section Search Protocol).

To ascertain that we performed a comprehensive search, we conducted an additional round of searches on Google Scholar for the same time period. Google Scholar does not provide an API for automated data retrieval, which is why it was not included in the core search rounds. This supplementary search employed more specific query terms aimed at broadening our exploration, including specific NLP downstream tasks alongside ``Greek'' and either ``Natural Language Processing'' or ``NLP''. Table ~\ref{tab:quality-assurance-search} summarizes this supplementary search process. 

We performed 23 queries, employing various alternatives, acronyms, and operators for selected NLP tasks from November 6th, 2023, to January 12th, 2024. Whenever the retrieved papers numbered less than 100, all were examined; otherwise, only the initial thirty were reviewed. The count of the top 30 papers seemed adequate, as beyond these, the subsequent papers didn't appear to be relevant. 

\begin{table*}[ht]
    \centering
    \caption{The queries and the number of papers retrieved from Google Scholar during the quality assurance search round.}
    \label{tab:quality-assurance-search}
    \begin{tabular}{|l | l |} 
    \hline
    \textbf{Query} & \textbf{Number of papers} \\ 
     \hline
    ``toxicity detection'' ``natural language processing'' greek & \sc 64\\ 
    \hline
    ``toxicity detection'' greek nlp & \sc 70\\ 
    \hline
    ``toxicity detection'' + greek + nlp & \sc 40\\ 
    \hline
    ``toxicity detection'' and ``greek'' and ``nlp'' & \sc 32\\ 
    \hline
    ``abusive language'' ``natural language processing'' greek & \sc 420\\ 
    \hline
    ``hateful language'' ``natural language processing'' greek & \sc 48\\ 
    \hline
    ``aggressive language'' ``natural language processing'' greek & \sc 44\\ 
    \hline
    ``authorship attribution'' ``natural language processing'' greek & \sc 803\\ 
    \hline
    ``authorship identification'' ``natural language processing'' greek & \sc 330\\ 
    \hline
    ``authorship analysis'' ``natural language processing'' greek & \sc 324\\ 
    \hline
    ``authorship detection'' ``natural language processing'' greek & \sc 61\\ 
    \hline
    ``sentiment analysis'' greek nlp & \sc 7,490\\ 
    \hline
    ``sentiment analysis'' ``natural language processing'' greek & \sc 7,050\\ 
    \hline
    ``sentiment analysis'' ``greek'' ``nlp'' & \sc 2,430\\ 
    \hline
    ``machine translation'' greek nlp & \sc 9,160\\ 
    \hline
    ``machine translation'' ``greek'' ``nlp'' & \sc 3,750\\ 
    \hline
    ``named entity recognition''  ``natural language processing'' greek & \sc 4350\\ 
    \hline
    ``named entity recognition'' greek & \sc 5,550\\ 
    \hline
    ``question answering'' ``natural language processing'' greek & \sc 4,840\\ 
    \hline
    summarization ``natural language processing'' greek & \sc 16,600\\ 
    \hline
    semantics ``natural language processing'' greek & \sc 17,100\\ 
    \hline
    syntax ``natural language processing'' greek & \sc 14,400\\ 
    \hline
    \end{tabular}
\end{table*}
The queries led us to review 698 publications. After removing duplicates both within the same round and across previous core search rounds, 357 publications were selected. Following the search process of the protocol, we kept those publications that referenced the term ``Greek'' in either the title or abstract. Additionally, for this particular round, we refined our approach to the term ``Natural Language Processing'' by including only those publications that explicitly mentioned the term within the publication's content, excluding those where it appeared only in the references. In the subsequent filtering process (see Section Filtering Strategy), applying the exclusion criteria yielded five additional publications, which were included in the final list of surveyed papers.

\section{Natural Language Understanding and Generation}
\label{sec:nlu-nlg}
This Appendix first discusses Natural Language Understanding (NLU), which enables machines to understand natural language, and then Natural Language Generation (NLG), which is the process of generating meaningful text.\cite{khurana2023natural}
\subsection{Natural Language Understanding}
NLU enables machines to understand, interpret, and derive meaning from human language in a way that is both accurate and contextually relevant. It addresses a wide range of linguistic phenomena, from lexical semantics concerning aspects of word meaning to the high-level reasoning and application of world knowledge.\cite{wang2019glue} These linguistic phenomena are foundational for various NLP tasks, as they provide the essential linguistic comprehension needed to analyse the textual data.\cite{bowman-dahl-2021-will} 

Traditional machine learning approaches represent texts by engineered features (e.g., based on POS tags or lexicons) or by character and word embeddings, encoding the linguistic knowledge into the input, before solving the task at hand.    
In the deep learning (DL) era, on the other hand, linguistic skills are encoded in neural network models via language modeling.\cite{belinkov2020interpretability} Therefore they are considered as an evaluation criterion, by assessing language models (LM) on  
datasets annotated for various linguistic phenomena. General-purpose NLU evaluation benchmark datasets exist primarily for English,\cite{bommasani2023holistic,chen-gao-2022-curriculum,wang2019superglue,dong2019unified,nie2019adversarial,wang2019glue} but also for other languages.\cite{urbizu2022basqueglue,shavrina-etal-2020-russiansuperglue,wilie-etal-2020-indonlu,xu2020clue,rybak-etal-2020-klej,ham2020kornli} Unfortunately, such monolingual NLU benchmarks are currently unavailable for Greek, based on our survey. There are multilingual NLU benchmarks, where the non-English language parts are translated either by humans, such as Belebele,\cite{bandarkar2023belebele} or automatically, such as ARC Greek.\cite{ilsp_arc_greek} However, even manual translations are distinct from naturally occurring data created by native speakers due to various factors, such as cultural nuances, idiomatic expressions, and context-specific references in native speakers’ natural language usage.\cite{Rogers2023} 

\subsection{Natural Language Generation}\label{ssec:nlg}

NLG involves the generation of human understandable natural language text from various data formats, such as text, structured data, video, and audio. NLG techniques are used in many downstream tasks, such as summarization, dialogue generation, Question Answering (QA), and Machine Translation (MT). Advancements in DL, particularly with Transformer-based LMs such as BART\cite{lewis-etal-2020-bart} (which follows the encoder-decoder neural architecture) and GPT-2\cite{radford2019language} (which uses a decoder-only architecture), as well as the latest developments in large language models, such as the GPT, Llama, and Palm families, have fueled rapid progress in NLG (Section The DL Era). Alongside the advancement of NLG models, attention toward their limitations and potential risks has also increased.\cite{ji2023survey} An example is output degeneration,\cite{holtzman2019curious,welleck2019neural} which refers to generated output that is bland, incoherent, or gets stuck in repetitive loops. Another example is the generation of nonsensical text, or text that is unfaithful to the provided source input,\cite{raunak2021curious,rohrbach2018object} also known as hallucination.\cite{maynez2020faithfulness}

\section{Sentiment Analysis and Argument Mining}\label{apx:sentiment}
Sentiment, opinion, and emotion can be approached by various annotation schemas. Our work presents only the ones employed in the retrieved papers. Binary Sentiment Analysis (SA), where the text's polarity is either positive or negative, is the primary approach used in most papers cited in this study.\cite{Bilianos2022,Kapoteli2022,Alexandridis2021Survey,Kydros2021,Braoudaki2020,Beleveslis2019,Medrouk2018, Medrouk2017,athanasiou2017,GIATSOGLOU2017, Charalampakis2016,markopoulos2015} However, several studies \cite{Alexandridis2021Survey,Drakopoulos2020,Beleveslis2019,tsakalidis2018building,Makrynioti2015,petasis2014sentiment} added a neutral class for texts carrying no opinion. 

Five-class sentiment classification \cite{spatiotis2020, Spatiotis2019, spatiotis2017examining,Spatiotis2016} increases the granularity, with scales from positive to negative. 
Subjectivity detection can precede SA, by addressing the binary task of whether the text comprises sentiment or not.\cite{Solakidis2014} 
Instead of the coarse level of sentiment and subjectivity, however, one can also focus on fine-grained emotions,\cite{Alexandridis2021,tsakalidis2018building,Chatzakou2017,Solakidis2014} using different emotion categories that have been suggested by theorists in psychology.\cite{kafetsios2012emotion,parrot2001emotions,ekman1982emotion,arnold1960emotion} Ekman has identified six emotions, i.e., anger, disgust, fear, joy, sadness, and surprise, as primary. 
A common approach, finally, concerns the selection of a single emotion or sentiment that is related to a specific condition, such as the anxiety during the COVID19 pandemic \cite{Kydros2021} or sarcasm/irony.\cite{tsakalidis2018building,antonakaki2017,Antonakaki2016}

A critical distinction between SA approaches is the level of granularity of the analysis. The coarser granularity concerns the document level, where the overall sentiment of an entire document is assessed. A finer granularity regards the sentence level, where the sentiment of short texts, such as tweets or the individual sentences within a document, is analyzed. SA can also be aspect-based, where the sentiment towards specific aspects or features of an entity is evaluated. These are described in detail below:

\begin{itemize}
    \item \textbf{Document-level SA} involves the detection of the emotions expressed in an opinion document (e.g., a lengthy product review) or its classification as positive or negative (i.e., the sentiment polarity). Analysis at this level focuses on the sentiment of the entire document, as a whole (e.g., a product review), without considering entities or aspects within the document. 
    \item \textbf{Sentence-level SA} concerns short texts, such as tweets or the sentences of a document. Although analysis at the sentence level is similar to that at the document level, the former can be more challenging due to the limited context contained. Also, classification at the sentence level cannot easily ignore the neutral class, because sentences with no opinion are more likely to appear. That is, even opinionated documents comprise sentences that bear no sentiment. Furthermore, a common (although implicit) assumption at research tasks focusing on the sentence level is that each sentence expresses a single sentiment,\cite{liu2020sentiment} which, however, is not always the case. 
    \item \textbf{Aspect-based SA}, also known as topic-based, entity-based, or target-based SA, as there is not a single term that sounds natural in every application domain,\cite{schouten2015survey} is the analysis that focuses on the sentiment expressed in a document or sentence with respect to a specific aspect of an entity. For example, in the sentence ``iPhone's battery lasts long'' the entity is iPhone and the aspect is battery. Closely related to aspect-based SA is the \textbf{stance detection} task,\cite{kuccuk2020stance} which aims to identify the stance (as in favor of or against) of the text author towards a target (person, organization, movement, policy, etc.) either explicitly mentioned or implied within the text.\cite{mohammad2016semeval,sobhani2017stance} Only a few published studies concern SA on Greek at this granularity.
    \item \textbf{Argument Mining} is the task of extracting natural language arguments and their relationships from text. The final objective is to provide machine-processable structured data for computational models of argument, to facilitate the automatic identification of reasoning capabilities upon the retrieved arguments and relations.\cite{cabrio2018five} 
\end{itemize}

\section{Toxicity Detection}
\label{apx:tox}
Our study uses the term toxicity as a broader umbrella term of unwanted user-generated content including hateful and violent speech.\cite{waseem2017understanding} Although no standard definition exists today and some researchers are against oversimplifying solutions,\cite{diaz2022accounting} our choice is driven simply by the practical need to cover all the subtypes in a single name. However, the definition of toxicity itself is controversial: different cultures, languages, dialects, social groups, minority groups do not necessarily agree on what constitutes toxicity; furthermore, aggressive language use is not always toxic, e.g., inside a group, swear words can function as solidarity markers.

\end{document}